\documentclass{article}

\usepackage{arxiv}

\usepackage{latexsym}
\usepackage{graphicx}
\usepackage{epstopdf}
\usepackage{multicol,multirow}
\usepackage{amsmath,amssymb,amsfonts}
\usepackage{mathrsfs}
\usepackage{amsthm}
\usepackage{rotating}
\usepackage{appendix}
\usepackage[authoryear]{natbib}
\usepackage{ifpdf}
\usepackage[T1]{fontenc}
\usepackage{times}
\usepackage{sourcesanspro}
\usepackage{newtxmath}
\usepackage{booktabs}
\usepackage{textcomp}
\usepackage{xcolor}
\usepackage{hyperref}
\usepackage{makecell}

\title{
Two-Scale Localized PCA-Net: Coarse-Global and Local-Residual
Representations for Artifact-Reduced PDE Operator Learning
}

\author{
  Mrigank Dhingra\thanks{Corresponding author: \texttt{mdhingra@vols.utk.edu}} \\
  Department of Mechanical \& Aerospace Engineering \\
  University of Tennessee \\
  Knoxville, Tennessee 37917, USA \\
  \And
  Jordan Stout \\
  Department of Mathematics and Statistics \\
  Boston University \\
  Boston, Massachusetts 02215, USA \\
  \And
  Omer San \\
  Department of Mechanical \& Aerospace Engineering \\
  University of Tennessee \\
  Knoxville, Tennessee 37917, USA
}

\begin{document}

\maketitle

\begin{abstract}
Localized dimensionality reduction provides an efficient alternative to
global representations for learning solution operators of high-dimensional
partial differential equations (PDEs), but independently decoded local output
patches can introduce block offsets, interface mismatches, and spurious
high-wavenumber content. We introduce Two-Scale Localized PCA-Net, a
representation-first extension of localized PCA-based operator learning that
decomposes the solution into a coarse-global component and local residual
corrections. A compact global PCA basis captures coherent domain-scale
structure, while nonoverlapping local PCA bases represent only the remaining
fine-scale residual. To train the coupled representation, we employ a
block-balanced latent objective that prevents the larger residual code from
dominating supervision of the compact global code. We further consider an
optional interface-aware fine-tuning stage that backpropagates reconstruction
and truth-referenced value and normal-derivative trace losses through a
differentiable PCA decoder and field assembler. On Poisson benchmarks across
multiple resolutions, the two-scale representation substantially reduces
reconstruction error relative to plain and overlap-based localized PCA-Net
variants while removing the dominant visible block-offset artifacts and
approximately halving PCA fitting cost relative to overlap. Experiments on
heterogeneous Darcy flow show strong reductions in interface and
discrete-residual errors, although gains in reconstruction error are more
modest. Ablation studies demonstrate that the dominant improvement arises from
the output representation itself, with interface-aware fine-tuning providing a
complementary continuity refinement. These results show that separating
globally coherent structure from localized residual detail provides a compact
and effective representation for artifact-reduced PDE operator learning.
\end{abstract}


% \section*{Impact Statement}

% Data-driven surrogate models can accelerate repeated simulations of
% engineering systems, but their usefulness depends on predictions that are both
% computationally efficient and spatially coherent. Localized reduced-order
% models improve scalability, yet independently reconstructed regions can create
% artificial boundaries that contaminate predicted fields and derived physical
% quantities. This work shows that these artifacts can be reduced by separating
% domain-wide structure from localized detail within the learned representation
% itself, rather than correcting predictions afterward. The resulting approach
% improves reconstruction quality and substantially lowers
% dimensionality-reduction cost relative to overlap-based localization, while
% preserving an inference pipeline. More broadly, the study demonstrates that
% matching representation scale to the spatial coherence of the underlying
% physics can produce more accurate, efficient, and reliable data-driven
% surrogates for engineering analysis.

\section{Introduction}
\label{sec1:introduction}

Many scientific and engineering tasks require repeated evaluation of a map
between spatial fields: a coefficient or forcing field is supplied as input,
and the solution of a partial differential equation (PDE) is required as
output. High-fidelity discretizations remain the source of truth, but their
cost can make parameter studies, optimization, uncertainty quantification, and
real-time decision workflows prohibitive. Data-driven PDE surrogates seek to
amortize that cost by learning the underlying solution operator from paired
fields. The central challenge is to retain the global coherence imposed by an
elliptic PDE while making both representation construction and learning
practical at high spatial resolution.

\subsection{Data-driven PDE surrogates and reduced representations}
\label{subsec11}

Neural networks can approximate nonlinear maps between function spaces
\cite{chen1995universal}, and operator-learning methods turn this observation
into trainable surrogates for PDE solution maps. DeepONet, Fourier neural
operators (FNOs), and the broader neural-operator framework have demonstrated
that one model can learn a family of field-to-field maps rather than a single
simulation trajectory \cite{lu2021deeponet,li2021fno,kovachki2023neuraloperator, SOJITRA2026114931, azizzadenesheli2024neuraloperators, subedi2026operatorlearning}.
Subsequent architectures have extended this direction to multigrid structure,
irregular meshes, message-passing representations, multiscale frequency
mixing, and general geometries
\cite{li2020mgno,pfaff2021meshgraphnets,brandstetter2022mpnn,rahman2022uno,tran2023ffno,tripura2022wno,li2023geofno, liu2024multiscaleoperator, kalimuthu2025loglofno}.
These methods provide powerful full-field baselines, but their dense spatial
representations can make training and inference increasingly demanding as the
grid is refined \cite{dehoop2022costaccuracy}. Benchmark efforts such as PDEBench further make clear that
resolution, geometry, and equation class materially affect the practical
comparison of learned PDE surrogates \cite{takamoto2023pdebench}.

Reduced-order modeling offers a complementary route. Proper orthogonal
decomposition (POD), principal component analysis (PCA), and reduced-basis
methods seek a low-dimensional coordinate system in which the dominant
variation of a solution family can be represented economically
\cite{sirovich1987pod,berkooz1993pod,quarteroni2016reducedbasis,hesthaven2016certifiedrb}.
Nonintrusive reduced models then learn a map between reduced input and output
coordinates, avoiding repeated high-dimensional solves
\cite{hesthaven2018nonintrusive,rowley2017modelreduction,benner2015survey, kramer2024operatorinference, duan2024nonintrusive, bhattacharya2021pcanet}.
However, a single global basis has a feature dimension proportional to the
number of grid points. Its construction and use can therefore become
unattractive for large fields, even when a modest number of PCA modes is
retained. Randomized matrix algorithms reduce the cost of truncated PCA
without changing this global feature-space scaling
\cite{halko2011randomized,martinsson2020rla,tropp2017sketching}.

\subsection{Localization as a scalability strategy}
\label{subsec12}

Localization addresses this dimensionality directly. Instead of fitting one
basis over the entire field, the domain is partitioned into spatial patches,
and a compact PCA representation is fit independently in each patch. This
idea aligns with a long numerical tradition: overlapping Schwarz methods,
coarse spaces, partition-of-unity constructions, and multiscale finite-element
methods all combine local computations with mechanisms that recover global
communication or compatibility
\cite{lions1988schwarz,xu1992subspace,toselli2005ddm,babuska1997pum,hou1997msfem,efendiev2009msfem, pan2024domain, gkimisis2025localizedrom}.

\begin{figure}[htbp]
    \centering
    \includegraphics[width=\linewidth]{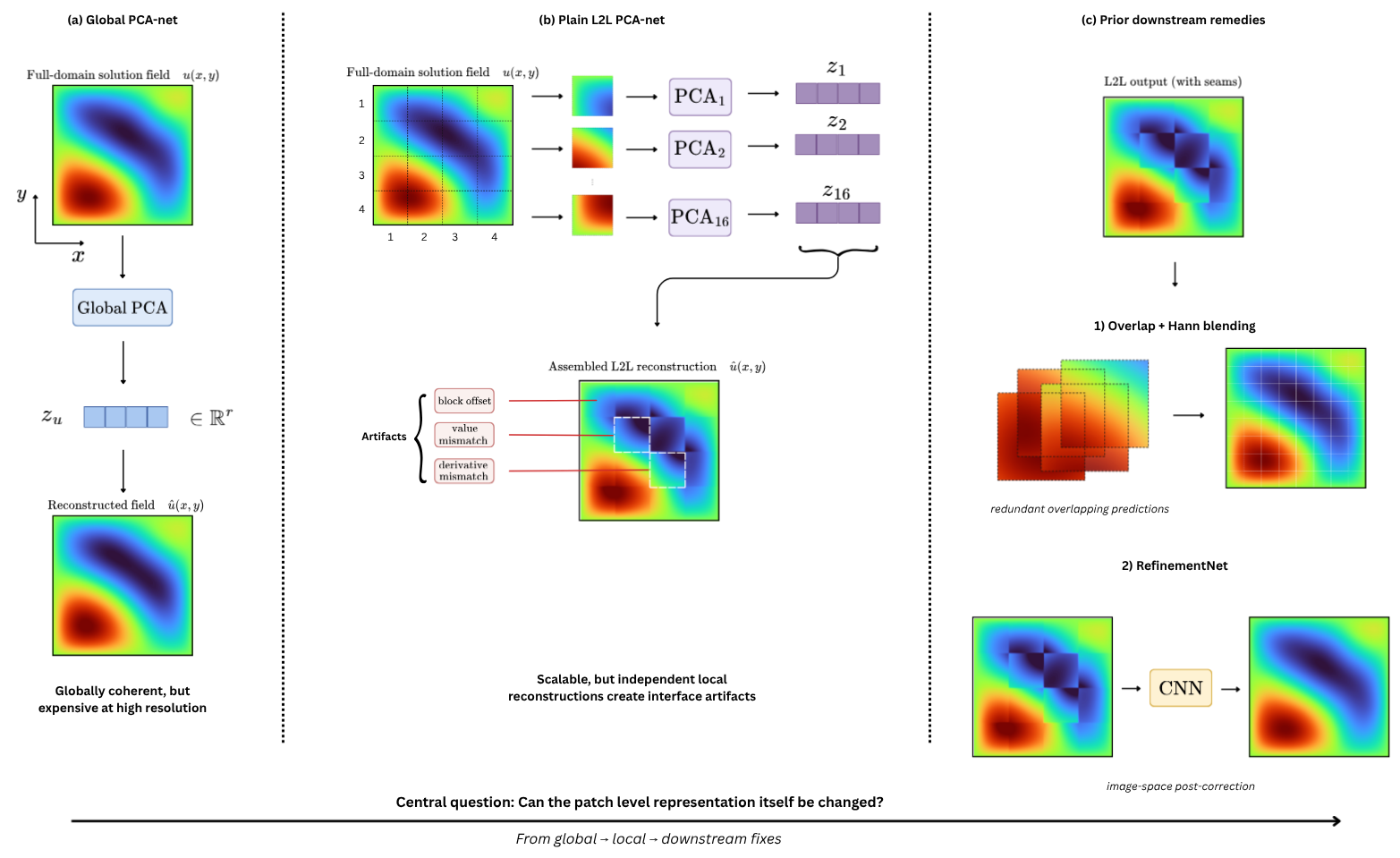}
    \caption{\textbf{Motivation for a representation-first treatment of patch-interface artifacts in localized PCA-Net.}
(a) Global PCA represents the complete solution field in a single reduced basis, preserving domain-wide coherence but becoming increasingly expensive as spatial resolution grows. (b) Plain local-to-local (L2L) PCA-Net instead decomposes the output field into independently encoded and decoded nonoverlapping patches. Although computationally attractive, independent local reconstruction can produce patch offsets, interface value mismatches, normal-derivative mismatches, and associated high-wavenumber artifacts. (c) The prior framework mitigated these effects downstream through overlapping output patches with Hann-weighted blending or through a learned image-space RefinementNet. The present work instead addresses the source of the artifact by redesigning the output representation itself}
    \label{fig:two-scale-motivation}
\end{figure}

Localized PCA-Net transfers the representation-side benefit of this principle
to a neural operator. It encodes input and output patches by independent PCA
models and learns a latent map between their concatenated scores
\cite{dhingra2026localized}. The resulting local-to-local (L2L) construction
can lower PCA fitting cost and keep latent representations compact, while
remaining compatible with an ordinary multilayer perceptron in score space.
Figure~\ref{fig:two-scale-motivation}(b) illustrates the corresponding output
path: independently represented patches are decoded and assembled into a
single field. The same locality that makes the method economical, however,
also creates its characteristic failure mode. Adjacent patches must agree on
a shared physical solution even though their output coordinates, PCA bases,
and reconstruction errors are local.

The interface issue is not unique to PCA-based surrogates. Domain-decomposed
neural solvers often introduce explicit value or derivative constraints at
subdomain boundaries \cite{jagtap2020cpinn,jagtap2020xpinn,moseley2023fbpinn,dolean2023fbpinnschwarz, dolean2024multilevel, hu2025schwarz}.
Those approaches typically decompose the PDE solve itself into interacting
neural subproblems. In contrast, our setting begins with one trained
field-to-field surrogate and asks how its compressed output coordinates should
be organized so that local reconstruction does not repeatedly recreate a
global compatibility problem.

\subsection{Limitations of downstream artifact correction}
\label{subsec13}

The preceding localized PCA-Net study evaluated two practical remedies for
patch artifacts: overlapping output patches with Hann-weighted assembly, and a
post-hoc RefinementNet \cite{dhingra2026localized}. Overlap gives a pixel
multiple decoded predictions and blends them, which can attenuate visible
value discontinuities. It also increases the number of patches, local PCA
fits, latent outputs, and decoded fields. A learned image-space correction can
further smooth the assembled reconstruction, in the same broad spirit as
convolutional image-to-image models \cite{ronneberger2015unet}; however, it is
tasked with repairing an error that has already been created by the target
representation. It can consequently add a second training stage and may alter
legitimate fine-scale content while suppressing seams.

These remedies are valuable baselines, not defects in the localization idea.
They reveal a more fundamental tension: a purely local output representation
requires each patch to reproduce both the global, low-wavenumber solution
structure and the genuinely local detail. Any small discrepancy in the former
is replicated at an interface. Averaging or refining after assembly can
reduce the symptom, but does not change which latent variables are responsible
for the shared field. This distinction motivates an intervention before
patchwise decoding, rather than only after it.

\subsection{Representation-first hypothesis}
\label{subsec14}

We test the following hypothesis: \emph{tiling artifacts are primarily a
representation-allocation problem, and they can be reduced by assigning
domain-wide structure to a shared coarse representation while reserving local
PCA models for residual detail.} The proposed two-scale localized PCA-Net
implements this hypothesis on the target side only. Each solution is first
restricted to a coarse grid and represented by one global PCA model. Its
decoded coarse field is bilinearly prolongated to the fine grid and subtracted
from the full solution. Local PCA models are then fitted to the residual
patches. The latent predictor receives the same localized input codes as L2L
and predicts one coarse global code together with the local residual codes.
At decoding, the dense coarse reconstruction and the assembled residual field
are added.

This arrangement is analogous in spirit to numerical methods that use a
coarse component for globally coupled behavior and local components for
fine-scale correction \cite{toselli2005ddm,hou1997msfem,efendiev2009msfem},
but it does not claim to be a domain-decomposition solver. It is a learned,
nonintrusive output representation for a fixed-grid operator surrogate. The
method also admits a complementary optional stage: with PCA bases and the
decoder frozen, we fine-tune the latent predictor through the differentiable
assembled field using truth-referenced interface value and normal-derivative
trace losses. This physical-field objective is informed by the broader
physics-informed learning literature, including its documented optimization
and loss-balancing challenges
\cite{raissi2019pinn,karniadakis2021piml,kharazmi2021vpinn,wang2021gradientpathologies,wang2022pinnntk,krishnapriyan2021failuremodes, hao2024pinnacle}.

Our goal is therefore not to claim a universally discretization-invariant
operator or exact interface continuity. Rather, we seek a more favorable
accuracy--continuity--cost tradeoff within the localized PCA-Net family: the
global branch carries structure that should not be independently inferred by
every patch, while the local branch retains the compressed detail that makes
localization useful.

\subsection{Contributions}
\label{subsec15}

The contributions of this work are as follows:
\begin{enumerate}
  \item We introduce a two-scale target representation for localized PCA-Net:
  a low-rank global PCA representation on a restricted grid plus independent
  local PCA representations of the fine-grid residual. The input-side
  localization and latent MLP are otherwise retained.
  \item We show that a shared, prolongated coarse field suppresses the
  dominant patch-aligned reconstruction artifact without requiring redundant
  overlapping output patches or an image-space post-processor.
  \item We develop an optional interface-aware physical-field fine-tuning
  stage with a differentiable patch assembler and truth-referenced value and
  normal-derivative trace losses. This stage is positioned as a
  continuity-oriented refinement, not as the default representation.
  \item We evaluate accuracy, structural fidelity, interface diagnostics, and
  stage-wise cost on Poisson and heterogeneous Darcy benchmarks across
  resolutions, sample budgets, coarse-representation choices, latent
  objectives, physical-field loss variants, and PCA solvers. The experiments
  establish two-scale localized PCA-Net with block-balanced latent supervision
  as the recommended default and quantify the optional fine-tuning tradeoff.
\end{enumerate}

\section{Localized latent operator learning and the interface problem}
\label{sec2:prob_statement}

\subsection{PDE operator learning problem}
\label{subsec21}

Let $\Omega\subset\mathbb{R}^2$ denote a bounded spatial domain and let
$x\in\mathcal{X}$ denote a field specifying a PDE instance, such as a forcing
field or heterogeneous coefficient field. The corresponding solution
$u\in\mathcal{U}$ is generated by an elliptic boundary-value problem and
defines an operator
\begin{equation}
  \mathcal{G}:\mathcal{X}\rightarrow\mathcal{U},
  \qquad
  u=\mathcal{G}(x).
  \label{eq:operator-learning-map}
\end{equation}
Given $m$ paired discretized samples
$\{(x^{(i)},u^{(i)})\}_{i=1}^{m}$ on a common $D\times D$ grid, we seek a
nonintrusive surrogate $\widehat{\mathcal{G}}_\theta$ that predicts the
solution field for an unseen input. This is the standard field-to-field
operator-learning setting
\cite{lu2021deeponet,li2021fno,kovachki2023neuraloperator,
kovachki2024operatorlearning,azizzadenesheli2024neuraloperators,
subedi2026operatorlearning}.

A direct full-field model acts on $D^2$ spatial degrees of freedom at both
input and output. Such models can provide high predictive accuracy, but dense
field representations become increasingly expensive to manipulate and learn
as resolution grows. The objective considered here is complementary: we seek
a compressed surrogate that learns the operator in reduced coordinates while
retaining the spatial coherence of the decoded PDE solution. This follows the
broader nonintrusive reduced-modeling paradigm in which dimensionality
reduction is separated from learning or identifying the reduced dynamics
\cite{hesthaven2018nonintrusive,rowley2017modelreduction,
kramer2024operatorinference,duan2024nonintrusive,dhingra2026localized}.

\subsection{PCA-Net in reduced coordinates}
\label{subsec22}

PCA, or equivalently POD for the snapshot data considered here, provides a
linear low-rank coordinate system for a field ensemble
\cite{sirovich1987pod,berkooz1993pod,holmes2012turbulence}. Let
$q\in\mathbb{R}^{D^2}$ be a vectorized field, $\mu_q$ and $\sigma_q$ be
training-set standardization statistics, and $\Phi_q\in\mathbb{R}^{D^2
\times k_q}$ contain the retained PCA components. Its reduced score is
\begin{equation}
  z_q=\Phi_q^\top S_q(q),
  \qquad
  S_q(q)=\left(q-\mu_q\right)\oslash\sigma_q,
  \label{eq:global-pca-encode}
\end{equation}
where $\oslash$ denotes elementwise division. The corresponding linear
decoder is
\begin{equation}
  \mathcal{D}_q(z_q)=
  \mu_q+\sigma_q\odot\left(\Phi_qz_q\right),
  \label{eq:global-pca-decode}
\end{equation}
with $\odot$ denoting elementwise multiplication. The rank $k_q$ can be
chosen by a retained-variance criterion or set explicitly. PCA is valuable
because it separates representation construction from the learned map and
makes the target dimension visible and controllable; randomized solvers make
its truncated construction practical for large snapshot matrices
\cite{halko2011randomized,martinsson2020rla,tropp2017sketching}.

A global PCA-Net fits separate PCA encoders for the input $x$ and target $u$
and trains a neural network in their score spaces,
\begin{equation}
  \widehat z_u=g_\theta(z_x),
  \qquad
  \widehat u=\mathcal{D}_u\!\left(g_\theta(z_x)\right).
  \label{eq:global-pcanet}
\end{equation}
The conventional latent objective is a mean-squared error between predicted
and true target scores. The global output basis in Eq.~\eqref{eq:global-pcanet}
decodes one field and therefore preserves global representation coherence.
Its drawback is representation scale: the PCA fit acts on vectors of length
$D^2$, and the input and output bases are dense over the full domain. This
motivates localization.

\subsection{Localized PCA-Net}
\label{subsec23}

Let $E_j$ extract the $j$th square patch from a field, with side length $p$,
stride $s$, and $j=1,\ldots,P$. For plain nonoverlapping L2L,
$s=p$; the configurations considered here use a fixed $4\times4$ patch
topology, giving $P=16$. Independent standardized PCA models are fitted to
each input and output patch,
\begin{equation}
  z^x_j=\Phi_j^{x\top}S_j^x(E_jx)
  \in\mathbb{R}^{k^x_j},
  \qquad
  z^u_j=\Phi_j^{u\top}S_j^u(E_ju)
  \in\mathbb{R}^{k^u_j}.
  \label{eq:localized-pca-codes}
\end{equation}
Because each patch selects its retained rank independently, the block
dimensions may differ across the domain. The complete input and target codes
are the ordered concatenations
\begin{equation}
  z_x=[z^x_1,\ldots,z^x_P],
  \qquad
  z_u=[z^u_1,\ldots,z^u_P],
  \label{eq:localized-concatenated-codes}
\end{equation}
and a single latent network predicts all target blocks jointly,
\begin{equation}
  \widehat z_u=g_\theta(z_x).
  \label{eq:l2l-latent-map}
\end{equation}
Each predicted block is decoded by its corresponding local PCA model and,
for plain L2L, the reconstructed patches are placed directly into the output
field. This is the localized PCA-Net construction introduced in our prior work
\cite{dhingra2026localized}.

The spatial decomposition is representational rather than a decomposition of
the PDE solve itself. Classical Schwarz and substructuring methods couple
local PDE problems through overlap, coarse spaces, or transmission conditions
\cite{lions1988schwarz,xu1992subspace,toselli2005ddm}; recent data-driven
reduced models have likewise explored domain-decomposed or spatially
localized representations with explicit coupling between regions
\cite{pan2024domain,gkimisis2025localizedrom}. Localized PCA-Net instead
learns one global map in the concatenated score space:
$g_\theta$ has access to every input-patch code when predicting every output
block. The source of the artifact is therefore not independent neural
subsolvers, but the use of independently parameterized \emph{output
reconstructions}.

\subsection{Origin of patch-interface artifacts}
\label{subsec24}

Elliptic solutions are globally coupled, whereas plain L2L parameterizes the
target field locally. Each output patch must therefore encode not only local
detail but also the long-wavelength level, amplitude, and curvature that
should remain compatible with neighboring patches. Although the latent network
predicts all patch scores jointly, no target-side coordinate is explicitly
shared across patch boundaries. Small errors in different score blocks are
consequently decoded through different patch bases and can appear after
assembly as block offsets and derivative mismatches. The resulting
patch-aligned structure is visible directly in the representative
reconstructions in Fig.~\ref{fig:poisson128-qualitative}; the mechanism ablation
in Table~\ref{tab:poisson128-headline} tests whether this error is removed
primarily by changing the representation or by imposing an interface loss
afterward.

To distinguish an artificial seam from legitimate variation in the target
field, we use truth-referenced interface traces. For a discrete field $v$ and
an internal vertical boundary at column $b$, define
\begin{equation}
  T_V(v;b)=v_{:,b}-v_{:,b-1},
  \label{eq:interface-value-trace-problem}
\end{equation}
and, with grid spacing $h$, the normal-derivative jump proxy
\begin{equation}
  T_F(v;b)=
  \frac{v_{:,b+1}-v_{:,b}}{h}
  -
  \frac{v_{:,b}-v_{:,b-1}}{h}.
  \label{eq:interface-flux-trace-problem}
\end{equation}
Horizontal interfaces are obtained by exchanging the grid axes. We compare
$T_V(\widehat u;b)$ and $T_F(\widehat u;b)$ with the corresponding traces of
the target solution rather than forcing either quantity to vanish. This is
important because a smooth solution can exhibit nonzero finite differences
across an arbitrary patch boundary.

The same need to communicate compatible information across artificial
subdomain boundaries appears in classical and learned domain-decomposition
methods. Nonoverlapping substructuring makes transmission conditions explicit
\cite{farhat1991feti,mandel1993bdd}, while neural decompositions have used
solution- and derivative-based interface coupling in both overlapping and
nonoverlapping settings
\cite{jagtap2020cpinn,jagtap2020xpinn,moseley2023fbpinn,
dolean2023fbpinnschwarz,dolean2024multilevel,hu2025schwarz}.
The distinction here is again representational: the PDE is predicted as one
operator map, but independently decoded local target coordinates can create
an artificial compatibility problem during reconstruction.

\subsection{Existing mitigation strategies}
\label{subsec25}

The preceding localized PCA-Net study considered two downstream strategies for
reducing this artifact \cite{dhingra2026localized}. In the first, the output
stride is reduced so that several decoded patches cover the same pixel, and
the final field is formed by normalized Hann-weighted overlap-add. This
partition-of-unity construction attenuates patch mismatch through redundant
local predictions
\cite{lions1988schwarz,babuska1997pum,toselli2005ddm}, at the cost of
additional patch PCA models, latent coordinates, and decoded fields.

The second strategy applies a convolutional RefinementNet to the assembled L2L
prediction. Such image-space architectures are effective spatial correction
models \cite{ronneberger2015unet}, but the correction occurs only after the
locally parameterized field has been reconstructed and introduces a separate
learned stage.

The measured tradeoffs of these two strategies are summarized alongside the
proposed representation in Table~\ref{tab:poisson128-headline}, while
Fig.~\ref{fig:tradeoff-common-seams} provides the geometry-controlled interface
comparison required when overlap and nonoverlap methods have different native
seam sets. These baselines motivate the question addressed next: can the
target coordinates themselves be reorganized so that coherent structure is
decoded once globally and only the remaining detail is reconstructed locally?
Section~\ref{sec:two-scale-localized-pcanet} introduces the resulting coarse-global plus
local-residual representation.

\section{Two-Scale Localized PCA-Net}
\label{sec:two-scale-localized-pcanet}

\begin{figure}[htbp]
    \centering
    \includegraphics[width=\linewidth]{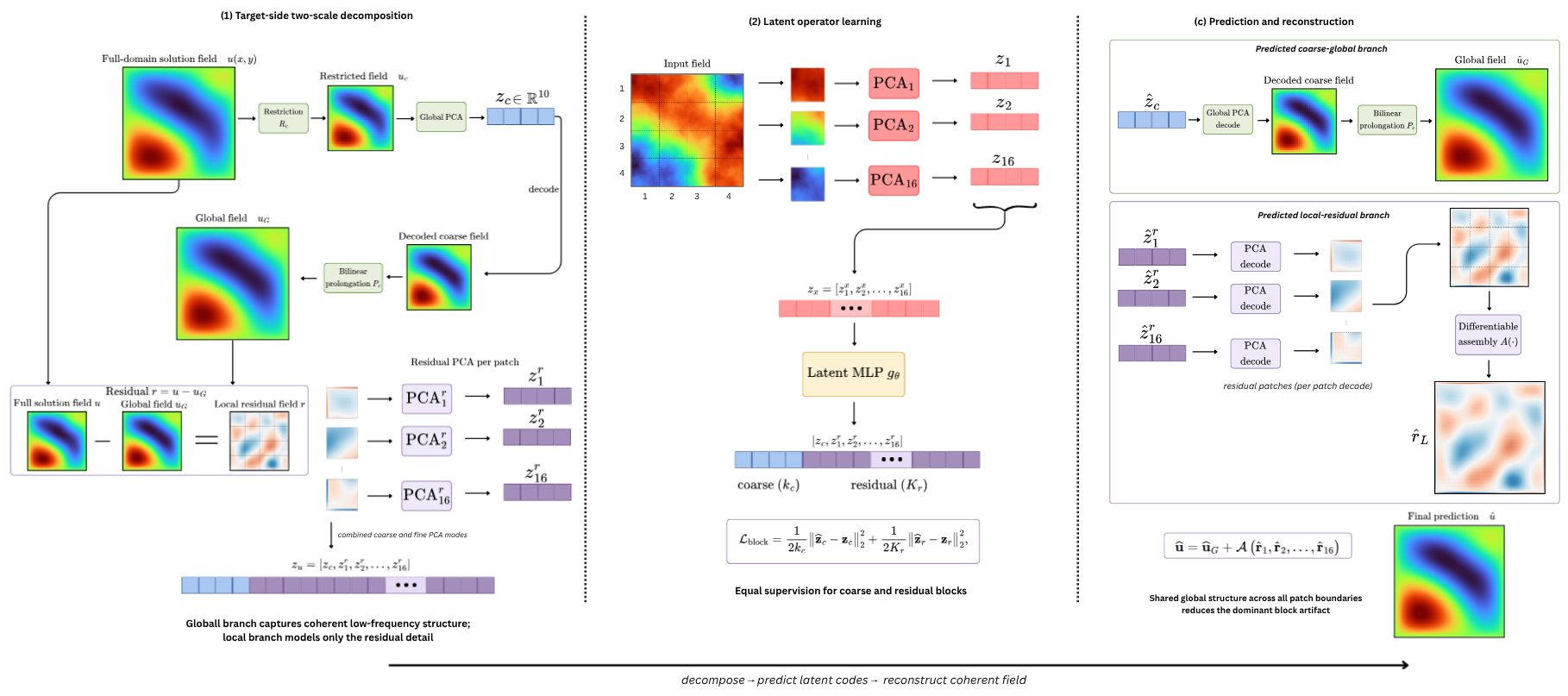}
    \caption{\textbf{Two-scale localized PCA-Net with coarse-global and local-residual output representations.} The target solution field (u) is first restricted to a coarse grid and represented using a compact global PCA code $(z_c)$. Decoding and bilinearly prolongating this code yields a coarse-global field $(u_G)$ that carries coherent domain-scale structure across all patch boundaries. The residual $(r=u-u_G)$ is subsequently partitioned into nonoverlapping patches and represented using independent local PCA bases, producing residual codes $(z_1^r,\ldots,z_P^r)$. The complete output representation is therefore $(z_u=[z_c,z_1^r,\ldots,z_P^r])$. The input representation remains local as in plain L2L PCA-Net, and the latent mapper $(g_\theta)$ predicts both the coarse and residual code blocks. A block-balanced objective gives equal aggregate supervision to the compact coarse block and the larger residual block. At inference, the predicted coarse-global field and assembled residual patches are combined as $(\widehat{u}=\widehat{u}_G+\mathcal{A}(\widehat{r}_1,\ldots,\widehat{r}_P))$. By assigning domain-wide structure to the shared global branch and only localized corrections to the residual branch, the representation reduces the dominant block-offset artifact without overlapping output patches or an image-space postprocessor}
    \label{fig:two-scale-schematic}
\end{figure}

The localized PCA-Net formulation in Section~\ref{sec2:prob_statement}
reduces representation cost by learning in patchwise PCA coordinates, but its
local-only target representation allows independently decoded patches to
reconstruct the domain-scale component of the solution separately. We address
this mismatch by changing only the target-side representation. The proposed
two-scale localized PCA-Net decomposes each solution into a coarse-global
component and local residual corrections, predicts both sets of coordinates
from the original localized input code, and combines their decoded fields.
Figure~\ref{fig:two-scale-schematic} summarizes the construction.

\subsection{Two-scale representation principle}
\label{subsec:two-scale-principle}

Let $x\mapsto u$ denote the operator of interest, with
$u\in\mathbb{R}^{D\times D}$. We represent the target as
\begin{equation}
  u \approx u_G + r_L,
  \label{eq:two-scale-decomposition}
\end{equation}
where $u_G$ is decoded from one global PCA basis fitted on a restricted grid,
and $r_L$ is assembled from independently decoded local residual patches. The
global branch carries coherent domain-scale structure, whereas the local branch
restores the detail not represented on the coarse grid.

This division of labor follows the broad multilevel principle of representing
globally coupled behavior through a coarse component and resolving remaining
structure locally
\cite{toselli2005ddm,hou1997msfem,efendiev2009msfem}.
Recent scientific-machine-learning and reduced-order approaches have likewise
shown benefits from combining globally communicating and locally specialized
representations
\cite{dolean2024multilevel,pan2024domain,gkimisis2025localizedrom,
kalimuthu2025loglofno}.
Those methods employ different domain-decomposition or architectural
mechanisms; here, the separation is imposed directly on the \emph{target
coordinates} of a nonintrusive PCA-based operator.

Importantly, the modification is output-side only. The localized input
representation and PCA-Net latent MLP are unchanged.

\subsection{Localized input representation}
\label{subsec:localized-input-representation}

The input representation is retained exactly from plain L2L PCA-Net.
Using the patch extraction and local PCA encoders introduced in
Section~\ref{subsec23}, the latent input is
\begin{equation}
  z_x=[z_1^x,\ldots,z_P^x],
  \label{eq:localized-input-vector}
\end{equation}
where the local ranks $k_j^x$ are selected independently by the prescribed
retained-variance criterion. In the primary experiments, the domain is divided
into a fixed $4\times4$ nonoverlapping topology, so $P=16$ and $s=p$.

No global input encoder is added. The distinction between global and local
scales enters only through the target representation that the latent map must
predict. This isolates the effect of changing the output coordinates from
changes to the input encoder or neural architecture.

\subsection{Coarse-global branch}
\label{subsec:coarse-global-pca}

Choose an integer restriction factor $c$ that divides $D$. The restriction
operator $R_c$ averages nonoverlapping $c\times c$ blocks,
\begin{equation}
  u_c=R_cu,
  \qquad
  u_c\in\mathbb{R}^{D_c\times D_c},
  \qquad
  D_c=D/c.
  \label{eq:coarse-restriction}
\end{equation}

We impose the design condition
\begin{equation}
  (D/c)^2<m,
  \label{eq:coarse-factor-invariant}
\end{equation}
where $m$ is the number of training fields. This is not a mathematical
requirement of PCA; rather, it keeps the coarse feature dimension below the
sample count and preserves the intended compact global branch. With
$m=8000$, the common setting used across the executed $64^2$, $128^2$, and
$256^2$ studies is $c=4$.

After standardization, one PCA model is fitted to the restricted solution
ensemble,
\begin{equation}
  z_c=
  \Phi_c^\top
  \left[
    S_c\!\left(\operatorname{vec}(u_c)\right)-m_c
  \right],
  \qquad
  z_c\in\mathbb{R}^{k_c},
  \label{eq:coarse-pca-code}
\end{equation}
where $m_c$ denotes the PCA centering vector in standardized coordinates.
The selected configuration retains $k_c=10$ coarse modes.

Decoding the coarse score and prolongating the result gives the dense global
field
\begin{equation}
  u_G(z_c)
  =
  P_c S_c^{-1}
  \left(m_c+\Phi_c z_c\right)
  \in\mathbb{R}^{D\times D},
  \label{eq:coarse-prolongation}
\end{equation}
where $P_c$ denotes bilinear prolongation. The latent dimension remains
$k_c$: prolongation changes the spatial realization of each coarse mode, not
the number of retained coordinates.

The moderate coarse rank and $c=4$ restriction are empirical design choices
rather than PCA-capacity optima. The coarse-representation study in
Section~\ref{sec:coarse-global-design} shows that increasing $k_c$
continues to lower the PCA-oracle error while eventually \emph{increasing}
learned reconstruction error. The selected rank therefore balances
representation capacity against latent-score predictability.

\subsection{Local residual branch}
\label{subsec:local-residual-representation}

For every training solution, we form the residual relative to its own decoded
coarse projection,
\begin{equation}
  r=u-u_G(z_c).
  \label{eq:local-residual}
\end{equation}
The residual is partitioned using the same nonoverlapping geometry as the
plain-L2L output. Each residual patch $r_j$ is then represented by an
independent PCA basis,
\begin{equation}
  z_j^r=
  \Phi_j^{r\top}
  \left[
    S_j^r(r_j)-m_j^r
  \right]
  \in\mathbb{R}^{k_j^r},
  \qquad j=1,\ldots,P,
  \label{eq:residual-pca-code}
\end{equation}
with $m_j^r$ denoting the corresponding PCA centering vector. The local
residual ranks are selected independently and may therefore differ across
patches.

The complete target representation is
\begin{equation}
  z_u=
  [z_c,z_1^r,\ldots,z_P^r]
  \in\mathbb{R}^{k_c+K_r},
  \qquad
  K_r=\sum_{j=1}^{P}k_j^r.
  \label{eq:two-scale-output-vector}
\end{equation}

The selected model retains 99.5\% variance in each residual PCA. A slightly
higher threshold than the legacy 99\% local-output setting is useful because
the residual contains lower-amplitude corrections whose tail components can
still affect interface fidelity. The coarse-representation ablation in
Section~\ref{sec:coarse-global-design} quantifies this choice and
shows that the additional residual capacity improves both reconstruction and
interface diagnostics at negligible measured end-to-end cost.

\subsection{Block-balanced latent learning}
\label{subsec:block-balanced-latent-learning}

The unchanged PCA-Net MLP predicts the full two-scale target,
\begin{equation}
  g_\theta(z_x)
  =
  [\widehat z_c,\widehat z_1^r,\ldots,\widehat z_P^r].
  \label{eq:two-scale-latent-map}
\end{equation}
A conventional MSE over the concatenated vector implicitly weights the two
representational blocks according to their numbers of coordinates. Because
$K_r\gg k_c$, the residual coordinates dominate the aggregate objective even
though the coarse block carries the shared global structure.

We therefore optimize
\begin{equation}
  \mathcal{L}_{\mathrm{block}}
  =
  \frac{1}{2}\operatorname{MSE}
  (\widehat z_c,z_c)
  +
  \frac{1}{2}\operatorname{MSE}
  (\widehat z_r,z_r),
  \label{eq:two-scale-block-balanced-loss}
\end{equation}
where each MSE averages over its own coordinate block and the minibatch. The
equal block weighting therefore balances the two representational
\emph{roles}, rather than weighting every latent coordinate identically \cite{kendall2018multitask}.

A score-normalized alternative is evaluated in
Section~\ref{sec:latent-objective}. Both structured objectives substantially
improve over ordinary vector MSE, but equal block balancing provides the more
consistent cross-problem physical-field behavior and is consequently used as
the default.

\subsection{Decoding, assembly, and artifact reduction}
\label{subsec:two-scale-decoding-assembly}

At inference, the residual patches are decoded as
\begin{equation}
  \widehat r_j=
  (S_j^r)^{-1}
  \left(m_j^r+\Phi_j^r\widehat z_j^r\right).
  \label{eq:residual-decode}
\end{equation}
Let $\mathcal{A}$ denote the differentiable weighted patch assembler,
\begin{equation}
  \mathcal{A}(\widehat r)_\xi
  =
  \frac{
  \sum_qW_q(\xi)\widehat r_q(\xi)}
  {\sum_qW_q(\xi)+\epsilon},
  \label{eq:weighted-residual-assembly}
\end{equation}
where $\xi$ indexes a fine-grid location and $W_q$ is the assembly weight of
a patch covering that location. For the default nonoverlapping topology,
$W_q=1$ and Eq.~\eqref{eq:weighted-residual-assembly} reduces to direct
placement. The same operator supports the weighted overlap baseline and has
the usual partition-of-unity interpretation
\cite{babuska1997pum,dhingra2026localized}.

The final prediction is
\begin{equation}
  \widehat u=
  u_G(\widehat z_c)
  +
  \mathcal{A}
  (\widehat r_1,\ldots,\widehat r_P).
  \label{eq:two-scale-final-decode}
\end{equation}

Equation~\eqref{eq:two-scale-final-decode} makes the intended division of
labor explicit. Plain L2L allows every patch to independently determine its
local mean, amplitude, and long-wavelength shape. Two-scale first writes one
coherent coarse field across the entire domain and then asks the local
decoders only for residual corrections. Errors in those residual coordinates
therefore perturb an already coherent global prediction rather than defining
the complete patch independently.

This mechanism does not impose exact interface continuity: independently
decoded residuals can still disagree in value or normal derivative. The
mechanism study in Section~\ref{sec:mechanism-ablation} directly tests this
distinction by applying the same interface-aware fine-tuning to both the
plain-L2L and two-scale representations. The large improvement produced by
the representation change, compared with the smaller effect of fine-tuning
plain L2L, supports the representation-first interpretation.

\subsection{Computational characteristics}
\label{subsec:two-scale-computational-characteristics}

The coarse branch is intentionally small: its spatial feature dimension is
$(D/c)^2$ and only $k_c=10$ coordinates are retained. Its PCA fit and latent
prediction therefore add a relatively small stage to the localized pipeline.
The dominant representation work remains the local input PCA and local
residual PCA.

The principal comparison is with Hann-overlap L2L. Under the fixed
four-patch-per-axis topology, plain and two-scale models use $4\times4=16$
patches, whereas halving the stride produces a $7\times7=49$-patch overlap
layout. Overlap consequently requires substantially more local PCA fits and
decoded patch predictions.

Two-scale additionally performs one coarse-field prolongation and one dense
field addition, both linear in the number of fine-grid points. We therefore
evaluate computational efficiency using the complete measured pipeline rather
than PCA fitting alone: PCA construction, latent transforms, neural-network
training, inference, and cumulative end-to-end time are reported separately.
The resolution study in Section~\ref{sec:resolution-scaling} confirms the
intended distinction: two-scale approximately halves the measured PCA-fit
cost relative to overlap across the tested resolutions, whereas the
end-to-end difference is smaller because neural-network training remains the
dominant stage.

\section{Interface-Aware Physical-Field Fine-Tuning}
\label{sec:interface-aware-physical-field-finetuning}

The two-scale representation substantially reduces the patch-scale bias, but
its independently decoded residual patches do not impose exact agreement of
values or derivatives at their interfaces. Related compatibility conditions
appear in classical domain decomposition and, more recently, in neural
subdomain methods that communicate solution and derivative information across
artificial interfaces
\cite{toselli2005ddm,farhat1991feti,dolean2024multilevel,
hu2025schwarz,olson2026twolevel}.

\begin{figure}[htbp]
    \centering
    \includegraphics[width=\linewidth]{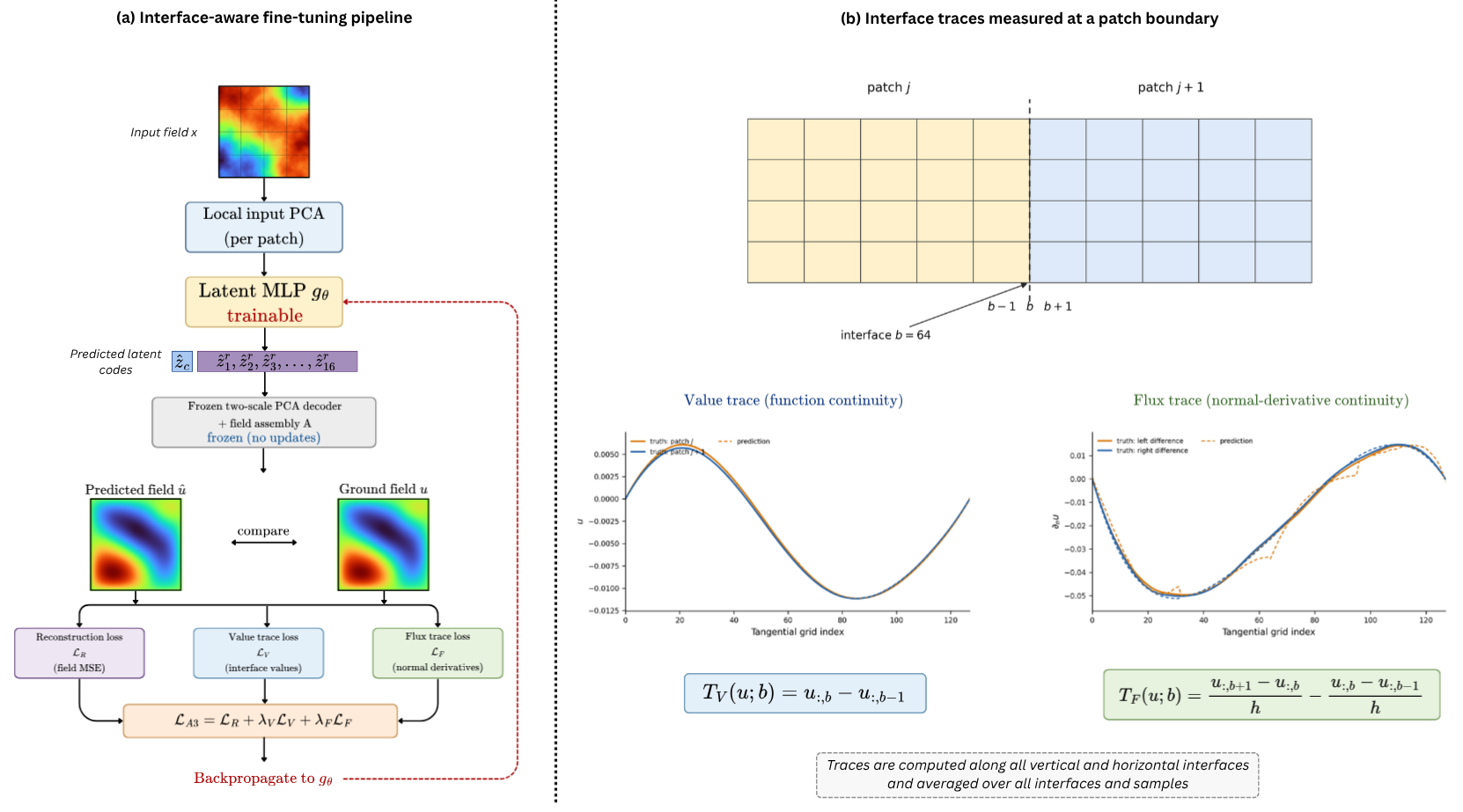}
    \caption{\textbf{Interface-aware physical-field fine-tuning of the pretrained two-scale model.}
(a) The trained latent mapper $(g_\theta)$ is warm-started from Method I and fine-tuned through a frozen differentiable two-scale PCA decoder and field assembler. The predicted latent codes are decoded to the assembled physical field $(\widehat{u})$, which is compared with the ground-truth field (u) using reconstruction and interface-aware objectives. The selected A3 objective is $(\mathcal{L}_{A3}=\mathcal{L}_{R}+\lambda_V\mathcal{L}_{V}+\lambda_F\mathcal{L}_{F})$. Gradients propagate through the fixed decoder and assembly operations, while only the parameters of $(g_\theta)$ are updated; the inference architecture is unchanged. (b) Interface supervision is evaluated along the vertical and horizontal boundaries of the nonoverlapping patch layout. The value trace $(T_V(u;b)=u_{:,b}-u_{:,b-1})$ measures the adjacent-field variation across interface (b), while the normal-derivative trace compares one-sided finite differences on either side of the interface. The losses match the predicted traces to their ground-truth counterparts rather than forcing them to vanish, thereby targeting the residual interface errors that remain after two-scale representation learning}
    \label{fig:interface-aware-finetuning}
\end{figure}

We therefore introduce an optional second stage that fine-tunes a trained
two-scale latent map using objectives evaluated on the assembled physical
field. The PCA representation and decoder remain fixed; only the parameters
of the latent predictor are updated. This distinction is important: the stage
is intended to refine the residual interface error left by the two-scale
representation, rather than replace the representation itself. The mechanism
study in Section~\ref{sec:mechanism-ablation} directly tests this distinction.
Figure~\ref{fig:interface-aware-finetuning} summarizes the gradient path and
selected interface objective.

\subsection{Differentiable physical-field decoder}
\label{subsec:differentiable-physical-field-decoder}

Starting from a trained block-balanced two-scale model, we freeze the local
input PCA models, coarse-global PCA basis, residual-patch PCA bases,
standardization statistics, and field assembler. For a localized input code
$z_x$, the trainable latent map produces
\begin{equation}
  \widehat z_u=g_\theta(z_x),
\end{equation}
and the fixed two-scale decoder introduced in
Section~\ref{subsec:two-scale-decoding-assembly} produces
\begin{equation}
  \widehat u=\mathcal{D}(\widehat z_u).
  \label{eq:differentiable-physical-decoder}
\end{equation}

The decoder performs the same inverse PCA, coarse prolongation, residual
decoding, patch assembly, and coarse--residual addition used at inference.
All of these operations are differentiable with respect to
$\widehat z_u$. Consequently, a loss evaluated on $\widehat u$ can update
$\theta$ while leaving the fitted representation unchanged.

Differentiating physical-space or physics-based objectives through an operator
decoder follows the broader physics-informed operator-learning paradigm
\cite{li2021pino,wang2021pideeponet,eshaghi2025vino}.
Unlike those approaches, however, the present stage does not retrain the
operator representation or introduce a new physical solver; it fine-tunes the
existing reduced-coordinate map through its fixed decoder.

\subsection{Assembled-field reconstruction and interface objectives}
\label{subsec:assembled-field-objectives}

The reconstruction term is the relative $L_2$ error of the decoded physical
field,
\begin{equation}
  \mathcal{L}_{R}
  =
  \frac{\|\widehat u-u\|_2}
       {\|u\|_2+\epsilon}.
  \label{eq:physical-reconstruction-loss}
\end{equation}
Unlike latent-space supervision, this objective is evaluated after inverse
PCA, coarse prolongation, residual decoding, and field assembly. It therefore
optimizes the same physical prediction used for evaluation.

The interface objectives use the truth-referenced traces defined in
Eqs.~\eqref{eq:interface-value-trace-problem}
and~\eqref{eq:interface-flux-trace-problem}. Let $\mathcal{B}$ denote the
internal horizontal and vertical interfaces induced by the nonoverlapping
patch topology. The value-trace loss is
\begin{equation}
  \mathcal{L}_{V}
  =
  \operatorname{mean}_{b\in\mathcal{B}}
  \left|
  T_V(\widehat u;b)-T_V(u;b)
  \right|,
  \label{eq:value-trace-loss}
\end{equation}
and the normal-derivative trace loss is
\begin{equation}
  \mathcal{L}_{F}
  =
  \operatorname{mean}_{b\in\mathcal{B}}
  \left|
  T_F(\widehat u;b)-T_F(u;b)
  \right|.
  \label{eq:flux-trace-loss}
\end{equation}

Both terms are deliberately truth referenced. They do not impose zero
variation at a patch boundary; instead, they penalize the difference between
the predicted and true variation across the same location. This distinction
prevents legitimate physical gradients from being treated as artificial
discontinuities.

Value and derivative transmission conditions have long been used to couple
subdomains in numerical methods and remain important in contemporary
neural-domain-decomposition formulations
\cite{mandel1993bdd,farhat1991feti,jagtap2020cpinn,
jagtap2020xpinn,hu2025schwarz,olson2026twolevel}.
Our use is narrower: Eqs.~\eqref{eq:value-trace-loss}
and~\eqref{eq:flux-trace-loss} supervise one assembled surrogate field rather
than coupling independently solved PDE subproblems.

The quantity $T_F$ remains a normal first-derivative jump proxy, not a
coefficient-weighted physical flux. In particular, for Darcy flow it is not
$a\,\partial_n u$; coefficient-weighted conservation is instead assessed by
the Darcy residual diagnostic. We therefore refer to
$\mathcal{L}_F$ as a normal-derivative trace loss throughout the manuscript.

% \subsection{Selected fine-tuning objective}
% \label{subsec:selected-finetuning-objective}

% The general differentiable loss infrastructure also supports radially binned
% spectral matching and generator-consistent Poisson or Darcy PDE-residual
% terms. Such physical residual losses are standard ingredients of
% physics-informed learning, although their optimization can be delicate
% \cite{raissi2019pinn,karniadakis2021piml,kharazmi2021vpinn,li2021pino}.
% The selected method, denoted A3, uses only reconstruction and the two
% truth-referenced interface terms:
% \begin{equation}
%   \boxed{\mathcal{L}_{\mathrm{A3}}=
%   \mathcal{L}_{R}+\lambda_V\mathcal{L}_{V}+
%   \lambda_F\mathcal{L}_{F}.}
%   \label{eq:selected-a3-objective}
% \end{equation}
% Hence, $\lambda_E=\lambda_P=0$ in the final configuration. The loss-ladder
% ablation motivates this selection: reconstruction-only fine-tuning is the
% pointwise-accuracy endpoint, while adding value and flux traces gives the
% selected accuracy--continuity tradeoff (see section~\ref{sec:interface-loss-ladder} for more information). The explicit spectral term is retained
% only for spectrum-focused ablations, and the tested domain-wide PDE-residual
% term is not part of A3. There is no RefinementNet, Hann post-processing, or
% additional learned module at inference; A3 changes the MLP parameters only.

\subsection{Selected interface-aware objective}
\label{subsec:selected-finetuning-objective}

The differentiable training infrastructure additionally supports a radially
binned spectral objective and generator-consistent Poisson or Darcy residual
terms. Physics- and variationally informed neural operators commonly augment
data objectives with such physical constraints
\cite{raissi2019pinn,karniadakis2021piml,li2021pino,
eshaghi2025vino}.
Their inclusion, however, does not guarantee a better multi-objective
optimization problem; competing loss terms and derivative scales can produce
substantial optimization imbalance
\cite{wang2021gradientpathologies,wang2022pinnntk,
hao2024pinnacle,chen2025brdr}.

The selected variant, denoted A3, therefore retains only reconstruction and
the two interface terms,
\begin{equation}
  \boxed{
  \mathcal{L}_{\mathrm{A3}}
  =
  \mathcal{L}_{R}
  +
  \lambda_V\mathcal{L}_{V}
  +
  \lambda_F\mathcal{L}_{F}.
  }
  \label{eq:selected-a3-objective}
\end{equation}
The spectral and PDE-residual weights are zero in the final configuration.

This selection is empirical rather than assumed a priori. The ordered loss
ablation in Section~\ref{sec:interface-loss-ladder} shows that
reconstruction-only fine-tuning provides the strongest pointwise-accuracy
endpoint, whereas adding value and normal-derivative traces produces the
selected accuracy--continuity tradeoff. The tested spectral and domain-wide
PDE-residual additions optimize their intended quantities but do not improve
the overall default tradeoff. We therefore treat them as ablation objectives,
not components of the final method.

A3 introduces no RefinementNet, Hann post-processing step, or additional
trainable inference module. It modifies only the parameters of
$g_\theta$. Consistent with this role, the results in
Section~\ref{sec:mechanism-ablation} show that A3 provides a comparatively
modest improvement when applied directly to plain L2L, but a much stronger
interface refinement after the two-scale representation has removed the
dominant global incompatibility.

\subsection{Loss calibration and optimization}
\label{subsec:loss-calibration-optimization}

The reconstruction and interface losses have different numerical scales, so
their raw coefficients are not transferable across datasets, resolutions, or
warm-started models. Composite physics-informed objectives are known to be
sensitive to this imbalance, motivating a range of adaptive or normalized
weighting strategies
\cite{wang2021gradientpathologies,wang2022pinnntk,
mcclenny2021sapinn,yu2022gpinn,hao2024pinnacle,chen2025brdr}.

For each warm-started run, we evaluate the active unweighted terms on up to
128 validation fields and set
\begin{equation}
  \lambda_t
  =
  \rho_t
  \frac{\mathcal{L}_{R,0}}
       {\mathcal{L}_{t,0}},
  \qquad
  t\in\{V,F\},
  \label{eq:interface-loss-calibration}
\end{equation}
where the subscript $0$ denotes the pretrained two-scale model. We use
\begin{equation}
  \rho_V=\rho_F=0.2,
\end{equation}
so that each interface term begins with a prescribed contribution relative to
the reconstruction objective. The resulting weights are resolved separately
for every run rather than treated as universal constants.

Fine-tuning warm-starts from the matching block-balanced two-scale checkpoint
and runs for 50 epochs with Adam at learning rate $10^{-4}$ and batch size 32.
The first five epochs use $\mathcal{L}_R$ alone. The interface weights are then
ramped linearly from zero to their calibrated values over ten epochs and
remain fixed thereafter. The best checkpoint under the full A3 validation
objective is retained.

The additional optimization requires differentiable full-field decoding,
assembly, and interface-stencil evaluation during training, but does not alter
the inference path. For this reason, all reported A3 costs are cumulative and
include the pretrained two-scale run. The mechanism and resolution studies in
Sections~\ref{sec:mechanism-ablation}
and~\ref{sec:resolution-scaling} quantify this offline
accuracy--continuity--cost tradeoff.

\section{Experimental Methodology}
\label{sec:experimental-methodology}

We evaluate the proposed representation through controlled comparisons of
reconstruction quality, interface fidelity, physical-field diagnostics, and
computational cost. Within each dataset--resolution--seed cell, compared
PCA-based methods use matched data splits, random seeds, PCA solvers, latent
architectures, and optimization budgets unless the corresponding study
explicitly changes one of these factors. This paired design is intended to
isolate the effect of the representation rather than differences in data or
training protocol. Such controlled benchmarking is particularly important
when comparing data-driven PDE surrogates with substantially different
representation and computational characteristics
\cite{takamoto2023pdebench,koumoutsakos2025benchmark}.

The benchmark generators and inherited PCA-Net baselines follow our preceding
localized PCA-Net study \cite{dhingra2026localized}. Retaining the same
operators also permits direct comparison with the previously proposed
overlap- and RefinementNet-based artifact mitigations.

\subsection{Benchmark operators}
\label{subsec:benchmark-operators}

We consider steady scalar elliptic operators on the unit square,
discretized on a uniform $D\times D$ grid including the boundary, with
$h=1/(D-1)$. Both problems impose homogeneous Dirichlet boundary conditions.

\subsubsection{Poisson equation}
\label{subsubsec:poisson-equation}

For Poisson, the input is a forcing field $f$ and the output satisfies the
generator convention
\begin{equation}
  \Delta u=f,
  \qquad
  u\rvert_{\partial\Omega}=0.
  \label{eq:poisson-benchmark}
\end{equation}
Forcing fields are sampled from a zero-mean Gaussian random field using the
DCT-based generator of the preceding study. With independent standard-normal
coefficients $\xi_{k\ell}$,
\begin{equation}
  \widehat f_{k\ell}
  =
  D\,\tau^{\alpha-1}
  \left[
    \pi^2(k^2+\ell^2)+\tau^2
  \right]^{-\alpha/2}
  \xi_{k\ell},
  \qquad
  \widehat f_{00}=0,
  \label{eq:grf-generator}
\end{equation}
followed by an orthonormal inverse DCT. We use
$(\alpha,\tau)=(2,3)$.

Solutions are obtained with the interior five-point discrete Laplacian using
a batched type-I discrete sine transform solve. Consistent with the data
generator, all reported Poisson residuals use
\begin{equation}
  \mathcal{R}_{P}(\widehat u)
  =
  \Delta_h\widehat u-f.
  \label{eq:poisson-eval-residual}
\end{equation}
No opposite-sign residual values are included in the manuscript results.

\subsubsection{Darcy equation}
\label{subsubsec:darcy-equation}

For the heterogeneous Darcy problem, the permeability field $a$ is the input
and the solution satisfies
\begin{equation}
  -\nabla\cdot(a\nabla u)=1,
  \qquad
  u\rvert_{\partial\Omega}=0.
  \label{eq:darcy-benchmark}
\end{equation}
A GRF $g$ is generated using the same $(\alpha,\tau)=(2,3)$ parameters and
thresholded pointwise,
\begin{equation}
  a(\mathbf{x})
  =
  \begin{cases}
    12, & g(\mathbf{x})\geq0,\\
    4,  & g(\mathbf{x})<0.
  \end{cases}
  \label{eq:darcy-coefficient}
\end{equation}
The discrete operator uses arithmetic face averages of $a$ and centered
second-order differences. The reported Darcy residual uses the identical
face-average discretization,
\begin{equation}
  \mathcal{R}_{D}(\widehat u)
  =
  -\nabla_h\cdot(a\nabla_h\widehat u)-1.
  \label{eq:darcy-eval-residual}
\end{equation}
Thus, the physical residual diagnostics for both benchmarks are evaluated
using the same discrete operators used to generate their solutions.

\subsection{Datasets, resolutions, and patch geometry}
\label{subsec:dataset-sizes-resolutions}

Each processed dataset contains 10,000 independently generated input--solution
pairs. For each run seed, the complete dataset is shuffled into
80/10/10\% training, validation, and test partitions, giving 8,000,
1,000, and 1,000 samples, respectively, in the standard experiments.
The same resolved split is used by all methods compared within a
dataset--resolution--seed cell.

Poisson is evaluated independently at
\begin{equation}
  D\in\{64,128,256\},
\end{equation}
while the heterogeneous Darcy transfer study uses $D=256$. The datasets at
different resolutions contain independently generated realizations.
Consequently, the resolution study in
Section~\ref{sec:resolution-scaling} evaluates robustness under increasing
spatial resolution rather than pointwise grid convergence of identical
samples.

All localized methods use a fixed $4\times4$ nonoverlapping topology. The
patch widths are therefore
\begin{equation}
  p\in\{16,32,64\}
\end{equation}
for $D=64$, 128, and 256, respectively. Plain L2L and two-scale use
$s=p$, whereas the Hann-overlap baseline uses $s=p/2$, producing a
$7\times7$ patch layout. The sample-efficiency study is the only deliberate
change in training-set size: on Poisson-$128^2$, it uses
\[
m\in\{1250,2000,4000,8000\}
\]
with matched validation and test protocols.

\subsection{Baselines and controlled training configuration}
\label{subsec:baselines-training}

We compare the proposed method with four PCA-Net variants and one full-field
neural-operator reference.

\begin{itemize}
  \item \textbf{Global PCA-Net} uses one input and one output PCA basis over
  the full field, each retaining 99\% variance.

  \item \textbf{Plain L2L} uses independent 99\%-variance input and output
  PCA models on the nonoverlapping patch topology introduced in
  Section~\ref{subsec23}.

  \item \textbf{L2L + Hann overlap} retains the same patch width but halves
  the stride and combines redundant local reconstructions using the safe
  Hann-weighted assembly of the preceding study
  \cite{dhingra2026localized}.

  \item \textbf{L2L + RefinementNet} applies the inherited four-layer
  convolutional correction network to assembled plain-L2L predictions
  \cite{dhingra2026localized}.

  \item \textbf{FNO} provides a higher-capacity full-field operator-learning
  reference \cite{li2021fno}. We use four Fourier layers, width 32, 12
  retained Fourier modes per spatial direction, and eight pixels of padding.
\end{itemize}

The complete Poisson-$128^2$ baseline comparison is reported in
Section~\ref{sec:poisson128-results}. Other studies retain only
the subset required to isolate their corresponding design question.

All PCA-Net variants use training-split standardization and randomized SVD
with oversampling 20 and four power iterations. Conventional input and output
PCA representations retain 99\% variance. Unless explicitly varied in an
ablation, two-scale uses the configuration selected in Section~7:
\[
c=4,
\qquad
k_c=10,
\qquad
99.5\%\ \text{residual variance}.
\]
The latent map is the same three-hidden-layer, width-128 ReLU MLP for Global
PCA-Net, L2L, overlap, and two-scale. It is trained with Adam
\cite{kingma2015adam} for at most 500 epochs using learning rate $10^{-3}$,
weight decay $10^{-4}$, and batch size 32.

Plain, global, and overlap models minimize ordinary output-score MSE.
Two-scale uses the block-balanced objective defined in
Eq.~\eqref{eq:two-scale-block-balanced-loss}. The interface-aware model uses
the A3 objective and warm-start protocol of
Section~\ref{sec:interface-aware-physical-field-finetuning}; these
methodological definitions are not changed by the experimental harness.

FNO is trained with Adam for at most 100 epochs at learning rate $10^{-3}$.
Poisson uses training/evaluation batch sizes 16/64, while Darcy-$256^2$ uses
4/8 because of the larger memory requirement. Validation-based learning-rate
reduction and early stopping are applied consistently within the FNO runs.

\subsection{Evaluation metrics}
\label{subsec:evaluation-metrics}

All predictive metrics are evaluated on the held-out test set in physical
field units. We report MSE and MAE together with the relative $L_2$ error
used throughout the PCA-Net studies,
\begin{equation}
  \operatorname{MRE}
  =
  \frac{\|\widehat u-u\|_2}
       {\|u\|_2}.
  \label{eq:test-mre}
\end{equation}
Values displayed as percentages in the results are
$100\times\operatorname{MRE}$. SSIM is computed independently for each test
field using the target field's dynamic range and then averaged
\cite{wang2004ssim}.

For every PCA-based method, we additionally evaluate a \emph{PCA oracle} by
encoding and decoding the true test solution through the fitted output
representation. Comparing its error with the learned prediction separates
representation capacity from latent-score prediction error. This distinction
is used explicitly in the coarse-rank, sample-efficiency, and heterogeneous
Darcy analyses.

Field-energy agreement is measured by the relative $L_2$ difference between
the mean radially binned two-dimensional Fourier energy spectra of predictions
and targets. Because radial averaging removes phase and spatial location, this
quantity is treated as a global spectral diagnostic rather than a direct
measure of tiling.

Patch-interface behavior is evaluated using the traces introduced in
Section~\ref{subsec24}. We report both the raw value jump and raw
normal-derivative jump at the configured seams, together with the
truth-referenced value- and normal-derivative trace errors. Because overlap
and nonoverlap models have different native seam locations, direct continuity
comparisons also use a common seam set corresponding to the plain-L2L stride.

Physical consistency is assessed by the mean-absolute and RMS values of
Eqs.~\eqref{eq:poisson-eval-residual}
and~\eqref{eq:darcy-eval-residual}.
The loss and evaluation implementations share the same discrete stencils, so
the physical quantities optimized during assembled-field fine-tuning coincide
with the quantities reported during evaluation.

Finally, computational cost is measured stage-wise:
\begin{equation}
  T_{\mathrm{total}}
  =
  T_{\mathrm{PCA}}
  +
  T_{\mathrm{transform}}
  +
  T_{\mathrm{train}}
  +
  T_{\mathrm{inference}},
\end{equation}
with PCA time further separated into input, coarse/global, local-output, and
residual-output components where applicable. Warm-started A3 results report
\emph{cumulative} cost, including construction and training of the base
two-scale model. This accounting is used in the headline and resolution
quality--cost comparisons.

\subsection{Statistical, timing, and reproducibility protocol}
\label{subsec:statistical-protocol}

Comparisons use paired random seeds. Within a given
dataset--resolution--seed cell, all methods receive the same train/validation/
test split and the same experiment seed; randomized PCA and neural-network
initialization are then generated deterministically from that seed. This
provides paired comparisons while allowing methods with different latent
dimensions or architectures to retain their appropriate parameter shapes.

The headline Poisson-$128^2$, mechanism, Darcy-$256^2$, and
full-versus-randomized-SVD studies use five paired seeds. The resolution,
coarse-representation, interface-loss, and sample-efficiency studies use
three paired seeds. Unless stated otherwise, aggregate results are reported as
mean $\pm$ standard deviation. The cross-problem latent-objective study uses
one matched seed per cell and is therefore interpreted only as a
design-selection experiment, not as a statistical ranking.

All wall-clock comparisons are executed on the same benchmark environment:
\begin{quote}
\textbf{[Intel Core i9-14900KF (16 cores, 32 threads), NVIDIA GeForce RTX 4090 (23.99~GiB), 31~GiB visible to the Linux runtime, PYTORCH/CUDA 2.5.0+cu121].}
\end{quote}
PCA fitting is performed on the CPU through NumPy/scikit-learn, whereas neural
training and inference use the stated GPU. No timing comparisons combine runs
from different hardware environments.

Each run archives the resolved experiment configuration, configuration hash,
split indices, random seed, retained PCA dimensions, model checkpoint, loss
history, evaluation metrics, software environment, and stage-wise timings.
Seed-zero predictions are retained for deterministic qualitative
visualization, while quantitative claims use all planned seeds. These records
provide the provenance required to reproduce the paired comparisons without
assuming bitwise identity across different CUDA or library versions.

\section{Results}
\label{sec6:results}

\subsection{Artifact reduction and reconstruction accuracy on Poisson-128}
\label{sec:poisson128-results}

We first test the central representation hypothesis on the
$128\times128$ Poisson benchmark using the full baseline set defined in
Section~\ref{sec:experimental-methodology}. Results are reported as mean and
standard deviation over five paired seeds. Unless otherwise stated, all
methods use the default configurations and matched experimental protocol
described in Section~\ref{sec:experimental-methodology}.

\begin{table*}[htbp]
  \centering
  \caption{Poisson-128 headline comparison over five paired seeds. Lower MRE
  and cumulative time are better; higher SSIM is better. The PCA-fit column
  isolates representation construction from neural-operator training. FNO
  does not require a PCA fitting stage.}
  \label{tab:poisson128-headline}
  \small
  \begin{tabular}{lrrrr}
    \toprule
    Method & MRE (\%) $\downarrow$ & SSIM $\uparrow$ & PCA fit (s) $\downarrow$ & Cumulative total (s) $\downarrow$ \\
    \midrule
    Global PCA-Net & $7.543 \pm 0.166$ & $0.9236 \pm 0.0021$ & $5.2 \pm 0.1$ & $192.5 \pm 11.5$ \\
    Plain L2L & $4.829 \pm 0.094$ & $0.9427 \pm 0.0013$ & $3.2 \pm 0.1$ & $181.2 \pm 2.9$ \\
    L2L + overlap & $2.308 \pm 0.050$ & $0.9850 \pm 0.0004$ & $9.9 \pm 0.2$ & $208.8 \pm 2.1$ \\
    L2L + RefinementNet & $3.210 \pm 0.058$ & $0.9747 \pm 0.0006$ & $3.2 \pm 0.1$ & $275.9 \pm 2.3$ \\
    FNO & $\mathbf{0.242 \pm 0.013}$ & $\mathbf{0.9998 \pm 0.0000}$ & -- & $725.4 \pm 4.0$ \\
    \midrule
    \textbf{Two-scale (ours)} & $1.149 \pm 0.022$ & $0.9966 \pm 0.0002$ & $5.0 \pm 0.2$ & $218.6 \pm 15.8$ \\
    Two-scale + interface & $1.083 \pm 0.023$ & $0.9967 \pm 0.0001$ & $5.0 \pm 0.2$ & $416.6 \pm 84.9$ \\
    \bottomrule
  \end{tabular}
\end{table*}

\begin{figure*}[htbp]
  \centering
  \includegraphics[width=\textwidth]{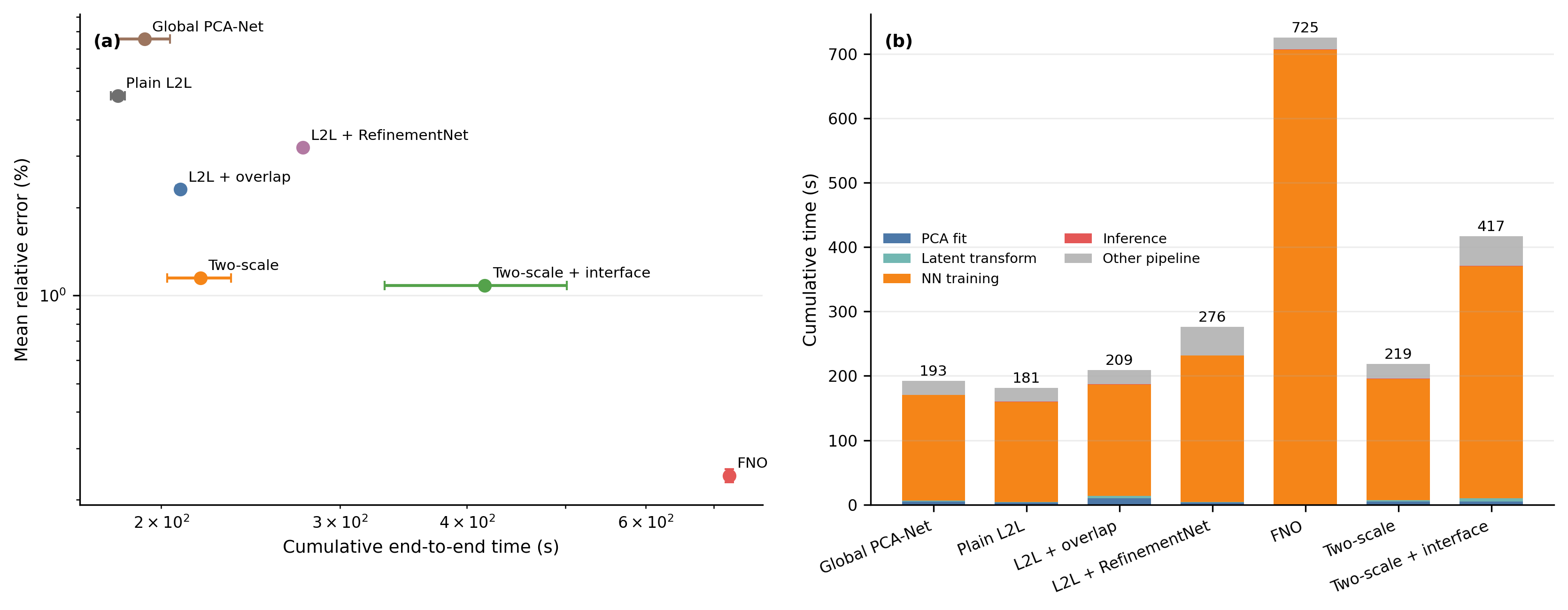}
  \caption{Poisson-128 quality--cost comparison.
  \textbf{Left:} mean relative error against cumulative end-to-end time;
  horizontal error bars denote one standard deviation in wall time.
  \textbf{Right:} stage-wise cumulative time. Two-scale occupies a favorable
  PCA-Net operating point, providing substantially lower reconstruction error
  than the local baselines at a cost close to overlap.}
  \label{fig:poisson128-quality-cost}
\end{figure*}

\paragraph{Two-scale localization substantially improves the PCA-Net family.}
Table~\ref{tab:poisson128-headline} and
Fig.~\ref{fig:poisson128-quality-cost} show a clear separation in
reconstruction quality within the PCA-based models. Two-scale achieves
$1.149\pm0.022\%$ MRE and $0.9966\pm0.0002$ SSIM, reducing MRE by
$76.2\%$ relative to plain L2L, $50.2\%$ relative to L2L with overlap, and
$64.2\%$ relative to RefinementNet. The same ordering is preserved across all
five paired seeds.

This gain is obtained without moving the method into a substantially different
cost regime. Two-scale requires $218.6$ s of cumulative wall time, only
$4.7\%$ more than overlap while approximately halving its MRE. It is also
$20.8\%$ faster than RefinementNet while providing markedly lower
reconstruction error. FNO remains the absolute accuracy reference at
$0.242\%$ MRE, but requires $3.32\times$ the cumulative time of two-scale.
The intended operating regime of two-scale is therefore a compact,
PCA-reduced surrogate rather than a replacement for a full-field neural
operator when accuracy alone is the objective.

\begin{figure*}[htbp]
  \centering
  \includegraphics[width=\textwidth]{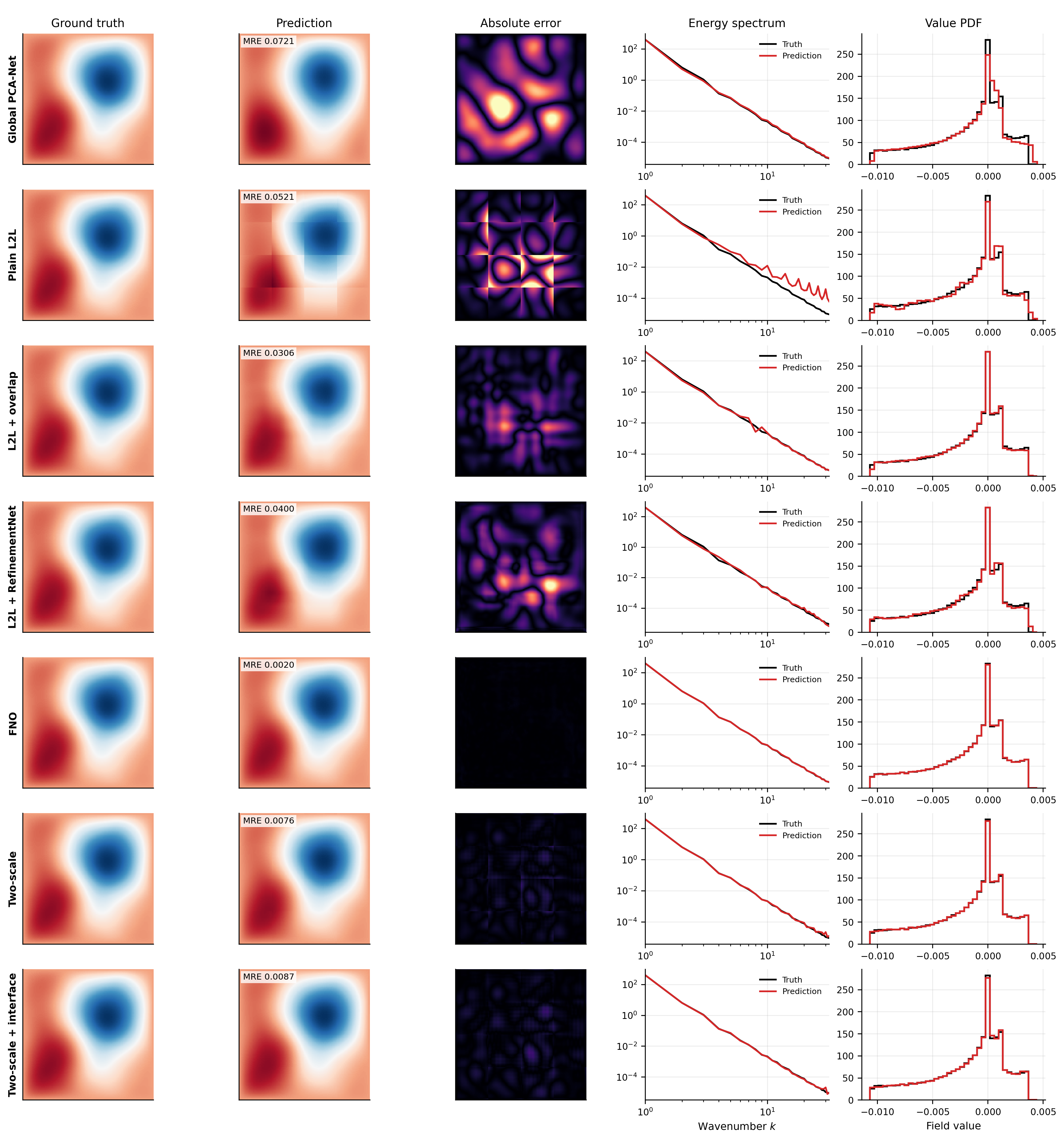}
  \caption{Representative Poisson-128 reconstruction, error, radial energy
  spectrum, and value-PDF diagnostics (seed 0, sample 5147). The example is
  selected deterministically as the seed-0 test field whose plain-L2L error is
  closest to the median. Plain L2L shows pronounced patch-aligned error,
  whereas two-scale removes the dominant tiled bias while preserving the
  global field structure. FNO is included as a high-capacity full-field
  reference.}
  \label{fig:poisson128-qualitative}
\end{figure*}

\paragraph{The dominant blockwise reconstruction error is strongly reduced.}
Figure~\ref{fig:poisson128-qualitative} shows the deterministically selected
seed-0 test case whose plain-L2L error is closest to the median. Plain L2L
exhibits piecewise offsets and bright error ridges aligned with the
$32\times32$ patch boundaries. Hann overlap suppresses the most prominent
seams, although its remaining error is still spatially structured. Two-scale
instead produces a substantially lower-amplitude and more spatially distributed
error field, with the dominant tiled pattern largely absent. Its predicted
value distribution also closely follows the target.

RefinementNet reduces visible patch boundaries but exhibits considerably
larger high-wavenumber spectral error, with a relative spectral error of
$3.02\times10^{-2}$ compared with $5.48\times10^{-3}$ for two-scale. FNO
provides the closest reconstruction overall and is retained as the
higher-capacity full-field reference.

\begin{figure*}[htbp]
  \centering
  \includegraphics[width=0.86\textwidth]{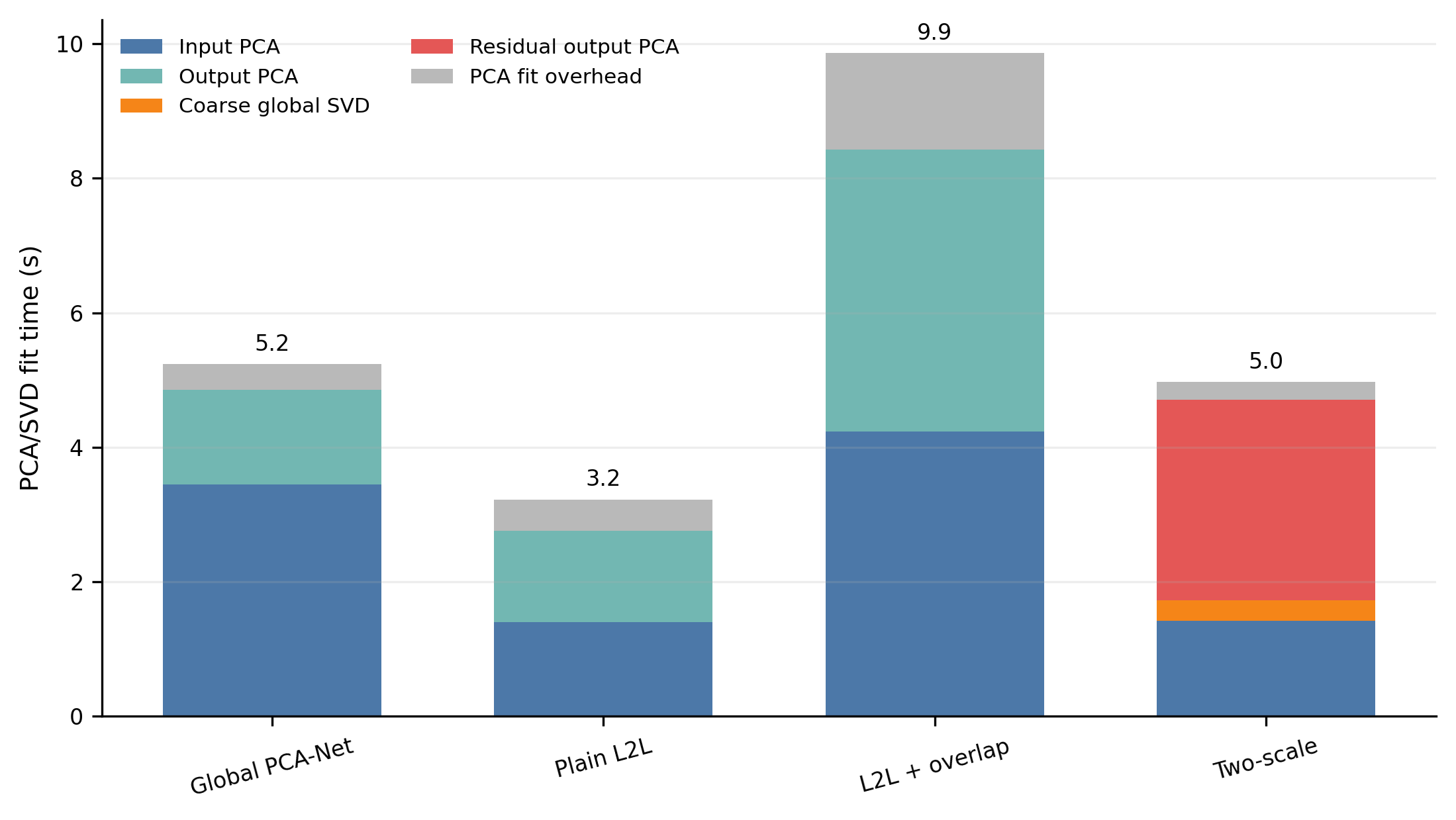}
  \caption{PCA/SVD fitting time on Poisson-128, decomposed by stage.
  Two-scale approximately halves the representation-construction cost of
  overlap while adding only a small coarse-global SVD.}
  \label{fig:poisson128-pca-breakdown}
\end{figure*}

\paragraph{Two-scale reduces representation-construction cost relative to overlap.}
The stage-wise PCA/SVD breakdown in
Fig.~\ref{fig:poisson128-pca-breakdown} shows that two-scale requires
$5.0\pm0.2$ s for representation construction, compared with
$9.9\pm0.2$ s for overlapping L2L. The added coarse-global SVD contributes
only a small fraction of this cost; the dominant two-scale output stage is
the residual PCA, whereas overlap incurs a larger local-output PCA cost from
its denser patch layout. Thus, the approximately twofold PCA-fit reduction
arises primarily from avoiding the redundant output representations required
by overlap.

Neural-network training remains the dominant end-to-end stage. Consequently,
the PCA-stage advantage is larger than the difference in cumulative runtime,
consistent with the distinction between representation-construction cost and
full-pipeline cost emphasized throughout this study.

\begin{figure}[htbp]
  \centering
  \includegraphics[width=\columnwidth]{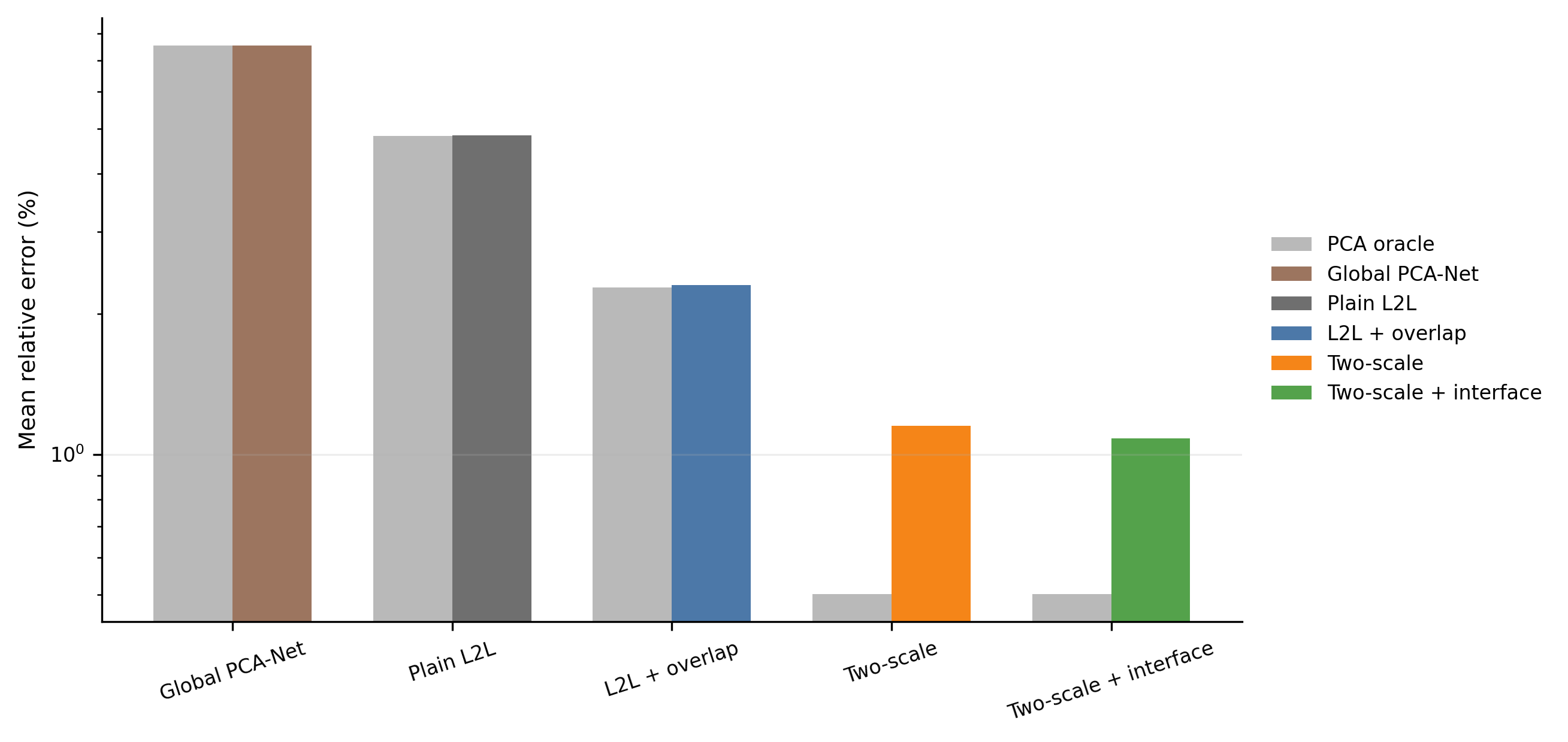}
  \caption{PCA-oracle and learned-operator MRE for the representational
  baselines. Two-scale has the lowest PCA reconstruction floor among the
  PCA-Net variants while retaining a substantial learned-to-oracle gap.}
  \label{fig:poisson128-representation-gap}
\end{figure}

\paragraph{The two-scale representation has a lower reconstruction floor but
retains latent-prediction headroom.}
Figure~\ref{fig:poisson128-representation-gap} separates output-representation
error from latent-score prediction error. When supplied with the true PCA
scores, the two-scale decoder reconstructs the test solutions at
$0.501\%$ MRE, compared with $1.149\%$ for the learned operator. The resulting
$2.29\times$ learned-to-oracle ratio indicates substantial remaining
headroom in the input-to-latent map.

By comparison, Global PCA-Net, plain L2L, and overlap operate much closer to
their respective PCA reconstruction floors. The two-scale representation
therefore provides both a lower attainable reconstruction error and additional
capacity that is not yet fully exploited by the present latent predictor. The
mechanism responsible for the learned improvement is examined directly in
Section~\ref{sec:mechanism-ablation}.

\paragraph{Interface-aware fine-tuning provides a secondary refinement.}
Starting from the two-scale model, interface-aware fine-tuning lowers MRE from
$1.149\pm0.022\%$ to $1.083\pm0.023\%$. It also reduces the
truth-referenced value-trace error by $66.5\%$ and the normal-derivative trace
error by $64.4\%$. These improvements are consistent across the five paired
seeds, but cumulative runtime increases to $416.6\pm84.9$ s. We therefore
retain plain two-scale as the default representation and treat the
interface-aware model as an optional accuracy--continuity refinement when the
additional offline cost is justified.

Interface fidelity nevertheless requires a geometry-controlled comparison
with overlap. Native overlap metrics are evaluated on stride-16 seams,
whereas the nonoverlapping methods use stride-32 seams, so those values should
not be interpreted as directly equivalent continuity tests. When all methods
are evaluated on the same stride-32 seam set, overlap retains the lowest
PCA-based value- and normal-derivative trace errors. Plain two-scale still
reduces both trace errors by more than $92\%$ relative to plain L2L while
providing the substantially lower global reconstruction error reported above.
The complete metric suite and common-seam audit are reported in
Appendix~\ref{app:poisson128-diagnostics}.

\subsection{Representation Change, Interface Refinement, and the
Accuracy--Continuity--Cost Tradeoff}
\label{sec:mechanism-ablation}
\label{sec:accuracy-continuity-cost}

The headline study establishes that two-scale PCA-Net substantially improves
Poisson-128 reconstruction. We next isolate the source of that improvement
and determine the role of the optional interface-aware objective. Specifically,
we compare plain L2L and two-scale under either their standard latent training
or the selected A3 physical-field fine-tuning objective. L2L with Hann overlap
is retained as a continuity-oriented reference. Results are aggregated over
five paired seeds.

Because this mechanism study was executed as a separate matched experiment,
its wall-clock values are interpreted within the timing set reported here
rather than compared directly with the independently executed headline timing
study in Section~\ref{sec:poisson128-results}.

\begin{table*}[htbp]
  \centering
  \caption{Mechanism ablation on Poisson-128 over five paired seeds. The
  interface objective is evaluated on identical output representations before
  and after fine-tuning. Trace metrics for overlap use its native stride-16
  seams, whereas the other methods use stride-32 seams; the common-stride
  comparison is reported later in this subsection.}
  \label{tab:mechanism-ablation-main}
  \small
  \begin{tabular}{lcccccc}
\toprule
Method
& MRE
& SSIM
& \makecell{Value\\trace}
& \makecell{Flux\\trace}
& \makecell{Spectral\\error}
& \makecell{Total\\time (s)} \\
\midrule

Plain L2L
& \makecell{$4.829$\\$\pm\,0.094$}
& \makecell{$0.9427$\\$\pm\,0.0013$}
& \makecell{$4.061{\times}10^{-4}$\\$\pm\,2.4{\times}10^{-6}$}
& \makecell{$5.428{\times}10^{-2}$\\$\pm\,3.1{\times}10^{-4}$}
& \makecell{$3.545{\times}10^{-3}$\\$\pm\,1.3{\times}10^{-4}$}
& \makecell{$179.9$\\$\pm\,3.5$}
\\

Plain L2L + interface
& \makecell{$4.574$\\$\pm\,0.089$}
& \makecell{$0.9465$\\$\pm\,0.0012$}
& \makecell{$3.289{\times}10^{-4}$\\$\pm\,2.0{\times}10^{-6}$}
& \makecell{$4.414{\times}10^{-2}$\\$\pm\,2.5{\times}10^{-4}$}
& \makecell{$4.058{\times}10^{-3}$\\$\pm\,2.9{\times}10^{-4}$}
& \makecell{$335.5$\\$\pm\,8.7$}
\\

Two-scale
& \makecell{$1.149$\\$\pm\,0.022$}
& \makecell{$0.9966$\\$\pm\,0.0002$}
& \makecell{$2.962{\times}10^{-5}$\\$\pm\,4.0{\times}10^{-7}$}
& \makecell{$4.304{\times}10^{-3}$\\$\pm\,5.4{\times}10^{-5}$}
& \makecell{$5.478{\times}10^{-3}$\\$\pm\,5.6{\times}10^{-4}$}
& \makecell{$202.6$\\$\pm\,2.7$}
\\

Two-scale + interface
& \makecell{$1.089$\\$\pm\,0.024$}
& \makecell{$0.9967$\\$\pm\,0.0001$}
& \makecell{$1.004{\times}10^{-5}$\\$\pm\,6.0{\times}10^{-7}$}
& \makecell{$1.549{\times}10^{-3}$\\$\pm\,9.2{\times}10^{-5}$}
& \makecell{$4.377{\times}10^{-3}$\\$\pm\,3.9{\times}10^{-4}$}
& \makecell{$364.4$\\$\pm\,13.2$}
\\

L2L + overlap
& \makecell{$2.308$\\$\pm\,0.050$}
& \makecell{$0.9850$\\$\pm\,0.0004$}
& \makecell{$8.204{\times}10^{-6}$\\$\pm\,3.1{\times}10^{-8}$}
& \makecell{$4.751{\times}10^{-4}$\\$\pm\,2.7{\times}10^{-6}$}
& \makecell{$1.522{\times}10^{-3}$\\$\pm\,3.1{\times}10^{-4}$}
& \makecell{$200.8$\\$\pm\,2.1$}
\\

\bottomrule
\end{tabular}
\end{table*}

\paragraph{The representation change is the dominant mechanism.}
Replacing the local-only output representation with the coarse-global plus
local-residual representation reduces MRE from
$4.829\pm0.094\%$ to $1.149\pm0.022\%$, a $76.2\%$ reduction
(Table~\ref{tab:mechanism-ablation-main}). Within the shared stride-32
geometry, the same change reduces the truth-referenced value- and
normal-derivative trace errors by $92.7\%$ and $92.1\%$, respectively.
These gains are achieved at a $12.7\%$ increase in cumulative runtime,
from $179.9\pm3.5$ s to $202.6\pm2.7$ s.

\begin{figure*}[htbp]
  \centering
  \includegraphics[width=\textwidth]
  {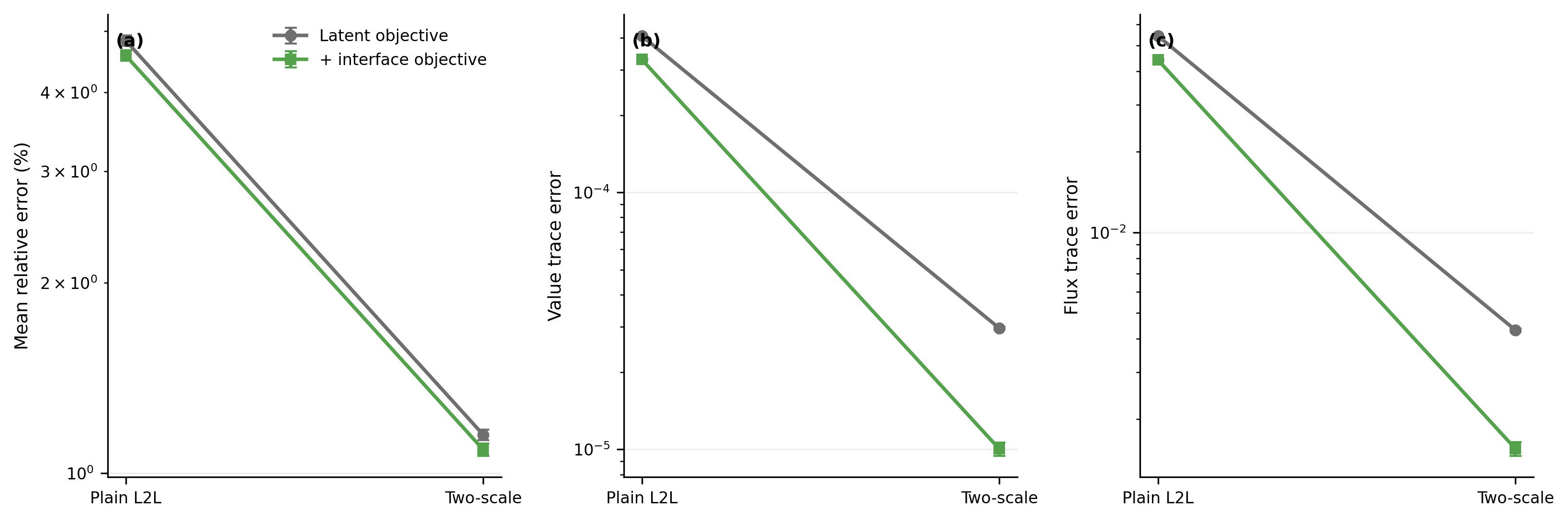}
  \caption{Interaction between output representation and interface-aware
  fine-tuning on Poisson-128. Gray circles denote latent-space training and
  green squares denote the same representation after A3 fine-tuning. The
  transition from plain L2L to two-scale produces the dominant reduction in
  MRE and interface errors; A3 then provides an additional refinement,
  particularly for the two-scale representation.}
  \label{fig:mechanism-interaction}
\end{figure*}

Figure~\ref{fig:mechanism-interaction} makes the interaction between
representation and fine-tuning explicit. Moving from plain L2L to two-scale
produces the dominant change in both reconstruction and interface metrics,
whereas fine-tuning acts as a secondary correction within a fixed output
representation.

\paragraph{Interface-aware fine-tuning cannot substitute for the representation
change.}
Applying A3 directly to plain L2L lowers MRE by only $5.3\%$, while reducing
the value- and normal-derivative trace errors by $19.0\%$ and $18.7\%$,
respectively. It also increases spectral error by $14.5\%$ and raises
cumulative runtime from $179.9\pm3.5$ s to $335.5\pm8.7$ s. Thus, the
physical-field objective can moderate the mismatch produced by the local
representation, but it does not remove its dominant tiled error structure.

\begin{figure*}[htbp]
  \centering
  \includegraphics[width=\textwidth]
  {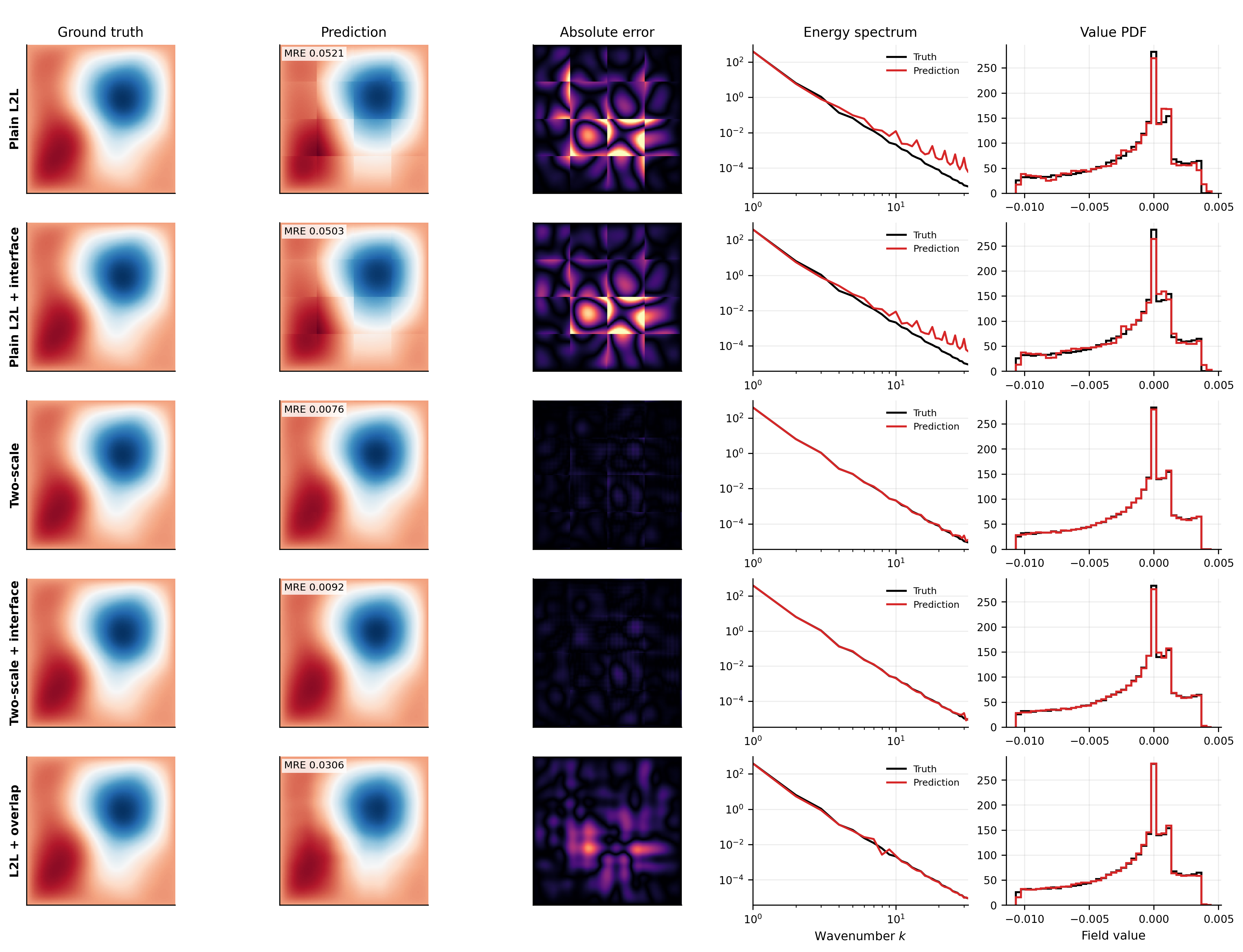}
  \caption{Matched Poisson-128 qualitative comparison (seed 0, sample 5147).
  Fine-tuning plain L2L attenuates some seam-aligned error but leaves the tiled
  reconstruction pattern visible. Two-scale removes the dominant blockwise
  bias, and its interface-aware refinement further reduces the remaining
  localized error.}
  \label{fig:mechanism-qualitative}
\end{figure*}

The qualitative comparison in Fig.~\ref{fig:mechanism-qualitative} shows the
same distinction directly. Fine-tuning plain L2L attenuates some seam-local
error but leaves the blockwise reconstruction pattern visible. Two-scale
removes the dominant patch-aligned bias before fine-tuning, after which A3
further reduces the remaining localized discrepancy.

\paragraph{A3 is most effective after the representation has been corrected.}
Starting from two-scale, A3 lowers MRE by a further $5.3\%$, from
$1.149\pm0.022\%$ to $1.089\pm0.024\%$. Its larger effect is on interface
fidelity: the value-trace error decreases by $66.1\%$ and the
normal-derivative trace error by $64.0\%$. Spectral error also decreases by
$19.3\%$. These improvements increase cumulative runtime from
$202.6\pm2.7$ s to $364.4\pm13.2$ s, corresponding to a $79.9\%$ cost
increase.

The resulting decision-oriented comparison is summarized in
Table~\ref{tab:tradeoff-decision}.

\begin{table*}[htbp]
  \centering
  \caption{Decision-oriented Poisson-128 comparison over five paired seeds.
  MRE and cumulative-time entries give mean $\pm$ standard deviation; trace
  entries give means, with uncertainties reported in the appendix. The A3
  changes are measured relative to the matching two-scale base model. Native
  overlap trace errors use stride-16 seams, whereas the nonoverlapping rows use
  stride-32 seams; Fig.~\ref{fig:tradeoff-common-seams} provides the
  geometry-controlled comparison.}
  \label{tab:tradeoff-decision}

  \small
  \setlength{\tabcolsep}{4pt}
  \renewcommand{\arraystretch}{1.08}

  \begin{tabular}{lccccc}
    \toprule
    Method
    & MRE (\%) $\downarrow$
    & \makecell{Value trace\\$\downarrow$}
    & \makecell{Normal-derivative\\trace $\downarrow$}
    & \makecell{Cumulative\\time (s) $\downarrow$}
    & \makecell{Change from\\two-scale} \\
    \midrule

    Plain L2L
    & $4.829 \pm 0.094$
    & $4.061{\times}10^{-4}$
    & $5.428{\times}10^{-2}$
    & $179.9 \pm 3.5$
    & -- \\

    L2L + overlap
    & $2.308 \pm 0.050$
    & $8.204{\times}10^{-6}{}^{\dagger}$
    & $4.751{\times}10^{-4}{}^{\dagger}$
    & $200.8 \pm 2.1$
    & -- \\

    \midrule

    \textbf{Two-scale (default)}
    & $1.149 \pm 0.022$
    & $2.962{\times}10^{-5}$
    & $4.304{\times}10^{-3}$
    & $202.6 \pm 2.7$
    & reference \\

    Two-scale + A3
    & $1.089 \pm 0.024$
    & $1.004{\times}10^{-5}$
    & $1.549{\times}10^{-3}$
    & $364.4 \pm 13.2$
    & \makecell{$-5.3\%$ MRE;\\$+79.9\%$ time} \\

    \bottomrule
  \end{tabular}
\end{table*}

\begin{figure*}[htbp]
  \centering
  \includegraphics[width=\textwidth]
  {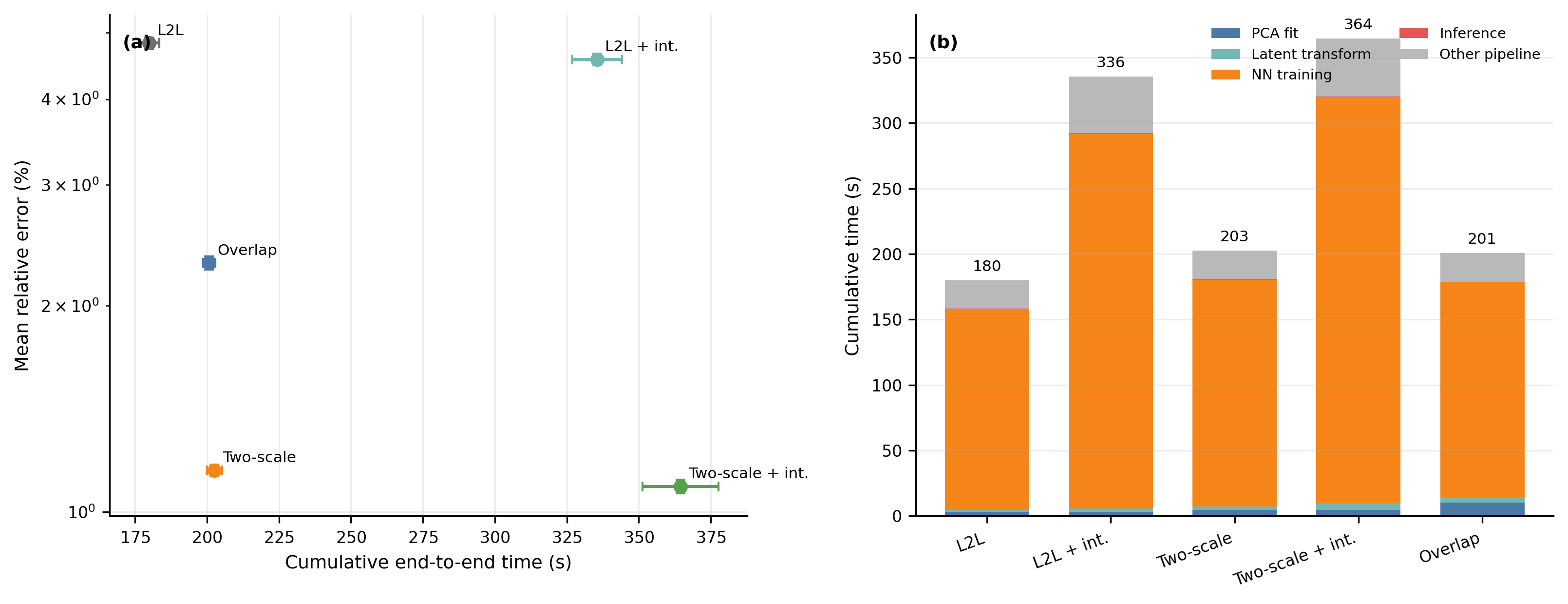}
  \caption{Accuracy--cost tradeoff for the Poisson-128 mechanism study.
  \textbf{Left:} MRE versus cumulative end-to-end time.
  \textbf{Right:} stage-wise cumulative cost. Two-scale provides substantially
  lower reconstruction error than overlap at nearly identical cumulative cost,
  whereas A3 is a separately priced refinement stage.}
  \label{fig:tradeoff-cost}
\end{figure*}

\paragraph{Two-scale is the default accuracy--cost operating point.}
Two-scale and overlap have essentially the same cumulative cost in this
matched study: $202.6\pm2.7$ s and $200.8\pm2.1$ s, respectively. Their
reconstruction errors, however, differ by approximately a factor of two:
$1.149\pm0.022\%$ for two-scale versus $2.308\pm0.050\%$ for overlap.
Figure~\ref{fig:tradeoff-cost} shows that this difference is not obtained by
adding a separate post-processing stage; both methods remain in the same
PCA-Net end-to-end cost regime, dominated by neural-network training.

A3 moves the two-scale model farther along the continuity axis, but at
substantial additional offline cost. Its $5.3\%$ MRE improvement is modest
relative to the original representation gain, whereas the $66.1\%$ and
$64.0\%$ reductions in value- and normal-derivative trace error are much more
pronounced. A3 is therefore most naturally selected when interface fidelity
has sufficient downstream value to justify the additional fine-tuning budget.

\begin{figure*}[htbp]
  \centering
  \includegraphics[width=\textwidth]
  {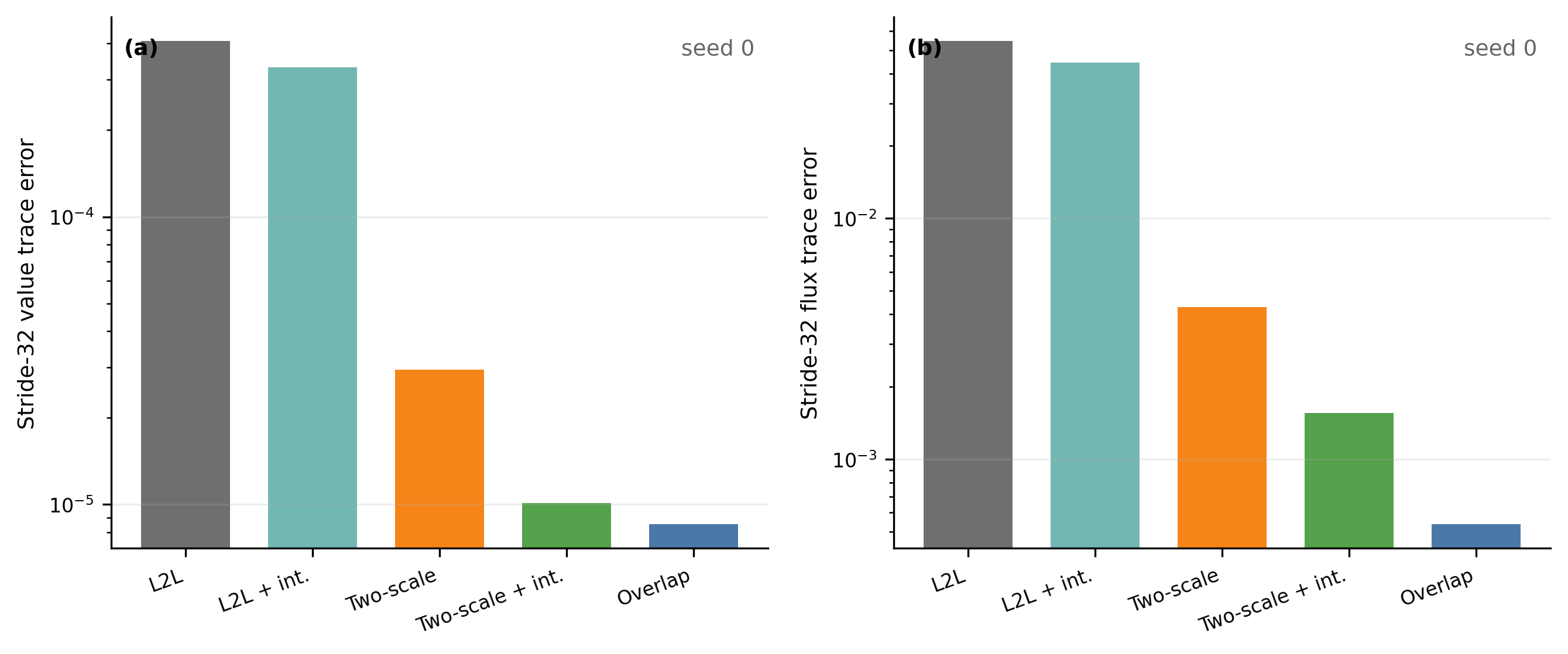}
  \caption{Geometry-controlled continuity audit on the retained seed-0
  predictions. Every method is evaluated on the same stride-32 interfaces.
  Interface-aware fine-tuning substantially improves the two-scale value- and
  normal-derivative trace errors, but overlap retains the lowest trace errors
  among the PCA-based methods on this common seam set.}
  \label{fig:tradeoff-common-seams}
\end{figure*}

\paragraph{Global reconstruction and interface fidelity remain distinct
objectives.}
Two-scale + A3 attains the lowest MRE among the PCA-Net variants in this study,
including a $52.8\%$ reduction relative to overlap. The geometry-controlled
audit in Fig.~\ref{fig:tradeoff-common-seams}, however, shows that overlap
retains lower value- and normal-derivative trace errors. This difference is
consistent with the objectives of the two approaches: overlap explicitly
averages redundant decoded predictions near each seam, whereas two-scale
primarily improves the global compatibility of the output representation.

The evidence therefore supports a representation-first operating rule:
two-scale PCA-Net is the default model when global reconstruction quality and
compact end-to-end cost are the primary objectives; A3 is an optional
continuity-oriented refinement; and overlap remains a strong reference when
minimizing interface-trace error is the dominant requirement.

\subsection{Resolution Scaling}
\label{sec:resolution-scaling}

We next evaluate whether the two-scale representation retains its advantage as
the spatial resolution increases. Independent Poisson datasets are considered
at $64\times64$, $128\times128$, and $256\times256$, with three paired seeds
at each resolution. The fixed $4\times4$ patch topology is preserved, so patch
width increases with grid resolution. Because the datasets at different
resolutions contain independently generated realizations, this experiment
measures robustness under spatial refinement rather than pointwise grid
convergence of identical samples.

\begin{table*}[htbp]
  \centering
  \caption{Poisson resolution sweep over three paired seeds. MRE, SSIM, and
  cumulative end-to-end time are reported at each resolution. The A3 entries
  include the cumulative cost of the warm-start two-scale model and subsequent
  interface-aware fine-tuning.}
  \label{tab:resolution-sweep-main}
  \scriptsize
  \resizebox{\textwidth}{!}{%
    \begin{tabular}{llllllllll}
\toprule
Method & MRE 64 (\%) & SSIM 64 & Time 64 (s) & MRE 128 (\%) & SSIM 128 & Time 128 (s) & MRE 256 (\%) & SSIM 256 & Time 256 (s) \\
\midrule
Global PCA-Net & 8.571 \ensuremath{\pm} 0.064 & 0.9057 \ensuremath{\pm} 0.0037 & 162.8 \ensuremath{\pm} 1.7 & 7.498 \ensuremath{\pm} 0.219 & 0.9239 \ensuremath{\pm} 0.0027 & 187.4 \ensuremath{\pm} 2.4 & 7.677 \ensuremath{\pm} 0.039 & 0.9301 \ensuremath{\pm} 0.0018 & 244.6 \ensuremath{\pm} 0.4 \\
Plain L2L & 4.739 \ensuremath{\pm} 0.043 & 0.9489 \ensuremath{\pm} 0.0017 & 167.6 \ensuremath{\pm} 3.6 & 4.794 \ensuremath{\pm} 0.110 & 0.9432 \ensuremath{\pm} 0.0014 & 190.7 \ensuremath{\pm} 5.7 & 4.835 \ensuremath{\pm} 0.037 & 0.9523 \ensuremath{\pm} 0.0010 & 235.5 \ensuremath{\pm} 4.1 \\
L2L + overlap & 2.365 \ensuremath{\pm} 0.025 & 0.9852 \ensuremath{\pm} 0.0007 & 177.7 \ensuremath{\pm} 2.3 & 2.292 \ensuremath{\pm} 0.063 & 0.9852 \ensuremath{\pm} 0.0005 & 210.1 \ensuremath{\pm} 3.6 & 2.289 \ensuremath{\pm} 0.017 & 0.9866 \ensuremath{\pm} 0.0004 & 288.6 \ensuremath{\pm} 2.3 \\
Two-scale & 1.060 \ensuremath{\pm} 0.023 & 0.9972 \ensuremath{\pm} 0.0003 & 190.2 \ensuremath{\pm} 11.8 & 1.155 \ensuremath{\pm} 0.028 & 0.9967 \ensuremath{\pm} 0.0002 & 213.1 \ensuremath{\pm} 1.0 & 1.221 \ensuremath{\pm} 0.099 & 0.9964 \ensuremath{\pm} 0.0004 & 273.3 \ensuremath{\pm} 3.6 \\
Two-scale + interface & 1.000 \ensuremath{\pm} 0.021 & 0.9973 \ensuremath{\pm} 0.0003 & 313.7 \ensuremath{\pm} 14.3 & 1.090 \ensuremath{\pm} 0.037 & 0.9968 \ensuremath{\pm} 0.0002 & 391.0 \ensuremath{\pm} 3.6 & 1.135 \ensuremath{\pm} 0.049 & 0.9967 \ensuremath{\pm} 0.0002 & 652.5 \ensuremath{\pm} 5.4 \\
FNO & 0.217 \ensuremath{\pm} 0.009 & 0.9999 \ensuremath{\pm} 0.0000 & 293.5 \ensuremath{\pm} 3.7 & 0.245 \ensuremath{\pm} 0.011 & 0.9998 \ensuremath{\pm} 0.0000 & 725.7 \ensuremath{\pm} 0.3 & 0.262 \ensuremath{\pm} 0.001 & 0.9998 \ensuremath{\pm} 0.0000 & 3417.2 \ensuremath{\pm} 0.6 \\
\bottomrule
\end{tabular}
  }
\end{table*}

\begin{figure*}[htbp]
  \centering
  \includegraphics[width=\textwidth]{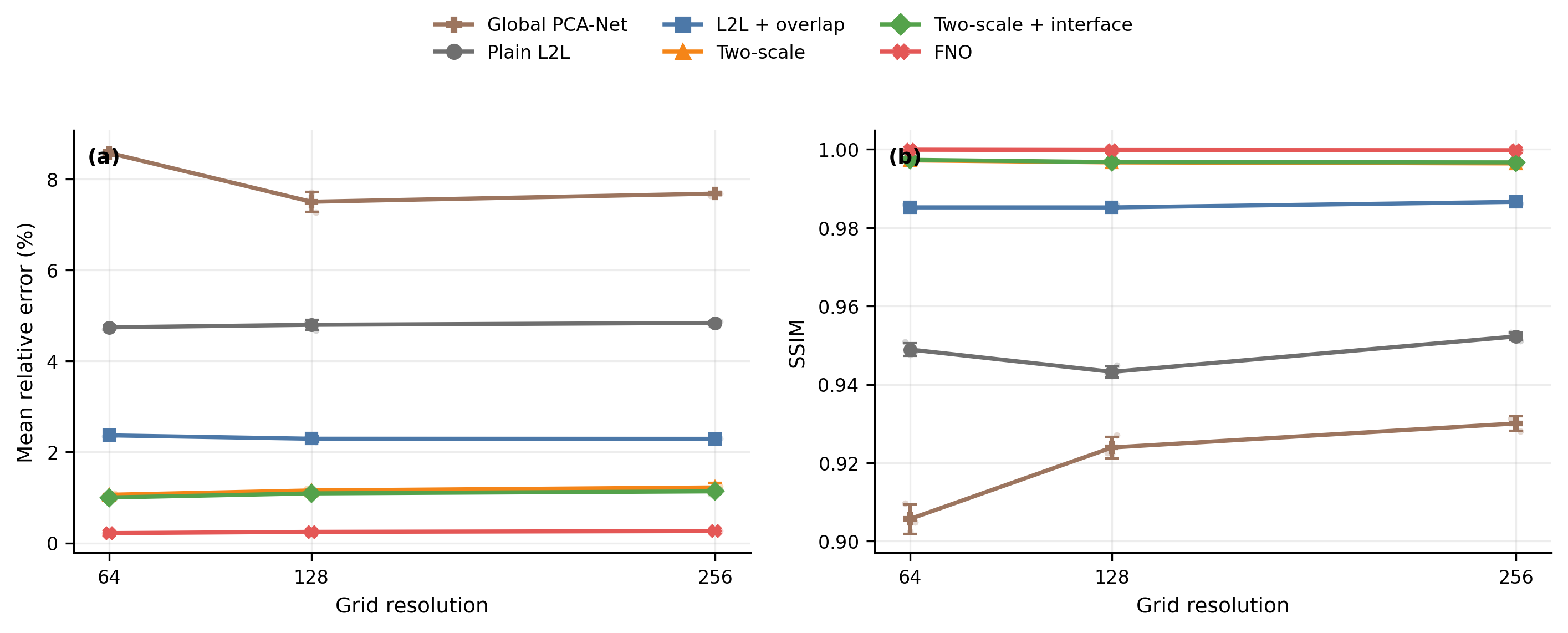}
  \caption{Accuracy across Poisson grid resolutions. Two-scale and its A3
  refinement retain low MRE and high SSIM from $64^2$ through $256^2$.
  Plain L2L and overlap remain at substantially higher reconstruction error,
  while FNO provides the full-field accuracy reference.}
  \label{fig:resolution-accuracy}
\end{figure*}

\paragraph{Two-scale preserves its reconstruction advantage under grid
refinement.}
Figure~\ref{fig:resolution-accuracy} and
Table~\ref{tab:resolution-sweep-main} show that two-scale maintains low
reconstruction error across the full resolution range:
$1.060\pm0.023\%$ MRE at $64^2$,
$1.155\pm0.028\%$ at $128^2$, and
$1.221\pm0.099\%$ at $256^2$, with SSIM remaining above $0.996$ throughout.
Plain L2L remains near $4.8\%$ MRE and overlap near $2.3\%$ over the same
range.

Relative to overlap, two-scale reduces MRE by $55.2\%$, $49.6\%$, and
$46.7\%$ at $64^2$, $128^2$, and $256^2$, respectively. Thus, the advantage
narrows modestly with increasing resolution but remains substantial at the
largest executed grid. The persistence of the gap also shows that simply
refining the grid does not eliminate the error associated with the
local-output baselines.

\begin{figure*}[htbp]
  \centering
  \includegraphics[width=\textwidth]{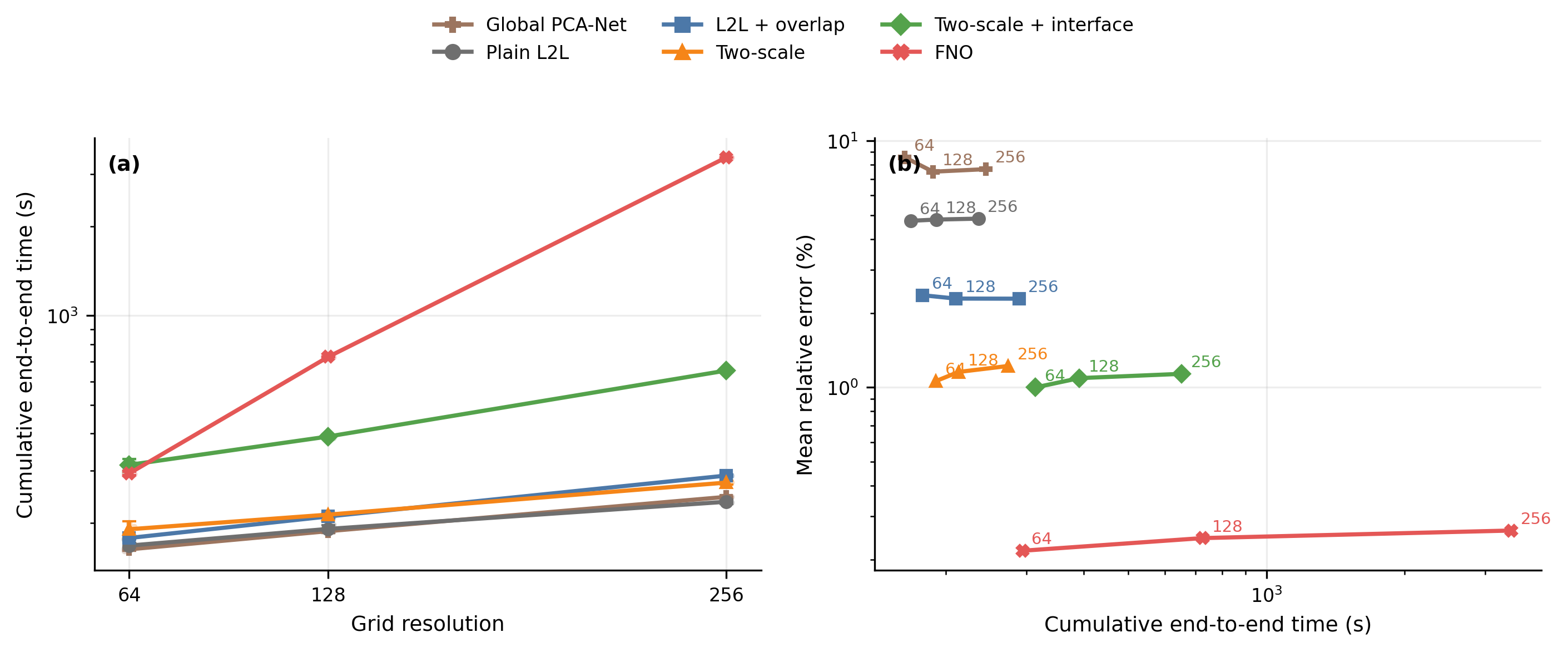}
  \caption{Measured Poisson resolution-scaling cost and quality--cost
  frontier. \textbf{Left:} cumulative end-to-end time versus grid resolution.
  \textbf{Right:} MRE versus cumulative time. At $256^2$, two-scale improves
  both reconstruction error and total runtime relative to overlap. A3 provides
  additional reconstruction improvement at a separately accumulated
  fine-tuning cost.}
  \label{fig:resolution-quality-cost}
\end{figure*}

\paragraph{Representation construction remains cheaper than overlap.}
The measured cost comparison in
Fig.~\ref{fig:resolution-quality-cost} places two-scale in the same compact
PCA-Net regime as the localized baselines. At $256^2$, two-scale requires
$273.3\pm3.6$ s cumulatively, compared with $288.6\pm2.3$ s for overlap,
giving a $5.3\%$ reduction in end-to-end time while simultaneously lowering
MRE by $46.7\%$.

The construction-stage difference is larger and systematic. Two-scale PCA
fitting requires $1.25$, $5.22$, and $19.79$ s at $64^2$, $128^2$, and
$256^2$, respectively, compared with $2.68$, $10.25$, and $39.72$ s for
overlap. Thus, two-scale requires approximately half the PCA fitting time
across all three grids. The coarse-global SVD contributes only a small portion
of this cost; the principal saving comes from avoiding the denser collection
of local PCA fits required by the overlapping patch layout. Because
neural-network training remains the dominant stage, the end-to-end advantage
is necessarily smaller than the PCA-stage advantage.

\begin{figure*}[htbp]
  \centering
  \includegraphics[width=0.88\textwidth]{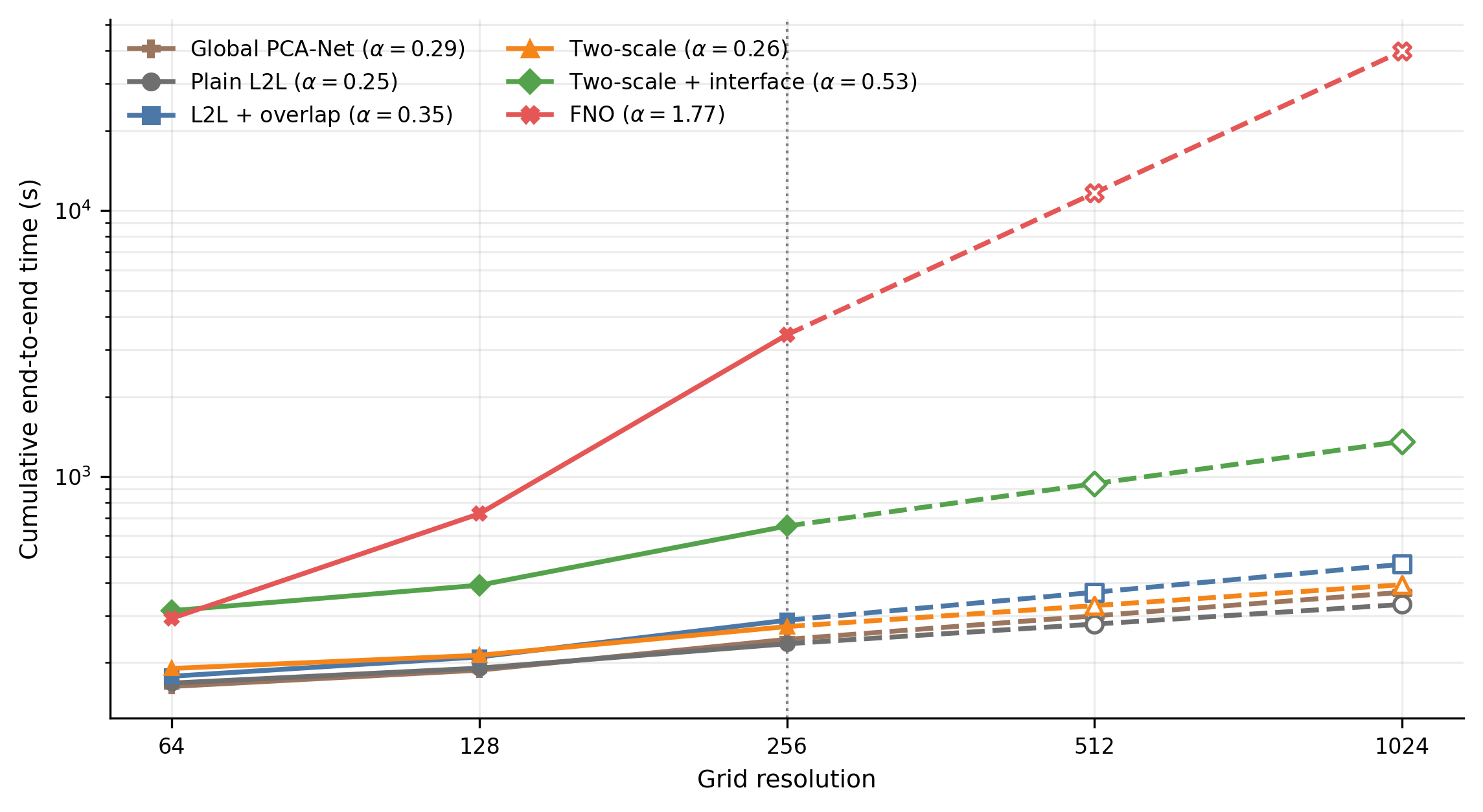}
  \caption{Measured and projected cumulative wall-time scaling. Solid segments
  and filled markers denote the three executed resolutions. Dashed segments
  and hollow markers extend least-squares log--log fits obtained from the
  measured $64^2$--$256^2$ results and anchored at the measured $256^2$
  value. The projected $512^2$ and $1024^2$ points are descriptive
  extrapolations and are not experimental measurements.}
  \label{fig:resolution-time-projection}
\end{figure*}

\paragraph{Measured timing trends favor the compact two-scale regime.}
To visualize the observed wall-time growth, Fig.~\ref{fig:resolution-time-projection}
fits a local power-law trend
\begin{equation}
  T(D)\propto D^\alpha
\end{equation}
to the three measured cumulative times for each method and extends the fitted
trend beyond the executed grids. The fitted exponent is $\alpha=0.26$ for
two-scale, compared with $0.35$ for overlap, $0.53$ for A3, and $1.77$ for
FNO under its fixed training budget.

Under the purely descriptive assumption that these local slopes persist with
unchanged sample counts, patch topology, and epoch budgets, the fitted curves
give projected cumulative times of 328 and 393 s for two-scale at $512^2$ and
$1024^2$, compared with 368 and 469 s for overlap. These values are
\emph{extrapolations only}: the scaling claim of this paper is restricted to
the measured $64^2$--$256^2$ range. The dashed continuation is included to
visualize the local slopes rather than to assert performance at unexecuted
resolutions.

\paragraph{A3 provides increasingly strong interface refinement at higher
resolution.}
The optional A3 stage lowers the two-scale MRE by $5.7\%$, $5.6\%$, and
$7.1\%$ at $64^2$, $128^2$, and $256^2$, respectively. Its larger effect is
on the remaining interface error. Relative to the corresponding two-scale
model, A3 reduces value-trace error by $51.7\%$, $66.2\%$, and $71.8\%$ and
normal-derivative trace error by $51.0\%$, $64.0\%$, and $70.5\%$ across the
three grids.

The added control is accompanied by substantial offline cost. Cumulative A3
time increases from $313.7\pm14.3$ s at $64^2$ to $652.5\pm5.4$ s at
$256^2$, compared with $190.2\pm11.8$ s and $273.3\pm3.6$ s for the base
two-scale models. The resolution sweep therefore reinforces the operating
rule established in Section~\ref{sec:mechanism-ablation}: two-scale is the
default scalable configuration, while A3 is an optional refinement when the
additional reduction in interface error warrants its training cost.

FNO remains the absolute accuracy reference, with MRE between $0.217\%$ and
$0.262\%$, but its cumulative time increases from approximately 294 s at
$64^2$ to 3,417 s at $256^2$. At the largest executed grid this is
$12.5\times$ the two-scale runtime. The purpose of two-scale PCA-Net is
therefore not to displace a higher-capacity full-field operator on absolute
accuracy, but to retain a substantial reconstruction advantage over prior
PCA-Net representations within a considerably lower-cost regime.

\subsection{Generalization to Heterogeneous Darcy Flow}
\label{sec:darcy-generalization}

\begin{figure*}[htbp]
  \centering
  \includegraphics[width=\textwidth]
  {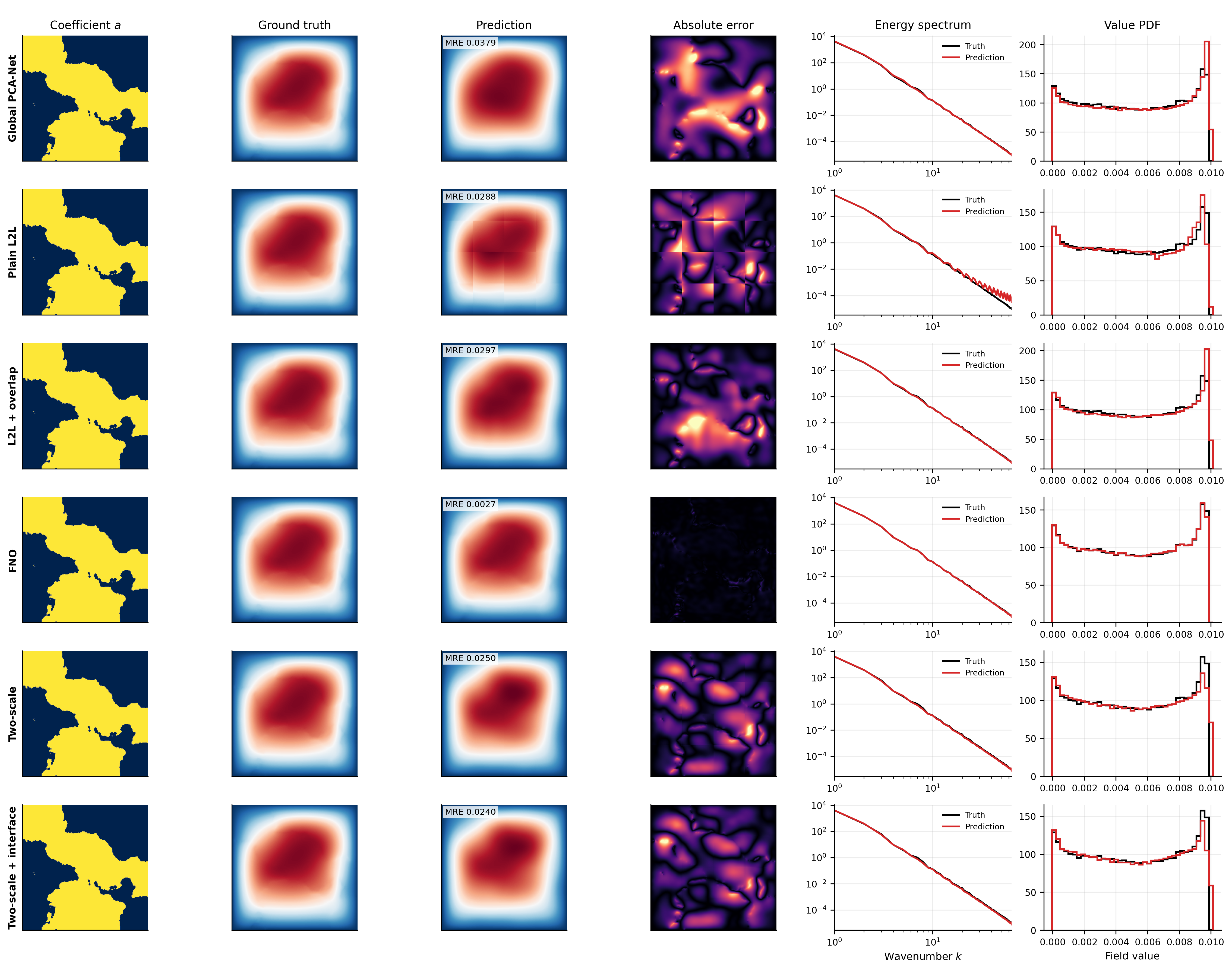}
  \caption{Representative heterogeneous Darcy reconstruction (seed 0,
  sample 7964). The discontinuous permeability field is shown in the first
  column. Plain L2L exhibits pronounced block-aligned error, whereas two-scale
  substantially suppresses the patchwise component while retaining broader
  reconstruction error. A3 further reduces localized interface error, and FNO
  remains the higher-capacity full-field accuracy reference.}
  \label{fig:darcy-qualitative}
\end{figure*}

We finally test whether the two-scale representation transfers to a more
challenging heterogeneous operator. The Darcy-$256^2$ benchmark uses
discontinuous permeability fields rather than the smooth forcing fields of the
Poisson studies; a representative coefficient realization is shown in
Fig.~\ref{fig:darcy-qualitative}. Results are reported over five paired seeds
using the baseline and training protocols defined in
Section~\ref{sec:experimental-methodology}.

\begin{table*}[htbp]
  \centering
  \caption{Heterogeneous Darcy results at $256^2$ over five paired seeds.
  Reconstruction, physical-field, and cumulative-cost metrics are reported
  together because two-scale substantially improves the localized model's
  interface and residual errors without dominating the Darcy MRE ranking.}
  \label{tab:darcy-main}
  \small
  \begin{tabular}{lcccccc}
\toprule
Method
& MRE (\%)
& SSIM
& \makecell{Normal derivative\\trace}
& \makecell{Spectral\\error}
& \makecell{Darcy residual\\RMS}
& \makecell{Total time\\(s)} \\
\midrule

Global PCA-Net
& 2.910
& 0.9945
& $2.521{\times}10^{-4}$
& $9.537{\times}10^{-4}$
& 3.38960
& 526.3 \\

Plain L2L
& 3.109
& 0.9900
& $3.197{\times}10^{-2}$
& $1.552{\times}10^{-3}$
& 21.07659
& 338.8 \\

L2L + overlap
& 2.693
& 0.9950
& $2.813{\times}10^{-4}$
& $1.920{\times}10^{-3}$
& 3.38720
& 717.2 \\

FNO
& 0.279
& 0.9996
& $4.269{\times}10^{-4}$
& $4.941{\times}10^{-4}$
& 4.00006
& 3185.7 \\

Two-scale
& 2.952
& 0.9946
& $3.577{\times}10^{-3}$
& $1.698{\times}10^{-3}$
& 4.13100
& 388.7 \\

Two-scale + interface
& 2.926
& 0.9947
& $6.077{\times}10^{-4}$
& $1.537{\times}10^{-3}$
& 3.44387
& 781.7 \\

\bottomrule
\end{tabular}
\end{table*}

\begin{figure*}[htbp]
  \centering
  \includegraphics[width=\textwidth]
  {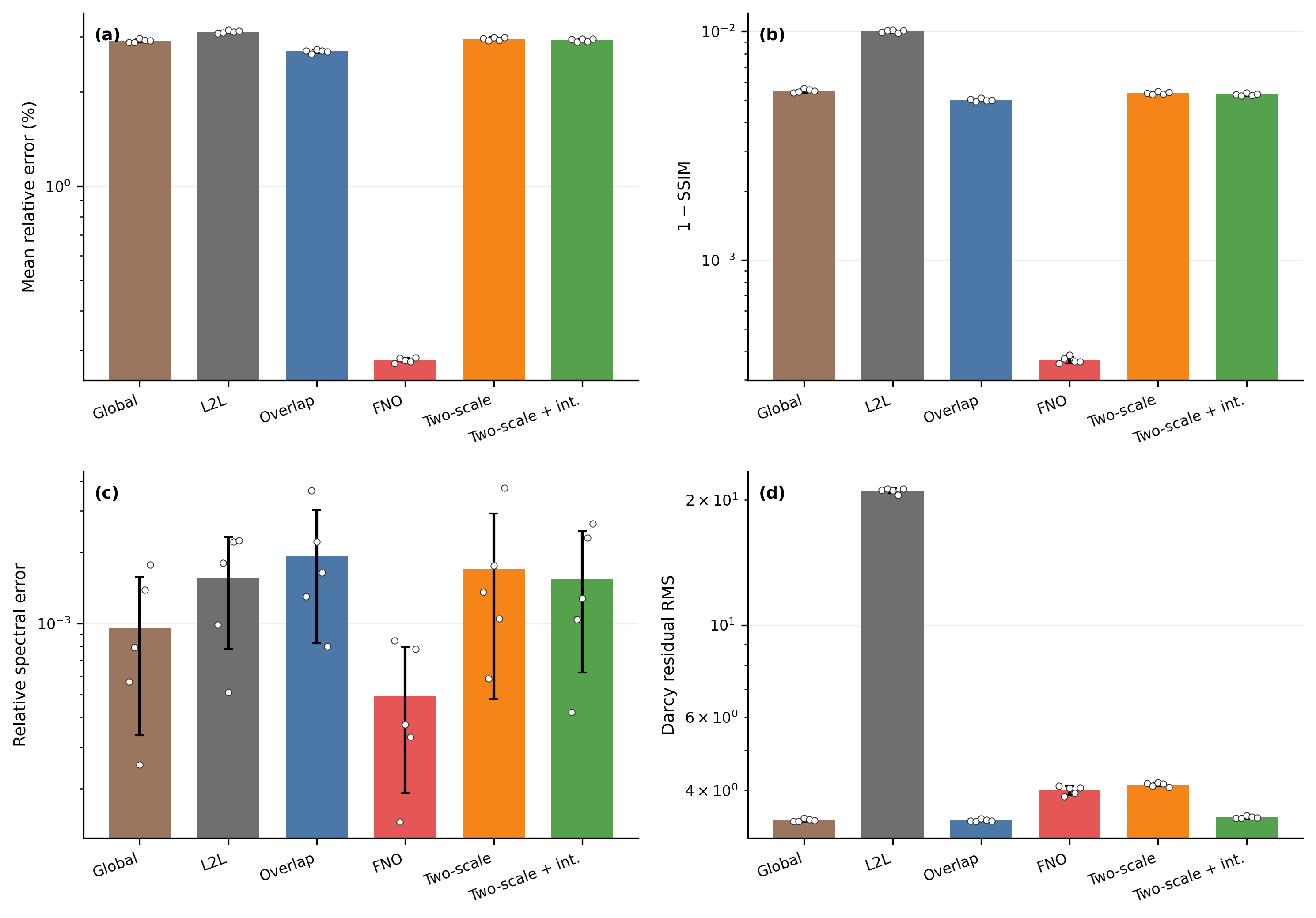}
  \caption{Aggregate diagnostics on heterogeneous Darcy flow at $256^2$.
  Two-scale provides only a modest MRE improvement over plain L2L, but
  substantially reduces the localized model's normal-derivative jump and
  Darcy residual error. A3 further improves these physical-field diagnostics
  while leaving global reconstruction error nearly unchanged.}
  \label{fig:darcy-accuracy-physics}
\end{figure*}

\paragraph{Two-scale transfers primarily as a physical-coherence improvement.}
Table~\ref{tab:darcy-main} and
Fig.~\ref{fig:darcy-accuracy-physics} show a more qualified result than on
Poisson. Two-scale attains $2.952\pm0.033\%$ MRE, a $5.0\%$ reduction from
plain L2L ($3.109\pm0.030\%$). Its reconstruction accuracy is comparable to
Global PCA-Net ($2.910\pm0.037\%$), but remains behind overlap
($2.693\pm0.034\%$) and the FNO reference ($0.279\pm0.005\%$).
Thus, two-scale is not the strongest Darcy reconstructor among the evaluated
baselines.

The improvement in physical-field diagnostics is substantially larger.
Relative to plain L2L, two-scale reduces the normal-derivative jump by
$88.8\%$, from $3.197\times10^{-2}$ to $3.577\times10^{-3}$, and lowers
Darcy residual RMS by $80.4\%$, from $21.08$ to $4.13$. SSIM increases from
$0.9900$ to $0.9946$. These gains require a $14.8\%$ increase in cumulative
runtime, from $338.8\pm8.6$ s to $388.7\pm5.3$ s. By comparison, overlap
increases cumulative cost by $111.8\%$ relative to plain L2L. The two-scale
representation therefore transfers to heterogeneous Darcy primarily by
suppressing the severe patch-scale physical inconsistency of the local-only
model at comparatively modest additional cost.

\paragraph{Interface-aware fine-tuning primarily improves physical
consistency.}
Starting from two-scale, A3 lowers MRE by only $0.9\%$, from $2.952\%$ to
$2.926\%$. Its effect on the physical-field diagnostics is much larger:
the normal-derivative jump decreases by $83.0\%$, to
$6.077\times10^{-4}$, while Darcy residual RMS decreases by $16.6\%$, to
$3.44$. The mean spectral error is also lower, but the paired changes vary
substantially across seeds; we therefore make no claim of a stable spectral
improvement on Darcy.

These gains increase cumulative runtime to $781.7\pm6.7$ s, approximately
twice the cost of the base two-scale model. A3 is consequently best interpreted
on this benchmark as a targeted interface and residual refinement rather than
a mechanism for reducing the dominant global prediction error.

\begin{figure}[htbp]
  \centering
  \includegraphics[width=\columnwidth]
  {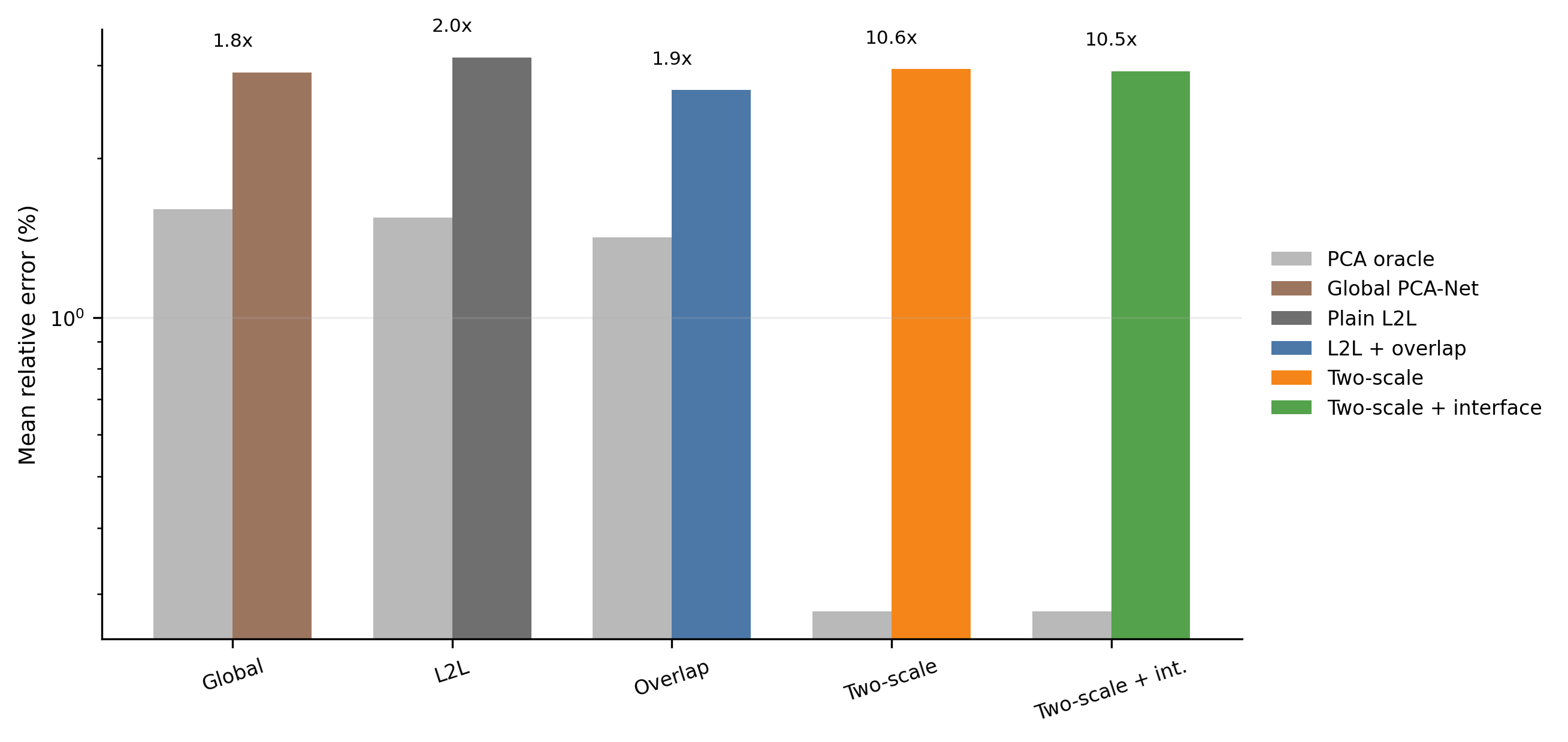}
  \caption{PCA-oracle and learned-operator MRE on heterogeneous Darcy flow.
  Two-scale has the lowest PCA reconstruction floor among the PCA-Net
  representations but the largest learned-to-oracle gap, identifying
  coefficient-to-latent-score prediction as the principal remaining
  bottleneck.}
  \label{fig:darcy-representation-gap}
\end{figure}

\paragraph{The output representation is not the dominant Darcy bottleneck.}
The PCA-oracle comparison in
Fig.~\ref{fig:darcy-representation-gap} separates representation capacity from
latent-score prediction. The two-scale oracle achieves
$0.278\pm0.002\%$ MRE, whereas the learned operator reaches
$2.952\pm0.033\%$, giving a $10.6\times$ learned-to-oracle gap. A3 shares
the same fitted representation and therefore the same oracle floor, with a
nearly identical $10.5\times$ learned-to-oracle ratio.

By contrast, the Global PCA-Net, plain-L2L, and overlap representations have
learned-to-oracle ratios of approximately $1.8$--$2.0\times$. The two-scale
output basis can therefore represent the heterogeneous Darcy solutions with
high fidelity; most of the remaining error arises in predicting its coarse
and residual coordinates from the discontinuous permeability input. This
identifies the latent map, rather than PCA reconstruction capacity, as the
principal limitation of the present Darcy model.

\paragraph{Two-scale retains its representation-construction advantage.}
Two-scale requires $93.1\pm5.4$ s for PCA fitting, compared with
$247.4\pm8.4$ s for overlap and $278.8\pm8.1$ s for Global PCA-Net.
The representation therefore remains substantially cheaper to construct than
the more redundant localized or full-field PCA alternatives on this
$256^2$ problem.

FNO demonstrates that substantially lower Darcy reconstruction error is
attainable, but at a cumulative cost of $3{,}185.7\pm3.2$ s, approximately
$8.2\times$ that of two-scale. Taken together, the Darcy results sharpen the
scope of the proposed method: two-scale improves the efficiency and physical
coherence of localized PCA-Net, while heterogeneous coefficients expose
latent-score prediction as the main remaining challenge for global accuracy.

\section{Design and Ablation Studies}
\label{sec:design_ablation_studies}

\subsection{Coarse-Global Representation Design}
\label{sec:coarse-global-design}

The two-scale representation has three coupled design choices: the number of
coarse global modes, the restriction factor $c$, and the residual-PCA retained
variance.  This study selects those choices on Poisson at $128^2$ and $256^2$.
All runs use 8,000 training examples, randomized SVD, unchanged local input
codes, and the block-balanced coarse/residual latent objective.  We report both
the PCA oracle, obtained by encoding and decoding the true solution, and the
learned operator.  Their separation distinguishes representation capacity from
the ability to infer its latent coordinates.

\begin{table*}[htbp]
  \centering
  \caption{Design choices for the coarse-global representation.  The selected
  configuration is used in subsequent two-scale experiments.  MRE values are
  means over three seeds.}
  \label{tab:coarse-design-selection}
  \small
  \begin{tabular}{p{0.19\textwidth}p{0.57\textwidth}p{0.16\textwidth}}
    \toprule
    Design element & Evidence and interpretation & Selected setting \\
    \midrule
    Coarse rank & The learned MRE is minimized at rank 10: $1.28\%$ at $128^2$
    and $1.29\%$ at $256^2$.  Ranks 20 and 40 reduce the oracle error, yet
    learned MRE rises to $3.79\%$/$3.93\%$ at $128^2$ and
    $5.69\%$/$7.57\%$ at $256^2$. & 10 modes \\
    Restriction factor & $c=2$ is strongest at $128^2$, but is illegal at
    $256^2$ because $(256/2)^2=16{,}384 > 8{,}000$.  $c=4$ is the best legal
    setting at $256^2$; $c=8$ removes too much intermediate-scale structure. &
    $c=4$ \\
    Residual retained variance & Raising the residual target from 99\% to
    99.5\% lowers learned MRE from $1.28\%$/$1.29\%$ to
    $1.15\%$/$1.15\%$ and reduces flux jump by 38\% at both resolutions,
    without a measurable end-to-end time penalty. & 99.5\% \\
    \bottomrule
  \end{tabular}
\end{table*}

\begin{figure*}[htbp]
  \centering
  \includegraphics[width=0.96\textwidth]{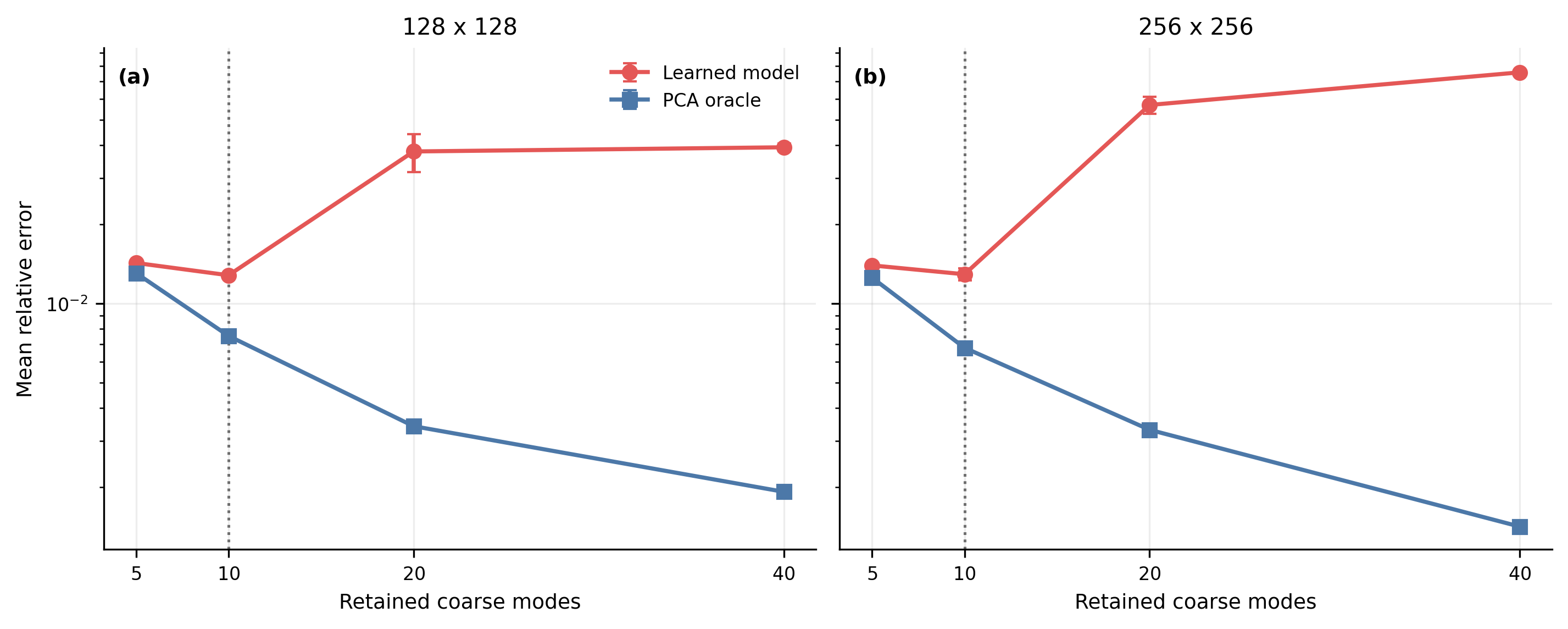}
  \caption{Coarse-rank trainability gap on Poisson.  More global PCA modes
  monotonically improve the oracle representation, but ranks above 10 make
  the input-to-latent regression substantially harder.  The selected rank is
  therefore determined by learned accuracy, not by the PCA reconstruction
  floor alone}
  \label{fig:coarse-rank-trainability}
\end{figure*}

\paragraph{A larger coarse subspace is not necessarily a more learnable
one.}
Figure~\ref{fig:coarse-rank-trainability} exposes the central result.  The
PCA oracle improves monotonically as coarse rank increases, as expected from
the nested global subspaces: at $256^2$, its MRE falls from $1.26\%$ at rank 5
to $0.14\%$ at rank 40.  The learned model instead reaches its best MRE at
rank 10 and degrades sharply afterwards.  In particular, rank 40 has an
excellent $0.14\%$ oracle MRE but a $7.57\%$ learned MRE at $256^2$.
The same reversal occurs at $128^2$.

This is a trainability limitation rather than evidence against a richer basis.
Increasing rank from 10 to 40 also expands the total output code from roughly
134 to 332 coordinates at $256^2$.  The additional coarse coordinates encode
lower-energy, more detailed global content whose amplitudes are harder to
recover from local input features; the altered residual requires more local
coordinates as well.  The adaptive 99\%-variance rule, capped at 20 coarse
modes, independently selected ten modes for every seed and resolution and
matched the fixed-rank result.  This supports rank 10 as a stable, rather than
incidental, operating point.

\paragraph{Choose the coarse grid for transferable fidelity, then retain
enough residual detail.}
At $128^2$, the finer $c=2$ coarse grid yields the best resolution-specific
MRE ($1.05\%$), while $c=8$ degrades to $1.51\%$.  A factor of two cannot be
used at $256^2$ under the required low-rank condition $(D/c)^2<m$, whereas
$c=4$ gives $1.29\%$ MRE and is the best legal setting there.  We consequently
use $c=4$ as the cross-resolution default.  The coarse-SVD itself costs less
than one second, so this choice is governed by representational fidelity and
the sample-size guard, not by a material runtime difference.

The residual branch should be moderately generous.  Relative to the 99\%
reference, retaining 99.5\% of local residual variance improves MRE to
$1.15\%$ at both resolutions, raises SSIM to $0.9967$, and lowers the flux
jump from $7.09\times10^{-3}$ to $4.37\times10^{-3}$ at $128^2$ and from
$1.32\times10^{-2}$ to $8.20\times10^{-3}$ at $256^2$.  It adds only about
24 residual coordinates and leaves cumulative time statistically unchanged.
The selected coarse-global representation is therefore rank 10, $c=4$, and a
99.5\% residual PCA target.  This is the configuration used as the default
two-scale representation in the subsequent studies.

\subsection{Latent Objective}
\label{sec:latent-objective}

The two-scale decoder predicts a short coarse-global code and a substantially
larger concatenation of local residual codes.  A vector MSE over the complete
code therefore weights these blocks in proportion to their coordinate counts,
which can leave the coarse field weakly supervised.  We compare three targets
while holding the two-scale basis, data split, network, and optimizer fixed:
plain vector MSE, equal block balancing, and equal block balancing after
dividing each score error by its training variance.  With coarse and residual
code dimensions $z_G$ and $z_R$, respectively, the latter two objectives are

\begin{equation}
\mathcal{L}_{\mathrm{block}} = \frac{1}{2z_G}\left\|\hat{\mathbf{z}}_G-
\mathbf{z}_G\right\|_2^2 + \frac{1}{2z_R}\left\|\hat{\mathbf{z}}_R-
\mathbf{z}_R\right\|_2^2,
\label{eq:block-balanced-latent-loss}
\end{equation}
\noindent and $\mathcal{L}_{\mathrm{score}}$ applies the same block average
after dividing coordinate $j$ by its PCA score variance $\sigma_j^2$.

This is a design-selection experiment over Poisson and Darcy at $128^2$ and
$256^2$, using one matched seed per cell.  Its purpose is to identify a robust
default, not to establish a multi-seed ranking between the two structured
objectives.

\begin{table*}[htbp]
  \centering
  \caption{Latent-objective selection.  All entries are matched seed-0 MRE
  (\%).  Both structured objectives improve on vector MSE in all four cells;
  block balancing is selected as the cross-problem default.}
  \label{tab:latent-objective-selection}
  \small
  \begin{tabular}{llrrrl}
    \toprule
    Dataset & Grid & MSE & Block & Score & Default \\
    \midrule
    Poisson & 128 & 2.244 & \textbf{1.256} & 1.572 & Block \\
    Poisson & 256 & 2.331 & \textbf{1.264} & 1.509 & Block \\
    Darcy & 128 & 7.958 & 4.571 & \textbf{3.662} & Block \\
    Darcy & 256 & 6.167 & 3.349 & \textbf{2.550} & Block \\
    \bottomrule
  \end{tabular}
\end{table*}

\begin{figure*}[htbp]
  \centering
  \includegraphics[width=0.96\textwidth]{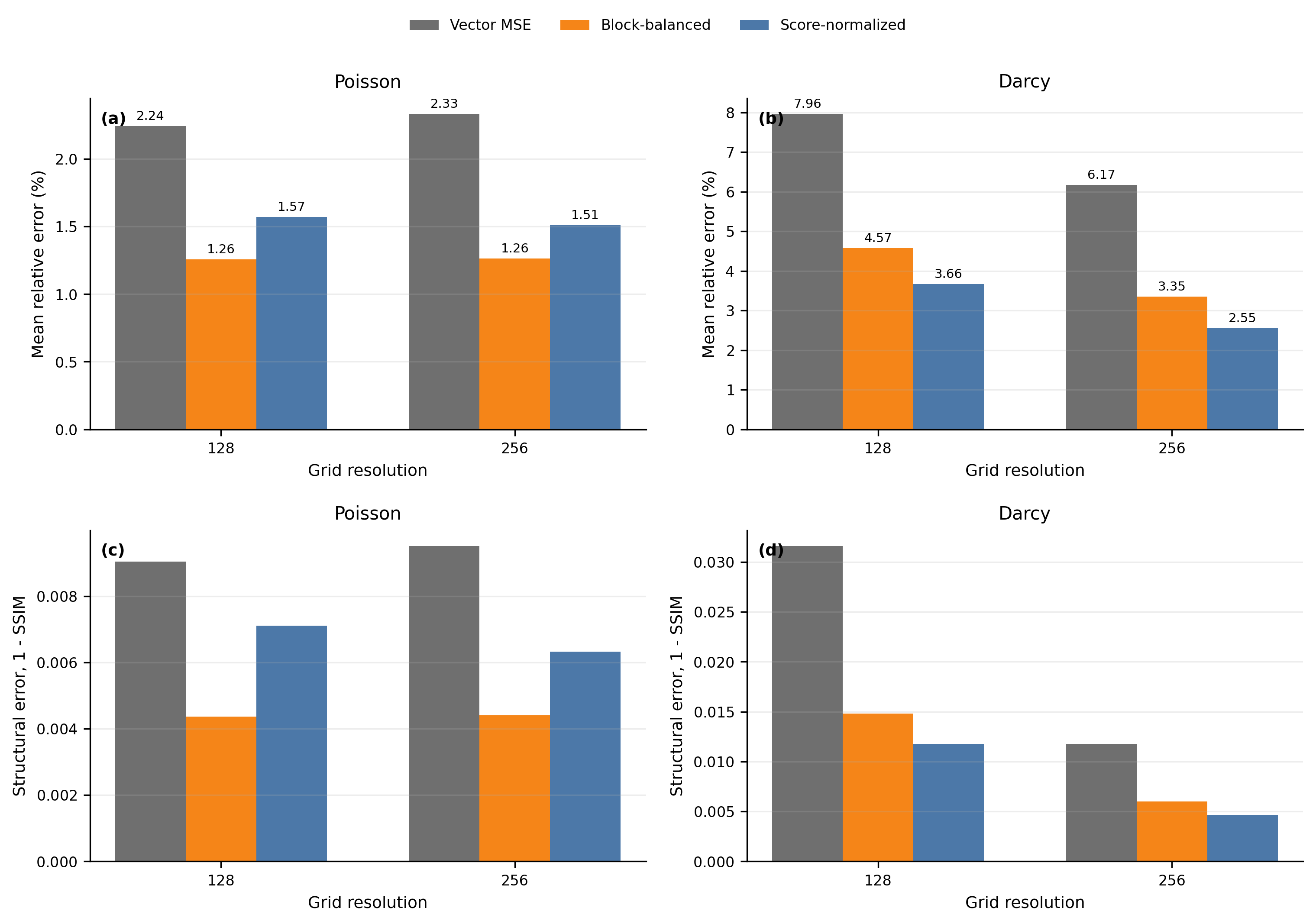}
  \caption{Latent-objective accuracy comparison for the fixed two-scale
  representation.  Both block-balanced and score-normalized objectives
  improve MRE and structural error over vector MSE.  Block balancing is best
  on Poisson, while score normalization gives the lowest Darcy MRE in this
  one-seed selection study}
  \label{fig:latent-objective-accuracy}
\end{figure*}

\paragraph{Structured supervision corrects the dominant latent imbalance.}
Figure~\ref{fig:latent-objective-accuracy} shows that both structured losses
improve MRE and SSIM in every dataset-resolution cell.  Block balancing lowers
MRE by $42$--$46\%$ relative to vector MSE across all four cells.  This is a
large effect for a change that leaves the representation, parameter count, and
inference path untouched.  In the Poisson cases, block balancing is also the
best learned target: it reaches $1.26\%$ MRE at both resolutions, compared
with $2.24\%$ and $2.33\%$ for vector MSE.

Score normalization improves Darcy MRE further, from $7.96\%$ to $3.66\%$ at
$128^2$ and from $6.17\%$ to $2.55\%$ at $256^2$.  However, it is not a
uniformly safer objective: on Darcy it increases the flux jump by $43\%$ and
$54\%$ relative to vector MSE, and increases the $256^2$ Darcy residual by
$23\%$.  Its Poisson physical-field diagnostics move in the favorable
direction, but this problem dependence makes per-score normalization a useful
dataset-specific option rather than the default objective.

\begin{figure*}[htbp]
  \centering
  \includegraphics[width=0.96\textwidth]{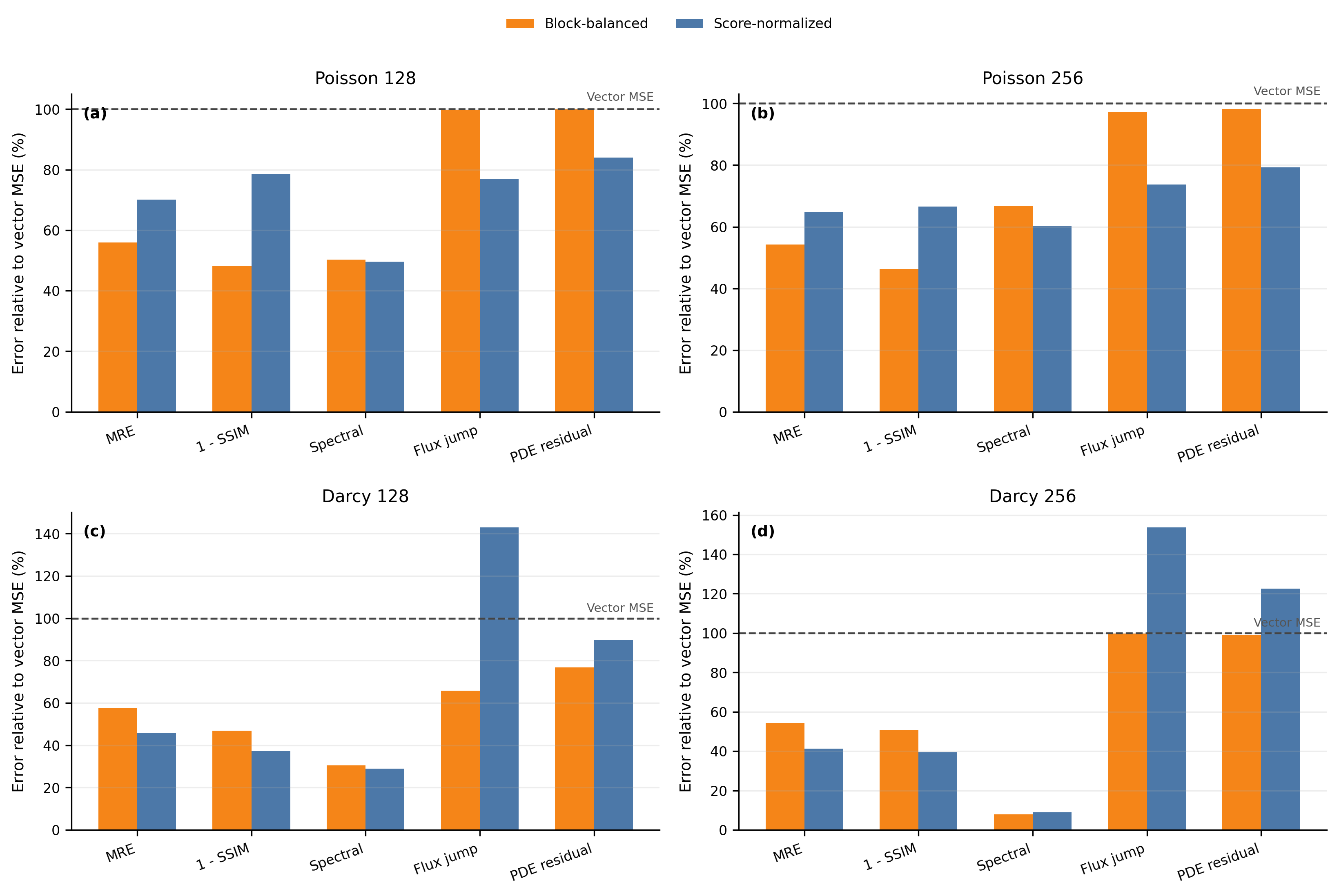}
  \caption{Physical-field error relative to vector MSE; values below 100\%
  indicate improvement.  Block balancing is consistently favorable or neutral
  across the metrics.  Score normalization improves aggregate Darcy accuracy
  but worsens Darcy interface flux mismatch and, at $256^2$, the PDE residual}
  \label{fig:latent-objective-tradeoffs}
\end{figure*}

\paragraph{Block balancing is the robust default.}
The residual block contains most output coordinates: for example, the Darcy
$256^2$ code has 20 coarse scores and 735 residual scores.  Plain vector MSE
therefore assigns only about $2.6\%$ of its average coordinate weight to the
coarse block.  Equal block averaging restores a direct learning signal for the
global component while preserving the PCA variance hierarchy within each
block.  Score normalization additionally amplifies low-variance tail scores;
this can lower aggregate field error, but can also overemphasize difficult
directions that degrade physical consistency.  We consequently use equal
coarse/residual block weights as the default for two-scale models and reserve
score normalization for problem-specific validation.

The structured objectives add roughly $14$--$23\%$ to neural-network training
time in this single-seed study because their validation trajectories train for
longer, not because they change the PCA representation or inference operator.
The corresponding end-to-end increases are modest on Poisson and remain
secondary to the quality gain.  Detailed stage costs, capacity checks, and
qualitative comparisons are provided in Appendix~\ref{app:latent-objective}.

\subsection{Interface-Loss Ladder}
\label{sec:interface-loss-ladder}

The two-scale representation removes the dominant patch-scale error before
fine-tuning begins, but it does not explicitly constrain the assembled field
at patch interfaces.  We therefore form an ordered in-loop loss ladder on
Poisson: A0 is the block-balanced latent two-scale model; A1 fine-tunes with
field reconstruction; A2 adds interface value traces; A3 adds interface flux
traces; A4 adds spectral matching; and A5 adds the Poisson residual.  Each
stage is warm-started from its matched A0 checkpoint, uses the same
rank-10/$c=4$/99.5\% representation, and is evaluated over three paired seeds
at $128^2$ and $256^2$.  Reported cumulative times include the original
two-scale training run and the fine-tuning stage.

All residuals in this version use the generator-consistent convention
$\Delta_h\widehat{u}-f$.  The A5 models were recalibrated and retrained under
this convention; the other variants contain no active PDE-residual loss and
are evaluated with the same corrected diagnostic.

\begin{figure*}[htbp]
  \centering
  \includegraphics[width=0.96\textwidth]{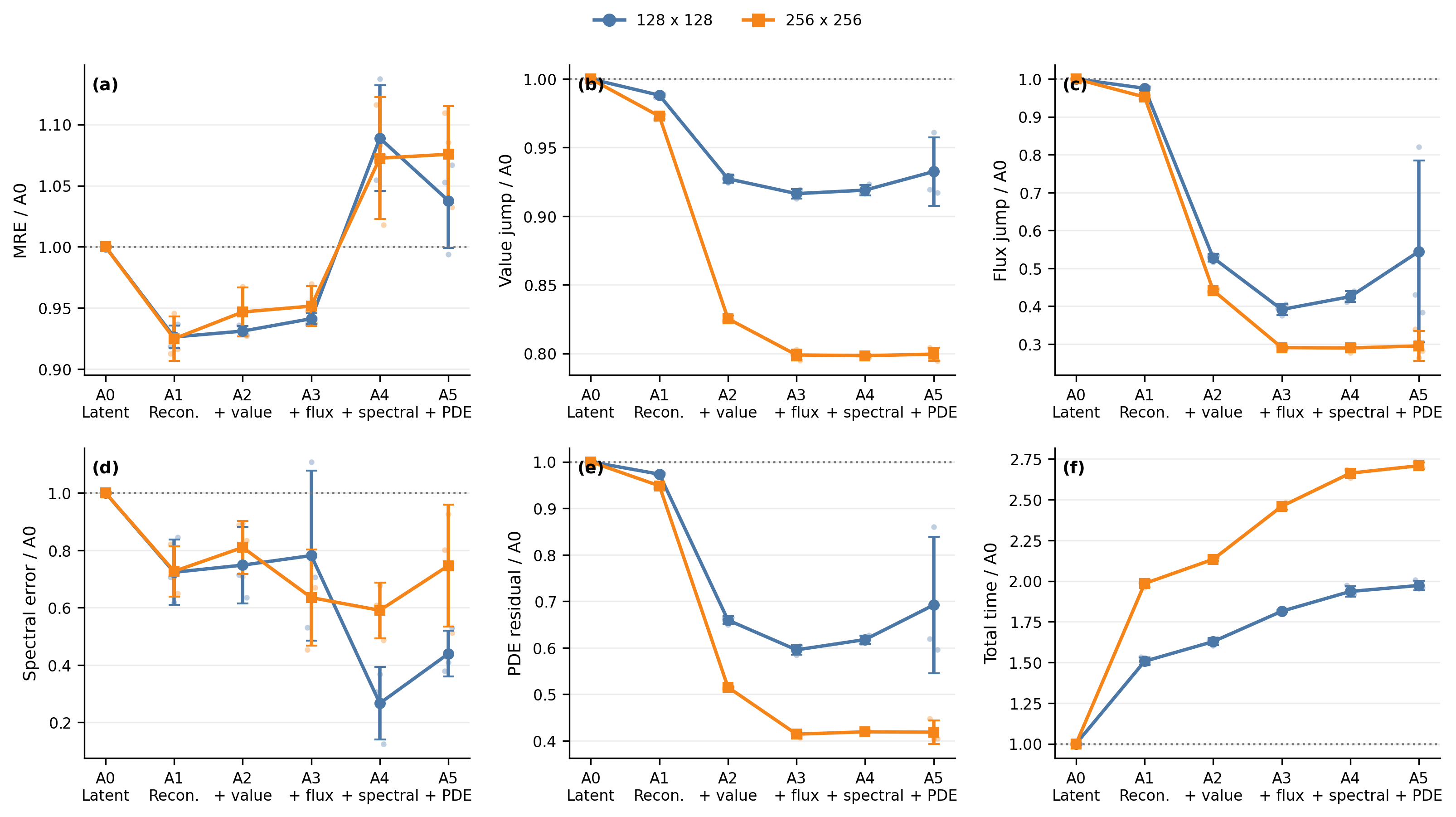}
  \caption{Ordered assembled-field loss ladder on Poisson.  All values are
  normalized by the A0 latent baseline at the corresponding resolution.
  Reconstruction improves MRE; interface value and flux terms produce the
  decisive continuity and residual reductions.  Spectral and PDE-residual
  additions raise cost without improving the default quality tradeoff}
  \label{fig:physics-loss-ladder}
\end{figure*}

\paragraph{Interface traces deliver the useful continuity correction.}
Figure~\ref{fig:physics-loss-ladder} shows a consistent sequence.  A1 gives
the best pointwise reconstruction, reducing MRE from $1.155\%$ to $1.069\%$
at $128^2$ and from $1.154\%$ to $1.067\%$ at $256^2$.  This reconstruction
gain alone barely changes flux mismatch.  Adding the value trace in A2 reduces
flux jump by $46\%$ at $128^2$ and $54\%$ at $256^2$ relative to A1.  Adding
the flux trace in A3 further reduces it to $1.71\times10^{-3}$ and
$2.38\times10^{-3}$, respectively.

Relative to A0, A3 lowers flux jump by $60.9\%$ at $128^2$ and $70.9\%$ at
$256^2$, while also lowering the corrected Poisson residual RMS by $40.4\%$
and $58.6\%$.  These residual improvements are obtained without an explicit
PDE term.  They show that derivative discontinuities at patch interfaces are
a substantial component of the discrete operator error, and that targeting
the interface is more direct than regularizing the domain-wide residual.

\begin{figure*}[htbp]
  \centering
  \includegraphics[width=0.96\textwidth]{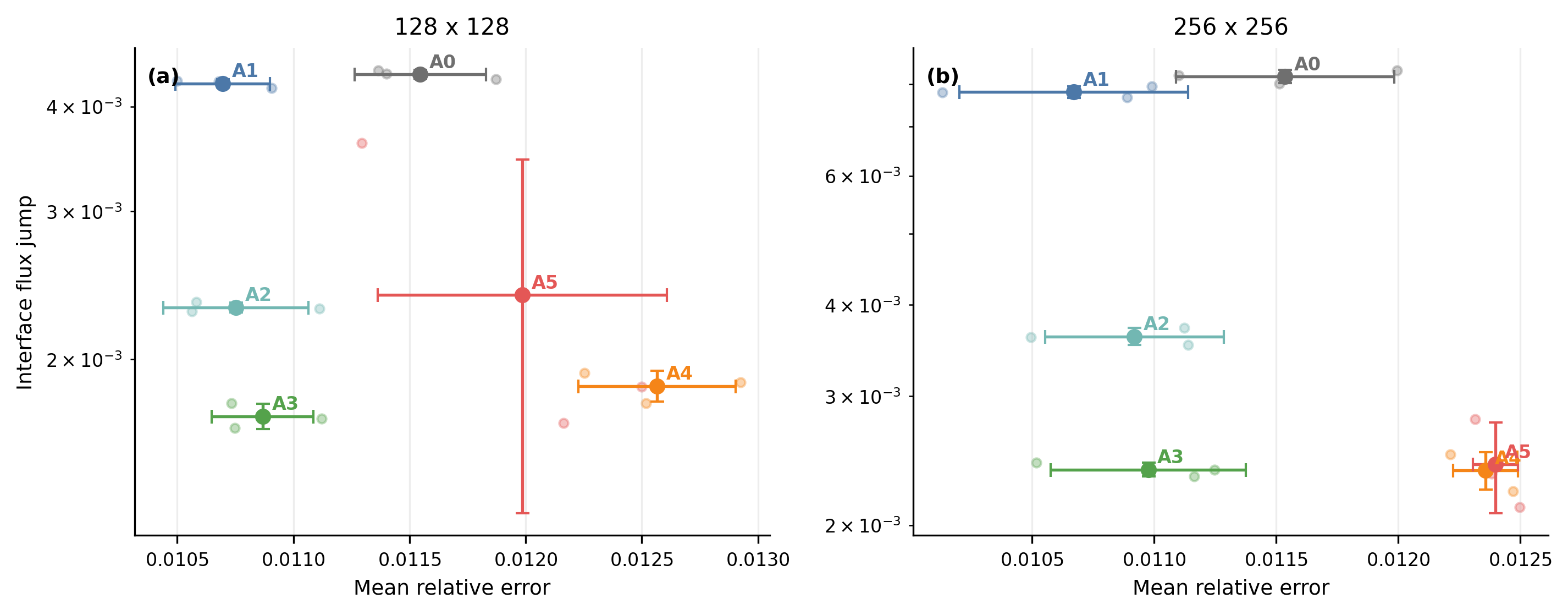}
  \caption{Accuracy--continuity tradeoff for the interface-loss ladder.
  A1 is the pointwise-MRE endpoint.  A3 accepts a small MRE increase in
  exchange for a substantially lower interface flux jump, making it the
  continuity-oriented fine-tuning choice}
  \label{fig:physics-accuracy-continuity-pareto}
\end{figure*}

\paragraph{A3 is the appropriate optional continuity refinement.}
Compared with A1, A3 increases MRE by only $1.6\%$ at $128^2$ and $2.9\%$ at
$256^2$, but reduces flux jump by $59.9\%$ and $69.5\%$ and residual RMS by
$38.8\%$ and $56.3\%$.  It also improves the value-jump diagnostic at both
resolutions.  The cost is meaningful: cumulative runtime rises from
$307.1\pm5.1$ s to $369.8\pm10.9$ s at $128^2$, and from
$507.3\pm2.9$ s to $628.7\pm4.4$ s at $256^2$.  Thus, A3 is not the universal
default for pointwise reconstruction; it is the optional refinement when
interface continuity and discrete residual are primary requirements.

\paragraph{The last two terms are specialized objectives, not default losses.}
A4 gives the best spectral error, reducing it by $63.9\%$ relative to A3 at
$128^2$, but raises MRE by $17.5\%$; at $256^2$ it improves spectral error by
$18.2\%$ but raises MRE by $15.8\%$.  It is therefore appropriate only when
radially averaged spectral agreement is the explicit downstream objective.
The correctly signed A5 PDE term does not improve on A3: its residual RMS is
$16.3\%$ higher at $128^2$ and slightly higher within three-seed variation at
$256^2$, while its MRE and cumulative time both increase.  The
tested domain-wide second-derivative loss is stiff and competes with the
reconstruction and interface objectives.  We therefore retain A3 as the
interface-aware option and do not include explicit spectral or PDE-residual
terms in the recommended fine-tuning recipe.  Appendix~\ref{app:interface-loss-ladder}
contains the calibrated weights, per-stage costs, and full diagnostics.

\subsection{Sample Efficiency}
\label{sec:sample-efficiency}

We next ask whether the coarse-global representation remains useful when
training data are limited.  On Poisson $128^2$, we train plain L2L,
Hann-overlap L2L, two-scale, and two-scale with the A3 reconstruction/value/
flux interface refinement on 1,250, 2,000, 4,000, and 8,000 examples.  Each
cell uses three paired seeds, randomized SVD, and a fixed validation/test split;
the smaller training sets are nested prefixes of the larger sets.  The lowest
budget is 1,250 because the two-scale construction requires
$(D/c)^2<m$, and $(128/4)^2=1{,}024$.

\begin{table*}[htbp]
  \centering
  \caption{Sample-efficiency crossover on Poisson $128^2$.  MRE is reported
  as a percentage over three seeds.  The two-scale variants surpass the
  full-data overlap reference once the latent map has sufficient training
  data.}
  \label{tab:sample-efficiency-crossover}
  \small
  \begin{tabular}{rlll}
    \toprule
    Training samples & L2L + overlap & Two-scale & Two-scale + interface \\
    \midrule
    1,250 & \textbf{$2.520\pm0.078$} & $3.283\pm0.153$ & $2.875\pm0.131$ \\
    2,000 & $2.399\pm0.071$ & $2.341\pm0.054$ & \textbf{$2.012\pm0.061$} \\
    4,000 & $2.321\pm0.060$ & $1.648\pm0.023$ & \textbf{$1.448\pm0.032$} \\
    8,000 & $2.292\pm0.063$ & $1.155\pm0.028$ & \textbf{$1.086\pm0.026$} \\
    \bottomrule
  \end{tabular}
\end{table*}

\begin{figure*}[htbp]
  \centering
  \includegraphics[width=0.96\textwidth]{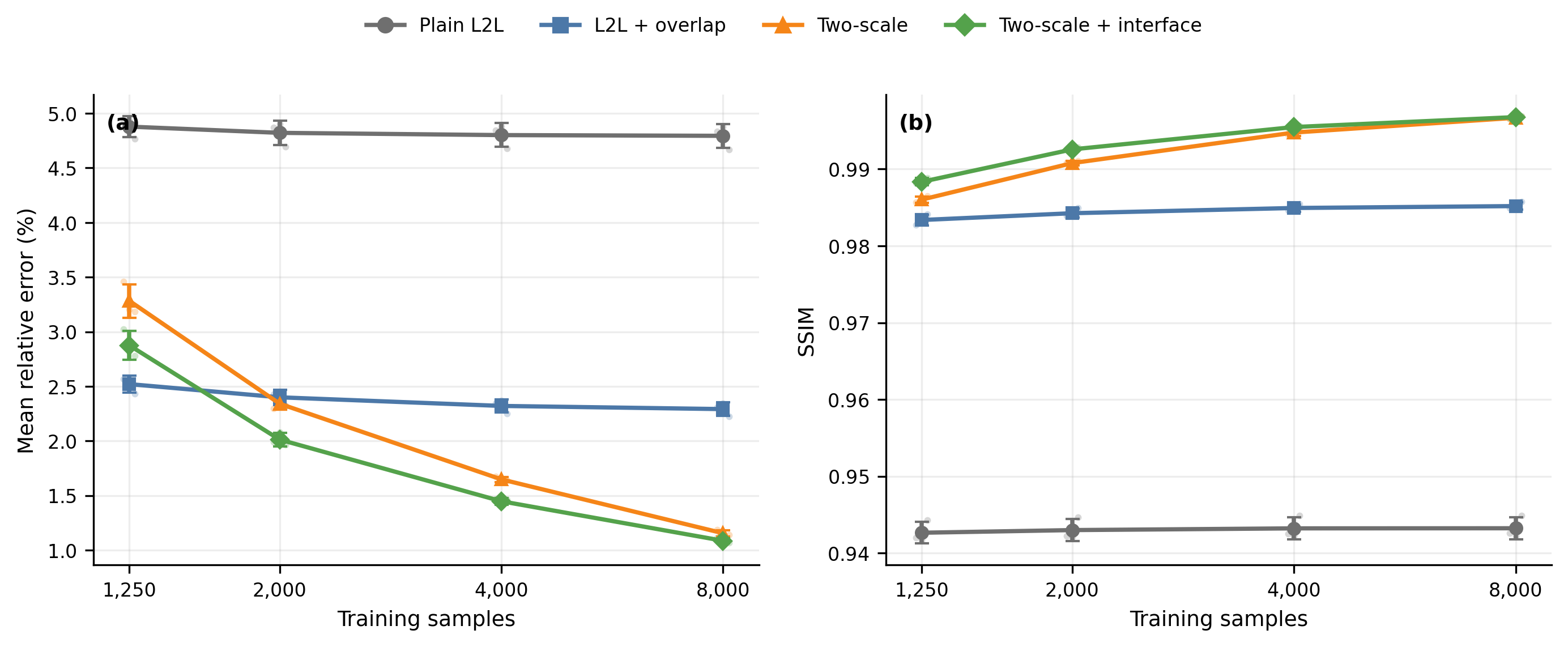}
  \caption{Poisson $128^2$ sample-efficiency curves.  Local PCA methods are
  nearly flat with additional data, whereas two-scale improves rapidly once
  the coarse and residual scores become learnable.  Interface fine-tuning
  improves the two-scale curve at every budget, at additional offline cost}
  \label{fig:sample-efficiency-accuracy}
\end{figure*}

\paragraph{Two-scale has a learnability threshold, then a markedly steeper
learning curve.}
Figure~\ref{fig:sample-efficiency-accuracy} shows that two-scale is not
uniformly superior in the smallest legal regime.  At 1,250 samples, overlap
reaches $2.520\pm0.078\%$ MRE, compared with $3.283\pm0.153\%$ for
two-scale.  This caveat matters: a compact global code does not eliminate the
need to learn its input-to-score map.  At 2,000 samples, reconstruction-only
two-scale is effectively tied with overlap in mean MRE, whereas the
interface-refined version reaches $2.012\pm0.061\%$ and wins all three paired
seeds.

By 4,000 samples, reconstruction-only two-scale reaches $1.648\pm0.023\%$
MRE, already better than overlap trained on all 8,000 examples
($2.292\pm0.063\%$).  The interface-refined variant reaches
$2.012\pm0.061\%$ using only 2,000 examples, likewise outperforming full-data
overlap with one quarter of the training data.  At 8,000 examples, two-scale
cuts overlap MRE by $49.6\%$ while retaining comparable end-to-end cost
($213.8\pm0.8$ s versus $210.0\pm4.2$ s).  The main data-efficiency claim is
therefore conditional but strong: after the low-data threshold, the
coarse-global representation reaches an overlap-level solution with
substantially fewer examples.

\begin{figure*}[htbp]
  \centering
  \includegraphics[width=0.96\textwidth]{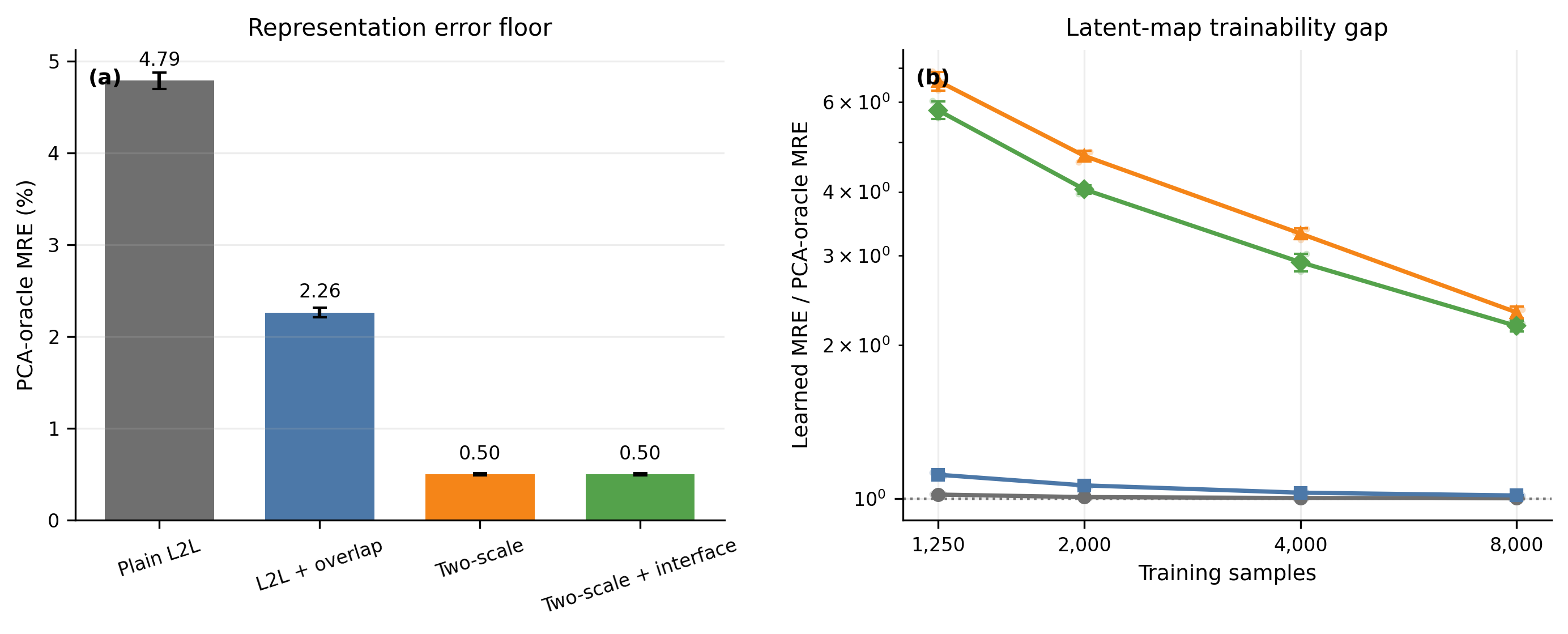}
  \caption{Representation floor and learned-to-oracle gap for the
  sample-efficiency study.  Two-scale has the lowest PCA reconstruction floor,
  but is regression-limited at small data budgets.  The closing gap explains
  the two-scale crossover with overlap}
  \label{fig:sample-efficiency-gap}
\end{figure*}

\paragraph{The different curves follow from different error bottlenecks.}
Figure~\ref{fig:sample-efficiency-gap} separates PCA reconstruction floor from
learned-operator error.  The oracle MRE is nearly constant with sample count:
$4.79\%$ for plain L2L, $2.26\%$ for overlap, and approximately $0.50\%$ for
both two-scale variants.  Plain L2L and overlap are close to these floors even
at 1,250 examples, so additional training data cannot remove their local
representation error.  Two-scale instead begins far above its lower floor:
its learned MRE is $6.6\times$ the oracle at 1,250 samples and falls to
$2.3\times$ by 8,000 samples.  The representation has headroom, and more data
is converted into accuracy by improving latent-score prediction.

Interface fine-tuning improves two-scale MRE at every budget by $6$--$14\%$
and sharply reduces its value and flux trace errors, but it doubles the
offline training cost at the smallest budgets and adds $77\%$ at 8,000
examples.  It is thus best viewed as an optional continuity refinement, not
the source of the data-efficiency effect.  Overlap remains the stronger native
flux-continuity baseline, while two-scale is the MRE/SSIM leader once enough
data are available.  The appendix reports the native and common-seam interface
audits, full metric curves, PCA-stage decomposition, costs, and representative
fields.

\subsection{Randomized PCA Solver}
\label{sec:randomized-pca-solver}

All preceding experiments use randomized SVD to obtain the retained PCA
subspaces.  We verify this choice on Poisson $128^2$ with 8,000 training
examples and five paired seeds, comparing full and randomized SVD for plain
L2L, overlap L2L, and two-scale.  Randomized SVD uses oversampling 20 and four
power iterations; the two-scale audit also includes a hybrid that randomizes
only the coarse global solve.

\begin{figure*}[htbp]
  \centering
  \includegraphics[width=0.49\textwidth]{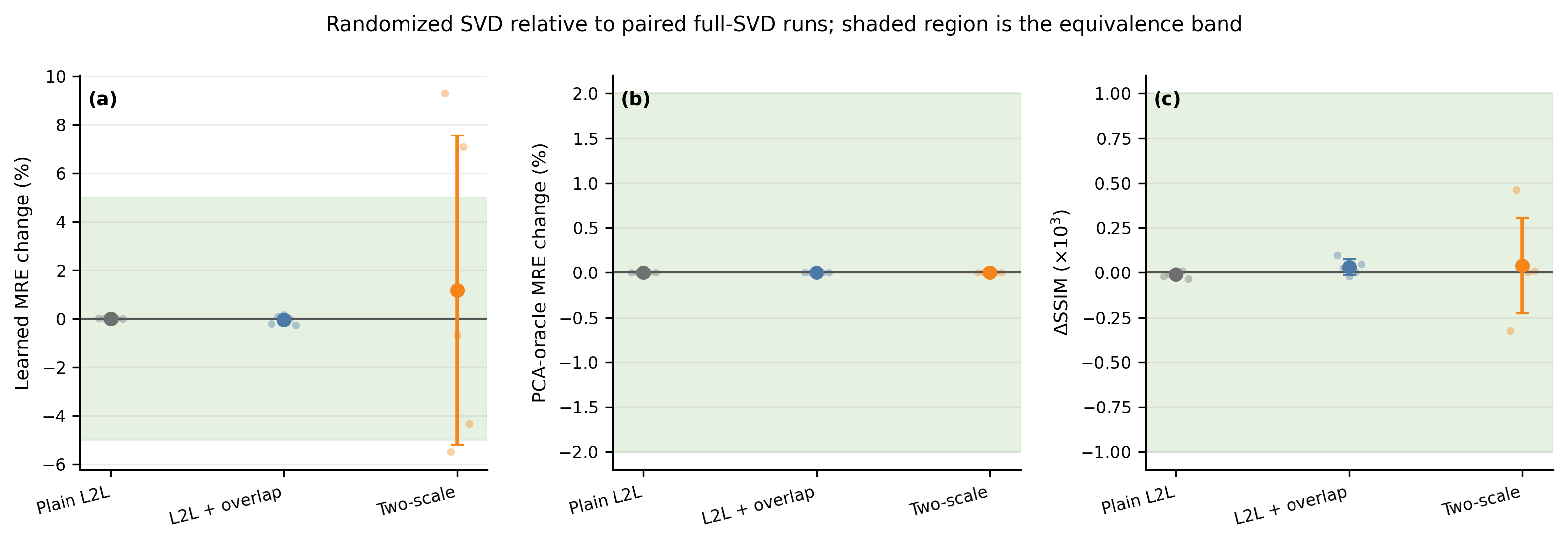}\hfill
  \includegraphics[width=0.49\textwidth]{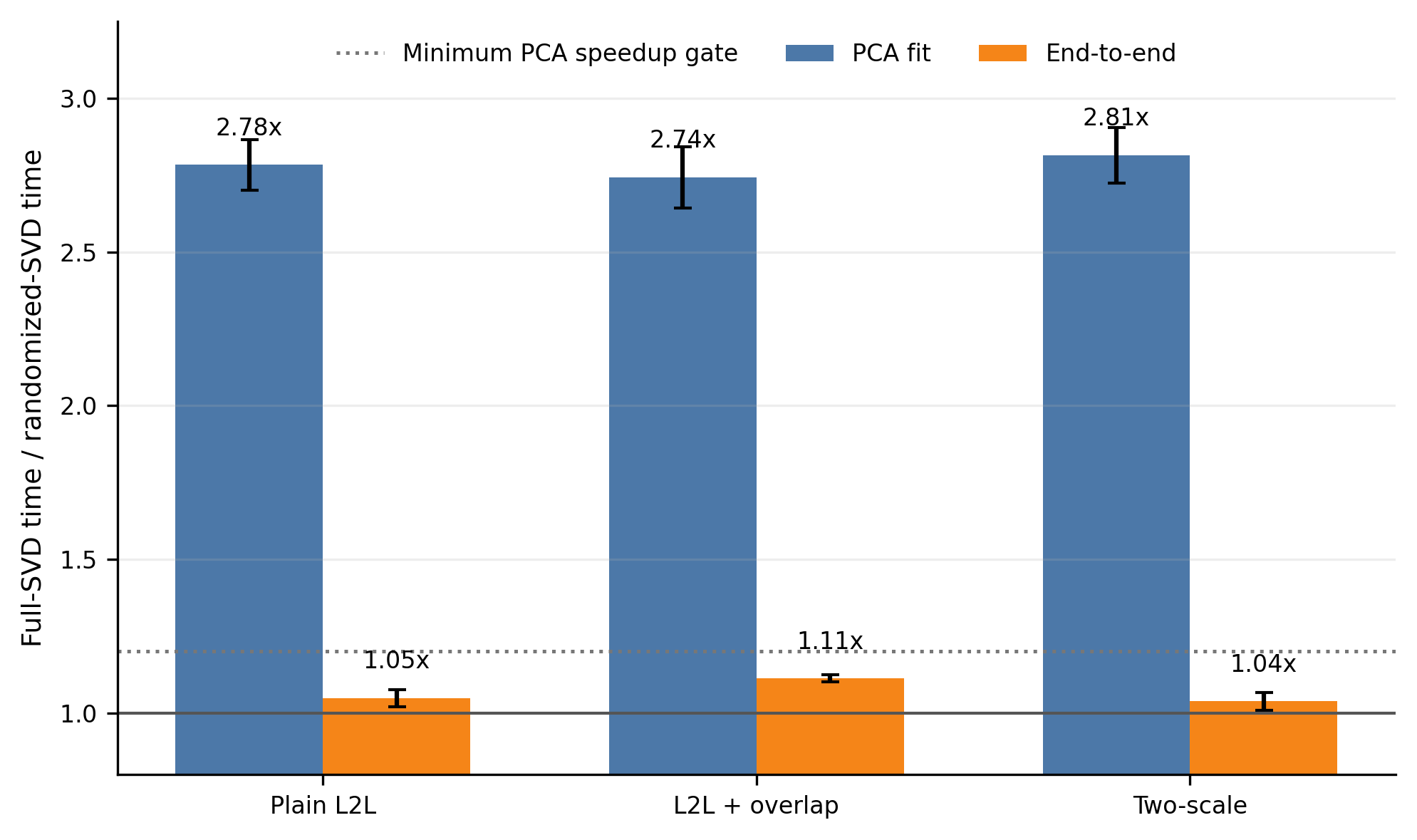}
  \caption{Randomized-SVD quality and cost audit.  \textbf{Left:} paired
  learned MRE, PCA-oracle MRE, and SSIM changes relative to full SVD; shaded
  regions denote the predeclared equivalence bands.  \textbf{Right:}
  randomized SVD reduces PCA-fit time by $2.74$--$2.81\times$ across the three
  representations, while the end-to-end gain is smaller because neural-network
  training dominates total cost}
  \label{fig:svd-solver-summary}
\end{figure*}

Figure~\ref{fig:svd-solver-summary} gives the result.  Randomization preserves
the PCA oracle to numerical precision and leaves mean learned accuracy
practically unchanged: the randomized-versus-full MRE changes are $+0.01\%$
for plain L2L, $-0.04\%$ for overlap, and $+1.17\%$ for two-scale.  Plain and
overlap satisfy the predeclared learned-MRE, PCA-oracle-MRE, and SSIM
equivalence tests.  The favorable overlap spectral change is formally
inconclusive under its symmetric tolerance.  For two-scale, the strict 90\%
learned-MRE interval
$[-5.20\%,+7.55\%]$ exceeds the $\pm5\%$ equivalence band, so its downstream
MRE decision is formally inconclusive.  Its oracle MRE, SSIM, interface
metrics, and PDE residual nevertheless satisfy their corresponding criteria,
and there is no evidence of a degraded retained representation.

The construction-time benefit is consistent: PCA fitting falls from
$9.0$ to $3.2$ s for plain L2L, $28.5$ to $10.4$ s for overlap, and $13.3$ to
$4.7$ s for two-scale.  End-to-end reductions are correspondingly modest,
between $3.6\%$ and $10.1\%$.  Randomizing only the two-scale coarse SVD is
ineffective because its local input and residual PCAs dominate the fit stage.
We therefore use fully randomized SVD as the default solver; full SVD is kept
only as the numerical reference.  Full paired diagnostics, representation
audits, and stage breakdowns are reported in Appendix~\ref{app:randomized-pca-solver}.

\section{Discussion}
\label{sec:discussion}

The central finding of this work is that patch artifacts in PCA-based operator
learning are principally a representation problem.  A local decoder gives each
patch enough freedom to reconstruct its interior, but asks independently
decoded patches to agree on low-frequency structure at their boundaries.  Hann
overlap reduces the visible consequence of that mismatch, but it duplicates
patch representations and does not change the fact that the learned output is
locally parameterized.  The two-scale representation changes that geometry:
a compact global coarse field carries the shared long-wavelength solution,
while local PCA bases encode only the residual detail.  Across the Poisson
studies, this representation change produces the largest and most durable
reduction in patch-aligned error relative to prior local PCA-Net baselines
(Secs.~\ref{sec:poisson128-results} and \ref{sec:resolution-scaling}).
Its advantage persists from $64^2$ through $256^2$ and becomes useful once a
modest amount of training data is available
(Sec.~\ref{sec:sample-efficiency}).  This supports a practical lesson that is
broader than the particular architecture: when independently decoded patches
must describe a coherent field, placing the globally shared component in the
representation can be more effective than attempting to repair coherence only
after decoding.

\paragraph{Representation first; interface optimization second.}
The ablations also distinguish two interventions that can otherwise appear
similar in qualitative reconstructions.  Two-scale is the default model
because it improves the representation floor and the learned reconstruction
without an assembled-field training stage.  A3, the reconstruction plus
interface-value plus interface-flux fine-tuning objective, is a different
tool: it directly improves residual trace agreement after the representation
has already removed the dominant blockwise error.  The loss ladder shows that
value and flux traces are the useful additions, whereas the tested radial
spectral and domain-wide PDE-residual losses do not improve the default
accuracy--continuity--cost tradeoff (Sec.~\ref{sec:interface-loss-ladder}).
The recommended interpretation is consequently not that physics losses
replace a suitable output representation.  Rather, the representation provides
the low-cost general-purpose model, and A3 is an optional refinement when
normal-derivative continuity or discrete residual is a primary requirement.

\paragraph{The remaining bottleneck is score prediction, not PCA decoding.}
The PCA-oracle experiments give the most useful diagnosis of the method's
remaining error.  The selected rank-10, $c=4$, 99.5\%-residual representation
has a low reconstruction floor, while increasing coarse rank can make the
learned score map harder to fit despite improving the oracle
(Sec.~\ref{sec:coarse-global-design}).  This separation is especially clear
for heterogeneous Darcy flow: the two-scale oracle is highly accurate, but
the learned model retains a large coefficient-to-score prediction gap
(Sec.~\ref{sec:darcy-generalization}).  Thus, simply adding more PCA modes is
unlikely to be the productive next step.  The more direct opportunity is a
stronger predictor of the existing coarse and residual scores, with explicit
conditioning between those blocks and with input features suited to rough,
discontinuous coefficient fields.  Multiresolution encoders, coefficient-aware
global context, and cross-block latent predictors are natural candidates, as
long as they retain the compact fixed-basis decoder that makes the method
attractive.

\paragraph{What the Darcy result does and does not establish.}
On heterogeneous Darcy flow, two-scale substantially improves flux-jump and
Darcy-residual diagnostics relative to plain L2L, while its MRE improvement is
modest and FNO remains the high-capacity accuracy reference
(Sec.~\ref{sec:darcy-generalization}).  This is an important qualification,
not a weakness to hide.  The representation transfers as a mechanism for
physical-field coherence, but the local input encoding and modest latent
operator are not yet competitive with a full-field neural operator for
discontinuous coefficient-to-solution maps.  The present interface term also
matches the normal derivative $\partial_n u$, whereas the conservative Darcy
quantity is the coefficient-weighted flux $a\,\partial_n u$.  A future Darcy
variant should impose this weighted trace directly, ideally using a
conservative discretization that shares its flux stencil with the residual
diagnostic.

\paragraph{Evaluation must match the claimed artifact.}
Tiling is spatially localized and derivative-sensitive.  We therefore treat
interface value and flux jumps, common-seam audits, and PDE residuals as the
primary artifact diagnostics, rather than relying on global MRE alone.  The
radially averaged spectral error remains useful for checking field-energy
content, but it discards phase and spatial location and can rank a visibly
tiled field favorably when its average energy spectrum is close to the target.
It should consequently be reported alongside, rather than substituted for,
interface diagnostics.  Direct overlap versus nonoverlap comparisons require
the stride-controlled common-seam audit because their native seam geometries
are different.  Finally, all Poisson residual results in this manuscript use
the generator-consistent $\Delta_h \widehat{u}-f$ convention; earlier
opposite-sign residual records are historical outputs and are not aggregated
with these results.

\paragraph{Scope and limitations.}
Several limitations delimit the current claims.  First, two-scale does not
give an exact continuity guarantee: independently decoded residual patches can
still disagree, and overlap remains a strong reference for strict seam
control.  Second, A3 improves continuity but adds a substantial offline
fine-tuning cost, so its favorable physical-field metrics should not be read
as a free inference-time improvement.  Third, the PCA bases are fitted for a
fixed grid resolution and patch topology; the method is not a
discretization-invariant operator and has not been tested on irregular meshes,
complex geometries, transient systems, or distribution shifts beyond the
Poisson and Darcy settings considered here.  Fourth, the measured scaling
study ends at $256^2$.  The $512^2$ and $1024^2$ time curves are explicitly
local log-log projections, not experimental evidence.  Finally, randomized
SVD was validated in detail on the Poisson-$128^2$ construction study
(Sec.~\ref{sec:randomized-pca-solver}); broader solver equivalence across
problems and resolutions remains a worthwhile systems check.

These limitations suggest a focused path forward.  The next method should
retain the two-scale decoder, improve coarse and residual score prediction for
heterogeneous inputs, and use conservative interface constraints only where
their added optimization cost is justified.  Extending the representation to
multiple coarse levels, adaptive or geometry-aware patches, and
coefficient-weighted flux supervision would test whether the same
representation-first principle can carry beyond regular steady-grid problems.
Within the present scope, the evidence supports a simple default: randomized
two-scale localized PCA-Net with block-balanced latent supervision, augmented
by A3 only when continuity-oriented physical diagnostics matter enough to pay
for fine-tuning.

\section{Conclusion}
\label{sec:conclusion}

Patch-based PCA operator learning offers an attractive low-cost alternative to
full-field neural operators, but independently decoded local output patches
create a structural source of tiling artifacts.  This work introduced
two-scale localized PCA-Net to address that source directly.  The method
represents the output as a globally decoded coarse field plus a mosaic of
locally decoded residual patches: the coarse channel carries shared,
long-wavelength structure, while the local channel preserves the compact
expressiveness of patch PCA.  This is a representation change, rather than a
post-processing rule or a larger local network.

Across Poisson problems from $64^2$ to $256^2$, the two-scale representation
consistently improves reconstruction quality and suppresses patch-aligned
artifacts relative to plain and overlapping localized PCA-Net baselines.  It
does so while retaining a lower PCA-fit cost than overlap, because it avoids
duplicated overlapping patch bases.  The construction study further shows that
randomized SVD preserves the relevant representation behavior while reducing
the PCA stage cost.  On heterogeneous Darcy flow, the same representation
substantially improves interface flux and discrete residual diagnostics, even
though its pointwise MRE gain is modest and a full-field FNO remains the
absolute-accuracy reference.  The PCA-oracle results identify the reason:
the two-scale decoder is highly expressive, whereas predicting its scores
from rough coefficient fields is the remaining challenge.

We also evaluated assembled-field fine-tuning.  Reconstruction plus
interface-value plus interface-flux supervision (A3) is an effective optional
continuity-oriented refinement: it reduces residual trace mismatch and PDE
residual beyond the base two-scale model.  Its additional training cost,
however, means it should be selected for applications where physical-field
coherence warrants that expense, rather than treated as the universal default.
The tested radial spectral and domain-wide residual terms did not improve the
overall accuracy--continuity--cost tradeoff.

The resulting practical recommendation is therefore simple: use randomized
two-scale localized PCA-Net with a rank-10 coarse basis, a $c=4$ restriction,
99.5\% residual retained variance, and block-balanced latent supervision as
the default PCA-Net configuration.  Add A3 only when interface continuity is
a first-class objective.  More broadly, the results show that local reduced
representations need not sacrifice global coherence: assigning global and
local structure to separate, complementary latent channels provides a compact
route to artifact-reduced PDE operator learning.  Future work should improve
coarse and residual score prediction for heterogeneous inputs, incorporate
coefficient-weighted conservative flux constraints, and extend the approach to
multilevel, geometry-aware, and time-dependent settings.

% \paragraph{Acknowledgments}
% We are grateful for the technical assistance of A. Author.

\paragraph{Funding Statement}
This work was supported  by the AFOSR Grant FA9550-24-1-0327.

% \paragraph{Competing Interests}
% A statement about any financial, professional, contractual or personal relationships or situations that could be perceived to impact the presentation of the work --- or `None' if none exist

\paragraph{Data and Code Availability}

The data supporting the findings of this study are available from the
corresponding author upon reasonable request. The source code and implementation
used in this study are publicly available at
\url{https://github.com/dmrigank/PAtch_PCA_methods_extensions}.

% \paragraph{Ethical Standards}
% The research meets all ethical guidelines, including adherence to the legal requirements of the study country.

% \paragraph{Author Contributions}
% Please provide an author contributions statement using the CRediT taxonomy roles as a guide {\verb+\url{https://www.casrai.org/credit.html}+}. Conceptualization: A.A; A.B. Methodology: A.A; A.B. Data curation: A.C. Data visualisation: A.C. Writing original draft: A.A; A.B. All authors approved the final submitted draft.

% \paragraph{Supplementary Material}
% State whether any supplementary material intended for publication has been provided with the submission.

\bibliographystyle{apalike}
%\bibliography{Sample-refs}

\bibliography{references}

\section{Appendix}

\subsection{Additional Poisson-128 Metrics and Interface Diagnostics}
\label{app:poisson128-diagnostics}

The headline comparison in
Table~\ref{tab:poisson128-headline} focuses on MRE, SSIM, representation
construction, and cumulative cost. Here we report the complementary
reconstruction, spectral, physical-residual, and interface diagnostics for the
same five-seed Poisson-128 study. All Poisson residuals use the
generator-consistent convention
$\Delta_h\widehat u-f$.

\begin{table*}[htbp]
  \centering
  \caption{Additional Poisson-128 reconstruction and physical-field diagnostics
  over five paired seeds. PCA-oracle MRE is reported only for models with an
  explicit PCA output representation. Two-scale and two-scale + interface have
  the same oracle because interface-aware fine-tuning modifies the latent
  predictor but not the fitted representation.}
  \label{tab:poisson128-additional-metrics}
  \small
  \begin{tabular}{lccccc}
\toprule
Method
& \makecell{PCA-oracle\\MRE (\%)}
& MSE
& MAE
& \makecell{Spectral\\error}
& \makecell{PDE residual\\RMS} \\
\midrule

Global PCA-Net
& \makecell{$7.539$\\$\pm\,0.166$}
& \makecell{$1.000{\times}10^{-7}$\\$\pm\,0.010{\times}10^{-7}$}
& \makecell{$2.353{\times}10^{-4}$\\$\pm\,0.012{\times}10^{-4}$}
& \makecell{$2.636{\times}10^{-3}$\\$\pm\,0.270{\times}10^{-3}$}
& \makecell{$0.08861$\\$\pm\,0.00015$} \\

Plain L2L
& \makecell{$4.825$\\$\pm\,0.094$}
& \makecell{$4.100{\times}10^{-8}$\\$\pm\,0.039{\times}10^{-8}$}
& \makecell{$1.293{\times}10^{-4}$\\$\pm\,0.005{\times}10^{-4}$}
& \makecell{$3.545{\times}10^{-3}$\\$\pm\,0.130{\times}10^{-3}$}
& \makecell{$3.16061$\\$\pm\,0.02330$} \\

L2L + overlap
& \makecell{$2.277$\\$\pm\,0.051$}
& \makecell{$9.361{\times}10^{-9}$\\$\pm\,0.054{\times}10^{-9}$}
& \makecell{$7.094{\times}10^{-5}$\\$\pm\,0.016{\times}10^{-5}$}
& \makecell{$1.522{\times}10^{-3}$\\$\pm\,0.310{\times}10^{-3}$}
& \makecell{$0.10271$\\$\pm\,0.00053$} \\

L2L + RefinementNet
& --
& \makecell{$1.811{\times}10^{-8}$\\$\pm\,0.036{\times}10^{-8}$}
& \makecell{$1.013{\times}10^{-4}$\\$\pm\,0.009{\times}10^{-4}$}
& \makecell{$3.017{\times}10^{-2}$\\$\pm\,0.500{\times}10^{-2}$}
& \makecell{$0.32290$\\$\pm\,0.04543$} \\

FNO
& --
& \makecell{$1.035{\times}10^{-10}$\\$\pm\,0.140{\times}10^{-10}$}
& \makecell{$7.466{\times}10^{-6}$\\$\pm\,0.370{\times}10^{-6}$}
& \makecell{$5.344{\times}10^{-4}$\\$\pm\,4.300{\times}10^{-4}$}
& \makecell{$0.10274$\\$\pm\,0.01463$} \\

Two-scale
& \makecell{$0.501$\\$\pm\,0.009$}
& \makecell{$2.325{\times}10^{-9}$\\$\pm\,0.130{\times}10^{-9}$}
& \makecell{$3.116{\times}10^{-5}$\\$\pm\,0.058{\times}10^{-5}$}
& \makecell{$5.478{\times}10^{-3}$\\$\pm\,0.560{\times}10^{-3}$}
& \makecell{$0.24481$\\$\pm\,0.00257$} \\

Two-scale + interface
& \makecell{$0.501$\\$\pm\,0.009$}
& \makecell{$2.066{\times}10^{-9}$\\$\pm\,0.140{\times}10^{-9}$}
& \makecell{$3.135{\times}10^{-5}$\\$\pm\,0.075{\times}10^{-5}$}
& \makecell{$4.058{\times}10^{-3}$\\$\pm\,0.690{\times}10^{-3}$}
& \makecell{$0.14525$\\$\pm\,0.00187$} \\

\bottomrule
\end{tabular}
\end{table*}

\begin{figure*}[htbp]
  \centering
  \includegraphics[width=0.48\textwidth]
  {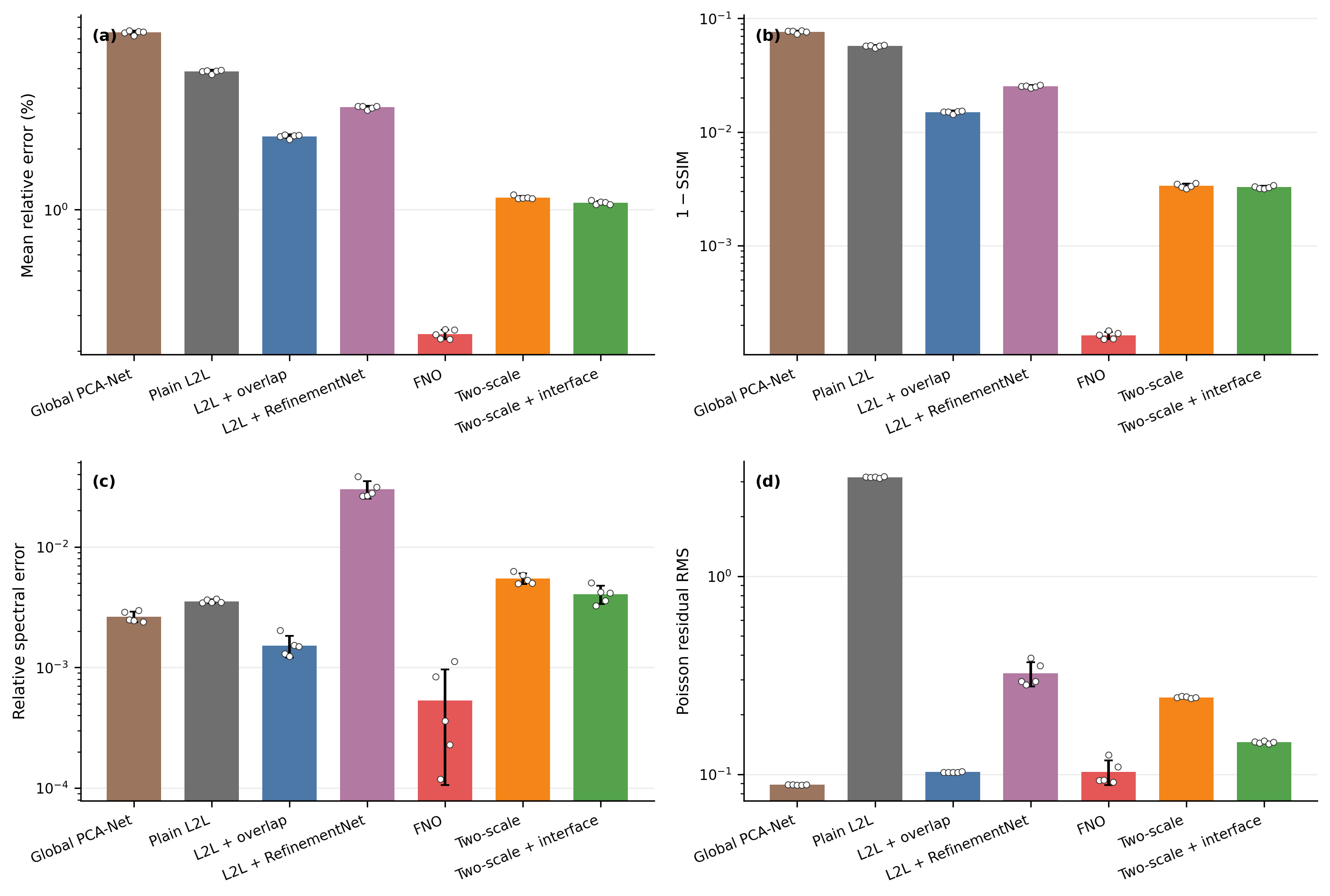}\hfill
  \includegraphics[width=0.48\textwidth]
  {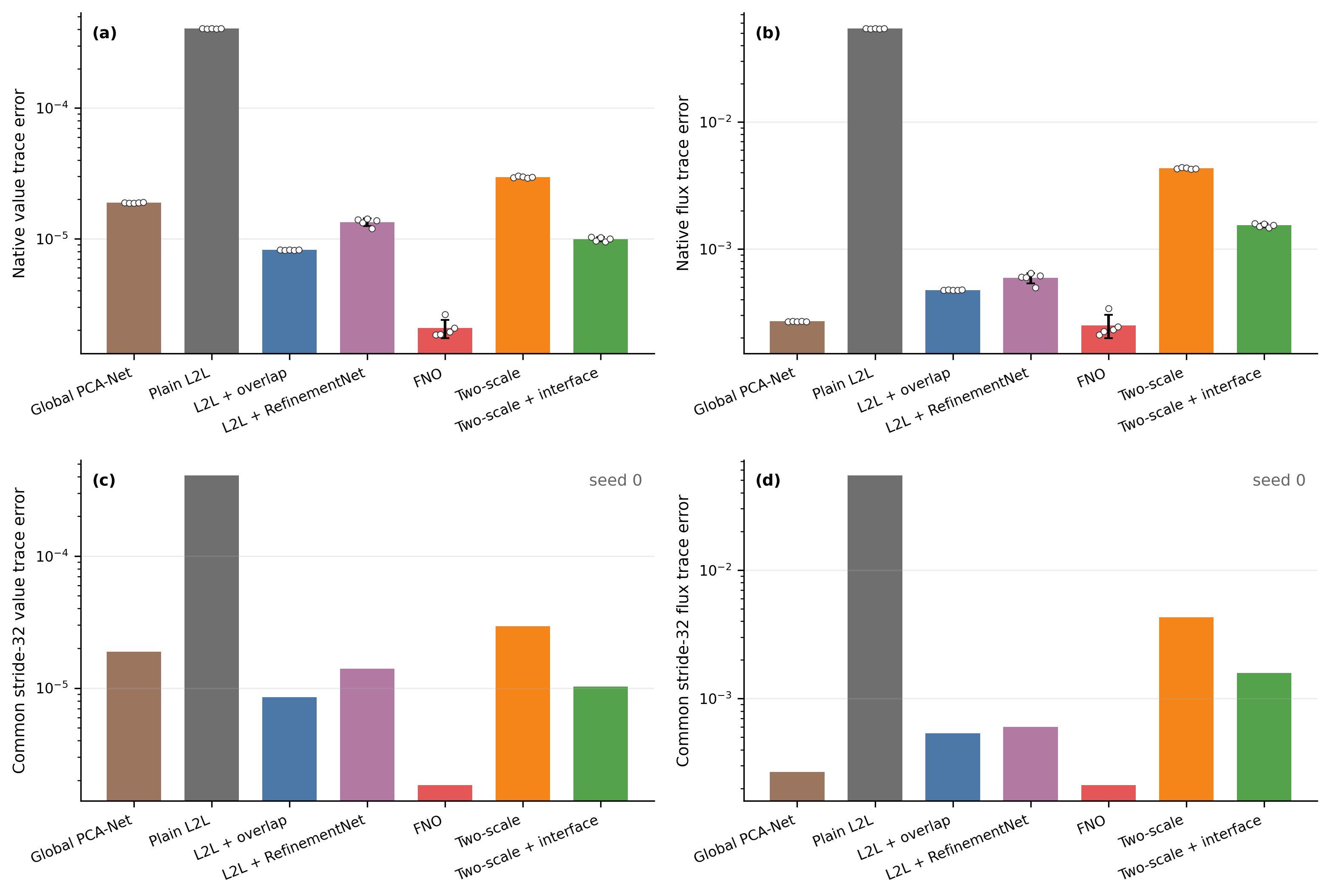}
  \caption{Additional Poisson-128 diagnostics.
  \textbf{Left:} supplementary reconstruction and distributional metrics.
  \textbf{Right:} interface diagnostics evaluated on each method's native seam
  geometry. Native quantities are directly comparable within a fixed geometry,
  but direct overlap-versus-nonoverlap continuity comparisons require the
  common-seam audit in Table~\ref{tab:poisson128-common-seam}}
  \label{fig:poisson128-appendix-diagnostics}
\end{figure*}

Figure~\ref{fig:poisson128-appendix-diagnostics} summarizes the complementary
accuracy and native-interface diagnostics. The native interface quantities
must be interpreted with care because overlap and nonoverlap methods do not
share the same seam geometry: overlap is evaluated at stride-16 boundaries,
whereas the nonoverlapping models use stride-32 boundaries.

Because full prediction arrays were retained only for seed 0, a separate
geometry-controlled diagnostic re-evaluates the retained predictions from
overlap, two-scale, and two-scale + interface on the same stride-32 seam set.

\begin{table}[htbp]
  \centering
  \caption{Common-seam Poisson-128 interface audit on retained seed-0
  predictions. All methods are evaluated on identical stride-32 interfaces.
  The comparison is diagnostic and is not treated as a multi-seed statistical
  ranking.}
  \label{tab:poisson128-common-seam}
  \small
  \begin{tabular}{lcc}
    \toprule
    Method
    & Value-trace error
    & Normal-derivative trace error \\
    \midrule
    L2L + overlap
    & $8.54\times10^{-6}$
    & $5.37\times10^{-4}$ \\
    Two-scale
    & $2.94\times10^{-5}$
    & $4.28\times10^{-3}$ \\
    Two-scale + interface
    & $1.03\times10^{-5}$
    & $1.58\times10^{-3}$ \\
    \bottomrule
  \end{tabular}
\end{table}

On this shared seam geometry, overlap retains the lowest value- and
normal-derivative trace errors among the PCA-based methods.
Interface-aware fine-tuning substantially narrows the corresponding
two-scale errors but does not reverse this ordering. This complements the
main-text result: two-scale is favored by global reconstruction accuracy and
representation-construction cost, whereas overlap remains the stronger
continuity reference.

\subsection{Supporting Diagnostics for the Representation--Fine-Tuning Study}
\label{app:mechanism-ablation}

The main-text mechanism study focuses on reconstruction, interface fidelity,
and cumulative cost. Here we provide complementary paired-effect and
native-interface diagnostics. All Poisson residual quantities use the
generator-consistent convention $\Delta_h\widehat u-f$.

Table~\ref{tab:mechanism-paired-effects} isolates the three controlled
mechanism comparisons for which both methods share the same stride-32
interface geometry. Negative values denote reductions in the corresponding
quantity. Cross-geometry interface changes involving overlap are excluded
from this table and evaluated instead through the common-seam audit below.

\begin{table*}[htbp]
  \centering
  \caption{Paired percentage effects in the Poisson-128 mechanism study.
  Each comparison uses identical stride-32 interface geometry, allowing the
  value- and normal-derivative trace changes to be interpreted directly.
  Negative values indicate reductions in the corresponding metric or cost.}
  \label{tab:mechanism-paired-effects}
  \small
  \begin{tabular}{lccccc}
\toprule
Comparison
& \makecell{MRE\\change}
& \makecell{Value-trace\\change}
& \makecell{Flux-trace\\change}
& \makecell{Spectral\\change}
& \makecell{Time\\change} \\
\midrule

\makecell[l]{Plain L2L + interface\\vs Plain L2L}
& $-5.3\%$
& $-19.0\%$
& $-18.7\%$
& $+14.5\%$
& $+86.6\%$ \\

\makecell[l]{Two-scale\\vs Plain L2L}
& $-76.2\%$
& $-92.7\%$
& $-92.1\%$
& $+54.9\%$
& $+12.7\%$ \\

\makecell[l]{Two-scale + interface\\vs Two-scale}
& $-5.3\%$
& $-66.1\%$
& $-64.0\%$
& $-19.3\%$
& $+79.9\%$ \\

\makecell[l]{L2L + overlap\\vs Plain L2L}
& $-52.2\%$
& $-98.0\%$
& $-99.1\%$
& $-57.0\%$
& $+11.6\%$ \\

\makecell[l]{Two-scale + interface\\vs L2L + overlap}
& $-52.8\%$
& $+22.4\%$
& $+226.0\%$
& $+198.9\%$
& $+81.6\%$ \\

\bottomrule
\end{tabular}
\end{table*}

\begin{figure*}[htbp]
  \centering
  \includegraphics[width=0.48\textwidth]
  {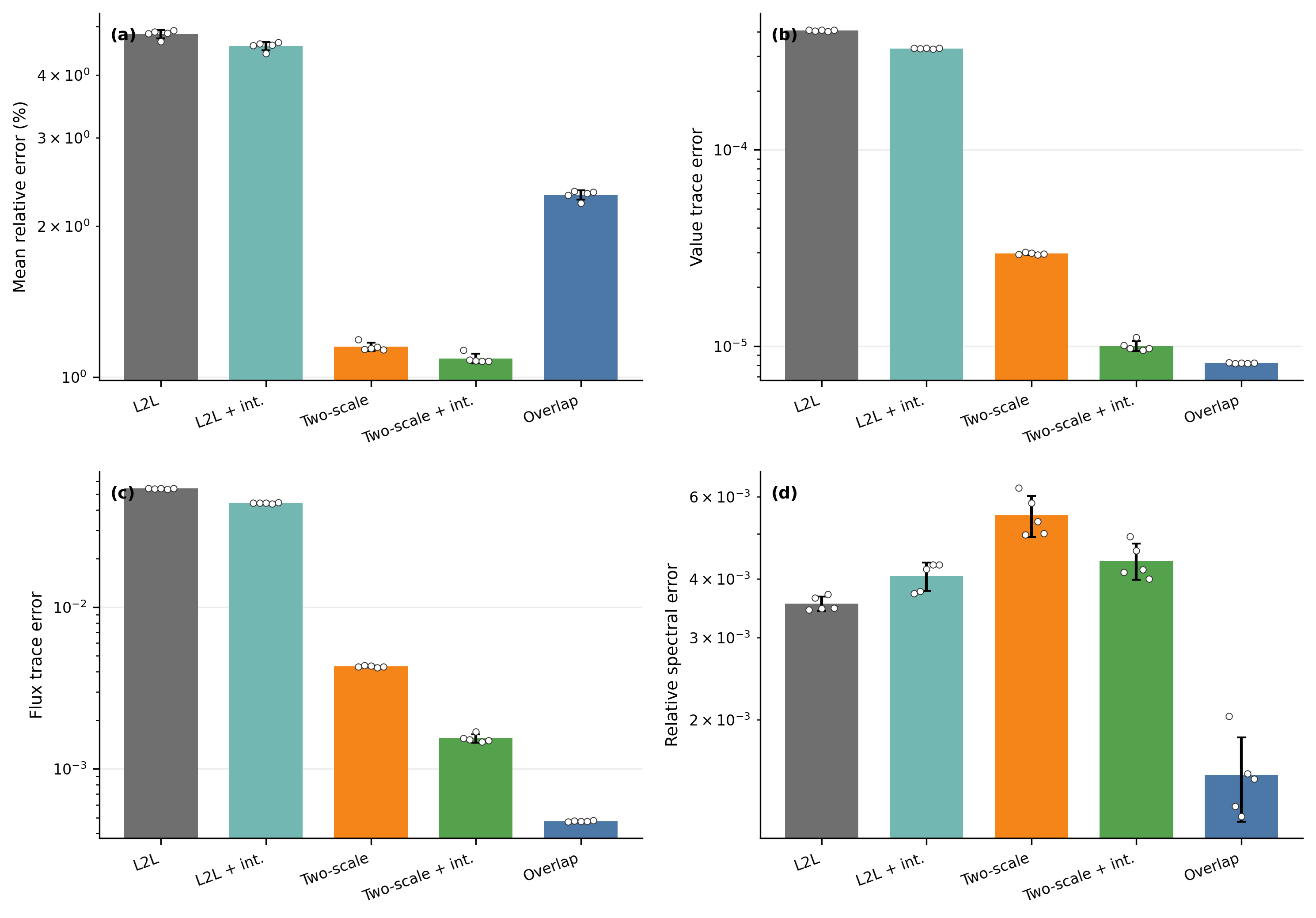}\hfill
  \includegraphics[width=0.48\textwidth]
  {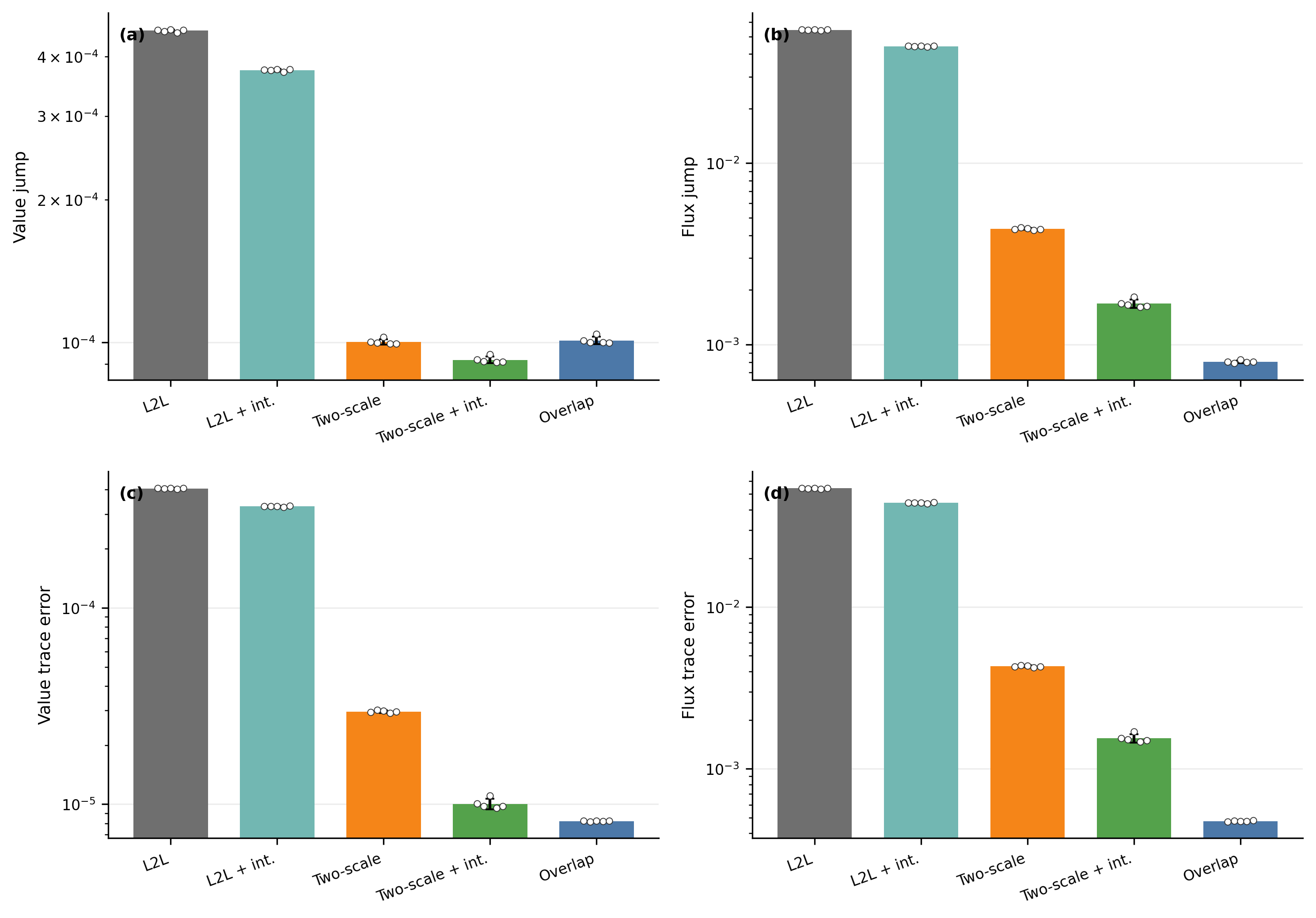}
  \caption{Supplementary Poisson-128 diagnostics for the mechanism study.
  \textbf{Left:} additional reconstruction and physical-field outcomes.
  \textbf{Right:} interface diagnostics evaluated on each method's native seam
  geometry. Overlap uses stride-16 interfaces, whereas the other methods use
  stride-32 interfaces. Direct overlap-versus-nonoverlap comparison therefore
  uses the geometry-controlled audit in
  Table~\ref{tab:mechanism-common-seams}}
  \label{fig:mechanism-native-diagnostics}
\end{figure*}

Figure~\ref{fig:mechanism-native-diagnostics} reports the supplementary
outcome and native-interface diagnostics. The overlap reference uses
stride-16 interfaces, whereas plain L2L and the two-scale variants use
stride-32 interfaces. The native quantities are therefore useful for
characterizing each representation on its own patch geometry but are not used
for direct continuity ranking across the two geometries.

For a direct continuity comparison, the retained seed-0 predictions are
re-evaluated on the same stride-32 seam set. Exact values are reported in
Table~\ref{tab:mechanism-common-seams}.

\begin{table}[htbp]
  \centering
  \caption{Geometry-controlled Poisson-128 interface audit on retained seed-0
  predictions. All methods are evaluated on identical stride-32 interfaces.
  Because full prediction arrays were retained only for seed 0, the comparison
  is treated as a diagnostic audit rather than a multi-seed statistical
  ranking.}
  \label{tab:mechanism-common-seams}
  \small
  \begin{tabular}{lcccccc}
\toprule
Method
& Seed
& Stride
& \makecell{Value\\jump}
& \makecell{Flux\\jump}
& \makecell{Value trace\\error}
& \makecell{Flux trace\\error} \\
\midrule

Plain L2L
& 0
& 32
& $4.554{\times}10^{-4}$
& $5.453{\times}10^{-2}$
& $4.078{\times}10^{-4}$
& $5.450{\times}10^{-2}$ \\

Plain L2L + interface
& 0
& 32
& $3.757{\times}10^{-4}$
& $4.431{\times}10^{-2}$
& $3.300{\times}10^{-4}$
& $4.428{\times}10^{-2}$ \\

Two-scale
& 0
& 32
& $1.003{\times}10^{-4}$
& $4.314{\times}10^{-3}$
& $2.938{\times}10^{-5}$
& $4.275{\times}10^{-3}$ \\

Two-scale + interface
& 0
& 32
& $9.201{\times}10^{-5}$
& $1.686{\times}10^{-3}$
& $1.010{\times}10^{-5}$
& $1.551{\times}10^{-3}$ \\

\bottomrule
\end{tabular}
\end{table}

The common-seam audit preserves the qualitative conclusion of the native
diagnostics: interface-aware fine-tuning substantially reduces the residual
two-scale trace errors, while overlap retains the lowest value- and
normal-derivative trace errors on the shared seam geometry. This distinction
supports the operating interpretation in
Section~\ref{sec:mechanism-ablation}: two-scale is favored for global
reconstruction quality at compact cost, whereas A3 is an optional
continuity-oriented refinement and overlap remains a strong interface-fidelity
reference.

\subsection{Supplementary Resolution-Scaling Diagnostics}
\label{app:resolution-scaling}

The figures below provide supplementary interface, representation-capacity,
timing, and qualitative evidence for the Poisson resolution-scaling study in
Section~\ref{sec:resolution-scaling}. All Poisson residuals use the
generator-consistent convention $\Delta_h\widehat u-f$. Native interface
diagnostics are reported on each method's own seam geometry; consequently,
overlap uses a denser seam set than the nonoverlapping methods at every
resolution and should be interpreted accordingly.

\begin{figure*}[htbp]
  \centering
  \includegraphics[width=\textwidth]{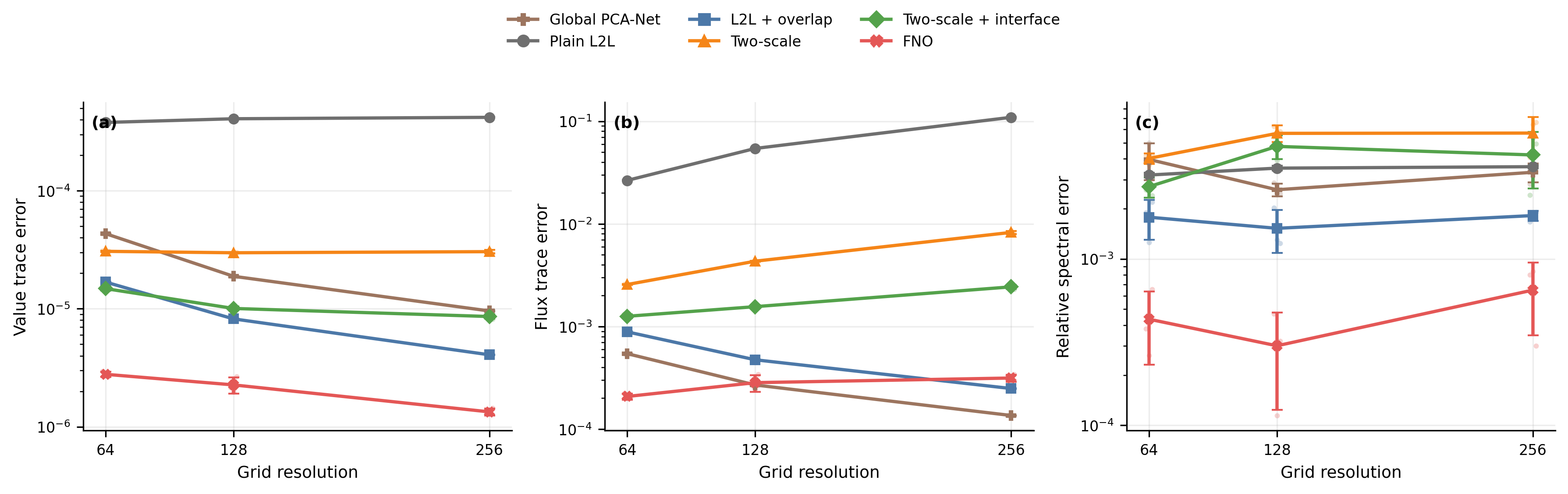}
  \caption{Supplementary interface and spectral diagnostics across Poisson
  grid resolutions. Native truth-referenced value-trace, normal-derivative
  trace, and radial spectral errors are shown for each method. A3 increasingly
  reduces the remaining two-scale interface errors with refinement. Because
  overlap uses a denser native seam geometry than the nonoverlapping methods,
  the interface quantities should be interpreted within a fixed geometry.}
  \label{fig:resolution-artifacts}
\end{figure*}

\begin{figure*}[htbp]
  \centering
  \includegraphics[width=0.48\textwidth]{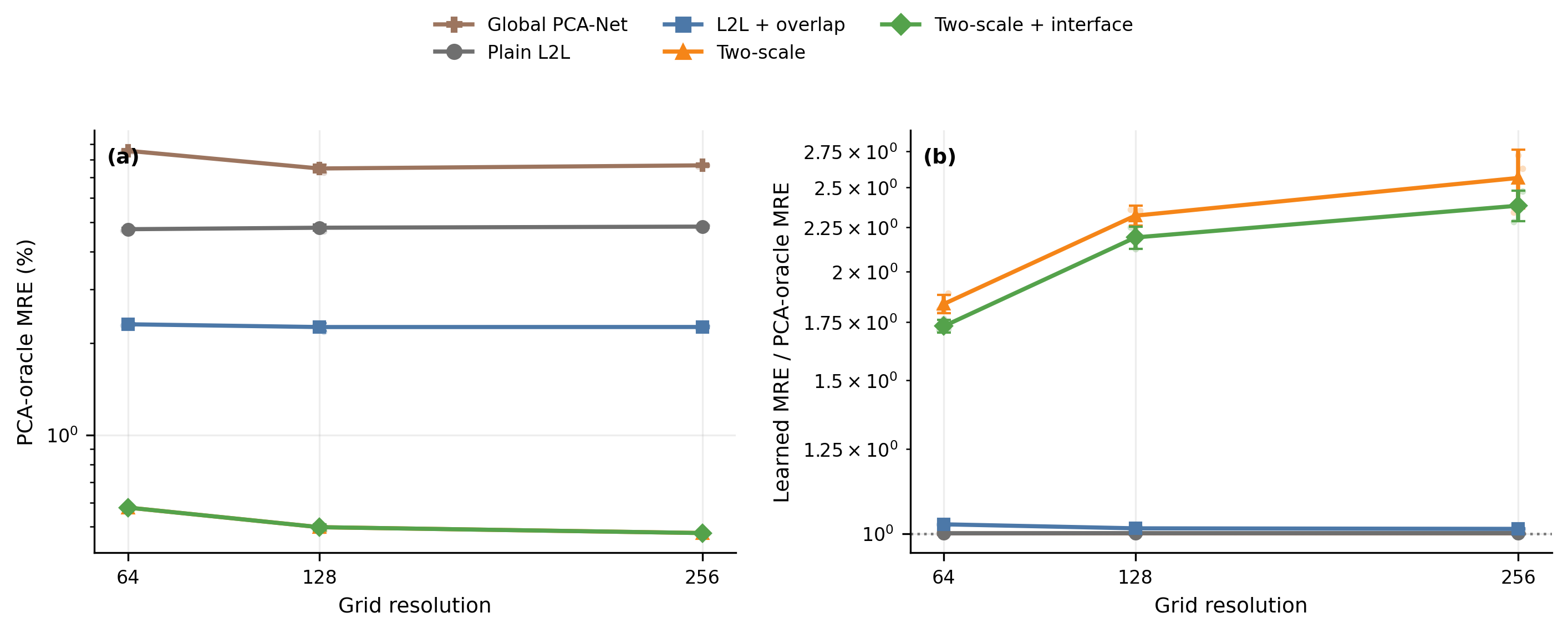}\hfill
  \includegraphics[width=0.48\textwidth]{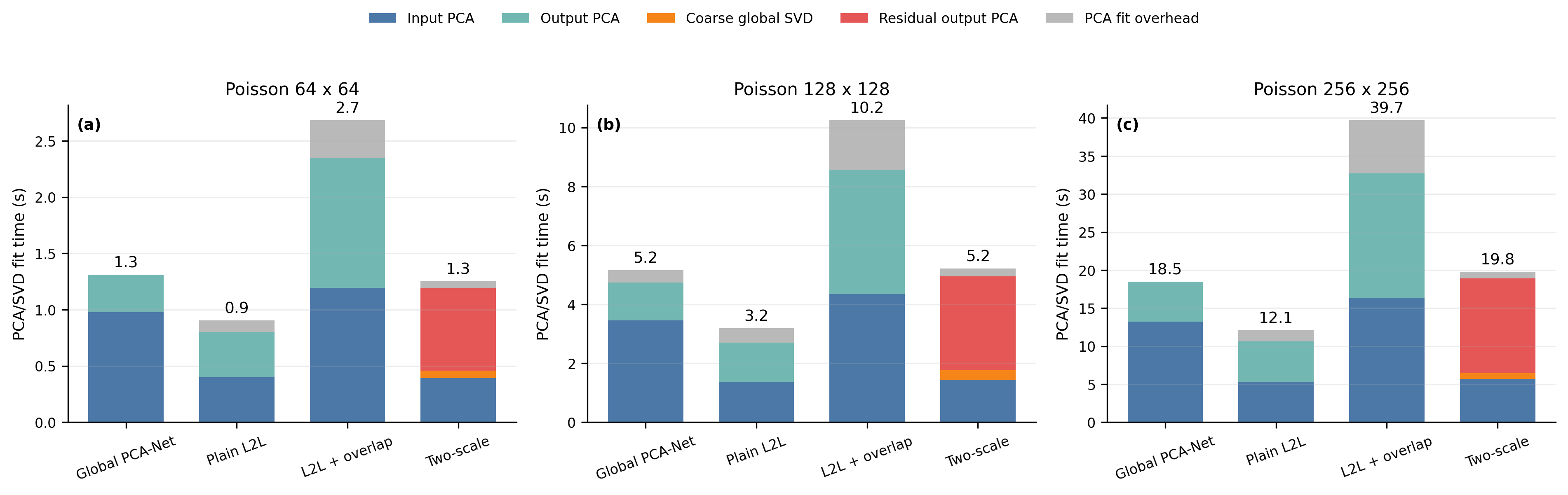}
  \caption{Representation capacity and PCA/SVD construction cost across
  resolution. \textbf{Left:} PCA-oracle and learned reconstruction error.
  The two-scale oracle floor remains low, while the learned-to-oracle gap
  grows with resolution, indicating that latent prediction becomes the more
  visible bottleneck. \textbf{Right:} stage-wise PCA/SVD fitting cost. The
  coarse-global SVD remains a small part of the two-scale cost, whereas
  overlap pays for substantially larger duplicated local PCA fits.}
  \label{fig:resolution-capacity-pca-cost}
\end{figure*}

\begin{figure*}[htbp]
  \centering
  \includegraphics[width=0.75\textwidth]{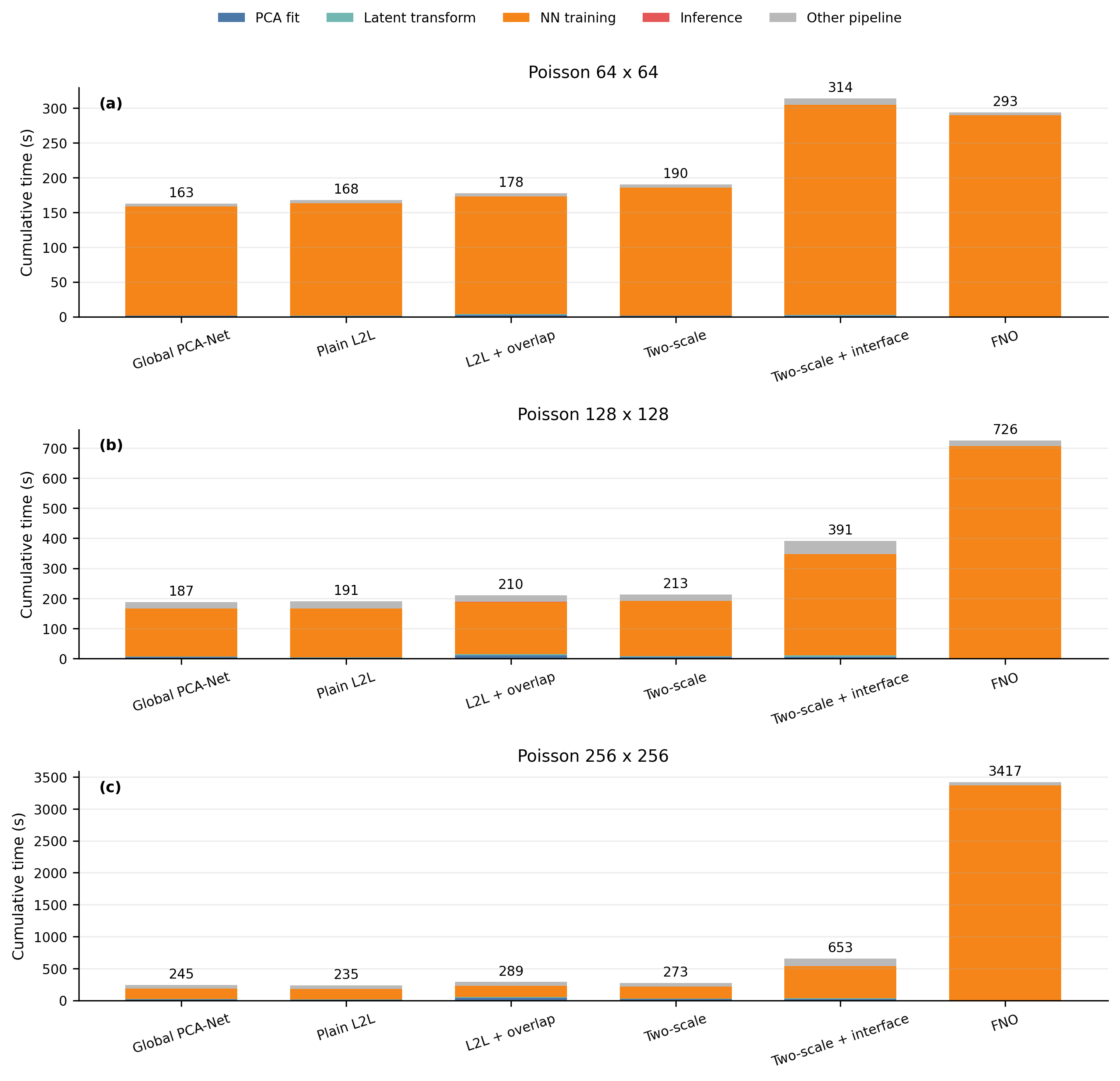}
  \caption{Cumulative stage-wise cost at each measured resolution. A3 totals
  include both the warm-start two-scale run and the subsequent
  assembled-field fine-tuning stage.}
  \label{fig:resolution-stage-costs}
\end{figure*}

\begin{figure*}[htbp]
  \centering
  \includegraphics[width=0.82\textwidth]{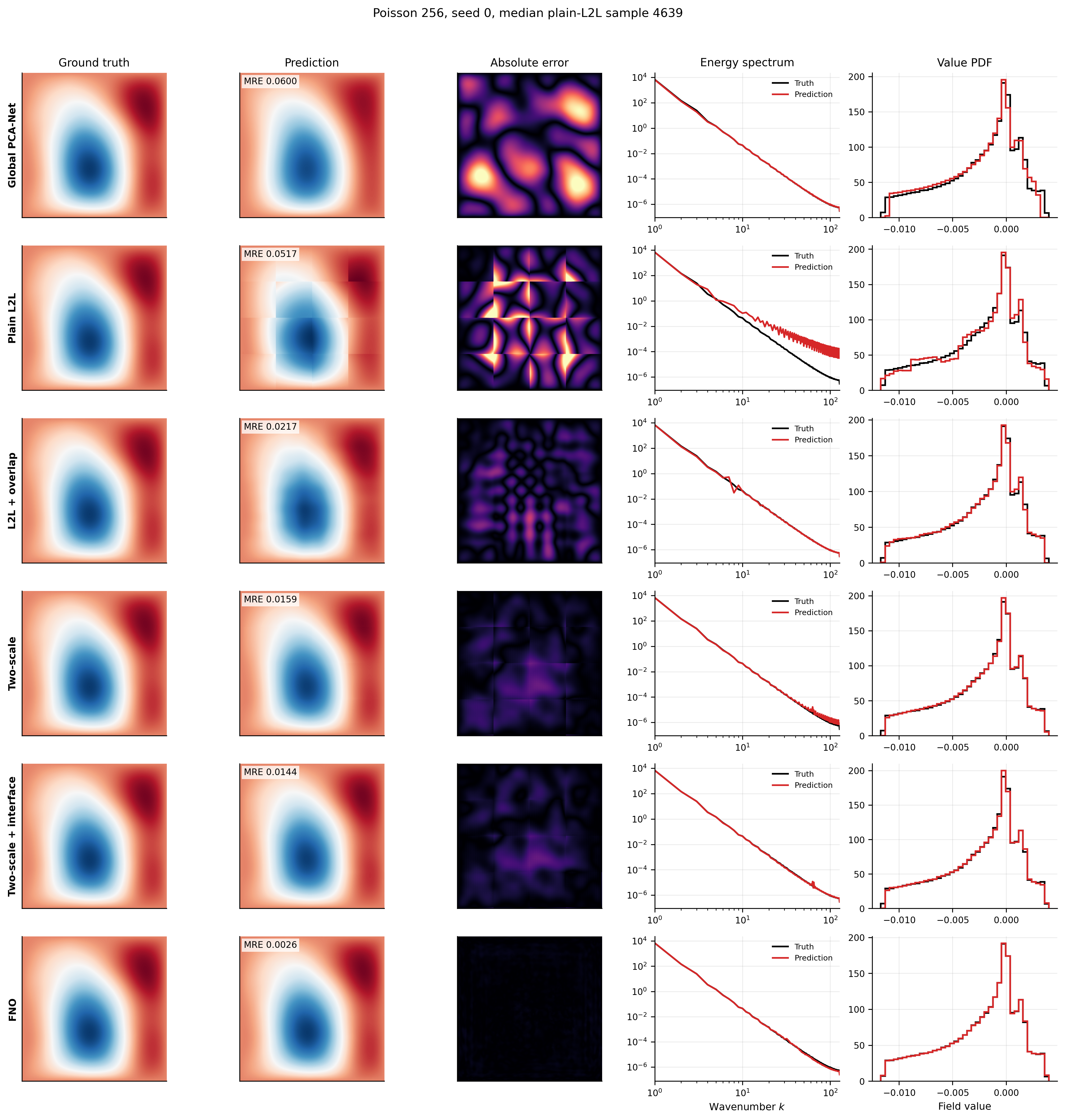}
  \caption{Representative qualitative diagnostics at resolution $256^2$.
  Plain L2L retains block-aligned error, two-scale suppresses the dominant
  tiled bias, A3 further reduces the remaining interface-local discrepancy,
  and FNO remains the highest-accuracy full-field reference.}
  \label{fig:resolution-qualitative-256}
\end{figure*}

\subsection{Supplementary Heterogeneous-Darcy Diagnostics}
\label{app:darcy-generalization}

The tables and figures below retain the complete Darcy metric matrix and
separate native-patch diagnostics from the common-seam comparison.  Overlap
uses stride-32 seams and the nonoverlapping methods use stride-64 seams in
their native evaluation; direct continuity ordering therefore relies on the
common-stride seed-0 audit.

% \begin{table*}[htbp]
%   \centering
%   \caption{Complete heterogeneous Darcy results at $256^2$ over five paired
%   seeds.  Fine-tuning totals include the warm-start two-scale run.}
%   \label{tab:darcy-full}
%   \scriptsize
%   \resizebox{\textwidth}{!}{%
%     \input{figures_and_results/darcy_generalization_full.tex}%
%   }
% \end{table*}

\begin{table*}[htbp]
  \centering
  \caption{Paired percentage effects for the heterogeneous Darcy study.
  Positive error changes denote degradation; positive time changes denote
  additional cost.}
  \label{tab:darcy-paired-effects}
  \small
  \begin{tabular}{lccccc}
\toprule
Comparison
& \makecell{MRE\\change}
& \makecell{Flux-jump\\change}
& \makecell{Spectral\\change}
& \makecell{Darcy-RMS\\change}
& \makecell{Time\\change} \\
\midrule

\makecell[l]{Global PCA-Net\\vs Plain L2L}
& $-6.4\%$ & $-99.2\%$ & $-41.5\%$ & $-83.9\%$ & $+55.4\%$ \\

\makecell[l]{L2L + overlap\\vs Plain L2L}
& $-13.4\%$ & $-99.1\%$ & $+75.6\%$ & $-83.9\%$ & $+111.8\%$ \\

\makecell[l]{Two-scale\\vs Plain L2L}
& $-5.0\%$ & $-88.8\%$ & $+45.7\%$ & $-80.4\%$ & $+14.8\%$ \\

\makecell[l]{Two-scale + interface\\vs Two-scale}
& $-0.9\%$ & $-83.0\%$ & $+14.4\%$ & $-16.6\%$ & $+101.1\%$ \\

\makecell[l]{Two-scale + interface\\vs L2L + overlap}
& $+8.6\%$ & $+116.0\%$ & $+18.1\%$ & $+1.7\%$ & $+9.0\%$ \\

\makecell[l]{FNO\\vs Two-scale + interface}
& $-90.5\%$ & $-29.6\%$ & $-42.5\%$ & $+16.2\%$ & $+307.6\%$ \\

\bottomrule
\end{tabular}
\end{table*}

\begin{table}[htbp]
  \centering
  \caption{Common-stride seed-0 continuity audit.  All predictions are
  evaluated on the same stride-64 interface set.}
  \label{tab:darcy-common-seams}
  \small
  \begin{tabular}{lcccccc}
\toprule
Method
& Seed
& Stride
& \makecell{Value\\jump}
& \makecell{Flux\\jump}
& \makecell{Value trace\\error}
& \makecell{Flux trace\\error} \\
\midrule

Global PCA-Net
& 0 & 64
& $3.870{\times}10^{-5}$
& $2.529{\times}10^{-4}$
& $7.028{\times}10^{-6}$
& $2.974{\times}10^{-4}$ \\

Plain L2L
& 0 & 64
& $1.336{\times}10^{-4}$
& $3.190{\times}10^{-2}$
& $1.211{\times}10^{-4}$
& $3.190{\times}10^{-2}$ \\

L2L + overlap
& 0 & 64
& $3.890{\times}10^{-5}$
& $2.470{\times}10^{-4}$
& $6.287{\times}10^{-6}$
& $3.169{\times}10^{-4}$ \\

FNO
& 0 & 64
& $3.896{\times}10^{-5}$
& $4.299{\times}10^{-4}$
& $1.796{\times}10^{-6}$
& $3.435{\times}10^{-4}$ \\

Two-scale
& 0 & 64
& $3.915{\times}10^{-5}$
& $3.660{\times}10^{-3}$
& $1.596{\times}10^{-5}$
& $3.742{\times}10^{-3}$ \\

Two-scale + interface
& 0 & 64
& $3.817{\times}10^{-5}$
& $6.340{\times}10^{-4}$
& $7.015{\times}10^{-6}$
& $6.909{\times}10^{-4}$ \\

\bottomrule
\end{tabular}
\end{table}

\begin{figure*}[htbp]
  \centering
  \includegraphics[width=0.48\textwidth]{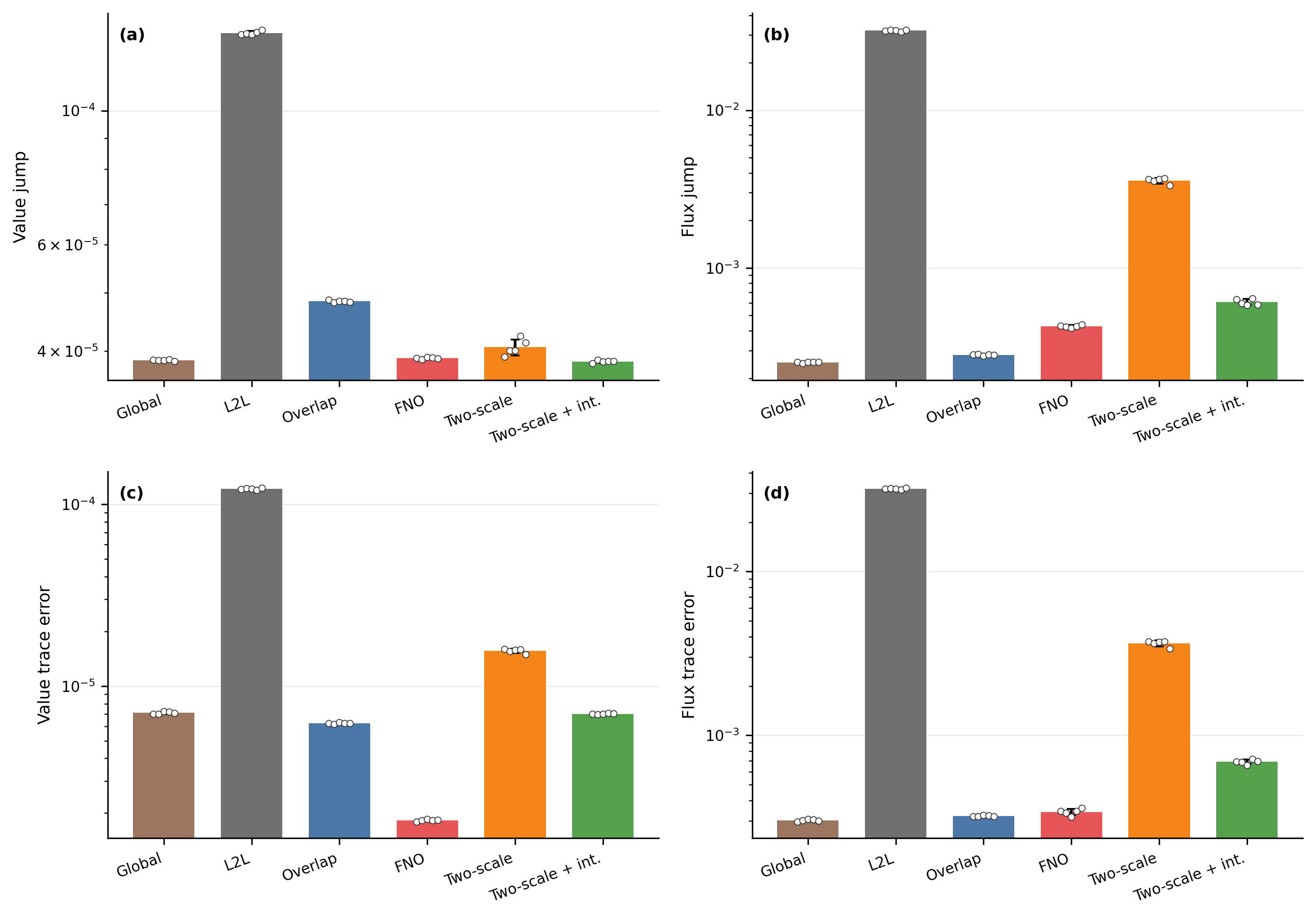}\hfill
  \includegraphics[width=0.48\textwidth]{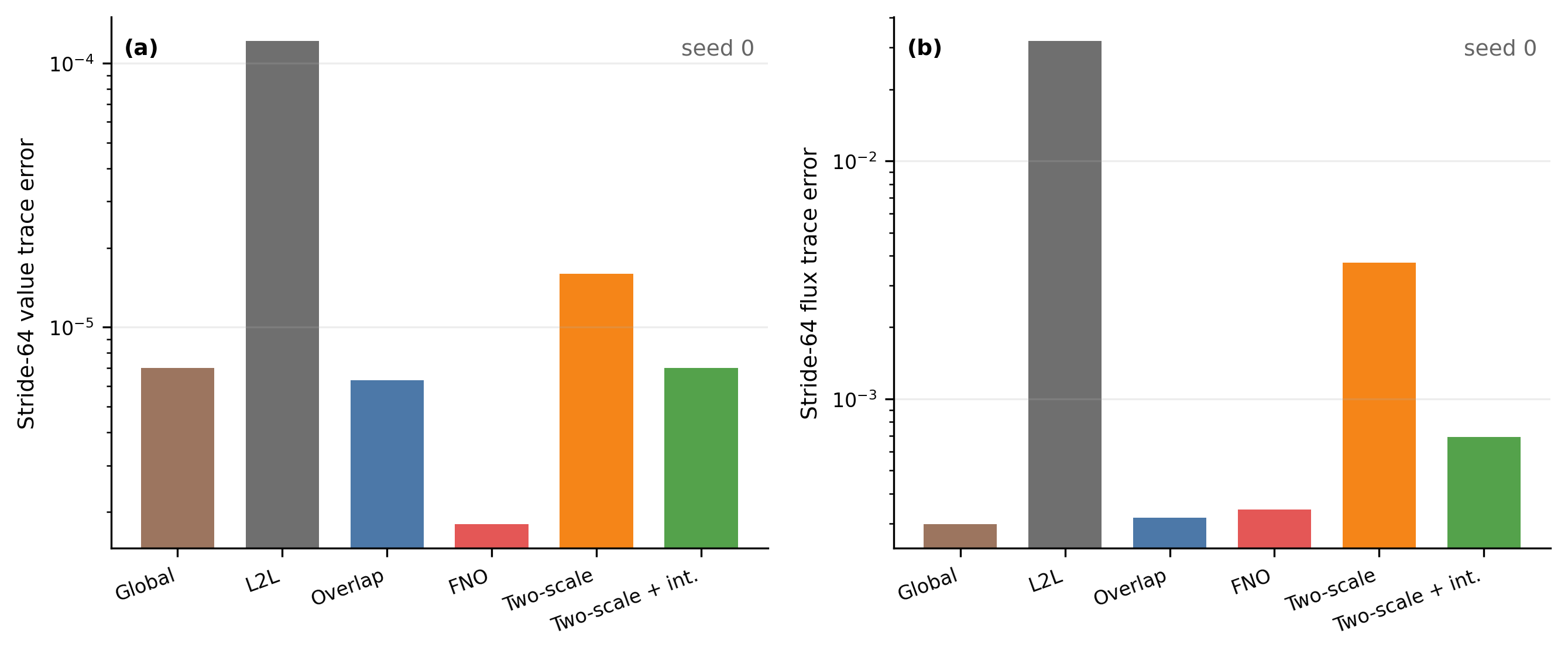}
  \caption{Darcy interface diagnostics.  \textbf{Left:} native interface
  values, which use different stride-induced seam sets for overlap and the
  nonoverlapping methods.  \textbf{Right:} the common-stride seed-0 audit.
  A3 sharply improves the two-scale flux diagnostic, while overlap remains the
  strict PCA-based reference for direct seam continuity}
  \label{fig:darcy-interface-diagnostics}
\end{figure*}

% \begin{figure*}[htbp]
%   \centering
%   \includegraphics[width=0.48\textwidth]{figures_and_results/darcy_quality_cost.png}\hfill
%   \includegraphics[width=0.48\textwidth]{figures_and_results/darcy_pca_breakdown.png}
%   \caption{Darcy cost diagnostics.  \textbf{Left:} MRE against cumulative
%   time and the stage-wise total.  \textbf{Right:} PCA/SVD fit-time breakdown.
%   The small coarse global solve contributes little to two-scale construction
%   time compared with the global and overlapping input-PCA fits.}
%   \label{fig:darcy-cost-diagnostics}
% \end{figure*}

\begin{table}[htbp]
  \centering
  \caption{PCA/SVD fit-time decomposition for the Darcy representation
  baselines.}
  \label{tab:darcy-pca-breakdown}
  \scriptsize
  \begin{tabular}{lcccccc}
\toprule
Method
& \makecell{Input\\PCA (s)}
& \makecell{Output\\PCA (s)}
& \makecell{Coarse global\\SVD (s)}
& \makecell{Residual output\\PCA (s)}
& \makecell{PCA fit\\overhead (s)}
& \makecell{Total PCA\\fit (s)} \\
\midrule

Global PCA-Net
& \makecell{$270.02$\\$\pm\,7.67$}
& \makecell{$8.73$\\$\pm\,0.40$}
& \makecell{$0.00$\\$\pm\,0.00$}
& \makecell{$0.00$\\$\pm\,0.00$}
& \makecell{$0.01$\\$\pm\,0.00$}
& \makecell{$278.76$\\$\pm\,8.06$} \\

Plain L2L
& \makecell{$71.75$\\$\pm\,2.59$}
& \makecell{$5.32$\\$\pm\,0.15$}
& \makecell{$0.00$\\$\pm\,0.00$}
& \makecell{$0.00$\\$\pm\,0.00$}
& \makecell{$1.99$\\$\pm\,0.69$}
& \makecell{$79.06$\\$\pm\,3.25$} \\

L2L + overlap
& \makecell{$225.16$\\$\pm\,6.36$}
& \makecell{$16.60$\\$\pm\,0.44$}
& \makecell{$0.00$\\$\pm\,0.00$}
& \makecell{$0.00$\\$\pm\,0.00$}
& \makecell{$5.62$\\$\pm\,1.78$}
& \makecell{$247.38$\\$\pm\,8.43$} \\

Two-scale
& \makecell{$71.58$\\$\pm\,2.58$}
& \makecell{$0.00$\\$\pm\,0.00$}
& \makecell{$0.77$\\$\pm\,0.06$}
& \makecell{$19.68$\\$\pm\,3.98$}
& \makecell{$1.07$\\$\pm\,0.14$}
& \makecell{$93.10$\\$\pm\,5.41$} \\

\bottomrule
\end{tabular}
\end{table}

\subsection{Supplementary Coarse-Global Representation Design Results}
\label{app:coarse-global-design}

This appendix provides the complete three-seed quality and cost matrices for
the coarse-rank, restriction-factor, and residual-variance sweeps.  The
reference point during the sweep used rank 10, $c=4$, and 99\% residual
variance; the main text selects 99.5\% residual variance after observing its
consistent reconstruction and continuity gains.  All runs use randomized SVD,
8,000 Poisson training examples, and block-balanced coarse/residual latent
losses.

\begin{table*}[htbp]
  \centering
  \caption{Complete coarse-global representation quality matrix at $128^2$.
  Values are means and sample standard deviations over three seeds.}
  \label{tab:coarse-quality-128}
  \scriptsize
  \resizebox{\textwidth}{!}{%
    \begin{tabular}{llrlllllll}
\toprule
Ablation & Setting & Seeds & MRE & Oracle MRE & SSIM & Value jump & Flux jump & Spectral error & Output dim. \\
\midrule
Coarse rank & rank 5 & 3 & 0.0143 \ensuremath{\pm} 0.0002 & 0.0131 \ensuremath{\pm} 0.0003 & 0.9923 \ensuremath{\pm} 0.0002 & 1.484e-04 \ensuremath{\pm} 9.6e-07 & 1.391e-02 \ensuremath{\pm} 1.2e-04 & 0.0029 \ensuremath{\pm} 0.0004 & 93.0 \ensuremath{\pm} 0.0 \\
Coarse rank & rank 10 (default) & 3 & 0.0128 \ensuremath{\pm} 0.0002 & 0.0075 \ensuremath{\pm} 0.0002 & 0.9956 \ensuremath{\pm} 0.0002 & 1.141e-04 \ensuremath{\pm} 2.3e-06 & 7.088e-03 \ensuremath{\pm} 2.3e-04 & 0.0054 \ensuremath{\pm} 0.0005 & 137.3 \ensuremath{\pm} 1.5 \\
Coarse rank & rank 20 & 3 & 0.0379 \ensuremath{\pm} 0.0062 & 0.0034 \ensuremath{\pm} 0.0001 & 0.9751 \ensuremath{\pm} 0.0072 & 9.346e-05 \ensuremath{\pm} 1.6e-06 & 2.468e-03 \ensuremath{\pm} 1.5e-05 & 0.0095 \ensuremath{\pm} 0.0008 & 204.0 \ensuremath{\pm} 0.0 \\
Coarse rank & rank 40 & 3 & 0.0393 \ensuremath{\pm} 0.0014 & 0.0019 \ensuremath{\pm} 0.0000 & 0.9734 \ensuremath{\pm} 0.0007 & 9.112e-05 \ensuremath{\pm} 1.6e-06 & 1.181e-03 \ensuremath{\pm} 1.5e-05 & 0.0142 \ensuremath{\pm} 0.0021 & 300.0 \ensuremath{\pm} 0.0 \\
Coarse factor & c = 2 & 3 & 0.0105 \ensuremath{\pm} 0.0005 & 0.0068 \ensuremath{\pm} 0.0002 & 0.9964 \ensuremath{\pm} 0.0002 & 1.105e-04 \ensuremath{\pm} 1.2e-06 & 6.819e-03 \ensuremath{\pm} 2.7e-05 & 0.0044 \ensuremath{\pm} 0.0004 & 134.0 \ensuremath{\pm} 0.0 \\
Coarse factor & c = 4 (default) & 3 & 0.0128 \ensuremath{\pm} 0.0002 & 0.0075 \ensuremath{\pm} 0.0002 & 0.9956 \ensuremath{\pm} 0.0002 & 1.141e-04 \ensuremath{\pm} 2.3e-06 & 7.088e-03 \ensuremath{\pm} 2.3e-04 & 0.0054 \ensuremath{\pm} 0.0005 & 137.3 \ensuremath{\pm} 1.5 \\
Coarse factor & c = 8 & 3 & 0.0151 \ensuremath{\pm} 0.0004 & 0.0085 \ensuremath{\pm} 0.0002 & 0.9939 \ensuremath{\pm} 0.0002 & 1.194e-04 \ensuremath{\pm} 1.6e-06 & 7.667e-03 \ensuremath{\pm} 8.3e-05 & 0.0055 \ensuremath{\pm} 0.0005 & 136.0 \ensuremath{\pm} 0.0 \\
Residual variance & 95\% & 3 & 0.0183 \ensuremath{\pm} 0.0006 & 0.0149 \ensuremath{\pm} 0.0004 & 0.9892 \ensuremath{\pm} 0.0003 & 1.641e-04 \ensuremath{\pm} 9.0e-07 & 1.654e-02 \ensuremath{\pm} 9.2e-05 & 0.0049 \ensuremath{\pm} 0.0006 & 86.0 \ensuremath{\pm} 0.0 \\
Residual variance & 99\% (default) & 3 & 0.0128 \ensuremath{\pm} 0.0002 & 0.0075 \ensuremath{\pm} 0.0002 & 0.9956 \ensuremath{\pm} 0.0002 & 1.141e-04 \ensuremath{\pm} 2.3e-06 & 7.088e-03 \ensuremath{\pm} 2.3e-04 & 0.0054 \ensuremath{\pm} 0.0005 & 137.3 \ensuremath{\pm} 1.5 \\
Residual variance & 99.5\% & 3 & 0.0115 \ensuremath{\pm} 0.0003 & 0.0050 \ensuremath{\pm} 0.0001 & 0.9967 \ensuremath{\pm} 0.0002 & 1.009e-04 \ensuremath{\pm} 1.5e-06 & 4.369e-03 \ensuremath{\pm} 5.2e-05 & 0.0057 \ensuremath{\pm} 0.0007 & 162.0 \ensuremath{\pm} 0.0 \\
Adaptive reference & 99\% variance (cap 20) & 3 & 0.0129 \ensuremath{\pm} 0.0004 & 0.0075 \ensuremath{\pm} 0.0002 & 0.9955 \ensuremath{\pm} 0.0003 & 1.141e-04 \ensuremath{\pm} 2.2e-06 & 7.082e-03 \ensuremath{\pm} 1.8e-04 & 0.0054 \ensuremath{\pm} 0.0007 & 137.3 \ensuremath{\pm} 1.5 \\
\bottomrule
\end{tabular}
  }
\end{table*}

\begin{table*}[htbp]
  \centering
  \caption{Complete coarse-global representation quality matrix at $256^2$.
  The $c=2$ configuration is omitted because it violates the required
  condition $(D/c)^2<m$ at this resolution.}
  \label{tab:coarse-quality-256}
  \scriptsize
  \resizebox{\textwidth}{!}{%
    \begin{tabular}{llrlllllll}
\toprule
Ablation & Setting & Seeds & MRE & Oracle MRE & SSIM & Value jump & Flux jump & Spectral error & Output dim. \\
\midrule
Coarse rank & rank 5 & 3 & 0.0140 \ensuremath{\pm} 0.0002 & 0.0126 \ensuremath{\pm} 0.0000 & 0.9927 \ensuremath{\pm} 0.0001 & 1.193e-04 \ensuremath{\pm} 1.7e-07 & 2.714e-02 \ensuremath{\pm} 4.8e-05 & 0.0034 \ensuremath{\pm} 0.0002 & 93.0 \ensuremath{\pm} 0.0 \\
Coarse rank & rank 10 (default) & 3 & 0.0129 \ensuremath{\pm} 0.0007 & 0.0068 \ensuremath{\pm} 0.0000 & 0.9956 \ensuremath{\pm} 0.0002 & 7.322e-05 \ensuremath{\pm} 3.6e-07 & 1.322e-02 \ensuremath{\pm} 1.7e-04 & 0.0064 \ensuremath{\pm} 0.0011 & 134.0 \ensuremath{\pm} 0.0 \\
Coarse rank & rank 20 & 3 & 0.0569 \ensuremath{\pm} 0.0041 & 0.0033 \ensuremath{\pm} 0.0000 & 0.9575 \ensuremath{\pm} 0.0045 & 5.010e-05 \ensuremath{\pm} 1.2e-07 & 4.965e-03 \ensuremath{\pm} 3.9e-05 & 0.0111 \ensuremath{\pm} 0.0004 & 204.0 \ensuremath{\pm} 0.0 \\
Coarse rank & rank 40 & 3 & 0.0757 \ensuremath{\pm} 0.0012 & 0.0014 \ensuremath{\pm} 0.0000 & 0.9390 \ensuremath{\pm} 0.0016 & 4.477e-05 \ensuremath{\pm} 1.6e-08 & 1.577e-03 \ensuremath{\pm} 5.5e-06 & 0.0132 \ensuremath{\pm} 0.0001 & 332.0 \ensuremath{\pm} 0.0 \\
Coarse factor & c = 4 (default) & 3 & 0.0129 \ensuremath{\pm} 0.0007 & 0.0068 \ensuremath{\pm} 0.0000 & 0.9956 \ensuremath{\pm} 0.0002 & 7.322e-05 \ensuremath{\pm} 3.6e-07 & 1.322e-02 \ensuremath{\pm} 1.7e-04 & 0.0064 \ensuremath{\pm} 0.0011 & 134.0 \ensuremath{\pm} 0.0 \\
Coarse factor & c = 8 & 3 & 0.0188 \ensuremath{\pm} 0.0006 & 0.0079 \ensuremath{\pm} 0.0000 & 0.9933 \ensuremath{\pm} 0.0003 & 7.742e-05 \ensuremath{\pm} 3.0e-07 & 1.379e-02 \ensuremath{\pm} 1.2e-04 & 0.0082 \ensuremath{\pm} 0.0004 & 135.0 \ensuremath{\pm} 0.0 \\
Residual variance & 95\% & 3 & 0.0180 \ensuremath{\pm} 0.0007 & 0.0143 \ensuremath{\pm} 0.0000 & 0.9899 \ensuremath{\pm} 0.0004 & 1.381e-04 \ensuremath{\pm} 4.8e-08 & 3.252e-02 \ensuremath{\pm} 1.2e-05 & 0.0055 \ensuremath{\pm} 0.0009 & 86.0 \ensuremath{\pm} 0.0 \\
Residual variance & 99\% (default) & 3 & 0.0129 \ensuremath{\pm} 0.0007 & 0.0068 \ensuremath{\pm} 0.0000 & 0.9956 \ensuremath{\pm} 0.0002 & 7.322e-05 \ensuremath{\pm} 3.6e-07 & 1.322e-02 \ensuremath{\pm} 1.7e-04 & 0.0064 \ensuremath{\pm} 0.0011 & 134.0 \ensuremath{\pm} 0.0 \\
Residual variance & 99.5\% & 3 & 0.0115 \ensuremath{\pm} 0.0004 & 0.0047 \ensuremath{\pm} 0.0000 & 0.9967 \ensuremath{\pm} 0.0001 & 5.835e-05 \ensuremath{\pm} 3.4e-07 & 8.198e-03 \ensuremath{\pm} 1.7e-04 & 0.0054 \ensuremath{\pm} 0.0012 & 158.0 \ensuremath{\pm} 0.0 \\
Adaptive reference & 99\% variance (cap 20) & 3 & 0.0125 \ensuremath{\pm} 0.0002 & 0.0068 \ensuremath{\pm} 0.0000 & 0.9958 \ensuremath{\pm} 0.0002 & 7.305e-05 \ensuremath{\pm} 1.6e-07 & 1.314e-02 \ensuremath{\pm} 7.1e-05 & 0.0056 \ensuremath{\pm} 0.0005 & 134.0 \ensuremath{\pm} 0.0 \\
\bottomrule
\end{tabular}
  }
\end{table*}

\begin{table*}[htbp]
  \centering
  \caption{Stage-wise cost matrix for the coarse-global representation
  design sweep.  Coarse SVD is explicitly separated from local residual PCA
  fit; it remains a small part of cumulative construction time.}
  \label{tab:coarse-costs}
  \scriptsize
  \resizebox{\textwidth}{!}{%
    \begin{tabular}{rllrllllllll}
\toprule
Resolution & Ablation & Setting & Seeds & Coarse modes & Residual dim. & Coarse SVD (s) & PCA fit (s) & Latent transform (s) & NN train (s) & Inference (s) & End-to-end (s) \\
\midrule
128 & Coarse rank & rank 5 & 3 & 5.0 \ensuremath{\pm} 0.0 & 88.0 \ensuremath{\pm} 0.0 & 0.3 \ensuremath{\pm} 0.0 & 4.7 \ensuremath{\pm} 0.1 & 2.3 \ensuremath{\pm} 0.1 & 176.2 \ensuremath{\pm} 2.2 & 0.3 \ensuremath{\pm} 0.0 & 204.0 \ensuremath{\pm} 2.0 \\
128 & Coarse rank & rank 10 (default) & 3 & 10.0 \ensuremath{\pm} 0.0 & 127.3 \ensuremath{\pm} 1.5 & 0.3 \ensuremath{\pm} 0.0 & 4.7 \ensuremath{\pm} 0.1 & 2.3 \ensuremath{\pm} 0.0 & 176.5 \ensuremath{\pm} 1.1 & 0.3 \ensuremath{\pm} 0.0 & 204.6 \ensuremath{\pm} 0.9 \\
128 & Coarse rank & rank 20 & 3 & 20.0 \ensuremath{\pm} 0.0 & 184.0 \ensuremath{\pm} 0.0 & 0.3 \ensuremath{\pm} 0.1 & 4.7 \ensuremath{\pm} 0.1 & 2.3 \ensuremath{\pm} 0.1 & 178.6 \ensuremath{\pm} 2.0 & 0.3 \ensuremath{\pm} 0.0 & 206.2 \ensuremath{\pm} 1.9 \\
128 & Coarse rank & rank 40 & 3 & 40.0 \ensuremath{\pm} 0.0 & 260.0 \ensuremath{\pm} 0.0 & 0.3 \ensuremath{\pm} 0.0 & 5.1 \ensuremath{\pm} 0.1 & 2.3 \ensuremath{\pm} 0.1 & 189.2 \ensuremath{\pm} 2.9 & 0.3 \ensuremath{\pm} 0.0 & 217.2 \ensuremath{\pm} 2.8 \\
128 & Coarse factor & c = 2 & 3 & 10.0 \ensuremath{\pm} 0.0 & 124.0 \ensuremath{\pm} 0.0 & 0.6 \ensuremath{\pm} 0.0 & 5.1 \ensuremath{\pm} 0.0 & 2.6 \ensuremath{\pm} 0.1 & 174.4 \ensuremath{\pm} 2.0 & 0.3 \ensuremath{\pm} 0.0 & 203.0 \ensuremath{\pm} 2.1 \\
128 & Coarse factor & c = 4 (default) & 3 & 10.0 \ensuremath{\pm} 0.0 & 127.3 \ensuremath{\pm} 1.5 & 0.3 \ensuremath{\pm} 0.0 & 4.7 \ensuremath{\pm} 0.1 & 2.3 \ensuremath{\pm} 0.0 & 176.5 \ensuremath{\pm} 1.1 & 0.3 \ensuremath{\pm} 0.0 & 204.6 \ensuremath{\pm} 0.9 \\
128 & Coarse factor & c = 8 & 3 & 10.0 \ensuremath{\pm} 0.0 & 126.0 \ensuremath{\pm} 0.0 & 0.2 \ensuremath{\pm} 0.0 & 4.6 \ensuremath{\pm} 0.1 & 2.3 \ensuremath{\pm} 0.0 & 178.4 \ensuremath{\pm} 4.5 & 0.3 \ensuremath{\pm} 0.0 & 206.2 \ensuremath{\pm} 4.3 \\
128 & Residual variance & 95\% & 3 & 10.0 \ensuremath{\pm} 0.0 & 76.0 \ensuremath{\pm} 0.0 & 0.3 \ensuremath{\pm} 0.0 & 4.7 \ensuremath{\pm} 0.0 & 2.3 \ensuremath{\pm} 0.1 & 175.8 \ensuremath{\pm} 3.5 & 0.3 \ensuremath{\pm} 0.0 & 203.4 \ensuremath{\pm} 3.1 \\
128 & Residual variance & 99\% (default) & 3 & 10.0 \ensuremath{\pm} 0.0 & 127.3 \ensuremath{\pm} 1.5 & 0.3 \ensuremath{\pm} 0.0 & 4.7 \ensuremath{\pm} 0.1 & 2.3 \ensuremath{\pm} 0.0 & 176.5 \ensuremath{\pm} 1.1 & 0.3 \ensuremath{\pm} 0.0 & 204.6 \ensuremath{\pm} 0.9 \\
128 & Residual variance & 99.5\% & 3 & 10.0 \ensuremath{\pm} 0.0 & 152.0 \ensuremath{\pm} 0.0 & 0.3 \ensuremath{\pm} 0.0 & 4.7 \ensuremath{\pm} 0.0 & 2.3 \ensuremath{\pm} 0.0 & 174.2 \ensuremath{\pm} 2.7 & 0.3 \ensuremath{\pm} 0.0 & 203.8 \ensuremath{\pm} 5.6 \\
128 & Adaptive reference & 99\% variance (cap 20) & 3 & 10.0 \ensuremath{\pm} 0.0 & 127.3 \ensuremath{\pm} 1.5 & 0.3 \ensuremath{\pm} 0.0 & 4.7 \ensuremath{\pm} 0.0 & 2.3 \ensuremath{\pm} 0.0 & 176.0 \ensuremath{\pm} 2.9 & 0.3 \ensuremath{\pm} 0.0 & 202.8 \ensuremath{\pm} 3.0 \\
256 & Coarse rank & rank 5 & 3 & 5.0 \ensuremath{\pm} 0.0 & 88.0 \ensuremath{\pm} 0.0 & 0.8 \ensuremath{\pm} 0.1 & 18.8 \ensuremath{\pm} 0.8 & 7.3 \ensuremath{\pm} 0.5 & 179.4 \ensuremath{\pm} 5.2 & 1.0 \ensuremath{\pm} 0.0 & 260.0 \ensuremath{\pm} 6.5 \\
256 & Coarse rank & rank 10 (default) & 3 & 10.0 \ensuremath{\pm} 0.0 & 124.0 \ensuremath{\pm} 0.0 & 1.0 \ensuremath{\pm} 0.3 & 18.7 \ensuremath{\pm} 1.0 & 7.5 \ensuremath{\pm} 0.3 & 178.7 \ensuremath{\pm} 2.1 & 0.9 \ensuremath{\pm} 0.0 & 258.2 \ensuremath{\pm} 1.3 \\
256 & Coarse rank & rank 20 & 3 & 20.0 \ensuremath{\pm} 0.0 & 184.0 \ensuremath{\pm} 0.0 & 0.7 \ensuremath{\pm} 0.0 & 18.4 \ensuremath{\pm} 0.6 & 8.1 \ensuremath{\pm} 0.2 & 178.7 \ensuremath{\pm} 2.9 & 1.0 \ensuremath{\pm} 0.0 & 259.4 \ensuremath{\pm} 2.6 \\
256 & Coarse rank & rank 40 & 3 & 40.0 \ensuremath{\pm} 0.0 & 292.0 \ensuremath{\pm} 0.0 & 0.8 \ensuremath{\pm} 0.0 & 19.6 \ensuremath{\pm} 0.3 & 7.8 \ensuremath{\pm} 0.7 & 188.9 \ensuremath{\pm} 2.0 & 0.9 \ensuremath{\pm} 0.0 & 268.8 \ensuremath{\pm} 3.6 \\
256 & Coarse factor & c = 4 (default) & 3 & 10.0 \ensuremath{\pm} 0.0 & 124.0 \ensuremath{\pm} 0.0 & 1.0 \ensuremath{\pm} 0.3 & 18.7 \ensuremath{\pm} 1.0 & 7.5 \ensuremath{\pm} 0.3 & 178.7 \ensuremath{\pm} 2.1 & 0.9 \ensuremath{\pm} 0.0 & 258.2 \ensuremath{\pm} 1.3 \\
256 & Coarse factor & c = 8 & 3 & 10.0 \ensuremath{\pm} 0.0 & 125.0 \ensuremath{\pm} 0.0 & 0.5 \ensuremath{\pm} 0.0 & 17.5 \ensuremath{\pm} 0.2 & 7.5 \ensuremath{\pm} 0.4 & 175.7 \ensuremath{\pm} 2.7 & 0.9 \ensuremath{\pm} 0.0 & 253.6 \ensuremath{\pm} 2.1 \\
256 & Residual variance & 95\% & 3 & 10.0 \ensuremath{\pm} 0.0 & 76.0 \ensuremath{\pm} 0.0 & 0.7 \ensuremath{\pm} 0.0 & 17.7 \ensuremath{\pm} 0.1 & 7.8 \ensuremath{\pm} 0.6 & 175.3 \ensuremath{\pm} 4.2 & 1.0 \ensuremath{\pm} 0.0 & 253.0 \ensuremath{\pm} 5.3 \\
256 & Residual variance & 99\% (default) & 3 & 10.0 \ensuremath{\pm} 0.0 & 124.0 \ensuremath{\pm} 0.0 & 1.0 \ensuremath{\pm} 0.3 & 18.7 \ensuremath{\pm} 1.0 & 7.5 \ensuremath{\pm} 0.3 & 178.7 \ensuremath{\pm} 2.1 & 0.9 \ensuremath{\pm} 0.0 & 258.2 \ensuremath{\pm} 1.3 \\
256 & Residual variance & 99.5\% & 3 & 10.0 \ensuremath{\pm} 0.0 & 148.0 \ensuremath{\pm} 0.0 & 0.8 \ensuremath{\pm} 0.1 & 18.3 \ensuremath{\pm} 0.4 & 8.3 \ensuremath{\pm} 0.5 & 176.4 \ensuremath{\pm} 4.1 & 0.9 \ensuremath{\pm} 0.0 & 255.7 \ensuremath{\pm} 3.3 \\
256 & Adaptive reference & 99\% variance (cap 20) & 3 & 10.0 \ensuremath{\pm} 0.0 & 124.0 \ensuremath{\pm} 0.0 & 0.8 \ensuremath{\pm} 0.0 & 18.1 \ensuremath{\pm} 0.3 & 7.6 \ensuremath{\pm} 0.2 & 176.8 \ensuremath{\pm} 0.7 & 1.0 \ensuremath{\pm} 0.0 & 255.0 \ensuremath{\pm} 0.9 \\
\bottomrule
\end{tabular}
  }
\end{table*}

\begin{figure*}[htbp]
  \centering
  \includegraphics[width=0.96\textwidth]{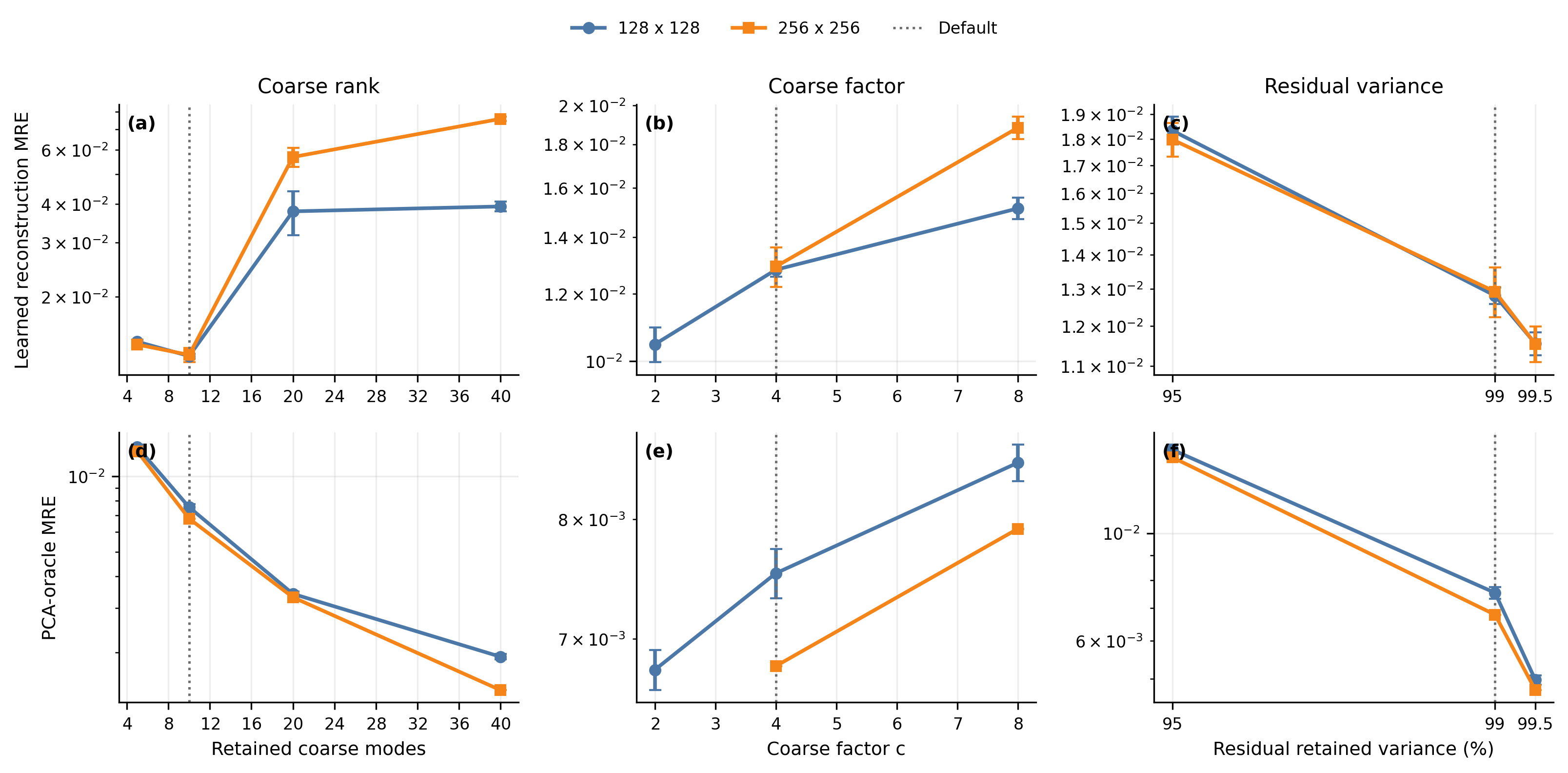}
  \caption{Learned and PCA-oracle reconstruction trends for each
  coarse-global design sweep.  Coarse rank reveals the strongest oracle versus
  learned divergence; residual retained variance improves both quantities}
  \label{fig:coarse-reconstruction-sweeps}
\end{figure*}

\begin{figure*}[htbp]
  \centering
  \includegraphics[width=0.96\textwidth]{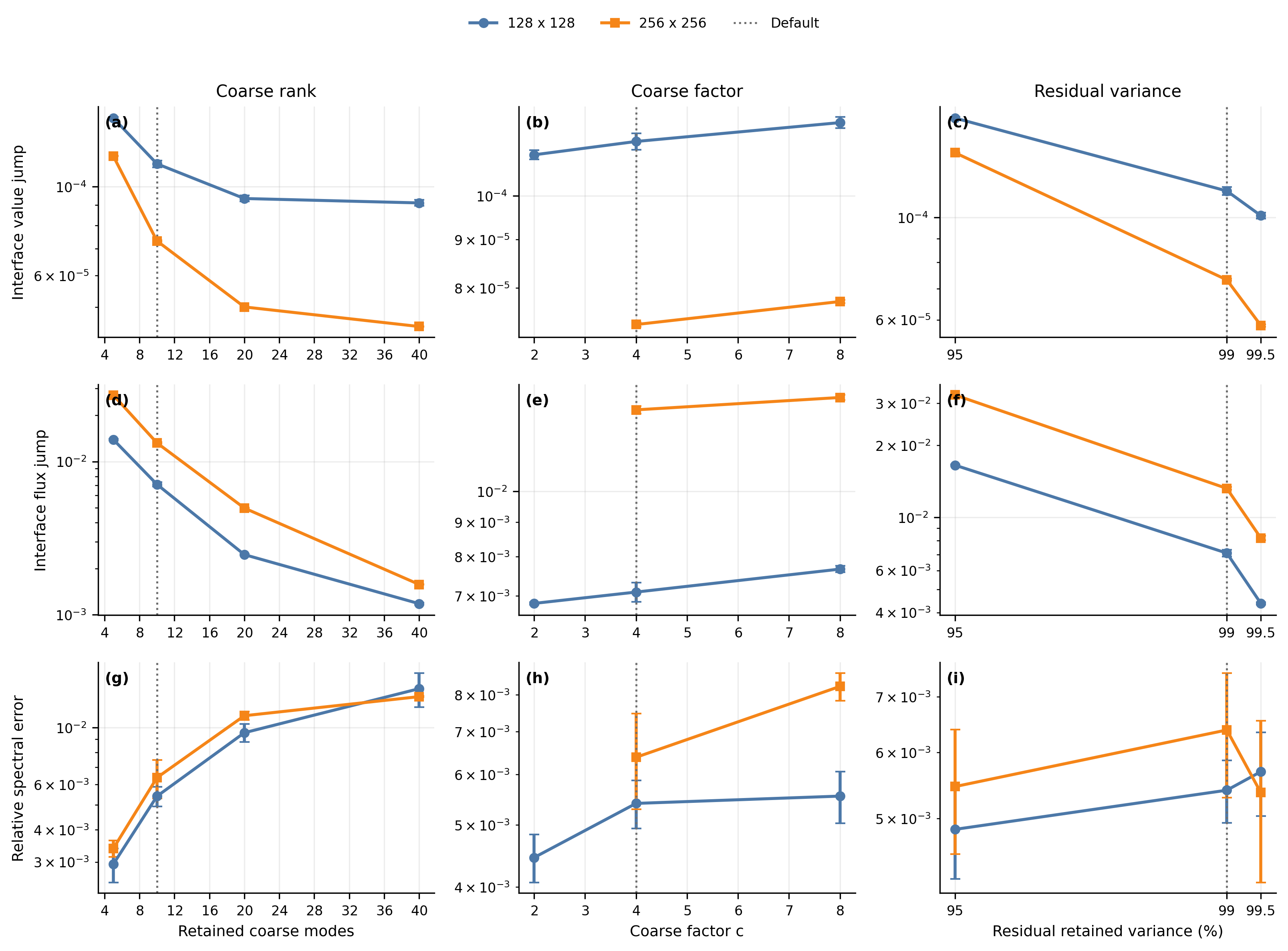}
  \caption{Interface value jump, interface flux jump, and spectral error
  across coarse-global design choices.  Higher coarse rank can reduce seam
  jumps even when learned MRE worsens, so continuity alone is not sufficient
  for model selection}
  \label{fig:coarse-artifact-sweeps}
\end{figure*}

\begin{figure*}[htbp]
  \centering
  \includegraphics[width=0.96\textwidth]{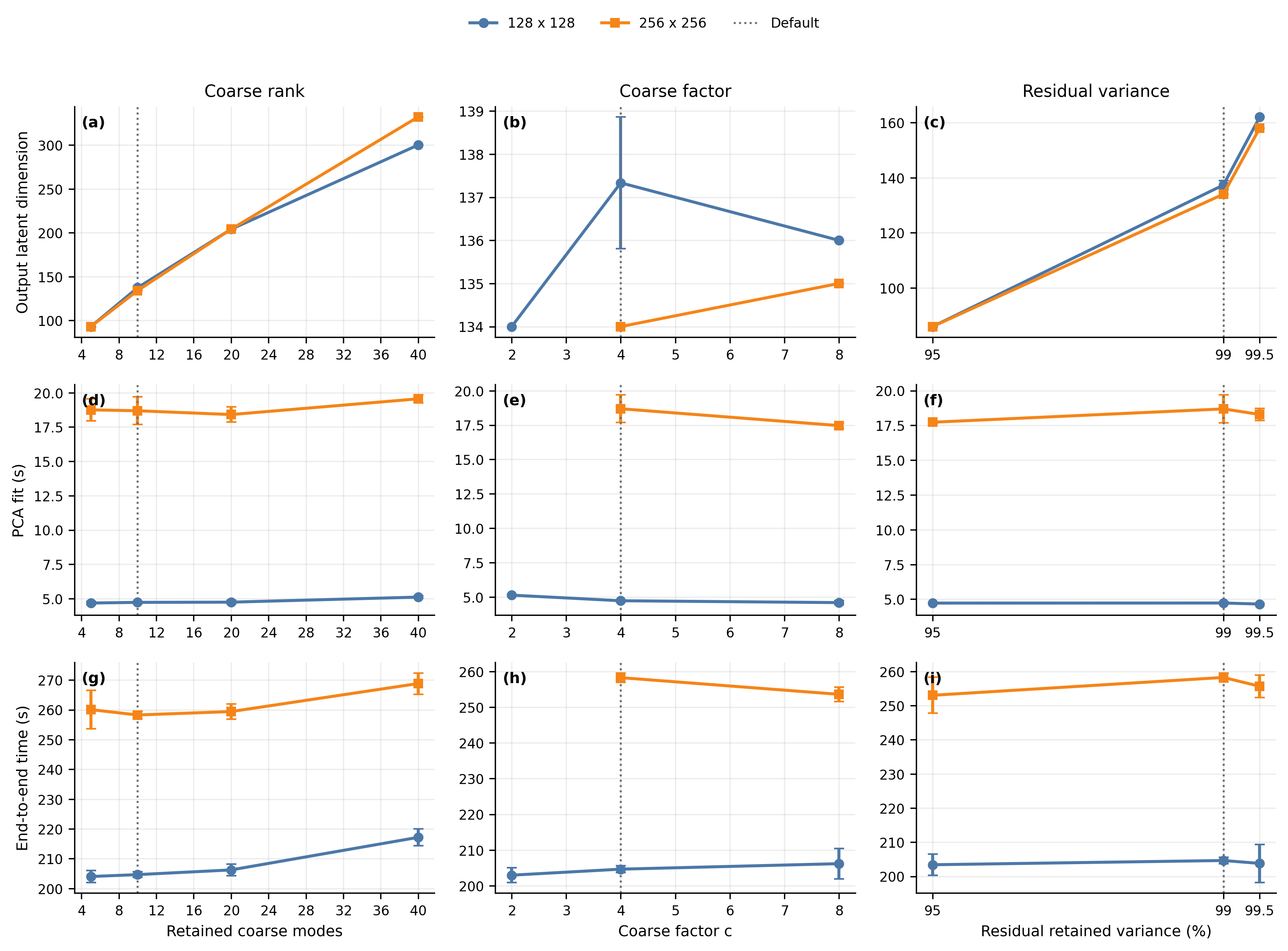}
  \caption{Output-code capacity and stage-wise cost for the
  coarse-global design sweep.  Increasing coarse rank substantially enlarges
  the learned output target, whereas the selected 99.5\% residual setting
  delivers improved field quality at essentially unchanged cumulative cost}
  \label{fig:coarse-capacity-cost}
\end{figure*}

\subsection{Supplementary Latent-Objective Results}
\label{app:latent-objective}

The latent-objective experiment holds the two-scale encoder, network
architecture, random split, and seed fixed within each dataset-resolution
cell.  It uses a single matched seed and is therefore a design-selection study
rather than a statistical comparison between block balancing and score
normalization.  The shared representation dimensions and parameter counts
confirm that the observed differences arise from the training objective.

% \begin{table*}[htbp]
%   \centering
%   \caption{Complete matched seed-0 latent-objective results.  The two-scale
%   representation is fixed across objectives; total time includes all pipeline
%   stages.}
%   \label{tab:latent-objective-quality}
%   \small
%   \input{figures_and_results/latent_loss_quality.tex}
% \end{table*}

% \begin{table*}[htbp]
%   \centering
%   \caption{Relative effects of each structured objective against vector MSE.
%   Negative error changes denote improvement; positive training-time changes
%   denote additional cost.}
%   \label{tab:latent-objective-effects}
%   \small
%   \input{figures_and_results/latent_loss_relative_effects.tex}
% \end{table*}

\begin{table*}[htbp]
  \centering
  \caption{Latent dimensions and parameter counts for the objective study.
  They are invariant to objective choice within each dataset-resolution cell.}
  \label{tab:latent-objective-capacity}
  \small
  \begin{tabular}{lclccccc}
\toprule
Dataset
& Grid
& Objective
& \makecell{Input\\latent dim.}
& \makecell{Coarse\\dim.}
& \makecell{Residual\\dim.}
& \makecell{Output\\latent dim.}
& Parameters \\
\midrule

Darcy & 128 & Block-balanced
& 6587 & 20 & 670 & 690 & 965298 \\

Darcy & 128 & Score-normalized
& 6587 & 20 & 670 & 690 & 965298 \\

Darcy & 128 & Vector MSE
& 6587 & 20 & 670 & 690 & 965298 \\

Darcy & 256 & Block-balanced
& 10558 & 20 & 735 & 755 & 1481971 \\

Darcy & 256 & Score-normalized
& 10558 & 20 & 735 & 755 & 1481971 \\

Darcy & 256 & Vector MSE
& 10558 & 20 & 735 & 755 & 1481971 \\

Poisson & 128 & Block-balanced
& 144 & 10 & 126 & 136 & 69128 \\

Poisson & 128 & Score-normalized
& 144 & 10 & 126 & 136 & 69128 \\

Poisson & 128 & Vector MSE
& 144 & 10 & 126 & 136 & 69128 \\

Poisson & 256 & Block-balanced
& 144 & 10 & 124 & 134 & 68870 \\

Poisson & 256 & Score-normalized
& 144 & 10 & 124 & 134 & 68870 \\

Poisson & 256 & Vector MSE
& 144 & 10 & 124 & 134 & 68870 \\

\bottomrule
\end{tabular}
\end{table*}

\begin{figure*}[htbp]
  \centering
  \includegraphics[width=0.96\textwidth]{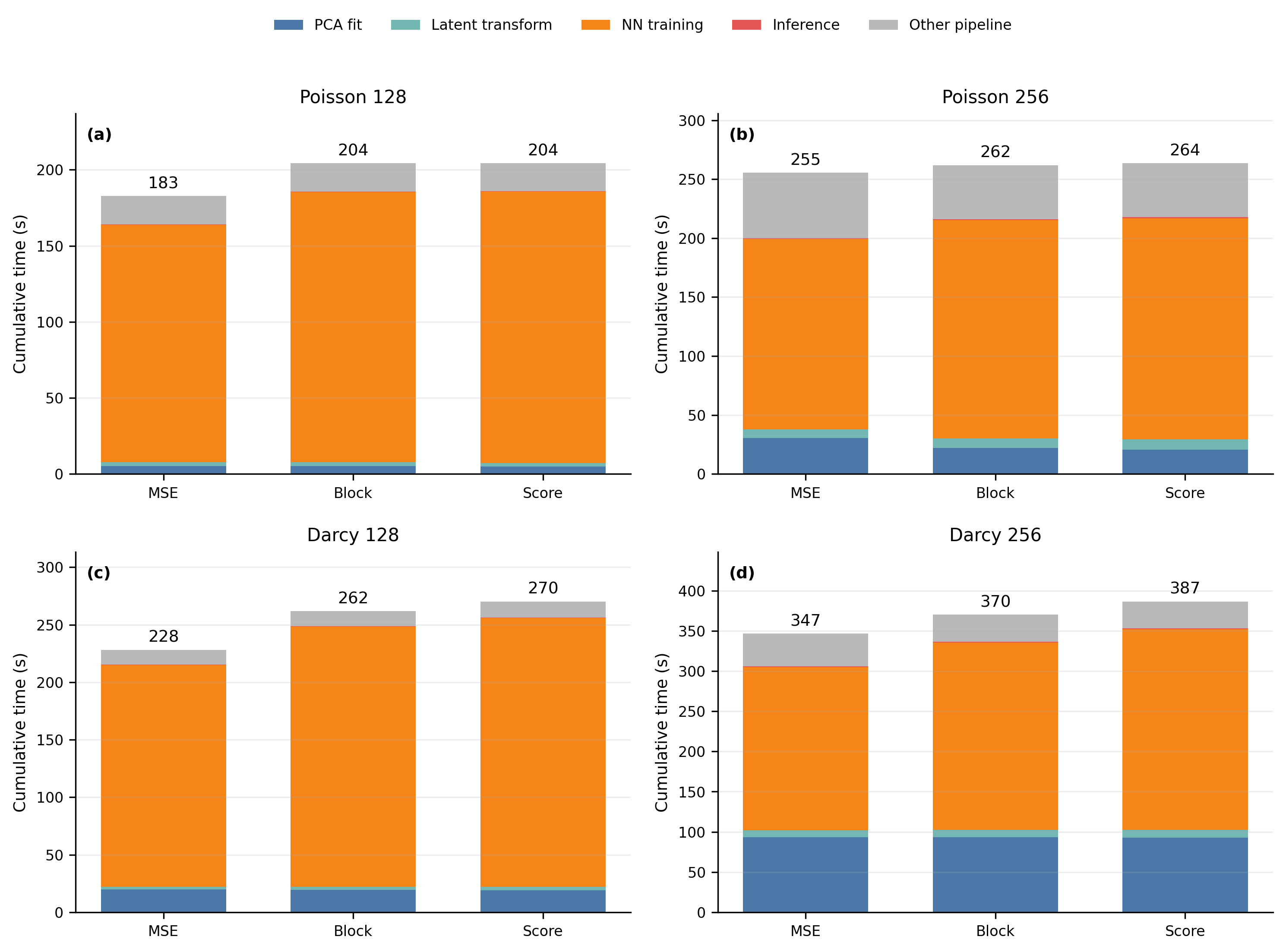}
  \caption{Stage-wise timing for the latent-objective study.  PCA fit and
  latent transformation are objective-independent up to runtime variation;
  the structured objectives can extend neural-network training through their
  validation and early-stopping trajectories}
  \label{fig:latent-objective-stage-costs}
\end{figure*}

\begin{figure*}[htbp]
  \centering
  \includegraphics[width=0.96\textwidth]{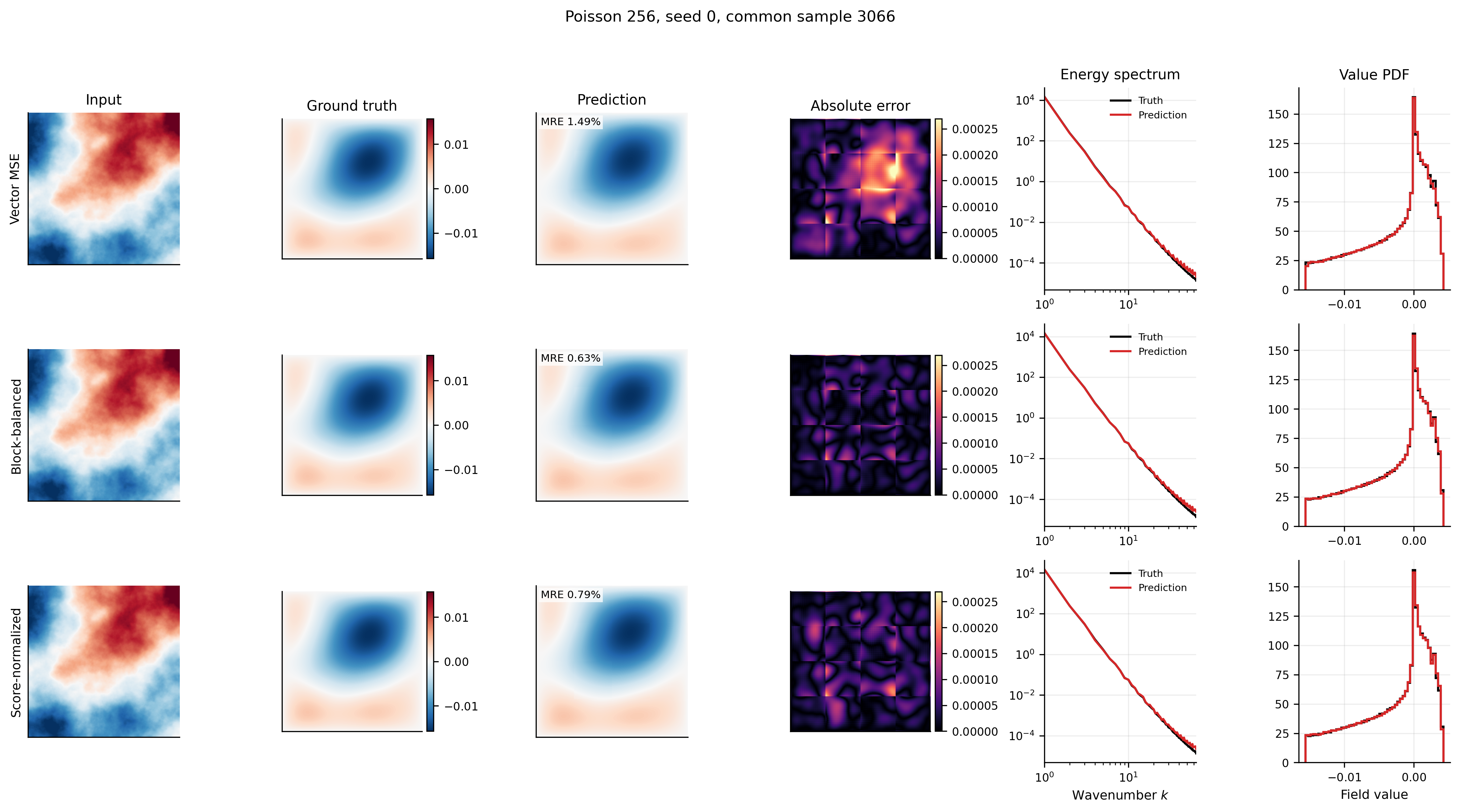}
  \caption{Poisson $256^2$ qualitative comparison for the latent-objective
  study.  The representation is held fixed, isolating the effect of the
  training target on the reconstructed field, spectrum, and value
  distribution}
  \label{fig:latent-objective-poisson-qualitative}
\end{figure*}

\begin{figure*}[htbp]
  \centering
  \includegraphics[width=0.96\textwidth]{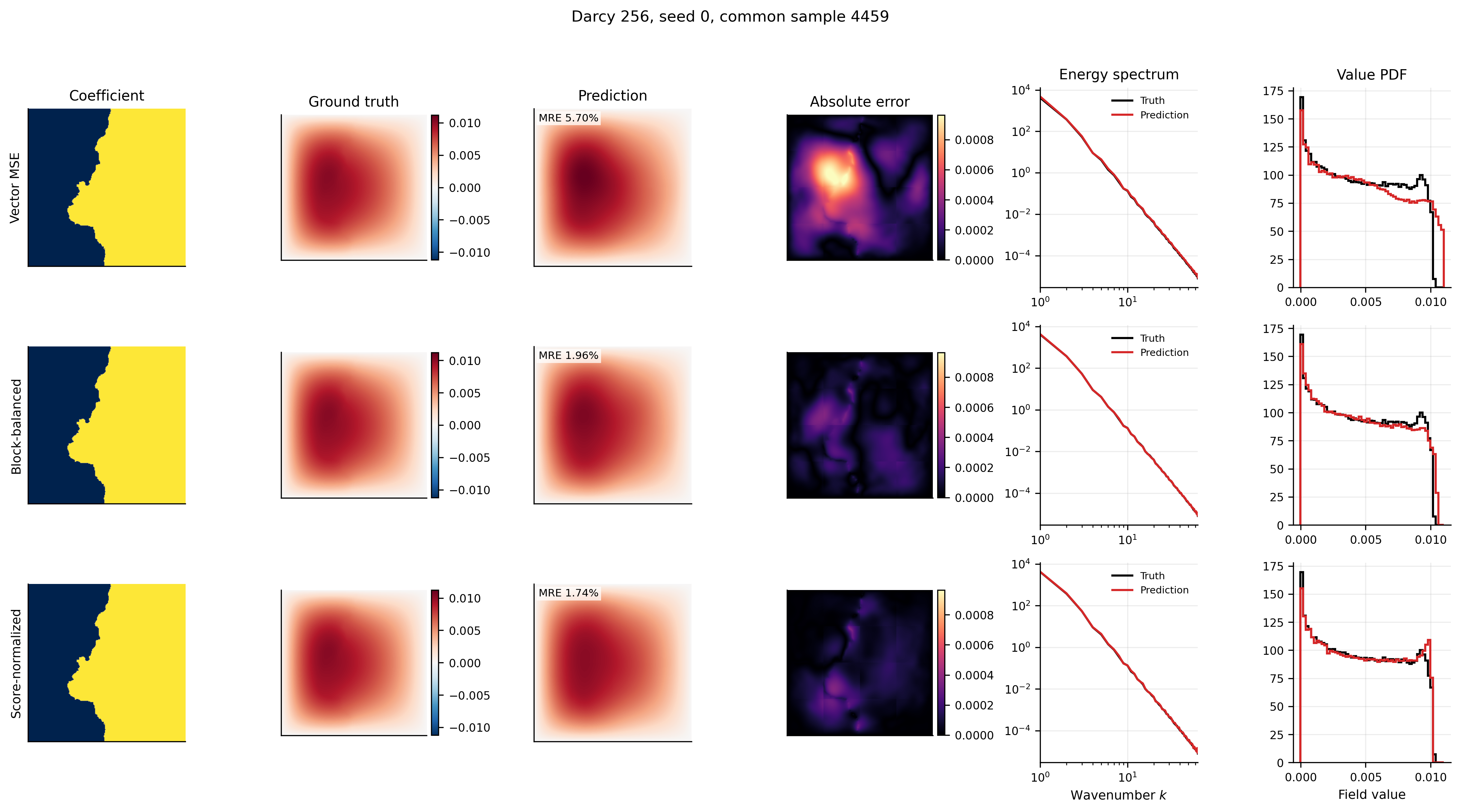}
  \caption{Darcy $256^2$ qualitative comparison for the latent-objective
  study.  Score normalization improves aggregate field error, whereas the
  diagnostic metrics in the main text motivate block balancing as the
  cross-problem default}
  \label{fig:latent-objective-darcy-qualitative}
\end{figure*}

\subsection{Supplementary Interface-Loss Ladder Results}
\label{app:interface-loss-ladder}

All results below use the corrected Poisson residual convention
$\Delta_h\widehat{u}-f$.  A0--A4 have no active PDE-residual training term and
retain their original checkpoints; all A5 runs were recalibrated and retrained
under the corrected convention.  Each cell contains three paired seeds.
A1--A5 warm-start from the corresponding A0 two-scale run, so their cumulative
times include representation construction, latent training, and fine-tuning.

\begin{table*}[htbp]
  \centering
  \caption{Interface-loss ladder at $128^2$.  Fine-tune time is the additional
  assembled-field optimization time after the A0 warm start.}
  \label{tab:physics-ladder-128}
  \scriptsize
  \resizebox{\textwidth}{!}{%
    \begin{tabular}{lrllllllll}
\toprule
Variant & Seeds & MRE & SSIM & Value jump & Flux jump & Spectral error & PDE residual RMS & Fine-tune time (s) & Total time (s) \\
\midrule
A0: latent baseline & 3 & 0.01155 \ensuremath{\pm} 0.00028 & 0.9967 \ensuremath{\pm} 0.0002 & 1.009e-04 \ensuremath{\pm} 1.5e-06 & 4.369e-03 \ensuremath{\pm} 5.2e-05 & 0.00569 \ensuremath{\pm} 0.00065 & 0.24621 \ensuremath{\pm} 0.00219 &  & 203.8 \ensuremath{\pm} 5.6 \\
A1: reconstruction & 3 & 0.01069 \ensuremath{\pm} 0.00020 & 0.9969 \ensuremath{\pm} 0.0001 & 9.971e-05 \ensuremath{\pm} 1.5e-06 & 4.262e-03 \ensuremath{\pm} 4.9e-05 & 0.00414 \ensuremath{\pm} 0.00096 & 0.23967 \ensuremath{\pm} 0.00224 & 103.3 \ensuremath{\pm} 2.6 & 307.1 \ensuremath{\pm} 5.1 \\
A2: + interface value & 3 & 0.01075 \ensuremath{\pm} 0.00031 & 0.9969 \ensuremath{\pm} 0.0001 & 9.357e-05 \ensuremath{\pm} 1.6e-06 & 2.305e-03 \ensuremath{\pm} 3.2e-05 & 0.00421 \ensuremath{\pm} 0.00044 & 0.16239 \ensuremath{\pm} 0.00187 & 127.9 \ensuremath{\pm} 3.1 & 331.7 \ensuremath{\pm} 6.3 \\
A3: + interface flux & 3 & 0.01087 \ensuremath{\pm} 0.00022 & 0.9968 \ensuremath{\pm} 0.0001 & 9.248e-05 \ensuremath{\pm} 1.7e-06 & 1.709e-03 \ensuremath{\pm} 5.9e-05 & 0.00432 \ensuremath{\pm} 0.00111 & 0.14663 \ensuremath{\pm} 0.00243 & 166.0 \ensuremath{\pm} 5.3 & 369.8 \ensuremath{\pm} 10.9 \\
A4: + spectral & 3 & 0.01257 \ensuremath{\pm} 0.00034 & 0.9959 \ensuremath{\pm} 0.0002 & 9.274e-05 \ensuremath{\pm} 1.7e-06 & 1.858e-03 \ensuremath{\pm} 7.8e-05 & 0.00150 \ensuremath{\pm} 0.00066 & 0.15208 \ensuremath{\pm} 0.00322 & 190.6 \ensuremath{\pm} 3.2 & 394.4 \ensuremath{\pm} 6.0 \\
A5: + PDE residual & 3 & 0.01199 \ensuremath{\pm} 0.00062 & 0.9963 \ensuremath{\pm} 0.0003 & 9.408e-05 \ensuremath{\pm} 1.8e-06 & 2.385e-03 \ensuremath{\pm} 1.1e-03 & 0.00250 \ensuremath{\pm} 0.00054 & 0.17048 \ensuremath{\pm} 0.03702 & 198.1 \ensuremath{\pm} 2.6 & 401.9 \ensuremath{\pm} 6.3 \\
\bottomrule
\end{tabular}
  }
\end{table*}

\begin{table*}[htbp]
  \centering
  \caption{Interface-loss ladder at $256^2$.}
  \label{tab:physics-ladder-256}
  \scriptsize
  \resizebox{\textwidth}{!}{%
    \begin{tabular}{lrllllllll}
\toprule
Variant & Seeds & MRE & SSIM & Value jump & Flux jump & Spectral error & PDE residual RMS & Fine-tune time (s) & Total time (s) \\
\midrule
A0: latent baseline & 3 & 0.01154 \ensuremath{\pm} 0.00045 & 0.9967 \ensuremath{\pm} 0.0001 & 5.835e-05 \ensuremath{\pm} 3.4e-07 & 8.198e-03 \ensuremath{\pm} 1.7e-04 & 0.00537 \ensuremath{\pm} 0.00118 & 0.61120 \ensuremath{\pm} 0.01083 &  & 255.7 \ensuremath{\pm} 3.3 \\
A1: reconstruction & 3 & 0.01067 \ensuremath{\pm} 0.00047 & 0.9970 \ensuremath{\pm} 0.0002 & 5.676e-05 \ensuremath{\pm} 3.4e-07 & 7.806e-03 \ensuremath{\pm} 1.4e-04 & 0.00383 \ensuremath{\pm} 0.00042 & 0.57961 \ensuremath{\pm} 0.00869 & 251.7 \ensuremath{\pm} 3.1 & 507.3 \ensuremath{\pm} 2.9 \\
A2: + interface value & 3 & 0.01092 \ensuremath{\pm} 0.00037 & 0.9969 \ensuremath{\pm} 0.0002 & 4.816e-05 \ensuremath{\pm} 1.8e-07 & 3.620e-03 \ensuremath{\pm} 9.8e-05 & 0.00432 \ensuremath{\pm} 0.00090 & 0.31495 \ensuremath{\pm} 0.00554 & 289.8 \ensuremath{\pm} 1.4 & 545.5 \ensuremath{\pm} 4.6 \\
A3: + interface flux & 3 & 0.01098 \ensuremath{\pm} 0.00040 & 0.9968 \ensuremath{\pm} 0.0001 & 4.661e-05 \ensuremath{\pm} 5.8e-08 & 2.382e-03 \ensuremath{\pm} 5.2e-05 & 0.00351 \ensuremath{\pm} 0.00145 & 0.25313 \ensuremath{\pm} 0.00202 & 373.0 \ensuremath{\pm} 2.8 & 628.7 \ensuremath{\pm} 4.4 \\
A4: + spectral & 3 & 0.01236 \ensuremath{\pm} 0.00013 & 0.9961 \ensuremath{\pm} 0.0001 & 4.658e-05 \ensuremath{\pm} 2.1e-07 & 2.375e-03 \ensuremath{\pm} 1.4e-04 & 0.00314 \ensuremath{\pm} 0.00067 & 0.25631 \ensuremath{\pm} 0.00533 & 425.1 \ensuremath{\pm} 6.4 & 680.7 \ensuremath{\pm} 8.2 \\
A5: + PDE residual & 3 & 0.01240 \ensuremath{\pm} 0.00009 & 0.9962 \ensuremath{\pm} 0.0001 & 4.665e-05 \ensuremath{\pm} 3.1e-07 & 2.420e-03 \ensuremath{\pm} 3.4e-04 & 0.00392 \ensuremath{\pm} 0.00112 & 0.25583 \ensuremath{\pm} 0.01665 & 436.8 \ensuremath{\pm} 8.7 & 692.5 \ensuremath{\pm} 11.5 \\
\bottomrule
\end{tabular}
  }
\end{table*}

\begin{table*}[htbp]
  \centering
  \caption{Relative change from A0 and from reconstruction-only A1.  Negative
  metric changes denote improvement; positive time changes denote additional
  cost.}
  \label{tab:physics-ladder-relative-effects}
  \scriptsize
  \resizebox{\textwidth}{!}{%
    \begin{tabular}{rlllllllllllll}
\toprule
Resolution & Variant & mre\_vs\_A0\_pct & interface\_jump\_vs\_A0\_pct & interface\_flux\_jump\_vs\_A0\_pct & relative\_spectrum\_error\_vs\_A0\_pct & poisson\_residual\_rms\_vs\_A0\_pct & end\_to\_end\_vs\_A0\_pct & mre\_vs\_A1\_pct & interface\_jump\_vs\_A1\_pct & interface\_flux\_jump\_vs\_A1\_pct & relative\_spectrum\_error\_vs\_A1\_pct & poisson\_residual\_rms\_vs\_A1\_pct & end\_to\_end\_vs\_A1\_pct \\
\midrule
128 & A0: latent baseline & +0.0\% & +0.0\% & +0.0\% & +0.0\% & +0.0\% & +0.0\% & +8.0\% & +1.2\% & +2.5\% & +37.3\% & +2.7\% & -33.6\% \\
128 & A1: reconstruction & -7.4\% & -1.2\% & -2.5\% & -27.2\% & -2.7\% & +50.7\% & +0.0\% & +0.0\% & +0.0\% & +0.0\% & +0.0\% & +0.0\% \\
128 & A2: + interface value & -6.9\% & -7.3\% & -47.2\% & -26.0\% & -34.0\% & +62.8\% & +0.5\% & -6.2\% & -45.9\% & +1.6\% & -32.2\% & +8.0\% \\
128 & A3: + interface flux & -5.9\% & -8.4\% & -60.9\% & -24.1\% & -40.4\% & +81.4\% & +1.6\% & -7.3\% & -59.9\% & +4.2\% & -38.8\% & +20.4\% \\
128 & A4: + spectral & +8.8\% & -8.1\% & -57.5\% & -73.7\% & -38.2\% & +93.6\% & +17.5\% & -7.0\% & -56.4\% & -63.9\% & -36.5\% & +28.5\% \\
128 & A5: + PDE residual & +3.8\% & -6.8\% & -45.4\% & -56.0\% & -30.8\% & +97.2\% & +12.1\% & -5.6\% & -44.0\% & -39.7\% & -28.9\% & +30.9\% \\
256 & A0: latent baseline & +0.0\% & +0.0\% & +0.0\% & +0.0\% & +0.0\% & +0.0\% & +8.1\% & +2.8\% & +5.0\% & +40.1\% & +5.5\% & -49.6\% \\
256 & A1: reconstruction & -7.5\% & -2.7\% & -4.8\% & -28.6\% & -5.2\% & +98.5\% & +0.0\% & +0.0\% & +0.0\% & +0.0\% & +0.0\% & +0.0\% \\
256 & A2: + interface value & -5.3\% & -17.5\% & -55.8\% & -19.6\% & -48.5\% & +113.4\% & +2.3\% & -15.2\% & -53.6\% & +12.6\% & -45.7\% & +7.5\% \\
256 & A3: + interface flux & -4.9\% & -20.1\% & -70.9\% & -34.6\% & -58.6\% & +145.9\% & +2.9\% & -17.9\% & -69.5\% & -8.3\% & -56.3\% & +23.9\% \\
256 & A4: + spectral & +7.1\% & -20.2\% & -71.0\% & -41.6\% & -58.1\% & +166.3\% & +15.8\% & -17.9\% & -69.6\% & -18.2\% & -55.8\% & +34.2\% \\
256 & A5: + PDE residual & +7.5\% & -20.0\% & -70.5\% & -27.0\% & -58.1\% & +170.9\% & +16.2\% & -17.8\% & -69.0\% & +2.3\% & -55.9\% & +36.5\% \\
\bottomrule
\end{tabular}
  }
\end{table*}

\begin{table*}[htbp]
  \centering
  \caption{Calibrated loss weights and validation-screen selection.  The
  screen uses Poisson $128^2$, seed 0, and a 5\% MRE guardrail relative to the
  reconstruction-only candidate.  The spectral and PDE screens did not pass
  this guardrail, consistent with their final tradeoff.}
  \label{tab:physics-ladder-weights}
  \scriptsize
  \resizebox{\textwidth}{!}{%
    \begin{tabular}{rlllll}
\toprule
Resolution & Variant & interface value & interface flux & spectral & pde residual \\
\midrule
128 & A1: reconstruction &  &  &  &  \\
128 & A2: + interface value & 7.649e+01 \ensuremath{\pm} 2.9e+00 &  &  &  \\
128 & A3: + interface flux & 7.649e+01 \ensuremath{\pm} 2.9e+00 & 5.263e-01 \ensuremath{\pm} 1.9e-02 &  &  \\
128 & A4: + spectral & 7.649e+01 \ensuremath{\pm} 2.9e+00 & 5.263e-01 \ensuremath{\pm} 1.9e-02 & 3.987e-01 \ensuremath{\pm} 1.0e-01 &  \\
128 & A5: + PDE residual & 7.649e+01 \ensuremath{\pm} 2.9e+00 & 5.263e-01 \ensuremath{\pm} 1.9e-02 & 3.987e-01 \ensuremath{\pm} 1.0e-01 & 4.635e-04 \ensuremath{\pm} 1.7e-05 \\
256 & A1: reconstruction &  &  &  &  \\
256 & A2: + interface value & 8.416e+01 \ensuremath{\pm} 6.3e+00 &  &  &  \\
256 & A3: + interface flux & 8.416e+01 \ensuremath{\pm} 6.3e+00 & 3.101e-01 \ensuremath{\pm} 2.3e-02 &  &  \\
256 & A4: + spectral & 8.416e+01 \ensuremath{\pm} 6.3e+00 & 3.101e-01 \ensuremath{\pm} 2.3e-02 & 3.506e-01 \ensuremath{\pm} 4.1e-02 &  \\
256 & A5: + PDE residual & 8.416e+01 \ensuremath{\pm} 6.3e+00 & 3.101e-01 \ensuremath{\pm} 2.3e-02 & 3.506e-01 \ensuremath{\pm} 4.1e-02 & 2.083e-04 \ensuremath{\pm} 1.5e-05 \\
\bottomrule
\end{tabular}
  }
\end{table*}

\begin{table*}[htbp]
  \centering
  \caption{Cumulative stage costs and the additional fine-tuning cost for the
  interface-loss ladder.}
  \label{tab:physics-ladder-stage-costs}
  \scriptsize
  \resizebox{\textwidth}{!}{%
    \begin{tabular}{rlrlllllll}
\toprule
Resolution & Variant & Seeds & PCA fit (s) & Latent transform (s) & NN train, cumulative (s) & Inference, cumulative (s) & Fine-tune NN (s) & Fine-tune total (s) & End-to-end, cumulative (s) \\
\midrule
128 & A0: latent baseline & 3 & 4.7 \ensuremath{\pm} 0.0 & 2.3 \ensuremath{\pm} 0.0 & 174.2 \ensuremath{\pm} 2.7 & 0.3 \ensuremath{\pm} 0.0 &  &  & 203.8 \ensuremath{\pm} 5.6 \\
128 & A1: reconstruction & 3 & 4.7 \ensuremath{\pm} 0.0 & 4.7 \ensuremath{\pm} 0.0 & 254.7 \ensuremath{\pm} 0.7 & 0.5 \ensuremath{\pm} 0.0 & 80.5 \ensuremath{\pm} 2.6 & 103.3 \ensuremath{\pm} 2.6 & 307.1 \ensuremath{\pm} 5.1 \\
128 & A2: + interface value & 3 & 4.7 \ensuremath{\pm} 0.0 & 4.8 \ensuremath{\pm} 0.1 & 278.6 \ensuremath{\pm} 6.0 & 0.5 \ensuremath{\pm} 0.0 & 104.4 \ensuremath{\pm} 3.5 & 127.9 \ensuremath{\pm} 3.1 & 331.7 \ensuremath{\pm} 6.3 \\
128 & A3: + interface flux & 3 & 4.7 \ensuremath{\pm} 0.0 & 4.8 \ensuremath{\pm} 0.1 & 316.3 \ensuremath{\pm} 7.1 & 0.5 \ensuremath{\pm} 0.0 & 142.1 \ensuremath{\pm} 5.0 & 166.0 \ensuremath{\pm} 5.3 & 369.8 \ensuremath{\pm} 10.9 \\
128 & A4: + spectral & 3 & 4.7 \ensuremath{\pm} 0.0 & 4.8 \ensuremath{\pm} 0.1 & 340.4 \ensuremath{\pm} 2.3 & 0.5 \ensuremath{\pm} 0.0 & 166.2 \ensuremath{\pm} 3.5 & 190.6 \ensuremath{\pm} 3.2 & 394.4 \ensuremath{\pm} 6.0 \\
128 & A5: + PDE residual & 3 & 4.7 \ensuremath{\pm} 0.0 & 5.1 \ensuremath{\pm} 0.1 & 348.2 \ensuremath{\pm} 2.0 & 0.5 \ensuremath{\pm} 0.0 & 174.0 \ensuremath{\pm} 2.3 & 198.1 \ensuremath{\pm} 2.6 & 401.9 \ensuremath{\pm} 6.3 \\
256 & A0: latent baseline & 3 & 18.3 \ensuremath{\pm} 0.4 & 8.3 \ensuremath{\pm} 0.5 & 176.4 \ensuremath{\pm} 4.1 & 0.9 \ensuremath{\pm} 0.0 &  &  & 255.7 \ensuremath{\pm} 3.3 \\
256 & A1: reconstruction & 3 & 18.3 \ensuremath{\pm} 0.4 & 18.5 \ensuremath{\pm} 1.0 & 364.9 \ensuremath{\pm} 3.5 & 1.7 \ensuremath{\pm} 0.0 & 188.4 \ensuremath{\pm} 2.1 & 251.7 \ensuremath{\pm} 3.1 & 507.3 \ensuremath{\pm} 2.9 \\
256 & A2: + interface value & 3 & 18.3 \ensuremath{\pm} 0.4 & 18.5 \ensuremath{\pm} 0.5 & 403.2 \ensuremath{\pm} 4.6 & 1.7 \ensuremath{\pm} 0.0 & 226.8 \ensuremath{\pm} 0.8 & 289.8 \ensuremath{\pm} 1.4 & 545.5 \ensuremath{\pm} 4.6 \\
256 & A3: + interface flux & 3 & 18.3 \ensuremath{\pm} 0.4 & 18.8 \ensuremath{\pm} 1.5 & 485.3 \ensuremath{\pm} 3.4 & 1.8 \ensuremath{\pm} 0.1 & 308.9 \ensuremath{\pm} 1.8 & 373.0 \ensuremath{\pm} 2.8 & 628.7 \ensuremath{\pm} 4.4 \\
256 & A4: + spectral & 3 & 18.3 \ensuremath{\pm} 0.4 & 18.3 \ensuremath{\pm} 1.2 & 536.5 \ensuremath{\pm} 6.7 & 1.7 \ensuremath{\pm} 0.1 & 360.1 \ensuremath{\pm} 3.0 & 425.1 \ensuremath{\pm} 6.4 & 680.7 \ensuremath{\pm} 8.2 \\
256 & A5: + PDE residual & 3 & 18.3 \ensuremath{\pm} 0.4 & 21.8 \ensuremath{\pm} 3.8 & 544.7 \ensuremath{\pm} 8.5 & 1.8 \ensuremath{\pm} 0.0 & 368.3 \ensuremath{\pm} 4.4 & 436.8 \ensuremath{\pm} 8.7 & 692.5 \ensuremath{\pm} 11.5 \\
\bottomrule
\end{tabular}
  }
\end{table*}

\begin{figure*}[htbp]
  \centering
  \includegraphics[width=0.96\textwidth]{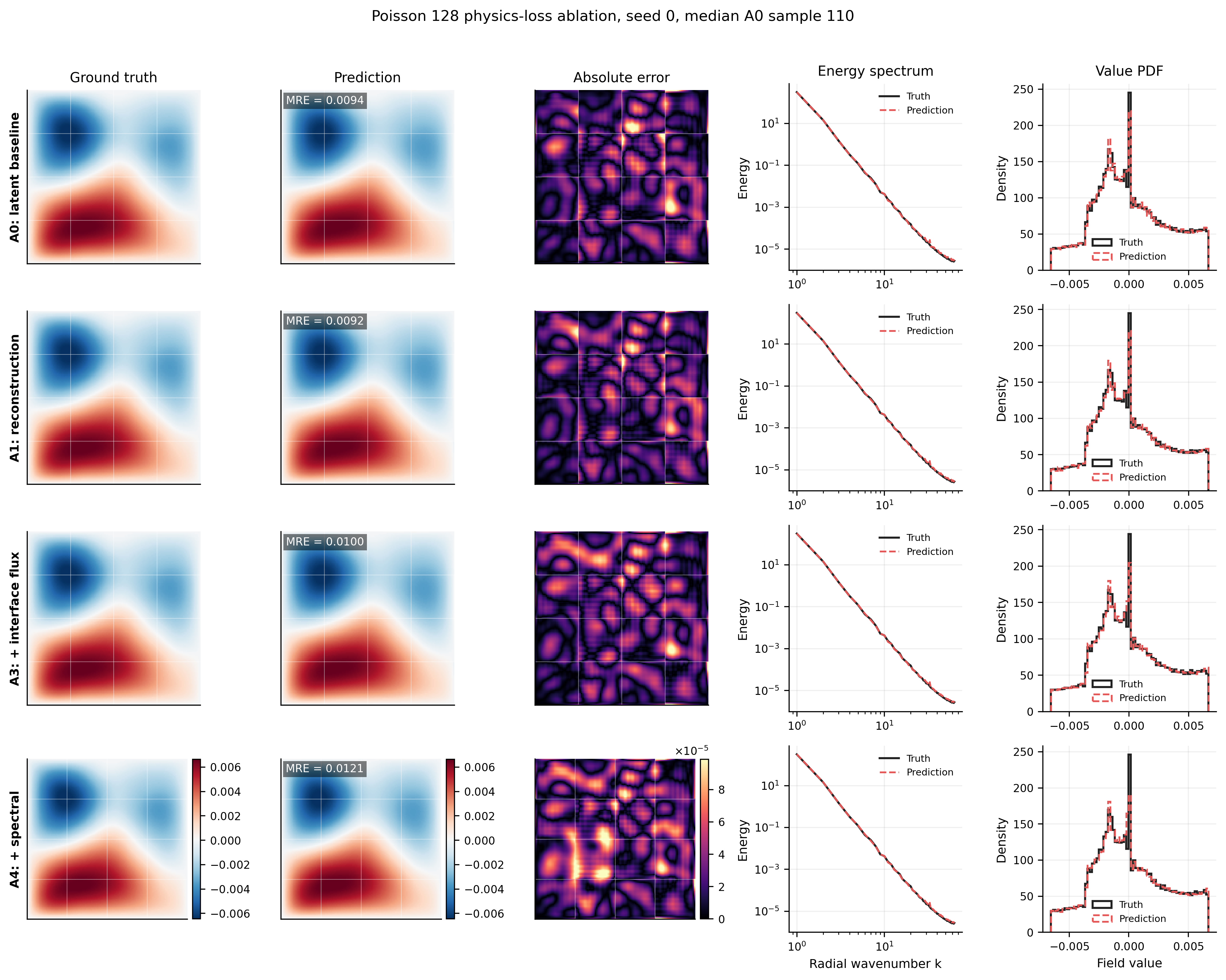}
  \caption{Poisson $128^2$ representative reconstruction, error, spectrum,
  and value-density comparison for selected ladder stages.  Interface-aware
  fine-tuning changes the seam-local error structure with only a small shift
  in pointwise MRE}
  \label{fig:physics-ladder-qualitative-128}
\end{figure*}

\begin{figure*}[htbp]
  \centering
  \includegraphics[width=0.96\textwidth]{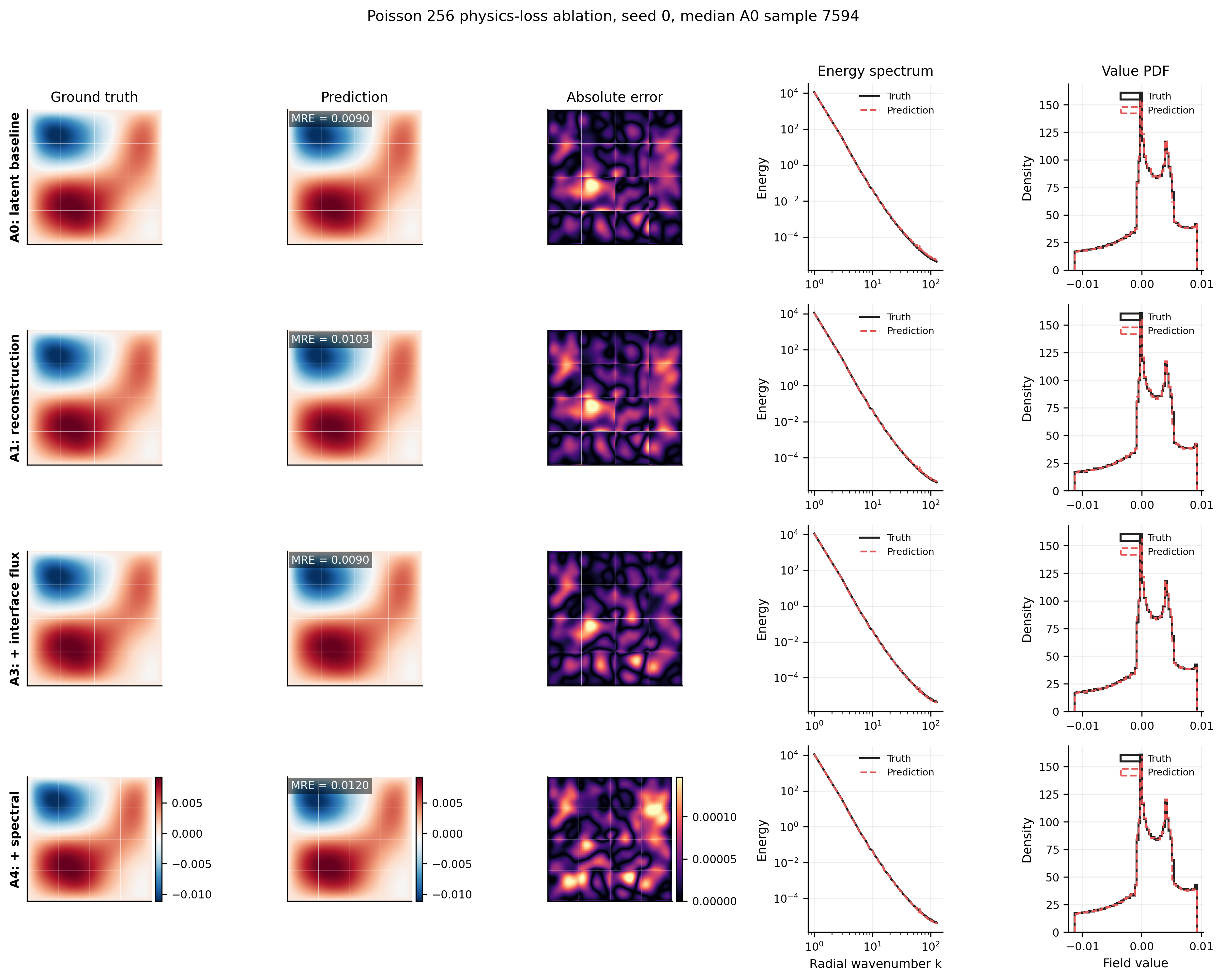}
  \caption{Poisson $256^2$ qualitative diagnostics for the interface-loss
  ladder}
  \label{fig:physics-ladder-qualitative-256}
\end{figure*}

\begin{figure*}[htbp]
  \centering
  \includegraphics[width=0.48\textwidth]{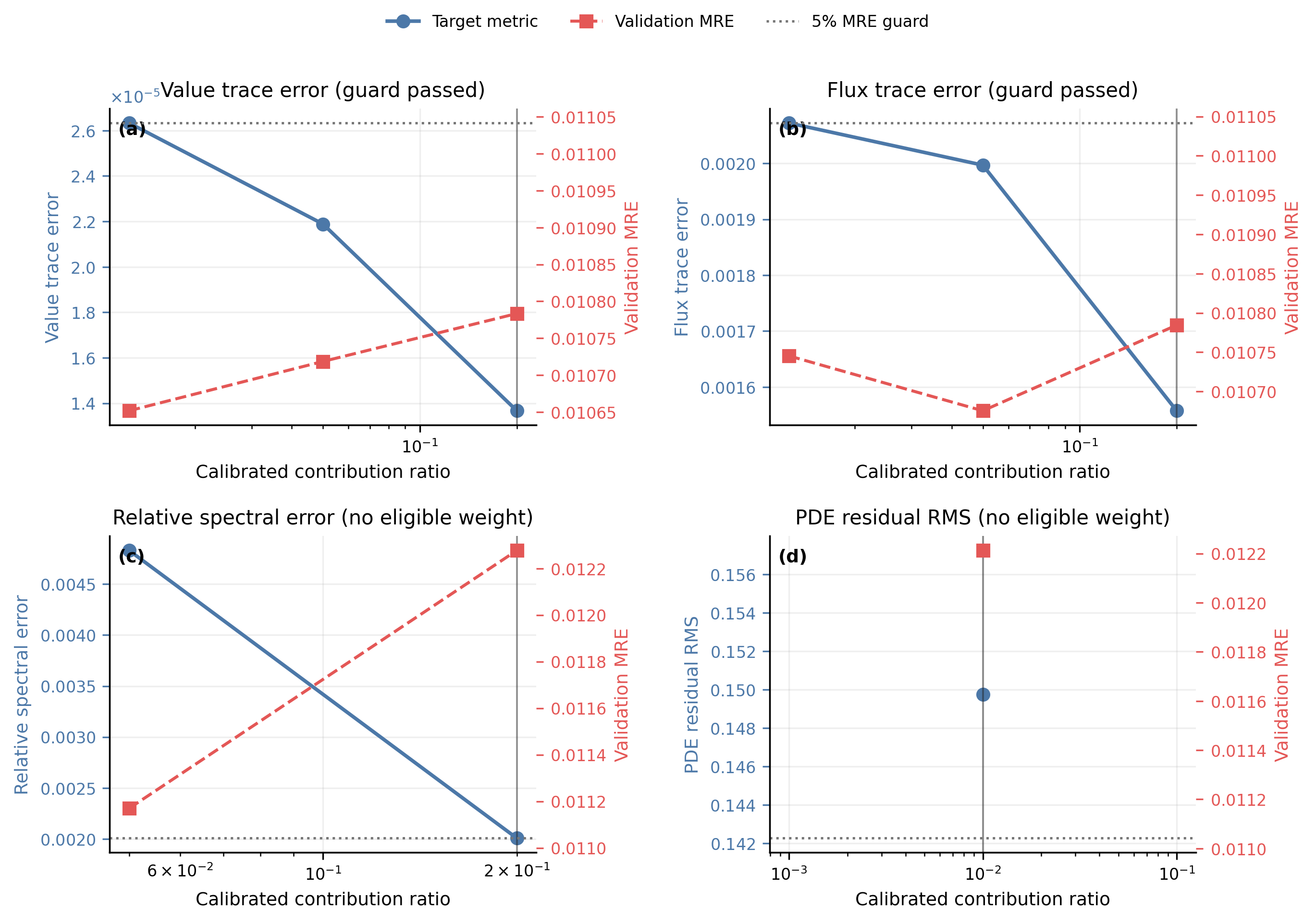}\hfill
  \includegraphics[width=0.48\textwidth]{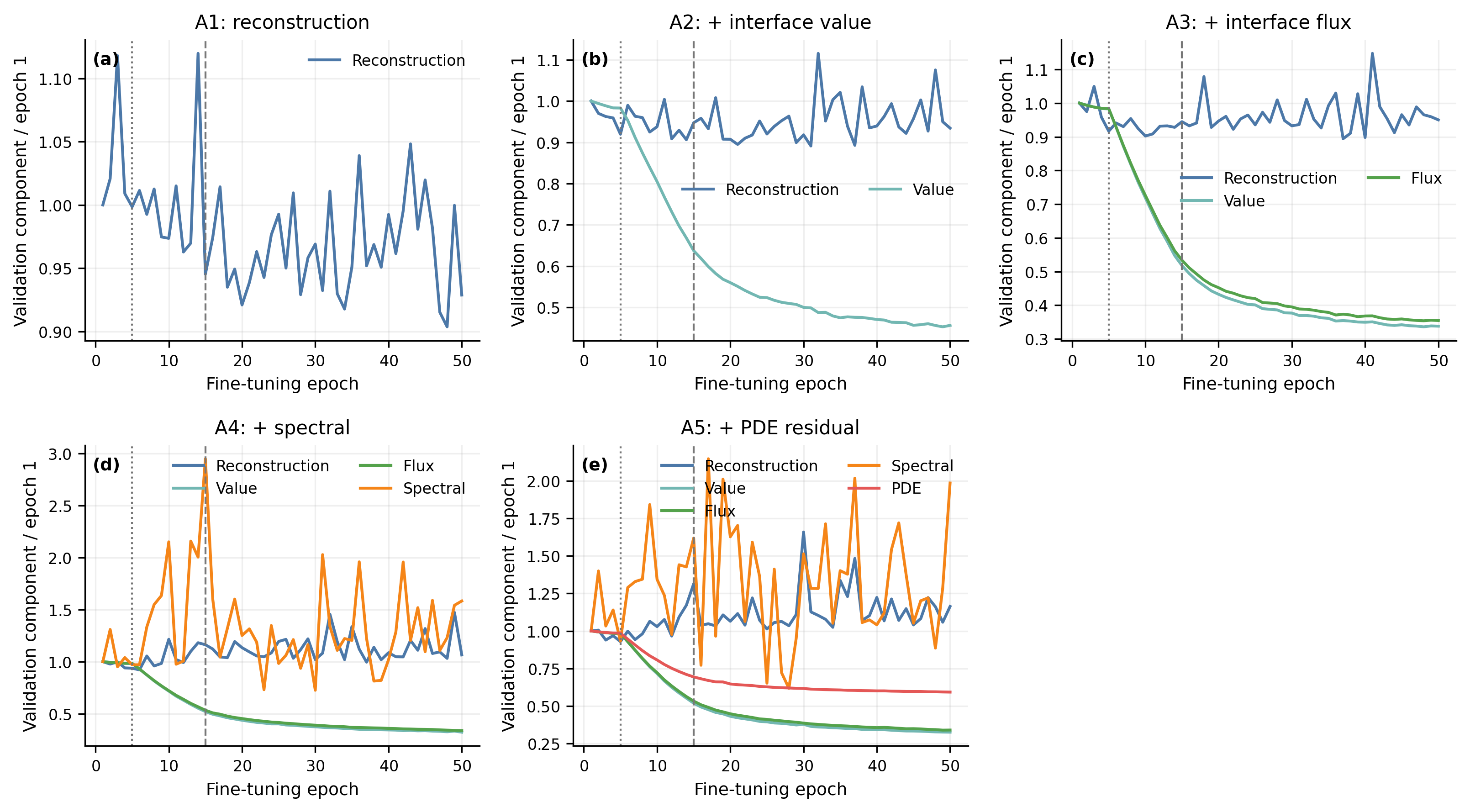}
  \caption{Loss-weight screening and training trajectories.  The calibrated
  interface terms satisfy the reconstruction guardrail; spectral and
  PDE-residual additions illustrate the quality tradeoffs observed in the
  final three-seed confirmation}
  \label{fig:physics-ladder-screen-curves}
\end{figure*}

\begin{figure*}[htbp]
  \centering
  \includegraphics[width=0.96\textwidth]{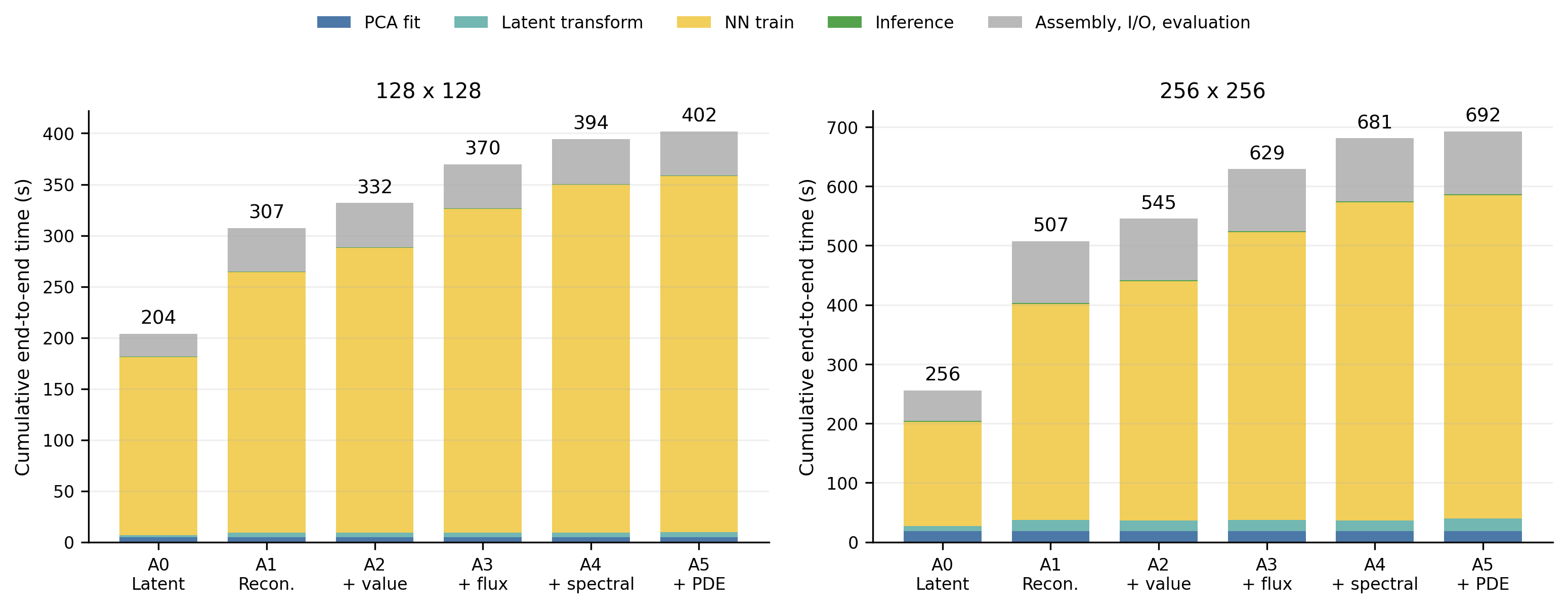}
  \caption{Stage-wise cumulative cost for the interface-loss ladder.  The
  extra cost is dominated by assembled-field fine-tuning rather than PCA
  fitting or inference}
  \label{fig:physics-ladder-cost-breakdown}
\end{figure*}

\subsection{Supplementary Sample-Efficiency Results}
\label{app:sample-efficiency}

This experiment contains 48 runs: four methods, four nested Poisson $128^2$
training-set sizes, and three paired seeds.  All methods use the same split
within each seed.  The interface model uses the A3 reconstruction/value/flux
fine-tuning protocol, and its cumulative time includes the two-scale warm start
plus its separate fine-tuning invocation.  Native overlap diagnostics use its
stride-16 seam grid; the other models use stride 32.  Direct continuity
comparisons therefore rely on the retained common-stride seed-0 audit.

\begin{table*}[htbp]
  \centering
  \caption{Complete Poisson $128^2$ sample-efficiency study.  Values are means
  and sample standard deviations over three paired seeds.}
  \label{tab:sample-efficiency-full}
  \scriptsize
  \resizebox{\textwidth}{!}{%
    \begin{tabular}{rlllllll}
\toprule
Training samples & Method & MRE (\%) & SSIM & Value trace error & Flux trace error & Spectral error & Total time (s) \\
\midrule
1250 & Plain L2L & 4.878 \ensuremath{\pm} 0.097 & 0.9426 \ensuremath{\pm} 0.0014 & 4.020e-04 \ensuremath{\pm} 1.9e-06 & 5.376e-02 \ensuremath{\pm} 2.4e-04 & 0.00999 \ensuremath{\pm} 0.00089 & 220.6 \ensuremath{\pm} 2.0 \\
1250 & L2L + overlap & 2.520 \ensuremath{\pm} 0.078 & 0.9834 \ensuremath{\pm} 0.0007 & 8.529e-06 \ensuremath{\pm} 7.7e-08 & 4.717e-04 \ensuremath{\pm} 2.2e-06 & 0.00413 \ensuremath{\pm} 0.00101 & 237.0 \ensuremath{\pm} 6.4 \\
1250 & Two-scale & 3.283 \ensuremath{\pm} 0.153 & 0.9860 \ensuremath{\pm} 0.0004 & 3.073e-05 \ensuremath{\pm} 4.8e-07 & 4.427e-03 \ensuremath{\pm} 8.0e-05 & 0.03634 \ensuremath{\pm} 0.00288 & 244.4 \ensuremath{\pm} 5.7 \\
1250 & Two-scale + interface & 2.875 \ensuremath{\pm} 0.131 & 0.9883 \ensuremath{\pm} 0.0005 & 9.025e-06 \ensuremath{\pm} 1.7e-07 & 1.225e-03 \ensuremath{\pm} 3.3e-05 & 0.03043 \ensuremath{\pm} 0.00296 & 459.8 \ensuremath{\pm} 14.9 \\
2000 & Plain L2L & 4.821 \ensuremath{\pm} 0.113 & 0.9430 \ensuremath{\pm} 0.0014 & 4.045e-04 \ensuremath{\pm} 2.0e-06 & 5.408e-02 \ensuremath{\pm} 2.4e-04 & 0.00624 \ensuremath{\pm} 0.00043 & 203.1 \ensuremath{\pm} 3.8 \\
2000 & L2L + overlap & 2.399 \ensuremath{\pm} 0.071 & 0.9843 \ensuremath{\pm} 0.0006 & 8.359e-06 \ensuremath{\pm} 5.6e-08 & 4.725e-04 \ensuremath{\pm} 2.5e-06 & 0.00211 \ensuremath{\pm} 0.00010 & 226.3 \ensuremath{\pm} 5.1 \\
2000 & Two-scale & 2.341 \ensuremath{\pm} 0.054 & 0.9908 \ensuremath{\pm} 0.0003 & 3.008e-05 \ensuremath{\pm} 2.7e-07 & 4.347e-03 \ensuremath{\pm} 4.7e-05 & 0.02128 \ensuremath{\pm} 0.00061 & 228.1 \ensuremath{\pm} 5.0 \\
2000 & Two-scale + interface & 2.012 \ensuremath{\pm} 0.061 & 0.9926 \ensuremath{\pm} 0.0003 & 8.470e-06 \ensuremath{\pm} 5.8e-08 & 1.225e-03 \ensuremath{\pm} 1.9e-05 & 0.01721 \ensuremath{\pm} 0.00026 & 417.9 \ensuremath{\pm} 5.8 \\
4000 & Plain L2L & 4.801 \ensuremath{\pm} 0.109 & 0.9432 \ensuremath{\pm} 0.0014 & 4.060e-04 \ensuremath{\pm} 1.8e-06 & 5.427e-02 \ensuremath{\pm} 2.4e-04 & 0.00433 \ensuremath{\pm} 0.00025 & 189.2 \ensuremath{\pm} 2.3 \\
4000 & L2L + overlap & 2.321 \ensuremath{\pm} 0.060 & 0.9849 \ensuremath{\pm} 0.0005 & 8.251e-06 \ensuremath{\pm} 5.1e-08 & 4.734e-04 \ensuremath{\pm} 2.1e-06 & 0.00133 \ensuremath{\pm} 0.00020 & 208.4 \ensuremath{\pm} 4.8 \\
4000 & Two-scale & 1.648 \ensuremath{\pm} 0.023 & 0.9948 \ensuremath{\pm} 0.0004 & 2.983e-05 \ensuremath{\pm} 1.9e-07 & 4.326e-03 \ensuremath{\pm} 2.1e-05 & 0.01143 \ensuremath{\pm} 0.00032 & 213.5 \ensuremath{\pm} 1.8 \\
4000 & Two-scale + interface & 1.448 \ensuremath{\pm} 0.032 & 0.9955 \ensuremath{\pm} 0.0002 & 8.751e-06 \ensuremath{\pm} 3.3e-07 & 1.329e-03 \ensuremath{\pm} 5.3e-05 & 0.00881 \ensuremath{\pm} 0.00061 & 396.2 \ensuremath{\pm} 17.8 \\
8000 & Plain L2L & 4.794 \ensuremath{\pm} 0.110 & 0.9432 \ensuremath{\pm} 0.0014 & 4.067e-04 \ensuremath{\pm} 1.9e-06 & 5.436e-02 \ensuremath{\pm} 2.3e-04 & 0.00351 \ensuremath{\pm} 0.00012 & 183.7 \ensuremath{\pm} 1.3 \\
8000 & L2L + overlap & 2.292 \ensuremath{\pm} 0.063 & 0.9852 \ensuremath{\pm} 0.0005 & 8.207e-06 \ensuremath{\pm} 3.9e-08 & 4.738e-04 \ensuremath{\pm} 2.1e-06 & 0.00153 \ensuremath{\pm} 0.00044 & 210.0 \ensuremath{\pm} 4.2 \\
8000 & Two-scale & 1.155 \ensuremath{\pm} 0.028 & 0.9967 \ensuremath{\pm} 0.0002 & 2.980e-05 \ensuremath{\pm} 4.0e-07 & 4.330e-03 \ensuremath{\pm} 5.1e-05 & 0.00569 \ensuremath{\pm} 0.00065 & 213.8 \ensuremath{\pm} 0.8 \\
8000 & Two-scale + interface & 1.086 \ensuremath{\pm} 0.026 & 0.9968 \ensuremath{\pm} 0.0001 & 1.043e-05 \ensuremath{\pm} 1.4e-07 & 1.613e-03 \ensuremath{\pm} 2.7e-05 & 0.00403 \ensuremath{\pm} 0.00041 & 378.3 \ensuremath{\pm} 4.8 \\
\bottomrule
\end{tabular}
  }
\end{table*}

\begin{table*}[htbp]
  \centering
  \caption{Relative changes of two-scale against overlap and of the
  interface-refined model against two-scale.  Positive error changes denote
  degradation; positive time changes denote additional cost.}
  \label{tab:sample-efficiency-relative-effects}
  \scriptsize
  \resizebox{\textwidth}{!}{%
    \begin{tabular}{rllllllllllll}
\toprule
Training samples & two\_scale\_vs\_overlap\_mre\_pct & interface\_vs\_two\_scale\_mre\_pct & two\_scale\_vs\_overlap\_ssim\_pct & interface\_vs\_two\_scale\_ssim\_pct & two\_scale\_vs\_overlap\_value\_trace\_pct & interface\_vs\_two\_scale\_value\_trace\_pct & two\_scale\_vs\_overlap\_flux\_trace\_pct & interface\_vs\_two\_scale\_flux\_trace\_pct & two\_scale\_vs\_overlap\_spectral\_pct & interface\_vs\_two\_scale\_spectral\_pct & two\_scale\_vs\_overlap\_total\_time\_pct & interface\_vs\_two\_scale\_total\_time\_pct \\
\midrule
1250 & -30.3\% & +12.4\% & +0.3\% & +0.2\% & -260.3\% & +70.6\% & -838.4\% & +72.3\% & -780.5\% & +16.3\% & -3.2\% & -88.1\% \\
2000 & +2.4\% & +14.0\% & +0.7\% & +0.2\% & -259.9\% & +71.8\% & -820.1\% & +71.8\% & -910.0\% & +19.2\% & -0.8\% & -83.2\% \\
4000 & +29.0\% & +12.2\% & +1.0\% & +0.1\% & -261.5\% & +70.7\% & -813.9\% & +69.3\% & -759.8\% & +22.9\% & -2.4\% & -85.5\% \\
8000 & +49.6\% & +6.0\% & +1.2\% & +0.0\% & -263.1\% & +65.0\% & -814.0\% & +62.7\% & -272.5\% & +29.2\% & -1.8\% & -77.0\% \\
\bottomrule
\end{tabular}
  }
\end{table*}

\begin{table*}[htbp]
  \centering
  \caption{Common-seam seed-0 audit at the low- and high-data endpoints.  All
  methods are evaluated on the same stride-32 interface set.}
  \label{tab:sample-efficiency-common-seams}
  \scriptsize
  \resizebox{\textwidth}{!}{%
    \begin{tabular}{rlrrllll}
\toprule
Training samples & Method & Seed & Diagnostic stride & Value jump & Flux jump & Value trace error & Flux trace error \\
\midrule
1250 & Plain L2L & 0 & 32 & 4.505e-04 & 5.395e-02 & 4.033e-04 & 5.392e-02 \\
1250 & L2L + overlap & 0 & 32 & 8.914e-05 & 6.749e-04 & 8.935e-06 & 5.340e-04 \\
1250 & Two-scale & 0 & 32 & 9.953e-05 & 4.551e-03 & 3.128e-05 & 4.518e-03 \\
1250 & Two-scale + interface & 0 & 32 & 8.969e-05 & 1.395e-03 & 9.168e-06 & 1.256e-03 \\
8000 & Plain L2L & 0 & 32 & 4.554e-04 & 5.453e-02 & 4.078e-04 & 5.450e-02 \\
8000 & L2L + overlap & 0 & 32 & 8.951e-05 & 6.740e-04 & 8.539e-06 & 5.365e-04 \\
8000 & Two-scale & 0 & 32 & 1.003e-04 & 4.314e-03 & 2.938e-05 & 4.275e-03 \\
8000 & Two-scale + interface & 0 & 32 & 9.203e-05 & 1.710e-03 & 1.031e-05 & 1.583e-03 \\
\bottomrule
\end{tabular}
  }
\end{table*}

\begin{table*}[htbp]
  \centering
  \caption{Stage-wise cumulative timing.  The interface rows include the
  two-scale run and the fine-tuning invocation.}
  \label{tab:sample-efficiency-costs}
  \scriptsize
  \resizebox{\textwidth}{!}{%
    \begin{tabular}{rllllllll}
\toprule
Training samples & Method & PCA fit (s) & Latent transform (s) & NN train (s) & Inference (s) & Invocation NN train (s) & Invocation total (s) & Cumulative total (s) \\
\midrule
1250 & Plain L2L & 0.5 \ensuremath{\pm} 0.0 & 0.3 \ensuremath{\pm} 0.0 & 199.7 \ensuremath{\pm} 2.2 & 0.1 \ensuremath{\pm} 0.0 & 199.7 \ensuremath{\pm} 2.2 & 220.6 \ensuremath{\pm} 2.0 & 220.6 \ensuremath{\pm} 2.0 \\
1250 & L2L + overlap & 1.6 \ensuremath{\pm} 0.1 & 1.0 \ensuremath{\pm} 0.0 & 213.6 \ensuremath{\pm} 6.6 & 0.3 \ensuremath{\pm} 0.0 & 213.6 \ensuremath{\pm} 6.6 & 237.0 \ensuremath{\pm} 6.4 & 237.0 \ensuremath{\pm} 6.4 \\
1250 & Two-scale & 0.8 \ensuremath{\pm} 0.0 & 0.6 \ensuremath{\pm} 0.0 & 222.2 \ensuremath{\pm} 6.1 & 0.3 \ensuremath{\pm} 0.0 & 222.2 \ensuremath{\pm} 6.1 & 244.4 \ensuremath{\pm} 5.7 & 244.4 \ensuremath{\pm} 5.7 \\
1250 & Two-scale + interface & 0.8 \ensuremath{\pm} 0.0 & 1.3 \ensuremath{\pm} 0.0 & 416.0 \ensuremath{\pm} 15.3 & 0.5 \ensuremath{\pm} 0.0 & 193.8 \ensuremath{\pm} 9.2 & 215.3 \ensuremath{\pm} 9.2 & 459.8 \ensuremath{\pm} 14.9 \\
2000 & Plain L2L & 0.8 \ensuremath{\pm} 0.0 & 0.5 \ensuremath{\pm} 0.0 & 181.7 \ensuremath{\pm} 4.0 & 0.1 \ensuremath{\pm} 0.0 & 181.7 \ensuremath{\pm} 4.0 & 203.1 \ensuremath{\pm} 3.8 & 203.1 \ensuremath{\pm} 3.8 \\
2000 & L2L + overlap & 2.3 \ensuremath{\pm} 0.1 & 1.3 \ensuremath{\pm} 0.0 & 200.1 \ensuremath{\pm} 4.5 & 0.3 \ensuremath{\pm} 0.0 & 200.1 \ensuremath{\pm} 4.5 & 226.3 \ensuremath{\pm} 5.1 & 226.3 \ensuremath{\pm} 5.1 \\
2000 & Two-scale & 1.2 \ensuremath{\pm} 0.0 & 0.8 \ensuremath{\pm} 0.1 & 205.7 \ensuremath{\pm} 5.1 & 0.3 \ensuremath{\pm} 0.0 & 205.7 \ensuremath{\pm} 5.1 & 228.1 \ensuremath{\pm} 5.0 & 228.1 \ensuremath{\pm} 5.0 \\
2000 & Two-scale + interface & 1.2 \ensuremath{\pm} 0.0 & 1.7 \ensuremath{\pm} 0.1 & 373.9 \ensuremath{\pm} 5.6 & 0.5 \ensuremath{\pm} 0.0 & 168.2 \ensuremath{\pm} 3.6 & 189.7 \ensuremath{\pm} 3.7 & 417.9 \ensuremath{\pm} 5.8 \\
4000 & Plain L2L & 1.5 \ensuremath{\pm} 0.0 & 0.7 \ensuremath{\pm} 0.0 & 166.5 \ensuremath{\pm} 2.2 & 0.1 \ensuremath{\pm} 0.0 & 166.5 \ensuremath{\pm} 2.2 & 189.2 \ensuremath{\pm} 2.3 & 189.2 \ensuremath{\pm} 2.3 \\
4000 & L2L + overlap & 4.6 \ensuremath{\pm} 0.1 & 2.1 \ensuremath{\pm} 0.1 & 180.7 \ensuremath{\pm} 4.4 & 0.3 \ensuremath{\pm} 0.0 & 180.7 \ensuremath{\pm} 4.4 & 208.4 \ensuremath{\pm} 4.8 & 208.4 \ensuremath{\pm} 4.8 \\
4000 & Two-scale & 2.4 \ensuremath{\pm} 0.0 & 1.3 \ensuremath{\pm} 0.0 & 189.4 \ensuremath{\pm} 1.6 & 0.3 \ensuremath{\pm} 0.0 & 189.4 \ensuremath{\pm} 1.6 & 213.5 \ensuremath{\pm} 1.8 & 213.5 \ensuremath{\pm} 1.8 \\
4000 & Two-scale + interface & 2.4 \ensuremath{\pm} 0.0 & 2.8 \ensuremath{\pm} 0.1 & 346.3 \ensuremath{\pm} 13.5 & 0.5 \ensuremath{\pm} 0.0 & 156.9 \ensuremath{\pm} 14.8 & 182.7 \ensuremath{\pm} 19.5 & 396.2 \ensuremath{\pm} 17.8 \\
8000 & Plain L2L & 3.2 \ensuremath{\pm} 0.1 & 1.2 \ensuremath{\pm} 0.1 & 158.8 \ensuremath{\pm} 1.5 & 0.1 \ensuremath{\pm} 0.0 & 158.8 \ensuremath{\pm} 1.5 & 183.7 \ensuremath{\pm} 1.3 & 183.7 \ensuremath{\pm} 1.3 \\
8000 & L2L + overlap & 10.1 \ensuremath{\pm} 0.3 & 3.8 \ensuremath{\pm} 0.1 & 174.7 \ensuremath{\pm} 4.0 & 0.3 \ensuremath{\pm} 0.0 & 174.7 \ensuremath{\pm} 4.0 & 210.0 \ensuremath{\pm} 4.2 & 210.0 \ensuremath{\pm} 4.2 \\
8000 & Two-scale & 5.3 \ensuremath{\pm} 0.1 & 2.4 \ensuremath{\pm} 0.1 & 184.9 \ensuremath{\pm} 0.7 & 0.3 \ensuremath{\pm} 0.0 & 184.9 \ensuremath{\pm} 0.7 & 213.8 \ensuremath{\pm} 0.8 & 213.8 \ensuremath{\pm} 0.8 \\
8000 & Two-scale + interface & 5.3 \ensuremath{\pm} 0.1 & 4.9 \ensuremath{\pm} 0.1 & 325.6 \ensuremath{\pm} 4.9 & 0.5 \ensuremath{\pm} 0.0 & 140.6 \ensuremath{\pm} 5.3 & 164.6 \ensuremath{\pm} 5.2 & 378.3 \ensuremath{\pm} 4.8 \\
\bottomrule
\end{tabular}
  }
\end{table*}

\begin{table*}[htbp]
  \centering
  \caption{PCA/SVD fit decomposition.  The two-scale construction avoids the
  duplicated overlapping input/output patch layouts and has roughly half the
  overlap PCA-fit time at 8,000 samples.}
  \label{tab:sample-efficiency-pca-breakdown}
  \scriptsize
  \resizebox{\textwidth}{!}{%
    \begin{tabular}{rlrrrrrrrrrrrr}
\toprule
Training samples & Method & Input PCA mean (s) & Input PCA std (s) & Output PCA mean (s) & Output PCA std (s) & Coarse global SVD mean (s) & Coarse global SVD std (s) & Residual output PCA mean (s) & Residual output PCA std (s) & PCA fit overhead mean (s) & PCA fit overhead std (s) & Total PCA fit mean (s) & Total PCA fit std (s) \\
\midrule
1250 & L2L + overlap & 0.689608 & 0.025273 & 0.681212 & 0.031940 & 0.000000 & 0.000000 & 0.000000 & 0.000000 & 0.199808 & 0.004654 & 1.570628 & 0.053803 \\
1250 & Plain L2L & 0.243689 & 0.037525 & 0.226141 & 0.021923 & 0.000000 & 0.000000 & 0.000000 & 0.000000 & 0.066490 & 0.000504 & 0.536319 & 0.030605 \\
1250 & Two-scale & 0.243654 & 0.018516 & 0.000000 & 0.000000 & 0.051294 & 0.004654 & 0.452990 & 0.020464 & 0.038957 & 0.002324 & 0.786896 & 0.032159 \\
2000 & L2L + overlap & 1.009946 & 0.044844 & 0.962589 & 0.050088 & 0.000000 & 0.000000 & 0.000000 & 0.000000 & 0.330295 & 0.010513 & 2.302830 & 0.089434 \\
2000 & Plain L2L & 0.331318 & 0.012669 & 0.322519 & 0.028946 & 0.000000 & 0.000000 & 0.000000 & 0.000000 & 0.107781 & 0.002883 & 0.761618 & 0.039172 \\
2000 & Two-scale & 0.333349 & 0.008913 & 0.000000 & 0.000000 & 0.077298 & 0.003770 & 0.695185 & 0.018343 & 0.059568 & 0.001110 & 1.165400 & 0.007481 \\
4000 & L2L + overlap & 1.991651 & 0.075542 & 1.956311 & 0.064044 & 0.000000 & 0.000000 & 0.000000 & 0.000000 & 0.655324 & 0.004241 & 4.603286 & 0.117761 \\
4000 & Plain L2L & 0.670569 & 0.008680 & 0.651024 & 0.031794 & 0.000000 & 0.000000 & 0.000000 & 0.000000 & 0.216974 & 0.005782 & 1.538566 & 0.043967 \\
4000 & Two-scale & 0.667759 & 0.015905 & 0.000000 & 0.000000 & 0.145260 & 0.017142 & 1.416259 & 0.018422 & 0.122667 & 0.004357 & 2.351946 & 0.018608 \\
8000 & L2L + overlap & 4.281089 & 0.021816 & 4.294998 & 0.115729 & 0.000000 & 0.000000 & 0.000000 & 0.000000 & 1.519751 & 0.157218 & 10.095838 & 0.252875 \\
8000 & Plain L2L & 1.360116 & 0.008930 & 1.340097 & 0.050376 & 0.000000 & 0.000000 & 0.000000 & 0.000000 & 0.453313 & 0.044779 & 3.153526 & 0.070466 \\
8000 & Two-scale & 1.428966 & 0.085808 & 0.000000 & 0.000000 & 0.365673 & 0.116292 & 3.191533 & 0.223303 & 0.266152 & 0.019994 & 5.252324 & 0.090627 \\
\bottomrule
\end{tabular}
  }
\end{table*}

\begin{figure*}[htbp]
  \centering
  \includegraphics[width=0.96\textwidth]{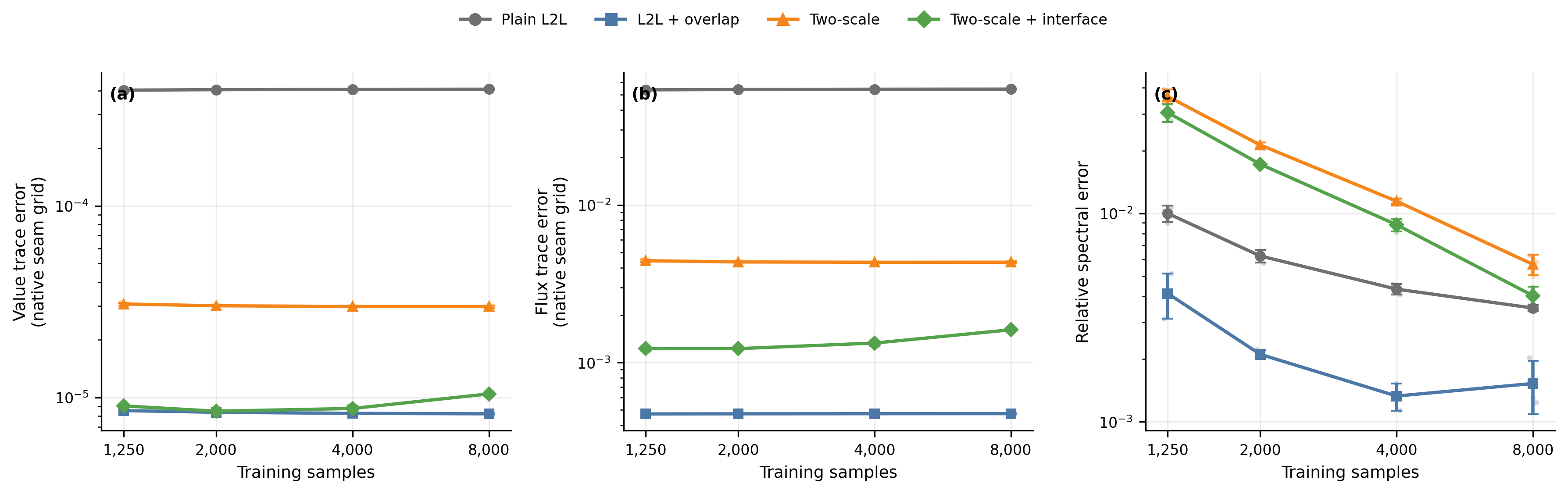}
  \caption{Native-seam artifact metrics across training-set size.  The
  interface refinement strongly improves two-scale traces, while overlap
  remains the strict flux-continuity reference.  Native seam geometries differ
  between overlap and nonoverlapping models; Table~\ref{tab:sample-efficiency-common-seams}
  provides the matched-stride audit}
  \label{fig:sample-efficiency-artifacts}
\end{figure*}

\begin{figure*}[htbp]
  \centering
  \includegraphics[width=0.96\textwidth]{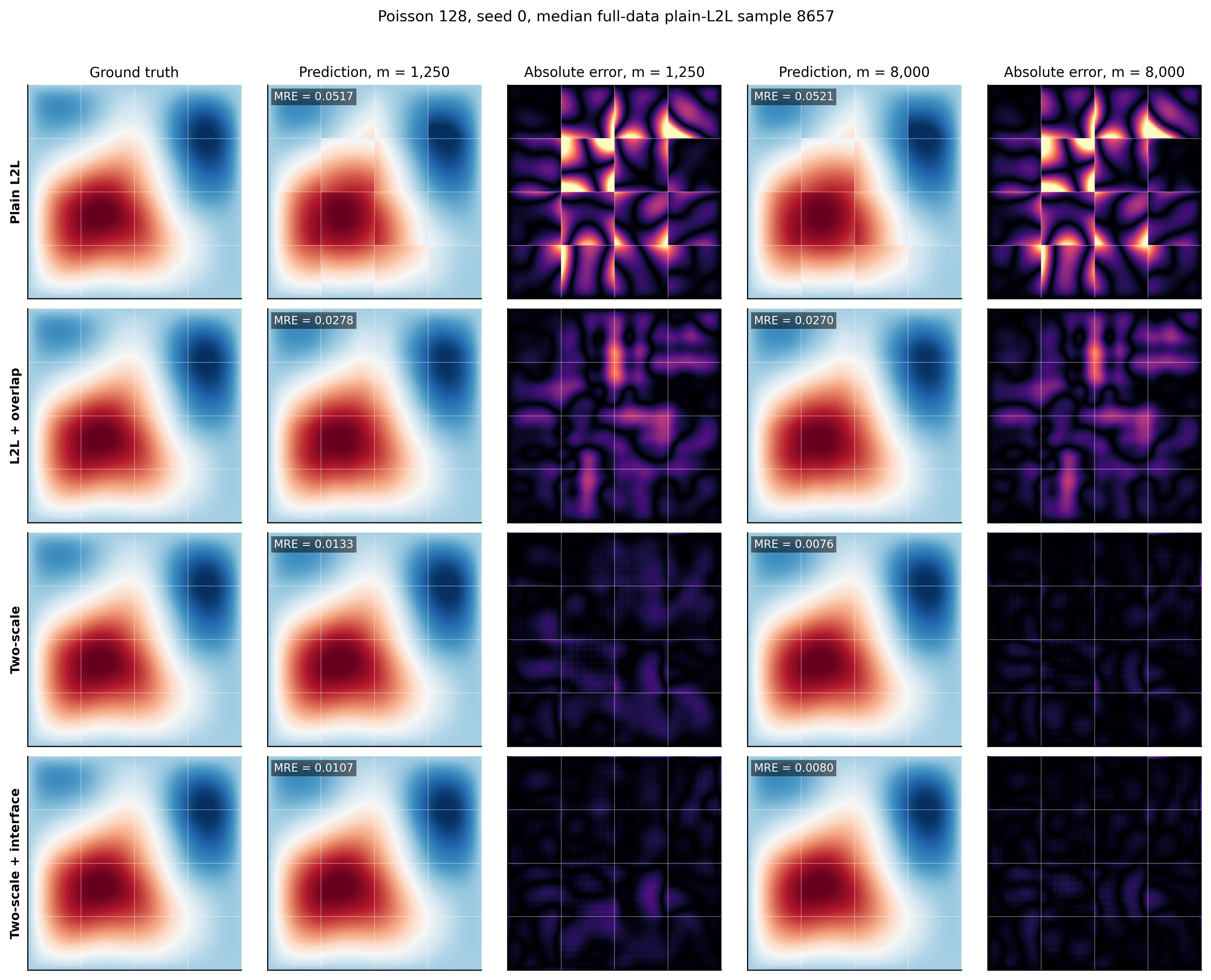}
  \caption{Representative fixed-sample reconstructions at 1,250 and 8,000
  training examples.  Two-scale field error contracts strongly with added
  data, consistent with the closing learned-to-oracle gap.}
  \label{fig:sample-efficiency-qualitative}
\end{figure*}

\begin{figure*}[htbp]
  \centering
  \includegraphics[width=0.48\textwidth]{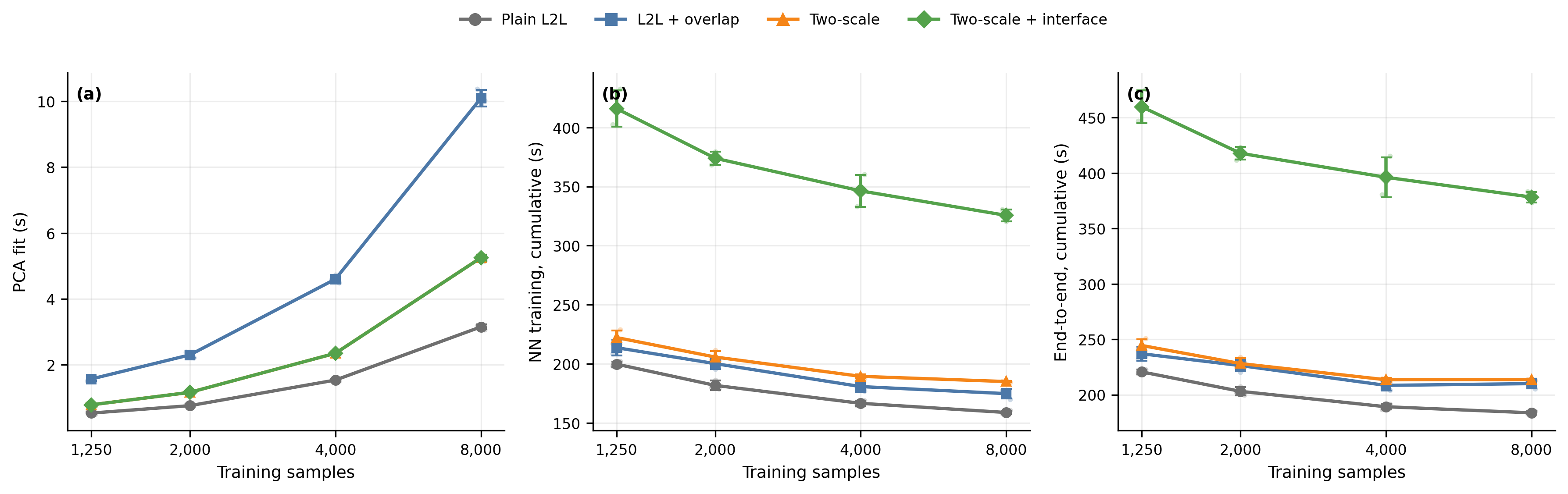}\hfill
  \includegraphics[width=0.48\textwidth]{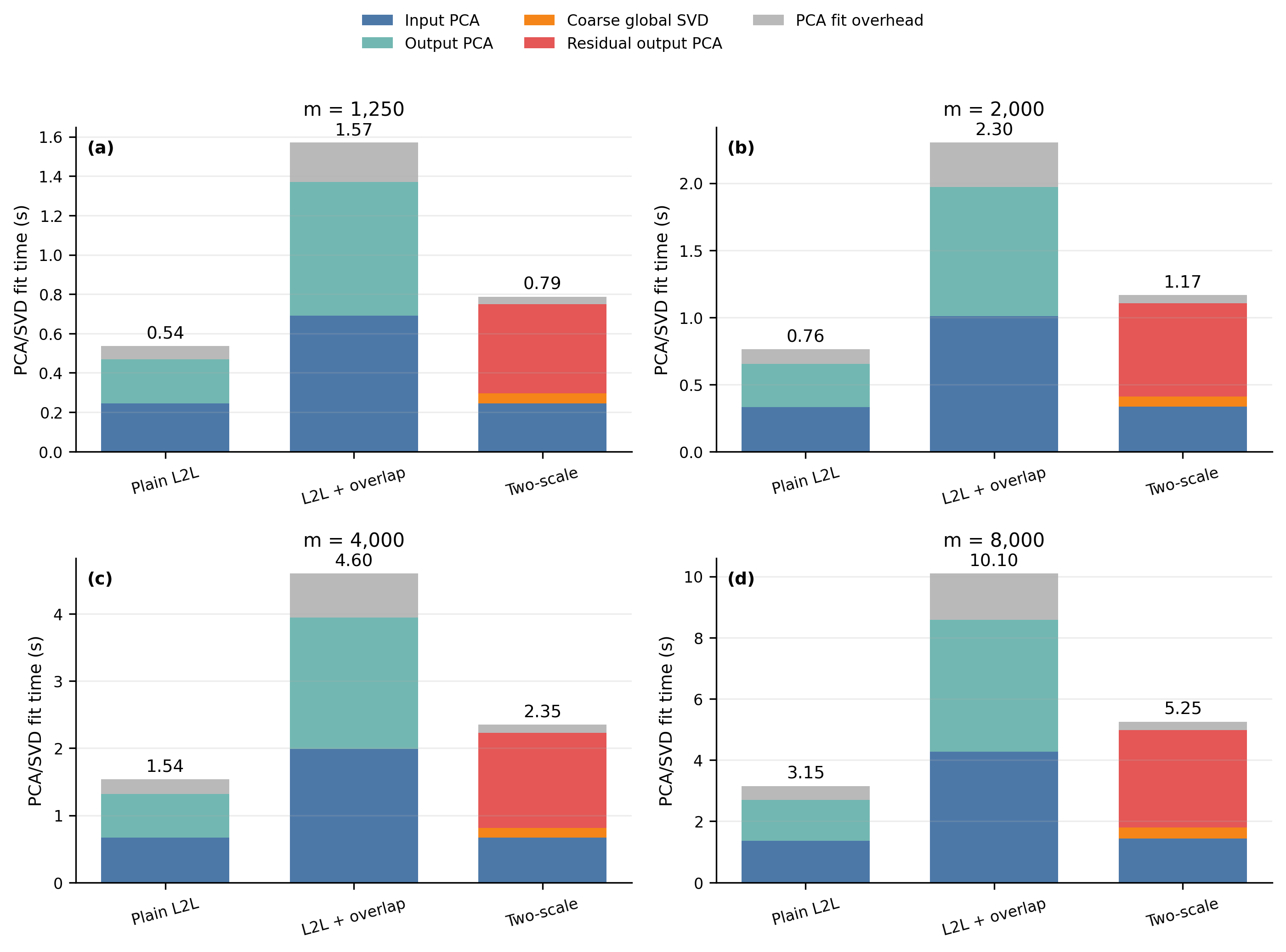}
  \caption{Sample-efficiency cost diagnostics.  The two-scale PCA advantage
  over overlap is visible in the fit-stage breakdown, although neural-network
  training dominates end-to-end time}
  \label{fig:sample-efficiency-cost-diagnostics}
\end{figure*}

\subsection{Supplementary Randomized-SVD Solver Results}
\label{app:randomized-pca-solver}

The solver audit uses five paired Poisson $128^2$ seeds with shared splits,
8,000 training examples, randomized-SVD oversampling 20, and four power
iterations.  The hybrid two-scale solver randomizes only the coarse global
PCA; it is included to isolate the contribution of this small construction
substage.

\begin{table*}[htbp]
  \centering
  \caption{Learned and PCA-oracle quality for the SVD solver audit.}
  \label{tab:svd-quality}
  \small
  \begin{tabular}{llcccc}
\toprule
Method
& Solver
& \makecell{Learned\\MRE}
& \makecell{PCA-oracle\\MRE}
& SSIM
& \makecell{Spectral\\error} \\
\midrule

Plain L2L
& Full
& \makecell{$0.04829$\\$\pm\,0.00093$}
& \makecell{$0.04825$\\$\pm\,0.00094$}
& \makecell{$0.94275$\\$\pm\,0.00125$}
& \makecell{$0.00349$\\$\pm\,0.00013$} \\

Plain L2L
& Randomized
& \makecell{$0.04829$\\$\pm\,0.00094$}
& \makecell{$0.04825$\\$\pm\,0.00094$}
& \makecell{$0.94274$\\$\pm\,0.00126$}
& \makecell{$0.00355$\\$\pm\,0.00013$} \\

L2L + overlap
& Full
& \makecell{$0.02309$\\$\pm\,0.00052$}
& \makecell{$0.02277$\\$\pm\,0.00051$}
& \makecell{$0.98496$\\$\pm\,0.00045$}
& \makecell{$0.00198$\\$\pm\,0.00019$} \\

L2L + overlap
& Randomized
& \makecell{$0.02308$\\$\pm\,0.00050$}
& \makecell{$0.02277$\\$\pm\,0.00051$}
& \makecell{$0.98499$\\$\pm\,0.00042$}
& \makecell{$0.00152$\\$\pm\,0.00031$} \\

Two-scale
& Full
& \makecell{$0.01139$\\$\pm\,0.00058$}
& \makecell{$0.00501$\\$\pm\,0.00009$}
& \makecell{$0.99659$\\$\pm\,0.00025$}
& \makecell{$0.00558$\\$\pm\,0.00085$} \\

Two-scale
& Hybrid
& \makecell{$0.01171$\\$\pm\,0.00084$}
& \makecell{$0.00501$\\$\pm\,0.00009$}
& \makecell{$0.99649$\\$\pm\,0.00036$}
& \makecell{$0.00577$\\$\pm\,0.00080$} \\

Two-scale
& Randomized
& \makecell{$0.01149$\\$\pm\,0.00022$}
& \makecell{$0.00501$\\$\pm\,0.00009$}
& \makecell{$0.99663$\\$\pm\,0.00015$}
& \makecell{$0.00548$\\$\pm\,0.00056$} \\

\bottomrule
\end{tabular}
\end{table*}

\begin{table*}[htbp]
  \centering
  \caption{Stage timing for full, hybrid, and randomized SVD.  PCA speedup is
  defined relative to the matching full-SVD representation.}
  \label{tab:svd-timing}
  \small
  \begin{tabular}{llccccc}
\toprule
Method
& Solver
& \makecell{PCA fit\\(s)}
& \makecell{PCA\\speedup}
& \makecell{Latent\\transform (s)}
& \makecell{NN train\\(s)}
& \makecell{End-to-end\\(s)} \\
\midrule

Plain L2L
& Full
& \makecell{$9.0$\\$\pm\,0.1$}
& $1.00\times$
& \makecell{$1.2$\\$\pm\,0.0$}
& \makecell{$157.9$\\$\pm\,3.8$}
& \makecell{$190.0$\\$\pm\,3.5$} \\

Plain L2L
& Randomized
& \makecell{$3.2$\\$\pm\,0.1$}
& $2.78\times$
& \makecell{$1.2$\\$\pm\,0.0$}
& \makecell{$155.7$\\$\pm\,1.8$}
& \makecell{$181.4$\\$\pm\,1.8$} \\

L2L + overlap
& Full
& \makecell{$28.5$\\$\pm\,0.3$}
& $1.00\times$
& \makecell{$4.0$\\$\pm\,0.2$}
& \makecell{$172.9$\\$\pm\,2.6$}
& \makecell{$227.5$\\$\pm\,2.1$} \\

L2L + overlap
& Randomized
& \makecell{$10.4$\\$\pm\,0.4$}
& $2.74\times$
& \makecell{$3.9$\\$\pm\,0.1$}
& \makecell{$168.0$\\$\pm\,2.5$}
& \makecell{$204.6$\\$\pm\,2.1$} \\

Two-scale
& Full
& \makecell{$13.3$\\$\pm\,0.2$}
& $1.00\times$
& \makecell{$2.3$\\$\pm\,0.1$}
& \makecell{$176.4$\\$\pm\,5.0$}
& \makecell{$214.0$\\$\pm\,5.2$} \\

Two-scale
& Hybrid
& \makecell{$12.9$\\$\pm\,0.4$}
& $1.03\times$
& \makecell{$2.3$\\$\pm\,0.1$}
& \makecell{$179.7$\\$\pm\,3.5$}
& \makecell{$216.6$\\$\pm\,3.6$} \\

Two-scale
& Randomized
& \makecell{$4.7$\\$\pm\,0.2$}
& $2.81\times$
& \makecell{$2.3$\\$\pm\,0.0$}
& \makecell{$177.0$\\$\pm\,2.9$}
& \makecell{$206.2$\\$\pm\,2.0$} \\

\bottomrule
\end{tabular}
\end{table*}

\begin{table*}[htbp]
  \centering
  \caption{Paired equivalence audit.  Relative MRE and PCA-oracle margins are
  $\pm5\%$ and $\pm2\%$, respectively; SSIM uses an absolute $\pm0.001$
  margin.  Inconclusive two-scale learned-MRE intervals reflect downstream
  neural-training variation, not an oracle-representation mismatch.}
  \label{tab:svd-equivalence}
  \scriptsize
  \resizebox{\textwidth}{!}{%
    \begin{tabular}{lllllll}
\toprule
Comparison & Metric & Mean change & 90\% CI lower & 90\% CI upper & Margin & Decision \\
\midrule
Plain: randomized vs full & Learned MRE & +0.01\% & -0.01\% & +0.02\% & +5.00\% & Equivalent \\
Plain: randomized vs full & PCA-oracle MRE & -0.00\% & -0.00\% & -0.00\% & +2.00\% & Equivalent \\
Plain: randomized vs full & SSIM & -0.00001 & -0.00003 & +0.00001 & +0.00100 & Equivalent \\
Plain: randomized vs full & MSE & +0.01\% & -0.01\% & +0.04\% & +10.00\% & Equivalent \\
Plain: randomized vs full & Spectral error & +1.68\% & -2.80\% & +6.15\% & +10.00\% & Equivalent \\
Plain: randomized vs full & Value trace error & +0.00\% & -0.03\% & +0.03\% & +10.00\% & Equivalent \\
Plain: randomized vs full & Flux trace error & -0.01\% & -0.04\% & +0.02\% & +10.00\% & Equivalent \\
Plain: randomized vs full & PDE residual RMS & -0.00\% & -0.03\% & +0.02\% & +10.00\% & Equivalent \\
Plain: randomized vs full & PCA-fit speedup & 2.78x & 2.70x & 2.87x & 1.20x & Pass \\
Overlap: randomized vs full & Learned MRE & -0.04\% & -0.21\% & +0.13\% & +5.00\% & Equivalent \\
Overlap: randomized vs full & PCA-oracle MRE & -0.00\% & -0.00\% & -0.00\% & +2.00\% & Equivalent \\
Overlap: randomized vs full & SSIM & +0.00003 & -0.00001 & +0.00008 & +0.00100 & Equivalent \\
Overlap: randomized vs full & MSE & -0.08\% & -0.43\% & +0.27\% & +10.00\% & Equivalent \\
Overlap: randomized vs full & Spectral error & -22.36\% & -41.08\% & -3.64\% & +10.00\% & Inconclusive \\
Overlap: randomized vs full & Value trace error & -0.03\% & -0.14\% & +0.08\% & +10.00\% & Equivalent \\
Overlap: randomized vs full & Flux trace error & -0.39\% & -0.59\% & -0.19\% & +10.00\% & Equivalent \\
Overlap: randomized vs full & PDE residual RMS & -0.08\% & -0.10\% & -0.05\% & +10.00\% & Equivalent \\
Overlap: randomized vs full & PCA-fit speedup & 2.74x & 2.64x & 2.84x & 1.20x & Pass \\
Two-scale: hybrid vs full & Learned MRE & +3.03\% & -6.05\% & +12.12\% & +5.00\% & Inconclusive \\
Two-scale: hybrid vs full & PCA-oracle MRE & +0.00\% & -0.00\% & +0.00\% & +2.00\% & Equivalent \\
Two-scale: hybrid vs full & SSIM & -0.00010 & -0.00047 & +0.00027 & +0.00100 & Equivalent \\
Two-scale: hybrid vs full & MSE & +6.88\% & -12.29\% & +26.06\% & +10.00\% & Inconclusive \\
Two-scale: hybrid vs full & Spectral error & +6.14\% & -19.78\% & +32.06\% & +10.00\% & Inconclusive \\
Two-scale: hybrid vs full & Value trace error & +0.40\% & -1.32\% & +2.12\% & +10.00\% & Equivalent \\
Two-scale: hybrid vs full & Flux trace error & +0.39\% & -1.36\% & +2.15\% & +10.00\% & Equivalent \\
Two-scale: hybrid vs full & PDE residual RMS & +0.30\% & -0.89\% & +1.50\% & +10.00\% & Equivalent \\
Two-scale: hybrid vs full & PCA-fit speedup & 1.03x & 1.01x & 1.06x & 1.20x & Inconclusive \\
Two-scale: randomized vs full & Learned MRE & +1.17\% & -5.20\% & +7.55\% & +5.00\% & Inconclusive \\
Two-scale: randomized vs full & PCA-oracle MRE & +0.00\% & -0.00\% & +0.00\% & +2.00\% & Equivalent \\
Two-scale: randomized vs full & SSIM & +0.00004 & -0.00023 & +0.00031 & +0.00100 & Equivalent \\
Two-scale: randomized vs full & MSE & +2.72\% & -10.28\% & +15.72\% & +10.00\% & Inconclusive \\
Two-scale: randomized vs full & Spectral error & +0.40\% & -18.48\% & +19.27\% & +10.00\% & Inconclusive \\
Two-scale: randomized vs full & Value trace error & +0.39\% & -2.11\% & +2.88\% & +10.00\% & Equivalent \\
Two-scale: randomized vs full & Flux trace error & +0.44\% & -2.08\% & +2.96\% & +10.00\% & Equivalent \\
Two-scale: randomized vs full & PDE residual RMS & +0.28\% & -1.54\% & +2.10\% & +10.00\% & Equivalent \\
Two-scale: randomized vs full & PCA-fit speedup & 2.81x & 2.72x & 2.90x & 1.20x & Pass \\
Two-scale: randomized vs hybrid & Learned MRE & -1.37\% & -9.41\% & +6.67\% & +5.00\% & Inconclusive \\
Two-scale: randomized vs hybrid & PCA-oracle MRE & +0.00\% & -0.00\% & +0.00\% & +2.00\% & Equivalent \\
Two-scale: randomized vs hybrid & SSIM & +0.00014 & -0.00030 & +0.00057 & +0.00100 & Equivalent \\
Two-scale: randomized vs hybrid & MSE & -2.14\% & -18.38\% & +14.09\% & +10.00\% & Inconclusive \\
Two-scale: randomized vs hybrid & Spectral error & -3.01\% & -23.54\% & +17.51\% & +10.00\% & Inconclusive \\
Two-scale: randomized vs hybrid & Value trace error & -0.00\% & -2.23\% & +2.23\% & +10.00\% & Equivalent \\
Two-scale: randomized vs hybrid & Flux trace error & +0.05\% & -2.13\% & +2.24\% & +10.00\% & Equivalent \\
Two-scale: randomized vs hybrid & PDE residual RMS & -0.02\% & -1.67\% & +1.63\% & +10.00\% & Equivalent \\
Two-scale: randomized vs hybrid & PCA-fit speedup & 2.72x & 2.63x & 2.81x & 1.20x & Pass \\
\bottomrule
\end{tabular}
  }
\end{table*}

\begin{table*}[htbp]
  \centering
  \caption{Retained dimensions, explained variance, and PCA-fit diagnostics.
  Full and randomized solvers retain the same dimensions and explained
  variance to the displayed precision.}
  \label{tab:svd-pca-diagnostics}
  \scriptsize
  \resizebox{\textwidth}{!}{%
    \begin{tabular}{lllllllllllll}
\toprule
Cell & Method & Solver & Input latent dimension & Output latent dimension & Coarse dimension & Residual latent dimension & Input explained variance & Output explained variance & Coarse explained variance & Residual explained variance & Internal PCA fit (s) & Coarse SVD (s) \\
\midrule
Plain L2L Full & Plain L2L & Full & 144.0 & 48.0 &  &  & 0.990830 & 0.997739 &  &  & 8.565 & 0.000 \\
Plain L2L Randomized & Plain L2L & Randomized & 144.0 & 48.0 &  &  & 0.990830 & 0.997739 &  &  & 2.772 & 0.000 \\
L2L + overlap Full & L2L + overlap & Full & 487.6 & 147.0 &  &  & 0.990863 & 0.997612 &  &  & 26.811 & 0.000 \\
L2L + overlap Randomized & L2L + overlap & Randomized & 487.6 & 147.0 &  &  & 0.990863 & 0.997612 &  &  & 8.794 & 0.000 \\
Two-scale Full & Two-scale & Full & 144.0 & 162.0 & 10.0 & 152.0 & 0.990830 &  & 0.991797 & 0.995962 & 13.072 & 0.633 \\
Two-scale Hybrid & Two-scale & Hybrid & 144.0 & 162.0 & 10.0 & 152.0 & 0.990830 &  & 0.991797 & 0.995962 & 12.636 & 0.274 \\
Two-scale Randomized & Two-scale & Randomized & 144.0 & 162.0 & 10.0 & 152.0 & 0.990830 &  & 0.991797 & 0.995962 & 4.489 & 0.284 \\
\bottomrule
\end{tabular}
  }
\end{table*}

\begin{table*}[htbp]
  \centering
  \caption{PCA-fit decomposition.  Randomizing the coarse global solve alone
  does not materially change two-scale fit time; randomizing local input and
  residual fits supplies the practical saving.}
  \label{tab:svd-pca-breakdown}
  \scriptsize
  \resizebox{\textwidth}{!}{%
    \begin{tabular}{lrrrrrrrrrrrr}
\toprule
Cell & Input PCA mean (s) & Input PCA std (s) & Output PCA mean (s) & Output PCA std (s) & Coarse global SVD mean (s) & Coarse global SVD std (s) & Residual output PCA mean (s) & Residual output PCA std (s) & PCA fit overhead mean (s) & PCA fit overhead std (s) & Total PCA fit mean (s) & Total PCA fit std (s) \\
\midrule
Plain L2L Full & 4.389318 & 0.099922 & 4.175717 & 0.061830 & 0.000000 & 0.000000 & 0.000000 & 0.000000 & 0.433713 & 0.011993 & 8.998747 & 0.140805 \\
Plain L2L Randomized & 1.397528 & 0.039351 & 1.374341 & 0.032062 & 0.000000 & 0.000000 & 0.000000 & 0.000000 & 0.463015 & 0.029646 & 3.234884 & 0.071645 \\
L2L + overlap Full & 13.711305 & 0.170269 & 13.100033 & 0.199288 & 0.000000 & 0.000000 & 0.000000 & 0.000000 & 1.683102 & 0.152137 & 28.494440 & 0.284104 \\
L2L + overlap Randomized & 4.499585 & 0.188495 & 4.294310 & 0.091798 & 0.000000 & 0.000000 & 0.000000 & 0.000000 & 1.607734 & 0.211967 & 10.401629 & 0.381887 \\
Two-scale Full & 4.442077 & 0.037829 & 0.000000 & 0.000000 & 0.633017 & 0.036377 & 7.997088 & 0.179841 & 0.252656 & 0.017432 & 13.324838 & 0.158261 \\
Two-scale Hybrid & 4.397816 & 0.049313 & 0.000000 & 0.000000 & 0.274243 & 0.032788 & 7.963469 & 0.311943 & 0.245828 & 0.008663 & 12.881357 & 0.375349 \\
Two-scale Randomized & 1.380544 & 0.027246 & 0.000000 & 0.000000 & 0.284135 & 0.022072 & 2.824340 & 0.126454 & 0.249844 & 0.010845 & 4.738862 & 0.156030 \\
\bottomrule
\end{tabular}
  }
\end{table*}

\begin{table*}[htbp]
  \centering
  \caption{Complete solver audit metric matrix.}
  \label{tab:svd-full-metrics}
  \scriptsize
  \resizebox{\textwidth}{!}{%
    \begin{tabular}{llllllllllllllll}
\toprule
Method & Solver & Learned MRE & PCA-oracle MRE & SSIM & Spectral error & MSE & MAE & Value trace error & Flux trace error & PDE residual RMS & PCA fit (s) & Latent transform (s) & NN train (s) & Inference (s) & End-to-end (s) \\
\midrule
Plain L2L & Full & 0.04829 \ensuremath{\pm} 0.00093 & 0.04825 \ensuremath{\pm} 0.00094 & 0.94275 \ensuremath{\pm} 0.00125 & 0.00349 \ensuremath{\pm} 0.00013 & 4.100e-08 \ensuremath{\pm} 3.9e-10 & 1.293e-04 \ensuremath{\pm} 4.8e-07 & 4.061e-04 \ensuremath{\pm} 2.3e-06 & 5.429e-02 \ensuremath{\pm} 2.9e-04 & 3.16076 \ensuremath{\pm} 0.02289 & 9.0 \ensuremath{\pm} 0.1 & 1.2 \ensuremath{\pm} 0.0 & 157.9 \ensuremath{\pm} 3.8 & 0.1 \ensuremath{\pm} 0.0 & 190.0 \ensuremath{\pm} 3.5 \\
Plain L2L & Randomized & 0.04829 \ensuremath{\pm} 0.00094 & 0.04825 \ensuremath{\pm} 0.00094 & 0.94274 \ensuremath{\pm} 0.00126 & 0.00355 \ensuremath{\pm} 0.00013 & 4.100e-08 \ensuremath{\pm} 3.9e-10 & 1.293e-04 \ensuremath{\pm} 4.6e-07 & 4.061e-04 \ensuremath{\pm} 2.4e-06 & 5.428e-02 \ensuremath{\pm} 3.1e-04 & 3.16061 \ensuremath{\pm} 0.02330 & 3.2 \ensuremath{\pm} 0.1 & 1.2 \ensuremath{\pm} 0.0 & 155.7 \ensuremath{\pm} 1.8 & 0.1 \ensuremath{\pm} 0.0 & 181.4 \ensuremath{\pm} 1.8 \\
L2L + overlap & Full & 0.02309 \ensuremath{\pm} 0.00052 & 0.02277 \ensuremath{\pm} 0.00051 & 0.98496 \ensuremath{\pm} 0.00045 & 0.00198 \ensuremath{\pm} 0.00019 & 9.369e-09 \ensuremath{\pm} 7.4e-11 & 7.096e-05 \ensuremath{\pm} 2.4e-07 & 8.207e-06 \ensuremath{\pm} 3.5e-08 & 4.770e-04 \ensuremath{\pm} 3.4e-06 & 0.10278 \ensuremath{\pm} 0.00055 & 28.5 \ensuremath{\pm} 0.3 & 4.0 \ensuremath{\pm} 0.2 & 172.9 \ensuremath{\pm} 2.6 & 0.3 \ensuremath{\pm} 0.0 & 227.5 \ensuremath{\pm} 2.1 \\
L2L + overlap & Randomized & 0.02308 \ensuremath{\pm} 0.00050 & 0.02277 \ensuremath{\pm} 0.00051 & 0.98499 \ensuremath{\pm} 0.00042 & 0.00152 \ensuremath{\pm} 0.00031 & 9.361e-09 \ensuremath{\pm} 5.4e-11 & 7.094e-05 \ensuremath{\pm} 1.6e-07 & 8.204e-06 \ensuremath{\pm} 3.1e-08 & 4.751e-04 \ensuremath{\pm} 2.7e-06 & 0.10271 \ensuremath{\pm} 0.00053 & 10.4 \ensuremath{\pm} 0.4 & 3.9 \ensuremath{\pm} 0.1 & 168.0 \ensuremath{\pm} 2.5 & 0.3 \ensuremath{\pm} 0.0 & 204.6 \ensuremath{\pm} 2.1 \\
Two-scale & Full & 0.01139 \ensuremath{\pm} 0.00058 & 0.00501 \ensuremath{\pm} 0.00009 & 0.99659 \ensuremath{\pm} 0.00025 & 0.00558 \ensuremath{\pm} 0.00085 & 2.287e-09 \ensuremath{\pm} 2.4e-10 & 3.112e-05 \ensuremath{\pm} 1.1e-06 & 2.952e-05 \ensuremath{\pm} 7.0e-07 & 4.287e-03 \ensuremath{\pm} 1.0e-04 & 0.24418 \ensuremath{\pm} 0.00437 & 13.3 \ensuremath{\pm} 0.2 & 2.3 \ensuremath{\pm} 0.1 & 176.4 \ensuremath{\pm} 5.0 & 0.3 \ensuremath{\pm} 0.0 & 214.0 \ensuremath{\pm} 5.2 \\
Two-scale & Hybrid & 0.01171 \ensuremath{\pm} 0.00084 & 0.00501 \ensuremath{\pm} 0.00009 & 0.99649 \ensuremath{\pm} 0.00036 & 0.00577 \ensuremath{\pm} 0.00080 & 2.424e-09 \ensuremath{\pm} 3.7e-10 & 3.180e-05 \ensuremath{\pm} 1.7e-06 & 2.963e-05 \ensuremath{\pm} 5.0e-07 & 4.303e-03 \ensuremath{\pm} 7.3e-05 & 0.24489 \ensuremath{\pm} 0.00350 & 12.9 \ensuremath{\pm} 0.4 & 2.3 \ensuremath{\pm} 0.1 & 179.7 \ensuremath{\pm} 3.5 & 0.3 \ensuremath{\pm} 0.0 & 216.6 \ensuremath{\pm} 3.6 \\
Two-scale & Randomized & 0.01149 \ensuremath{\pm} 0.00022 & 0.00501 \ensuremath{\pm} 0.00009 & 0.99663 \ensuremath{\pm} 0.00015 & 0.00548 \ensuremath{\pm} 0.00056 & 2.325e-09 \ensuremath{\pm} 1.3e-10 & 3.116e-05 \ensuremath{\pm} 5.8e-07 & 2.962e-05 \ensuremath{\pm} 4.0e-07 & 4.304e-03 \ensuremath{\pm} 5.4e-05 & 0.24481 \ensuremath{\pm} 0.00257 & 4.7 \ensuremath{\pm} 0.2 & 2.3 \ensuremath{\pm} 0.0 & 177.0 \ensuremath{\pm} 2.9 & 0.3 \ensuremath{\pm} 0.0 & 206.2 \ensuremath{\pm} 2.0 \\
\bottomrule
\end{tabular}
  }
\end{table*}

\begin{figure*}[htbp]
  \centering
  \includegraphics[width=0.48\textwidth]{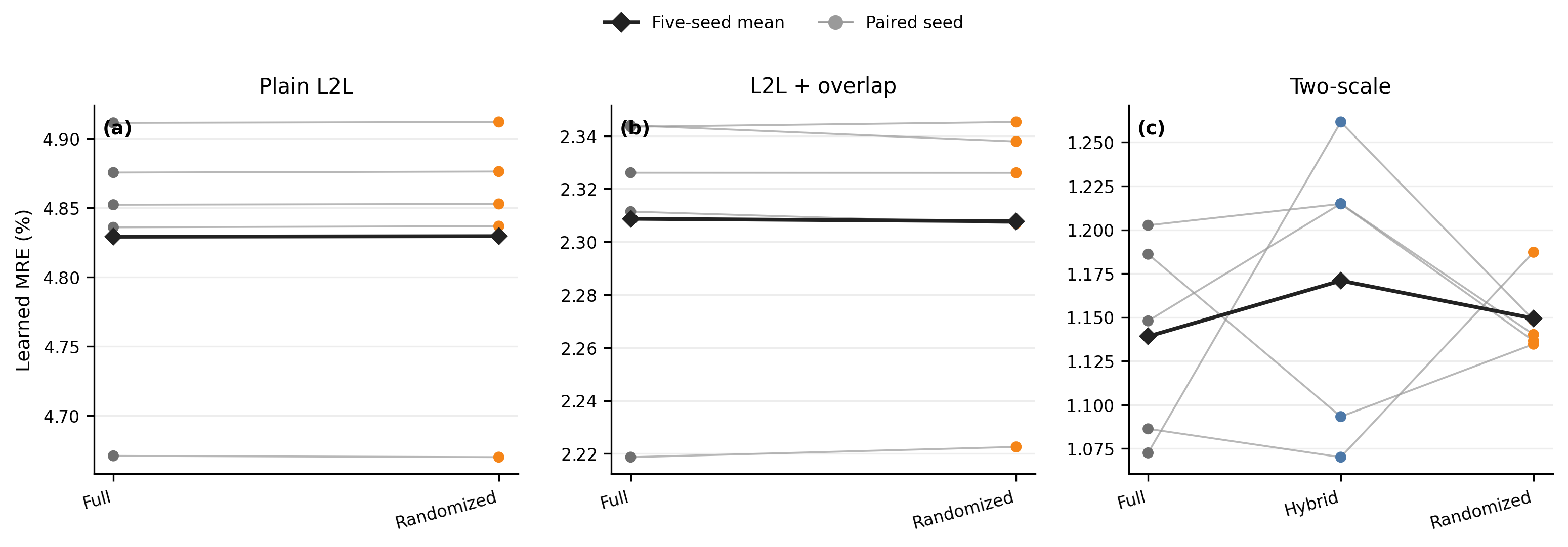}\hfill
  \includegraphics[width=0.48\textwidth]{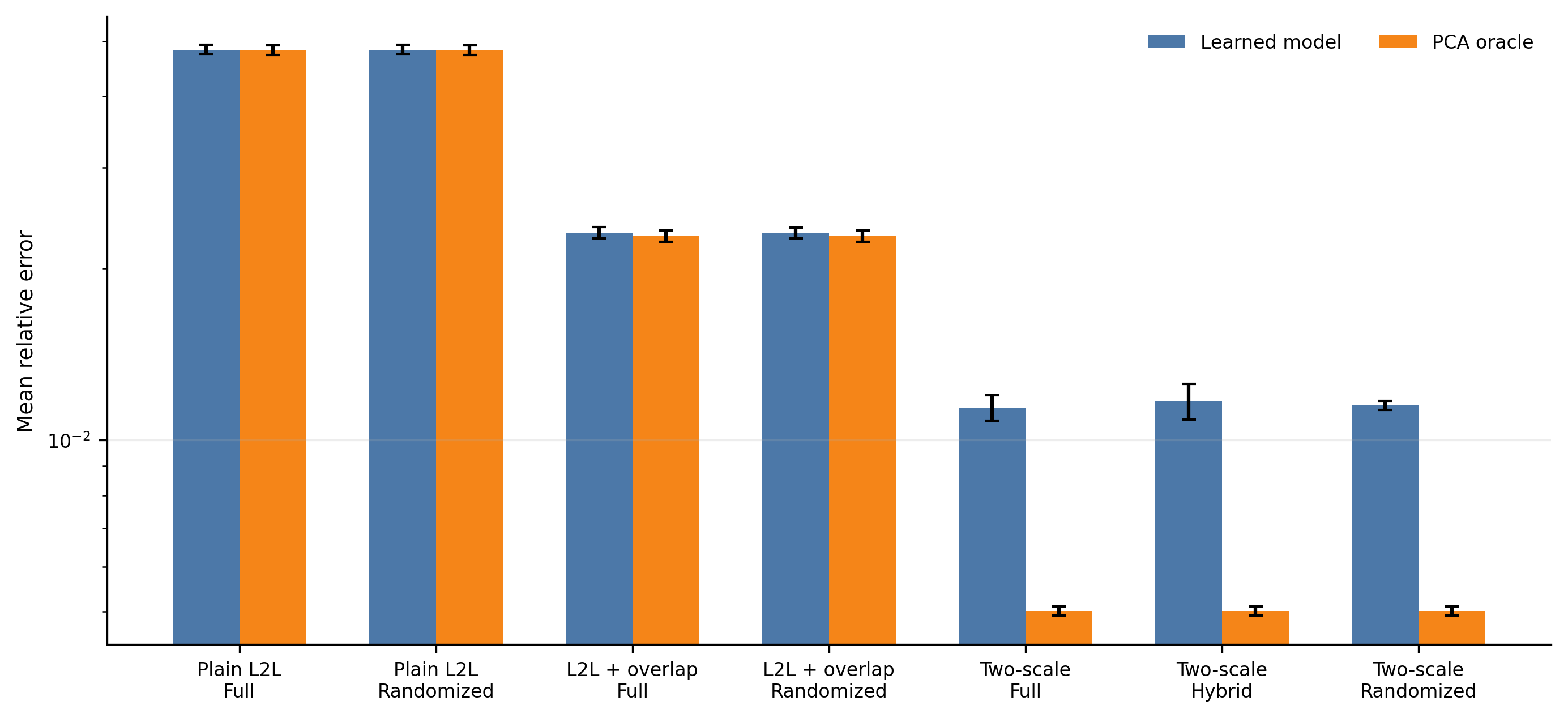}
  \caption{Per-seed learned-MRE variation and learned-to-oracle gap.  The
  two-scale representation retains a large trainability gap under either
  solver, confirming that its remaining error is not an SVD-accuracy effect}
  \label{fig:svd-paired-and-gap}
\end{figure*}

\begin{figure*}[htbp]
  \centering
  \includegraphics[width=0.48\textwidth]{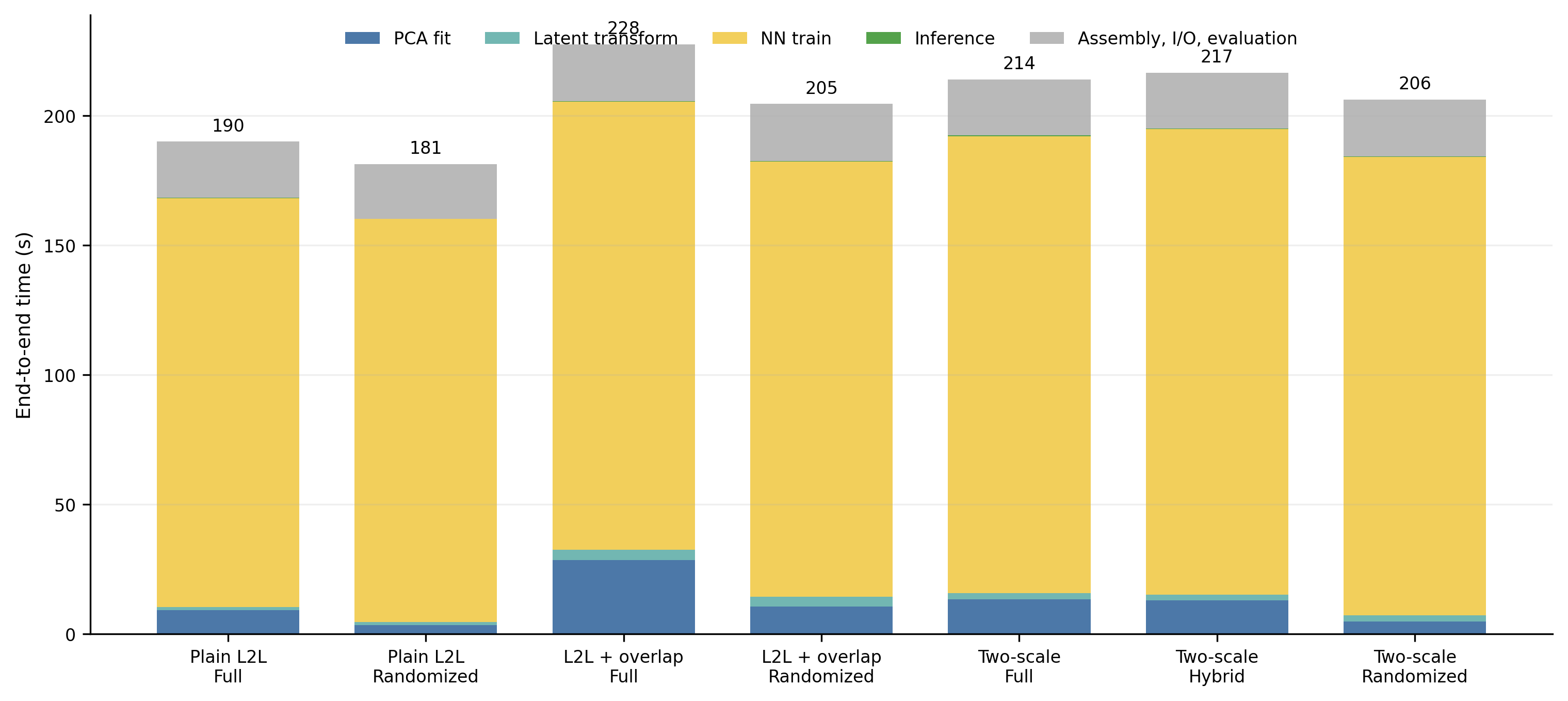}\hfill
  \includegraphics[width=0.48\textwidth]{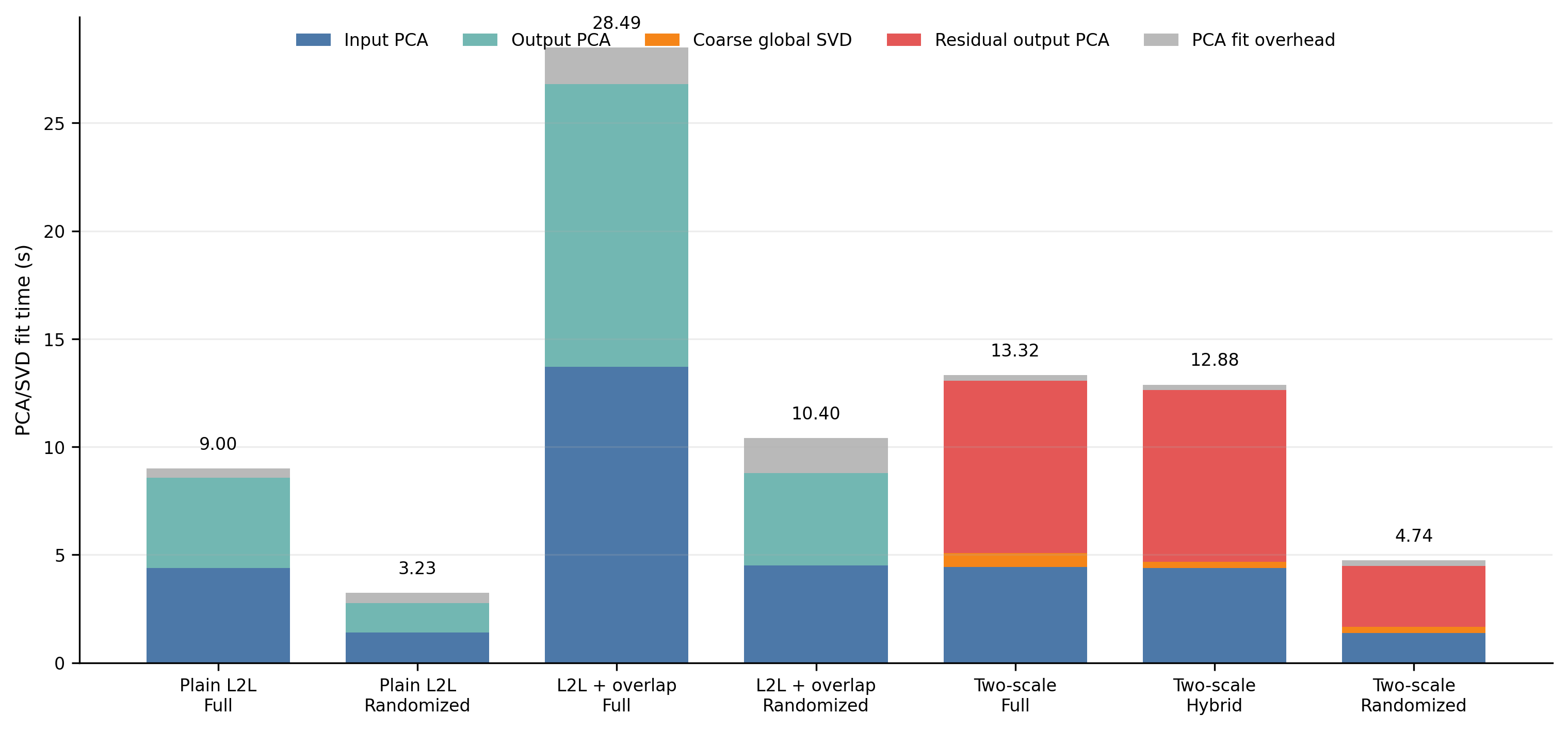}
  \caption{Cumulative stage costs and PCA-fit decomposition for the solver
  audit.  The end-to-end effect is limited by neural-network training, while
  the full randomized configuration captures the available PCA-stage saving}
  \label{fig:svd-cost-diagnostics}
\end{figure*}

\end{document}